\documentclass[a4paper]{article}[12pts]

\usepackage[english]{babel}
\usepackage[utf8x]{inputenc}
\usepackage[T1]{fontenc}

\normalsize

\usepackage[a4paper,top=1in,bottom=1in,left=1in,right=1in,marginparwidth=1in]{geometry}

\usepackage{natbib}
\usepackage{setspace}
\usepackage{amsmath}
\usepackage{amssymb} 
\usepackage{amsthm}
\usepackage{amsfonts, mathtools}
\usepackage{algorithm,algpseudocode}
\usepackage{bm}
\usepackage{bbm}
\usepackage{comment}
\usepackage{graphicx}
\usepackage{multirow, booktabs}
\usepackage{caption}
\usepackage{subcaption}
\usepackage[colorinlistoftodos]{todonotes}
\usepackage{hyperref}
\hypersetup{colorlinks=true, allcolors=blue}

\usepackage{multirow}
\usepackage{textgreek}
\usepackage{adjustbox}

\theoremstyle{plain}
\newtheorem{theorem}{Theorem}[section]
\newtheorem{proposition}[theorem]{Proposition}
\newtheorem{lemma}[theorem]{Lemma}
\newtheorem{corollary}[theorem]{Corollary}
\theoremstyle{definition}

\newtheorem{assumption}[theorem]{Assumption}
\newtheorem{example}[theorem]{Example}
\newtheorem{claim}[theorem]{Claim}
\theoremstyle{remark}

\definecolor{rose}{rgb}{1.0, 0.33, 0.64}

\DeclareMathOperator*{\argmax}{arg\,max}
\DeclareMathOperator*{\argmin}{arg\,min}

\title{OMGPT: A Sequence Modeling Framework for Data-driven Operational Decision Making}

\author{Hanzhao Wang$^\dagger$,  Guanting Chen$^\ddagger$, Kalyan Talluri$^\dagger$, Xiaocheng Li$^\dagger$}
\date{\small $^\dagger$
Imperial College Business School, Imperial College London, \\
$\{$h.wang19, kalyan.talluri, xiaocheng.li$\}$@imperial.ac.uk\\
$^\ddagger$Department of Statistics and Operations Research, University of North Carolina at Chapel Hill, guanting@unc.edu}

\begin{document}
\maketitle
\onehalfspacing

\begin{abstract}
We build a Generative Pre-trained Transformer (GPT) model from scratch to solve sequential decision making tasks arising in contexts of operations research and management science which we call OMGPT. We first propose a general sequence modeling framework to cover several operational decision making tasks as special cases, such as dynamic pricing, inventory management, resource allocation, and queueing control. Under the framework, all these tasks can be viewed as a sequential prediction problem where the goal is to predict the optimal future action given all the historical information. Then we train a transformer-based neural network model (OMGPT) as a natural and powerful architecture for sequential modeling. This marks a paradigm shift compared to the existing methods for these OR/OM tasks in that (i) the OMGPT model can take advantage of the huge amount of pre-trained data; (ii) when tackling these problems, OMGPT does not assume any analytical model structure and enables a direct and rich mapping from the history to the future actions. Either of these two aspects, to the best of our knowledge, is not achieved by any existing method. We establish a Bayesian perspective to theoretically understand the working mechanism of the OMGPT on these tasks, which relates its performance with the pre-training task diversity and the divergence between the testing task and pre-training tasks. Numerically, we observe a surprising performance of the proposed model across all the above tasks.
\end{abstract}

\section{Introduction}
\label{sec:intro}
We study the problem of \textit{sequential decision making}, where a decision maker interacts iteratively with an unknown environment. At each time period $t=1,\ldots,T$, the decision maker observes a context vector, makes a decision, and receives a reward and possibly other feedback. The decision maker's objective is to maximize cumulative reward. Examples of such problems include the network revenue management problem with unknown demands \citep{jasin2015performance}, the newsvendor problem with unknown demand distribution \citep{huh2009nonparametric}, and the dynamic pricing problem with unknown demand function \citep{den2015dynamic}.

A central challenge in these problems is balancing learning about the environment (exploration) and earning rewards (exploitation). This is often referred to as the exploration-exploitation trade-off. Most existing works tackle this problem through the \textit{online learning} paradigm, where they assume specific structures in the underlying problem to design decision algorithms or policies. These algorithms often come with theoretical performance guarantees and can be optimal under certain conditions and definitions. However, this approach has two limitations: (i) it relies on assumptions of the underlying problem, and more relaxed assumptions often require the design of more complex algorithms, and (ii) it cannot easily incorporate the decision maker’s prior knowledge about the environment into the decision making process.

 Alternatively, in this work, we propose a \textit{supervised learning} paradigm to solve sequential decision making problems, which leverages generative artificial intelligence (GenAI) techniques to train a deep learning model. We re-frame the decision making problem as a sequence modeling task, where the goal is to predict optimal actions based on historical data. Specifically, by utilizing the decision maker's prior knowledge, we can generate pre-training data and train a transformer/GPT-based neural network, named the \textit{Operational Management Generative Pre-trained Transformer} (OMGPT), to sequentially predict the next optimal action. The key contributions of this work are as follows: 
\begin{itemize} 
\item We propose a new supervised learning paradigm for sequential decision making problems using GenAI techniques. Unlike existing methods, our paradigm does not assume any specific problem structure and can incorporate the decision maker’s prior knowledge. Specifically, we design a GPT-based architecture, OMGPT, with a unique pre-training scheme for solving sequential decision making problems. 
\item In numerical experiments, OMGPT (i) consistently outperforms well-established benchmarks across multiple operational decision making tasks, (ii) demonstrates strong generalization ability when the tested environments differ from the decision maker's prior knowledge, (iii) handles non-stationary environments and multiple environment types, showcasing its strong capability, and (iv) can also provide valuable side information beyond predicting optimal actions, such as demand forecasting. 
\item Analytically, we study the behavior of OMGPT as a decision algorithm, showing that (i) asymptotically, it behaves as a Bayes-optimal decision maker, (ii) the pre-training loss functions can serve as surrogates for regrets, which are defined as the reward gaps between the OMGPT and the optimal decisions, and (iii) prediction errors due to finite pre-training samples can provide an inherent exploration mechanism, leading to performance guarantees. 
\end{itemize}

The remainder of the paper is organized as follows. Section \ref{sec:literature} presents a brief review of the literature; Section \ref{sec:setup} gives the formal definition of the problem, along with performance measures and examples; Section \ref{sec:OMGPT} proposes the supervised learning paradigm using OMGPT, introducing its architecture, pre-training, and testing. Section \ref{sec:analysis} analyzes pre-trained OMGPT as a decision algorithm, focusing on its working mechanism and performance analysis. Finally, Section \ref{sec:numerical} provides numerical experiments to evaluate its performance and capability with interpretation results, and Section \ref{sec:conclusion} concludes the paper.

\section{Literature Review}
\label{sec:literature}

\subsection{Sequential Decision Making}
\label{sec:seq_dec_mak}

Sequential decision making encompasses many well-studied problems in operations management, including revenue management \citep{talluri2006theory, gallego2019revenue}, dynamic pricing \citep{den2015dynamic, keskin2014dynamic}, inventory management \citep{axsater2015inventory}, queuing control \citep{chen2001fundamentals}, etc. Below, we briefly review two of them as examples:
\begin{itemize}
    \item \textbf{Dynamic pricing}  involves setting prices to discern the underlying revenue function and make revenues, typically defined as the product of demand and price. Algorithms that rely purely on the exploitation of historical data tend to be suboptimal, and they may converge to non-optimal prices with positive probability \citep{keskin2014dynamic}. Therefore, random exploration is crucial to more accurately understand the demand function and to avoid settling on suboptimal prices. Various algorithms address this exploration-exploitation trade-off to achieve near-optimal regret under different settings. See \cite{harrison2012bayesian, den2014simultaneously, besbes2015surprising, cohen2020feature, wang2021dynamic} for examples of such algorithms, and \cite{den2015dynamic} for a comprehensive overview of dynamic pricing.
    \item \textbf{Inventory management} (or the newsvendor problem) requires the decision maker to decide on the quantity of inventory to order to meet a random demand with an unknown distribution. There are various settings, such as perishable inventory, where the unsold inventory expires at the end of each period, and non-perishable inventory, where the unsold inventory can be carried to the next period.  Additionally, the achieved demand can be censored or uncensored. In an uncensored setting, the decision maker observes the actual demand in each period. In a censored setting, when demand exceeds inventory levels, unmet demand is not observed, leading to lost sales. In this case, the optimal decision policy must balance the exploration of observing the realized demand and the exploitation of reducing the cost, sometimes requiring higher inventory levels to gather more demand data, but not so high that exploration costs become excessive. Near-optimal policies with $O(\sqrt{T})$ regret bounds have been developed under various assumptions about lead times and demand models, as shown in works such as \cite{burnetas2000adaptive, godfrey2001adaptive, zhang2020closing, agrawal2022learning}.
\end{itemize}

\subsection{Deep Learning in Operations Management}

Another related line of literature is about using deep learning tools to address problems in operations management. For example, \cite{bentz2000neural, aouad2022representing} apply neural networks to study customer choice behavior and predict customer decisions, and \cite{wang2023neural} develop a neural network architecture to both predict customer choice and optimize assortment decisions. Additionally, \cite{qi2023practical, gijsbrechts2022can,oroojlooyjadid2022deep} develop deep learning frameworks for solving inventory control problems,  \cite{li2024deep} adapt deep reinforcement learning to dynamic assortment planning, \cite{meng2024reinforcement}  propose using reinforcement learning for choice-based network revenue management, and \cite{han2023deep} provide a theoretical framework for analyzing theoretical guarantees in terms of excess risk bounds in newsvendor problems. \cite{chen2022using} propose a general framework that leverages deep neural networks to solve data-driven stochastic optimization problems.

The main contribution of our proposed method to this body of literature is that our neural network is trained as an algorithm, meaning it does not require any data from the testing environment and can operate across different environments. In contrast, while the works mentioned above may also tackle sequential decision making problems like inventory management, they typically rely on collecting data from the testing environment and learning a policy that works only in that specific environment. In this sense, our model is more aligned with generative artificial intelligence: it can sequentially generate different (good) decisions for various (unknown) environments.

\subsection{Transformer for Reinforcement Learning}

The earliest efforts to apply transformers and supervised learning approaches to reinforcement learning can be found in the area of offline reinforcement learning \citep{janner2021offline, chen2021decision}. In this paradigm, offline reinforcement learning is effectively converted into a supervised learning problem through pre-training, where actions are predicted based on trajectories in the offline dataset. During the evaluating/testing phase, actions are generated by the pre-trained transformer, conditioned on the so-called ``return-to-go'', which is often chosen as the cumulative reward from ``good'' trajectories in the dataset. This approach allows the transformer to retrieve information from training data trajectories with similar return-to-go values and perform imitation learning from high-reward offline trajectories. Building on this idea, several works further examine the validity of return-to-go conditioning and explore its mechanisms \citep{emmons2021rvs, ajay2022conditional, brandfonbrener2022does, lin2023transformers}, and proposed alternative methods to improve or extend the approach to other settings \citep{laskin2022context, badrinath2024waypoint, zheng2022online, wu2024elastic}. 

The most closely related work to ours is \cite{lee2024supervised}, which also proposes utilizing transformers to predict optimal actions for reinforcement learning problems. However, our proposed paradigm differs in several key aspects, including the studied underlying tasks, the data generation/pre-training step, and the neural network architecture. These differences are tailored specifically to operational decision makings and provide us with the ability to analyze the working mechanisms and performance. We elaborate more discussions in Appendix \ref{appx:architecture}.

\section{Problem Setup}

\label{sec:setup}

\subsection{Sequential Decision Making}
In this section, we introduce a general formulation for sequential decision making problems. A decision maker makes decisions sequentially in an environment with a \textit{horizon} $T$. At each time $t=1,\ldots,T$, the decision maker first observes a \textit{context} vector $X_t\in\mathcal{X}$ (also known as covariates or side information), takes an \textit{action/decision} $a_t\in\mathcal{A}$, and then receives an \textit{observation} $O_t\in\mathcal{O}$, where $\mathcal{X}, \mathcal{A}$, and $\mathcal{O}$ denote the spaces of contexts, actions, and observations, respectively. Specifically, at each time $t=1,\ldots,T$, 
    
\begin{enumerate}
    \item The decision maker observes a context $X_t \in \mathcal{X}$.
    \item Based the historical information $H_{t}$ (including the context $X_t$) available at $t$, 
    $$H_{t}\coloneqq (X_1,a_1,O_1,\ldots,X_{t-1},a_{t-1},O_{t-1},X_t),$$
    the decision maker makes a (non-anticipatory) decision/action $a_t \in \mathcal{A}$ following an \textit{algorithm} $f$, i.e., $a_t=f(H_t)$.
    \item Next, the decision maker receives an observation $O_t\in\mathcal{O}$ as a random outcome of $X_t$ and $a_t$. 
\end{enumerate}
We remark that the algorithm $f$ is very general in that it takes the whole history which is of a varying length as its input. Also, unlike the development of many online learning algorithms the implementation and analysis of which require the horizon $T$ to be large, our model still works even when $T=1$, i.e., a static decision making problem.

\paragraph{Environment Dynamics.}\

Throughout the paper, we assume contexts $X_t$'s are independently and identically distributed (i.i.d.) across $t$ and observations $O_t$'s follow a conditional distribution given $X_t$ and $a_t$
\begin{equation}
    X_{t} \overset{\text{i.i.d.}}{\sim} \mathbb{P}_{\gamma}(\cdot), \ O_t \sim \mathbb{P}_{\gamma}(\cdot|X_t, a_t),
    \label{eqn:law_X}
\end{equation}
where the \textit{environment} parameter $\gamma$ encapsulates all the parameters for these two probability distributions. In particular, the parameter $\gamma$ is unknown to the decision maker.

At each time $t$, the decision maker collects a random \textit{reward} $R_{t}=R(X_t,a_t)$ which depends on $X_t$, $a_t,$ and also some possible exogenous randomness. The observation $O_t$ includes $R_t$ as its first coordinate and it may or may not include other observed information as well. The goal of the decision maker is to maximize the cumulative reward over the decision horizon
$$\sum_{t=1}^T R_t = \sum_{t=1}^T R(X_t,a_t).$$
We define the expected reward by $r(X_t,a_t)=\mathbb{E}[R(X_t,a_t)|X_t,a_t]$. Specifically, this reward function $r(\cdot,\cdot)$ is unknown to the decision maker and it also depends on the underlying environment $\gamma$.

\paragraph{Performance Metric.} \ 

The performance of the decision maker's algorithm $f$ is measured by the notion of \textit{regret}, which is defined by
\begin{equation}
\text{Regret}(f;\gamma) \coloneqq \mathbb{E}\left[\sum_{t=1}^T r(X_{t},a_t^*)-r(X_t,a_t)\right] 
\label{eqn:regret_def}
\end{equation}
where the action $a_t^*$ is the optimal action that maximizes the expected reward
\begin{equation}
a_t^* \coloneqq \argmax_{a\in\mathcal{A}}\  r(X_{t},a)
\label{eqn:act_opt}
\end{equation}
and the optimization assumes the knowledge of the underlying environment, i.e., the parameter $\gamma$. In the regret definition \eqref{eqn:regret_def}, the expectation is taken with respect to the underlying probability distribution \eqref{eqn:law_X} and possible randomness in the decision algorithm $f$. We use the arguments $f$ and $\gamma$ to reflect the dependency of the regret on the algorithm $f$ and the environment $\gamma.$ Also, we note that the benchmark is defined as a dynamic oracle where $a_t^*$ may be different over time.

\subsection{Examples}
Here we provide two well-studied sequential decision making problems as examples.
\begin{itemize}
%\item Stochastic multi-armed bandits: There is no context, i.e., $X_t=\text{null}$ for all $t.$ The action $a_t\in \mathcal{A}=\left\{1,2,\ldots,k\right\}$ denotes the index of the arm played at time $t$. The random reward $R(X_t,a_t)$ is generated following the distribution $P_{a_{t}}$ and the observation $O_t=R(X_t,a_t)$.  The parameter $\gamma$ encapsulates the distributions $P_{1},\ldots,P_k.$
%\item Linear bandits: There is no context, i.e., $X_t=\text{null}$ for all $t.$ The action $a_t\in \mathcal{A} \subset \mathbb{R}^{d_a}$ is selected from some pre-specified domain $\mathcal{A}$. The random reward $R(X_t, a_t)=w^\top a_t+\epsilon_t$ where $\epsilon_t$ is some zero-mean noise random variable and $w\in \mathbb{R}^{d_a}$ is a vector unknown to the decision maker. The observation $O_t=R(X_t, a_t)$ and the expected reward $r(X_t, a_t)= \mathbb{E}\left[R(X_t, a_t)|X_t,a_t\right]=w^\top a_t$. The parameter $\gamma$ encapsulates the vector $w$ and the noise distribution.
\item Dynamic pricing: The context vector $X_t$ describes the market-related information at time $t$. Upon the reveal of $X_t$, the decision maker takes the action $a_t\in \mathcal{A} \subset \mathbb{R}^{+}$ as the pricing decision. Then the decision maker observes the demand $D_t = d(X_t,a_t)+\epsilon_t$ where $\epsilon_t$ is some zero-mean noise random variable and $d(\cdot,\cdot)$ is a demand function unknown to the decision maker. The random reward $R(X_t,a_t)=D_t\cdot a_t$ is the revenue collected from the sales at time $t$, and the observation $O_t=(R_t,D_t)$. The expected reward $r(X_t, a_t)= \mathbb{E}\left[R(X_t, a_t)|X_t,a_t\right]=d(X_t,a_t)\cdot a_t.$ The parameter $\gamma$ governs the generation of $X_t$'s and also contains information about the demand function $d$ and the noise distribution. 
\item Newsvendor problem: As the dynamic pricing problem, the context vector $X_t$ describes the market-related information at time $t$. Upon its reveal, the decision maker takes the action $a_t$ which represents the number of inventory prepared for the sales at time $t$. Then the decision maker observes the demand $D_t = d(X_t)+\epsilon_t$ where $\epsilon_t$ is some noise random variable and $d(\cdot)$ is a demand function unknown to the decision maker. The random reward is the negative cost $R(X_t,a_t)=-h\cdot (a_t-D_t)^+ - l\cdot (D_t-a_t)^+$ where $(\cdot)^+$ is the positive-part function, $h$ is the left-over/overage cost, and $l$ is the lost-sale/underage cost. The  observation $O_t=(R_t,D_t)$ and the expected reward $r(X_t,a_t)=\mathbb{E}\left[R(X_t,a_t)|X_t,a_t\right]$ with the expectation taken with respect to $\epsilon_t$. The parameter $\gamma$ governs the generation of $X_t$'s and encodes the demand function $d$, the noise distribution, and the (possibly known) cost parameters $h$ and $l$.  

\end{itemize}

\begin{comment}
 \textbf{Remark:} Here we assume $X_t$ are i.i.d. across $t$ only for notation simplicity, which can be relaxed (and the definition of $a^*_t$ should be accordingly changed). For example, $X_t\sim \mathbb{P}_{\gamma}(\cdot|X_{t-1},a_{t-1})$, i.e., its distribution depends on $t-1$'s context and action.   
\end{comment}

\section{Supervised Learning for a (Good) Algorithm Using OMGPT}
\label{sec:OMGPT}

In the literature on sequential decision making problems, a common paradigm, often known as \textit{online or reinforcement learning}, is to assume some structure for the underlying environment $\gamma$ and/or the reward function $r$.  Based on these assumptions, decision algorithms are then accordingly \textit{designed}.  This means that different algorithms are crafted for different sets of assumptions. In contrast, this work proposes a paradigm for \textit{learning}  policies instead. In this section, we introduce the Operations Management Generative Pre-trained Transformer (OMGPT) as a method for solving sequential decision making problems.  It is important to note that OMGPT is not merely a model or an algorithm; rather, it represents a paradigm for potentially solving all sequential decision making problems using a supervised learning approach. Similar to a common supervised learning task like regression, this approach involves optimizing a family of GPT models to predict the optimal actions $a^*_t$ based on the historical data, and the optimization is based on minimizing the empirical training loss over the generated training samples.

\subsection{A (Good) Algorithm as a Sequence Function}
 Recall that at each time step $t$, the decision $a_t$ made by an algorithm $f$ is given by $a_t = f(H_t)$. Thus, an algorithm $f$ can be viewed as a \textit{sequence function} whose input is $H_t$, a sequence of any length. To illustrate this, we first consider the case where the environment $\gamma$ is known and given. Ideally, we aim to find a sequence function $f^*$ such that for any $t=1,\ldots,T$ and any possibly sampled $H_t$ as defined in Section \ref{sec:setup},  
$$f^*(H_t)=a^{*}_t,$$
i.e.,  $f^*$ can predict the optimal action $a^{*}_t$ based on any possible $H_t$ at each $t$. Under the assumption the environment $\gamma$ and thus the reward function $r$ is known, this can be satisfied by setting $f^*(H_t)=\argmax_{a\in\mathcal{A}} r(X_t,a)$ (recall $X_t$ is included in $H_t$).

More generally, we can formulate a sequence function optimization task aimed at predicting optimal actions. Given a candidate family $\mathcal{F}$ of sequence functions, our goal is to find a sequence function $\hat{f} \in \mathcal{F}$ that minimizes the divergence between the predicted action $\hat{f}(H_t)$ and the target (optimal) action $a_t^*$: 
\begin{equation}
    \hat{f}=\argmin_{f\in \mathcal{F}}  \sum_{t=1}^T \mathbb{E}_{(H_t,a^*_t)}\left[l(f(H_t),a^{*}_t)\right],
    \label{eqn:hat_f_single}
\end{equation}
where the expectations are with respect to $(H_t,a^*_t)$'s (recall $a^*_t$ depends on $X_t$ in $H_t$) and $l(a,a')$ is the loss function to measure the divergence between $a$ and $a'$: it should be non-negative and only be zero when $a=a'$. For example, the $L_2$ loss (i.e., $l(a,a')=\|a-a'\|_2$) or the absolute ($L_1$) loss (i.e., $l(a,a')=\|a-a'\|_1$).  

In an environment where the reward function is ``smooth'' with respect to the action---meaning that the single-period regret incurred at time $t$ decreases as the action $a_t$ approaches the optimal action $a^*_t$---a sequence function $\hat{f}$ which is optimal in minimizing the prediction loss for the optimal actions should also perform well in terms of regret.  In particular, if $f^*\in \mathcal{F}$, we have $\hat{f}=f^*$ and $\text{Regret}(f^*;\gamma)=0$.

However, in sequential decision making problems, the environment $\gamma$ is unknown, making the formulation in \eqref{eqn:hat_f_single}
 invalid.  To address this, one approach is to consider a distribution $\mathcal{P}_{\gamma}$ such that $\gamma \sim \mathcal{P}_{\gamma}$. Here, $\mathcal{P}_{\gamma}$  can be viewed as the decision maker's prior knowledge or belief about the environment. The existence of such prior knowledge is a typical assumption in operational sequential decision making problems (e.g., \cite{bensoussan2009bayesian,harrison2012bayesian,ferreira2018online}). In this case, we define the expected loss of the decision function $f$ under the environment specified by $\gamma$ as
 \begin{equation}
     L\left(f;\gamma\right) \coloneqq \sum_{t=1}^T \mathbb{E}_{(H_t^{(\gamma)},a^{(\gamma)*}_t)} \left[l(f(H^{(\gamma)}_t),a^{(\gamma)*}_t)\right]
     \label{eqn:test_loss}
 \end{equation}
 where $(H^{(\gamma)}_t,a^{(\gamma)*}_t)$'s are derived with respect to the environment $\gamma$ and $f$. Then we can formulate the sequence function optimization task as:
\begin{equation}
    \hat{f}=\argmin_{f\in \mathcal{F}} \mathbb{E}_{\gamma\sim \mathcal{P}_{\gamma}} \left[ L\left(f;\gamma\right) \right].
    \label{eqn:hat_f}
\end{equation}
 The redefined $\hat{f}$ should approximate the optimal actions for each possible environment  $\gamma\sim \mathcal{P}_{\gamma}$ (as much as possible) to become the minimizer of \eqref{eqn:hat_f}, thereby handling unknown environments. However, there are still several challenges for applying \eqref{eqn:hat_f} to find a (good) algorithm $\hat{f}$:

\begin{enumerate}
    \item  \textbf{Choosing $\mathcal{F}$}: We need $\mathcal{F}$ to include a sequence function $\hat{f}$ that can accurately predict the optimal actions for different and unknown $\gamma$'s. However, even with a known $\gamma$, mapping a high-dimensional and unstructured input $H_t$ approximately to $a^*_t$ is complex, requiring $\mathcal{F}$ to have a large capacity. Additionally, $\mathcal{F}$ must be a family of sequence functions, which can handle variable-length input sequences.
    \item \textbf{Solving \eqref{eqn:hat_f}}: There are multiple challenges in solving  \eqref{eqn:hat_f}. First, this stochastic optimization problem involves complex randomness in both the environment sampling $\gamma\sim \mathcal{P}_{\gamma}$ and the environment dynamics $H_t$'s. Second, the dynamics $H_t$'s themselves depend on the decision variable $f$, and this dependence is also complex. Third, optimizing over the complex function family $\mathcal{F}$ is difficult. These difficulties collectively make it challenging to even approximately solve \eqref{eqn:hat_f}.
\item \textbf{Understanding $\hat{f}$}: It is crucial to understand the algorithm we intend to use, particularly to avoid extreme decisions and to evaluate its performance. Even if we assume that \eqref{eqn:hat_f} can be solved, we still need to know what the resulting algorithm $\hat{f}$ will be and how it performs in terms of regret.
\end{enumerate}

In the following subsections, we will address these challenges.

\subsection{Preliminaries on (Original) GPTs and OMGPT}
In this work, we select the GPT architecture, which stands for Generative Pre-trained Transformer \citep{radford2018improving}, as the sequence function family $\mathcal{F}$ to learn a good algorithm. GPT, a variant of the transformer architecture for deep learning \citep{vaswani2017attention}, was originally designed as a language model due to its two key properties: high expressiveness and the ability to model sequences. Specifically, GPT can be efficiently scaled to billions of parameters, providing a high degree of expressiveness and the capacity to capture complex patterns, such as those in natural language. Additionally, its structure allows it to handle inputs of varying lengths, making it well-suited for sequence modeling. These properties align precisely with the requirements for the function family $\mathcal{F}$ in \eqref{eqn:hat_f}. 
Here, we provide a brief introduction to the original GPT architecture, followed by an overview of the OMGPT architecture, with detailed information deferred to Appendix \ref{appx:architecture}.

\textbf{Attention Mechanism.} The key component of the GPT architecture is the \textit{attention mechanism/function}. The input to an attention function is a set of embedding vectors $\{z_{\tau}\}_{\tau=1}^t$, all sharing the same dimension, and the output is a set of processed embedding vectors with the same dimension, denoted by $\{\tilde{z}_{\tau}\}_{\tau=1}^t$. Let $Z=[z_1,z_2,\ldots,z_t]^{\top}$ be the matrix composed of the embedding vectors $\{z_{\tau}\}_{\tau=1}^t$. The attention function is then defined as
\begin{equation*}
f^{\text{Att}}(Z;W_Q,W_K,W_V) \coloneqq \text{Softmax}\left(Z^\top W_Q W_K Z \right)Z^\top W_V.
\label{eqn:self_att}
\end{equation*}
Here, the Softmax function is applied to each row of the matrix $Z^\top W_Q W_K Z$, and $ W_Q, W_K, W_V$  are trainable square parameter  matrices, commonly referred to as the query weight, key weight, and value weight, respectively, which have the same dimensions as the embedding vectors $z_{\tau}$'s.

Intuitively, the attention mechanism connects each element from the input sequence using the Softmax function and the weight matrices, allowing each output vector to embed information from the entire sequence. Additionally, the attention mechanism supports variable-length inputs.

\textbf{GPT architecture.} The GPT architecture is built with multiple layers that share the same structure. Each layer consists of an attention function $f^{\text{Att}}$ followed by an element-wise, fully connected feed-forward network sub-layer. Each embedding vector $\tilde{z}_{\tau}$ from the attention function  ,for $\tau=1,\ldots,t$, is individually processed through this sub-layer, which uses trainable matrices $W_1, W_2$ and vectors $b_1, b_2$, all matching the input dimension of $\tilde{z}_{\tau}$:
    \begin{equation}
        z'_{\tau}=W_2 \phi (W_1\tilde{z}_{\tau}+b_1)+b_2,
        \label{eqn:linear_layer}
    \end{equation}
    where $\phi$ is an activation function to add the non-linearity (like ReLU or GELU \citep{hendrycks2016gaussian}) and the outputs  $z'_{\tau}$'s form the input sequence for the next layer. By stacking multiple layers, GPT gains the capacity to ``understand'' more complex relationships within the data.

The GPT architecture also includes a layer at the beginning to transform input elements (e.g., words/tokens in natural language or actions in sequential decision making tasks) into embedding vectors, and an output layer at the end to convert these embedding vectors (typically the last one) into the desired output elements (e.g., tokens or actions). These layers can be structured as linear layers similar to \eqref{eqn:linear_layer}, with dimensions of trainable parameters adjusted to match the inputs and outputs.

\textbf{OMGPT.} Unlike the original GPT, which is designed for natural language modeling with a single type of input and output (words/tokens), OMGPT is tailored for sequential decision making tasks and must handle different types of elements as input (actions, contexts, observations) and output only actions. Therefore, modifications to the OMGPT architecture are required to accommodate these different elements. Details on these modifications are provided in Appendix \ref{appx:architecture}. Additionally, other neural network architectures, such as Long Short-Term Memory (LSTM) \citep{hochreiter1997long}, can also be applied as $\mathcal{F}$, as they are also capable of sequence modeling and can handle inputs of varying lengths. In Appendix \ref{appx:LSTM}, we evaluate the performance of OMGPT by replacing the transformer with LSTM, and the results demonstrate that the transformer performs better. This finding is consistent with observations from the literature on natural language modeling/large language models \citep{zhao2023survey}.

\subsection{Supervised Learning for OMGPT}
\label{sec:supervised}
We denote the OMGPT architecture by $\texttt{TF}_{\theta}$, where $\theta$ encapsulates all the trainable parameters (we assume that the hyperparameters, such as the number of layers and the dimension of the embedding space, are given and not included in $\theta$). We assume there exists a parameter space $\Theta$ for $\theta$ and set $\mathcal{F}=\{\texttt{TF}_{\theta}\}_{\theta\in \Theta}$. The sequence function learning task \eqref{eqn:hat_f} for the OMGPT architecture can then be reformulated as:
\begin{equation}
    \min_{\theta\in \Theta } \mathbb{E}_{\gamma\sim \mathcal{P}_{\gamma}} \left[L\left(\texttt{TF}_{\theta};\gamma\right)\right].
\label{eqn:hat_TF}
\end{equation}

As mentioned earlier, this stochastic optimization problem is challenging to solve (even approximately) due to the complex randomness in $\gamma$ and $H_t$,  as well as the high dimensionality of the parameter space $\Theta$, which can consist of millions or even billions of dimensions. To address this, we can adopt a supervised pre-training approach, similar to the training of large language models (e.g., ChatGPT), to learn a parameter $\hat{\theta}$ that serves as an approximate solution to \eqref{eqn:hat_TF}.

\subsubsection{Pre-training Data Generation}  
The supervised pre-training first constructs a pre-training dataset
$$\mathcal{D}_{\text{PT}} \coloneqq \left\{\left(H_1^{(\gamma_i)},a_1^{(\gamma_i)*}\right),\left(H_2^{(\gamma_i)},a_2^{(\gamma_i)*}\right),\ldots,\left(H_T^{(\gamma_i)},a_T^{(\gamma_i)*}\right)\right\}_{i=1}^n.$$
In particular, we assume the parameter $\gamma$ (see \eqref{eqn:law_X}) that governs the generation of $X_t$ and $O_t$ are generated from the environment distribution $\mathcal{P}_{\gamma}$. Then, we generate $n$ environments denoted by $\gamma_1,\ldots,\gamma_n.$ There are two key aspects regarding the generation of $\mathcal{D}_{\text{PT}}$:
\begin{itemize}
    \item \textbf{Target action $a_t^{(\gamma_i)*}$}: For this paper, we generate $a_t^{(\gamma_i)*}$ as the optimal action specified by \eqref{eqn:act_opt}, which requires knowledge of the underlying environment $\gamma_i.$ A common property of all these sequential decision making problems is that the optimal action can be often easily or analytically computed from the generated $\gamma_i$'s. Alternatively, $a_t^{(\gamma_i)*}$ can also be generated based on expert algorithms such as Thompson sampling or upper confidence bound (UCB) algorithms \citep{lattimore2020bandit,wang2021dynamic}. In this case, the action $a_t^{(\gamma_i)*}=\texttt{Alg}(H_t^{(\gamma_i)})$ where $\texttt{Alg}(\cdot)$ represents the algorithm that maps from $H_t^{(\gamma_i)}$ to the action without utilizing the knowledge of $\gamma_i$. These target actions can be viewed as  ``labels'' to predict in the supervised learning task.
    \item \textbf{History} $H_{t}^{(\gamma_i)}=\left(X_1^{(\gamma_i)},a_1^{(\gamma_i)},O_1^{(\gamma_i)},\ldots,X_{t-1}^{(\gamma_i)},a_{t-1}^{(\gamma_i)},O_{t-1}^{(\gamma_i)},X_t^{(\gamma_i)}\right)$: Here, $X_{\tau}^{(\gamma_i)}$'s and $O_{\tau}^{(\gamma_i)}$'s are generated based on the parameter $\gamma_i$ following \eqref{eqn:law_X}. The action $a_{\tau}^{(\gamma_i)}$ is generated based on some pre-specified decision function $\tilde{f}$ in a recursive manner:
\begin{equation}
a_t^{(\gamma_i)} = \tilde{f}\left(H_{t}^{(\gamma_i)}\right), \ \ H_{t+1}^{(\gamma_i)}=\left(H_{t}^{(\gamma_i)}, a_t^{(\gamma_i)}, O_t^{(\gamma_i)}, X_{t+1}^{(\gamma_i)}\right)
\label{eqn:g}
\end{equation}
for $t=1,\ldots,T.$  It is important to note that  $a_t^{(\gamma_i)}$  can differ from the target action $a_t^{(\gamma_i)*}$. The pre-specified decision function $\tilde{f}$ can be chosen to randomly select an action or as another expert algorithm. In general, the history $H_{t}^{(\gamma_i)}$ generated by $\tilde{f}$ should be representative of the histories encountered during the testing phase where we will deploy the learned/pre-trained OMGPT. Intuitively, the pre-training data should resemble the testing situations to ensure that the pre-trained OMGPT generalizes well during testing. These histories can be viewed as  ``features'' in the supervised learning task. Further discussion on the choice of $\tilde{f}$ and its influence on pre-training $\texttt{TF}_{\theta}$ is provided in Appendix \ref{sec:algo}. 
\end{itemize}

Formally, we define a distribution $\mathcal{P}_{\gamma,\tilde{f}}$ that generates $H_t$ and $a_t^*$ as
\begin{equation}
X_\tau\sim \mathbb{P}_{\gamma}(\cdot), \ a_\tau=\tilde{f}(H_\tau), \ O_\tau\sim \mathbb{P}_{\gamma}\left(\cdot|X_\tau, a_\tau\right),\ H_{\tau+1}=\left(H_{\tau}, a_\tau, O_\tau, X_{\tau+1}\right) 
\label{eqn:P_ga_g}
\end{equation}
for $\tau=1,\ldots,t-1$ and $a_t^*$ is specified by \eqref{eqn:act_opt}.
The parameter $\gamma$ and the function $\tilde{f}$ jointly parameterize the distribution $\mathcal{P}_{\gamma,\tilde{f}}$. The parameter $\gamma$ governs the generation of $X_t$ and $O_t$ just as in \eqref{eqn:law_X}. The decision function $\tilde{f}$  maps from the history to the action and it governs the generation of $a_\tau$'s in the sequence $H_t$. In this way, the pre-training data is generated with the following flow:
$$\mathcal{P}_{\gamma} \rightarrow \gamma_i \rightarrow \mathcal{P}_{\gamma_i,\tilde{f}} \rightarrow \left\{\left(H_1^{(\gamma_i)},a_1^{(\gamma_i)*}\right),\ldots,\left(H_T^{(\gamma_i)},a_T^{(\gamma_i)*}\right)\right\}.$$

\subsubsection{Supervised Pre-training} 
\begin{figure}[ht!]
    \centering
    \includegraphics[width=0.7\linewidth]{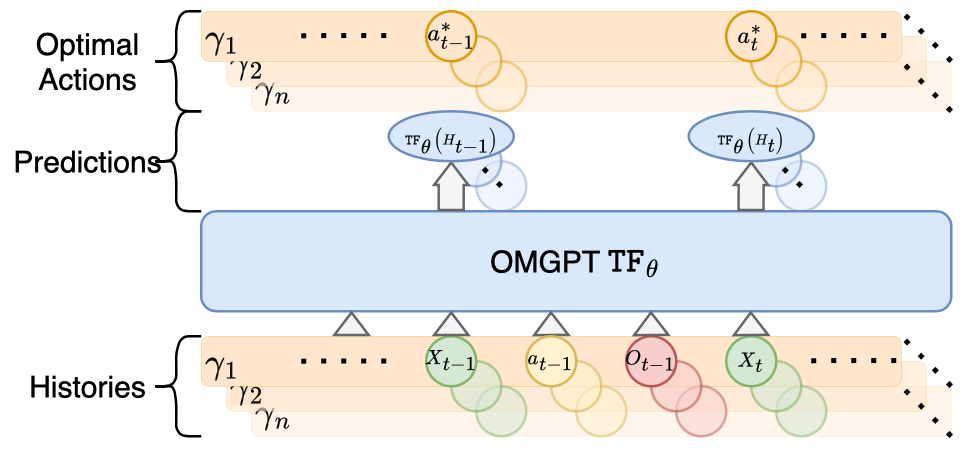}
    \caption{The pre-training phase of OMGPT. We first generate different environments $\gamma_i$. For each $\gamma_i$, we generate the corresponding histories and optimal actions. The histories serve as the inputs to the OMGPT, which then uses them to predict the corresponding optimal actions. The goal of the pre-training is to minimize the empirical prediction errors.}
    \label{fig:pretrain}
\end{figure}
Figure \ref{fig:pretrain} illustrates the supervised pre-training phase of OMGPT.  Based on the dataset $\mathcal{D}_{\text{PT}}$, the supervised pre-training refers to the learning of the sequence function $\texttt{TF}_{\hat{\theta}}$ through minimizing the following empirical loss
\begin{equation}
\hat{\theta}  \coloneqq  \argmin_{\theta\in\Theta} \frac{1}{nT} \sum_{i=1}^n \sum_{t=1}^T l\left(\texttt{TF}_{\theta}\left(H_t^{(\gamma_i)}\right),a_t^{(\gamma_i)*}\right) 
\label{eqn:erm_theta}
\end{equation}
where the loss function $l(\cdot, \cdot):\mathcal{A}\times \mathcal{A}\rightarrow \mathbb{R}$ specifies the prediction loss. Thus the key idea of supervised pre-training is to formulate the sequential decision making task as a \textit{sequence modeling} task that predicts the optimal action $a_{t}^{(\gamma_i)*}$ given the history $H_t^{(\gamma_i)}$.  Further discussions on how different loss function $l$ influences $\texttt{TF}_{\hat{\theta}}$ and how the pre-training loss relates to the regret are deferred to Section  \ref{sec:analysis}.

\subsubsection{Testing}
\begin{figure}[ht!]
    \centering
    \includegraphics[width=0.7\linewidth]{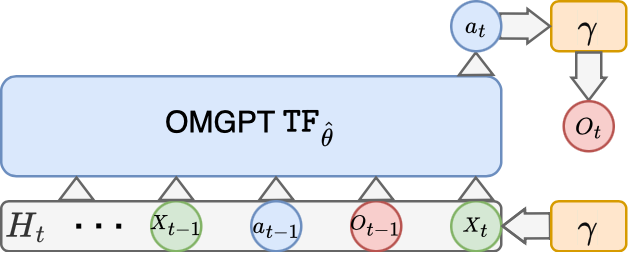}
    \caption{The testing/application phase of the pre-trained OMGPT. we iteratively input the history $H_t$ into the pre-trained OMGPT $\texttt{TF}_{\hat{\theta}}$ to obtain an action $a_t$. This action is then applied to interact with the testing environment $\gamma$, generating a new observation $O_t$. The cycle is then repeated with the updated history $H_{t+1}$.}
    \label{fig:testing}
\end{figure}

Figure \ref{fig:testing} illustrates the testing (or application) phase of a pre-trained OMGPT.  In the testing phase, a new environment $\gamma$ (which is unknown) is sampled from $\mathcal{P}_{\gamma}$ and we apply the learned decision function $\texttt{TF}_{\hat{\theta}}$. The testing procedure is described by the following dynamics:
\begin{equation}
    X_\tau\sim \mathbb{P}_{\gamma}(\cdot), \ a_\tau=\texttt{TF}_{{\hat{\theta}}}(H_\tau), \ O_\tau\sim \mathbb{P}_{\gamma}\left(\cdot|X_\tau, a_\tau\right),\ H_{\tau+1}=\left(H_{\tau}, a_\tau, O_\tau, X_{\tau+1}\right).
    \label{eqn:test_dynamics}
\end{equation}
The only difference between the test dynamics \eqref{eqn:test_dynamics} and the training dynamics \eqref{eqn:P_ga_g} is that the action $a_t$ (or $a_{\tau}$) is generated by $\texttt{TF}_{{\hat{\theta}}}$ instead of the pre-specified decision function $\tilde{f}$. 

\subsection{Difference from Online/Reinforcement Learning}
In this subsection, we discuss the key differences between the proposed OMGPT paradigm and traditional online/reinforcement learning approaches for sequential decision makings in the following aspects:

\begin{itemize}
\item \textbf{Learning-based Paradigm:}  OMGPT represents a learning-based approach, while most online/reinforcement learning algorithms are manually designed. Through a supervised pre-training phase, OMGPT learns an algorithm, $\texttt{TF}_{\hat{\theta}}$, that can function like a decision algorithm when interacting with an unknown environment, and can even achieve better performances than well-established online algorithms, as demonstrated in Section \ref{sec:numerical}. This means OMGPT discovers new algorithms during pre-training and has the potential to offer valuable insights and guidance for the design of future online learning algorithms.
    \item \textbf{Utilization of Prior Knowledge and Pre-training Data:} Many online/reinforcement learning algorithms rely on setting hyperparameters and can only utilize limited prior knowledge of the testing environments, such as parameter bounds. They only learn the optimal actions through the data collected from the testing task. This restricts their ability to incorporate more detailed prior knowledge, such as a prior distribution of the testing environment. While some algorithms, like Thompson Sampling, can use a prior distribution, embedding more complex prior knowledge (e.g., demand functions can change unpredictably during the horizon) can be challenging. OMGPT, however, can handle such complex prior knowledge by generating pre-training samples based on simulations. This ensures that the pre-trained $\texttt{TF}_{\hat{\theta}}$ can effectively leverage the prior knowledge when making decisions based on the data collected from the testing task. Numerical results demonstrating this advantage are provided in Appendix \ref{appx:prior_benefit}.
    \item  \textbf{Flexibility:}  Unlike online algorithms designing, which are based on specific assumptions and often require redesign when those assumptions are not met, OMGPT serves as a universal paradigm. Regardless of the assumptions or prior knowledge at hand, the same framework can be applied to pre-train an OMGPT model and utilize it effectively, thanks to its large capacity. This universality means OMGPT can work across various problems without relying on predefined analytical model structures or assumptions. Our numerical experiments in Section \ref{sec:numerical} demonstrate its strong versatility across various problems and structures.
\end{itemize}

\subsection{Discussions and Extensions} 

We provide the following discussions and extensions to the proposed supervised learning paradigm.

    \subsubsection{Extending to Longer Testing Horizon}
\label{sec:disc_longer_horizon}
Our paradigm can be generalized to a longer testing horizon (beyond the length of the pre-training data) by introducing a \textit{context window}. Specifically, we define a context window size $W$ and limit the input to the last $\min\{W, t\}$ timesteps of the input sequence $H_t$ to predict $a^*_t$ during both the pre-training and testing phases. By considering only the latest $\min\{W, t\}$ timesteps, we enable the model to handle testing sequences of any length, even when $t > T$. This technique is commonly used in the literature (e.g., \citet{chen2021decision}).

In general, a larger context window provides more information for predicting the optimal action but requires more training data, longer training time, and potentially a larger model size. We provide numerical performance results in Section \ref{sec:numerical}, and believe that its theoretical analysis (e.g., determining the optimal window size for a random testing horizon) is an important and interesting future research direction.

\subsubsection{Extending to State-dependent Context}
\label{sec:disc_state_dependent}
    
Although in Section \ref{sec:setup} we assume that the contexts $X_t$'s are i.i.d. generated, the OMGPT paradigm can be extended to handle Markov Decision Processes (MDPs), where $X_t$ encapsulates the state of the decision process and is sampled through a Markov process:
    $$X_t\sim \mathbb{P}(\cdot|X_{t-1},a_t).$$ 
For this extension, no changes to the paradigm are required to accommodate such non-i.i.d. $X_t$'s, and we can follow the same procedure to generate pre-training data, as well as to conduct pre-training and testing. In Section \ref{sec:numerical}, we provide numerical experiments that demonstrate the performance of OMGPT in common operational decision making problems involving states, including queuing control and revenue management.

\subsubsection{Extending to Observation Prediction}
\label{sec:dics_obs_predict}
In addition to predicting optimal actions, OMGPT can be pre-trained to also predict observations, $O_t$, with some modifications. These predictions can be valuable for decision makers in certain scenarios. For example, in a dynamic pricing task, while the main target is to find the optimal prices, the decision maker may also need to predict the resulting demands to plan manufacturing or storage.

To enable it, an additional output layer is added to OMGPT to predict observations, while the original output layer continues to predict optimal actions as usual. Specifically, with this added output layer, OMGPT becomes a sequence function that maps a sequence $H$ to both an action $a \in \mathcal{A}$ and an observation $O \in \mathcal{O}$, i.e., $\texttt{TF}_{\theta}(H) = (a, O)$.

When predicting observations, the input sequence is modified to $H_t \cup \{a_t\}$, incorporating the taken action to provide the necessary information for the observation prediction, rather than just $H_t$ as used for predicting optimal actions. Additionally, during pre-training, the training loss function must account for the prediction error on observations:

\begin{equation}
\hat{\theta}  \coloneqq  \argmin_{\theta\in\Theta} \frac{1}{nT} \sum_{i=1}^n \sum_{t=1}^T l\left(\hat{a}_t^{(\gamma_i)*},a_t^{(\gamma_i)*}\right)+l'\left(\hat{O}_t^{(\gamma_i)},O_t^{(\gamma_i)}\right).
\label{eqn:obs_predict}
\end{equation}
Here, $l'(\cdot, \cdot)$ is a loss function that measures the prediction error of observations. As defined earlier, $\hat{a}_t^{(\gamma_i)*} = \texttt{TF}_{\theta,1}(H_t^{(\gamma_i)})$ is the predicted action, and $\hat{O}_t^{(\gamma_i)} = \texttt{TF}_{\theta,2}(H_t^{(\gamma_i)} \cup {a_t^{(\gamma_i)}})$ is the predicted observation, where the subscripts 1 and 2 in $\texttt{TF}$ indicate the first and second outputs, respectively.

Compared to the original loss function in \eqref{eqn:erm_theta}, which only accounts for predicting optimal actions, the revised loss function in \eqref{eqn:obs_predict} includes an additional term to capture the prediction error on observations, which ensures that the pre-trained OMGPT can effectively predict the observations through the pre-training.

\section{Understanding Pre-trained $\texttt{TF}_{\hat{\theta}}$ as an Algorithm}
\label{sec:analysis}
In the previous section, we discuss using a supervised learning paradigm to learn a pre-trained  $\texttt{TF}_{\hat{\theta}}$ as an algorithm. Now,  we return to the perspective of sequential decision making and analyze the properties and performance of $\texttt{TF}_{\hat{\theta}}$ when being deployed as a decision algorithm. The proofs in this section are deferred to Appendix \ref{appx:proofs}.

\subsection{$\texttt{TF}_{\hat{\theta}}$ as a Bayes-optimal Decision Maker}
\label{sec:Bayes_decision_maker}
For a given distribution $\mathcal{P}_{\gamma}$, we define the \textit{Bayes-optimal decision function} $\texttt{Alg}^*$ as follows: 
\begin{align}
\texttt{Alg}^*(H) &\coloneqq \argmin_{a\in\mathcal{A}} \mathbb{E}_{\gamma}\left[l(a, a^{(\gamma)*}_t) | H\right]   \label{eqn: alg_star} \\
&= \argmin_{a\in\mathcal{A}} \int_{\gamma} l(a, a^{(\gamma)*}_t) \mathcal{P}(\gamma|H) \mathrm{d} \gamma \nonumber\\
& = \argmin_{a\in\mathcal{A}} \int_{\gamma} l(a, a^{(\gamma)*}_t) \mathbb{P}(H|\gamma) \mathrm{d}\mathcal{P_{\gamma}} \nonumber
\end{align}
for any history $H\in \mathcal{H}$, where
$$\mathcal{H}=\left\{ (X_1,a_1,O_1,\ldots,X_{t-1},a_{t-1},O_{t-1},X_{t})| X_{t} \in \mathcal{X}, a_{t} \in \mathcal{A}, O_{t} \in \mathcal{O}, t=1,\ldots, T \right\},$$
is the space of all possible histories.

In the definition of $\texttt{Alg}^*$, recall $\mathcal{P}_{\gamma}$ is the environment distribution that generates the environment $\gamma$, and the optimal action $a^{(\gamma)*}_t$ depends on both the history $H$ (or more precisely, $X_t$) and the environment $\gamma$ (as defined in \eqref{eqn:act_opt}). $\mathbb{P}(H|\gamma)$ is proportional to the likelihood of observing the history $H$ under the environment $\gamma$
$$\mathbb{P}(H|\gamma) \propto \prod_{\tau=1}^{t} \mathbb{P}_\gamma(X_{\tau})\cdot \prod_{\tau=1}^{t-1}\mathbb{P}_\gamma(O_{\tau}|X_{\tau},a_{\tau})$$
where $H=(X_1,a_1,O_1,\ldots,X_{t-1},a_{t-1},O_{t-1},X_t)$ for some $t\in\{1,\ldots,T\}$ and the distributions $\mathbb{P}_\gamma(\cdot)$ and $\mathbb{P}_\gamma(\cdot|X_{\tau},a_{\tau})$ are as in \eqref{eqn:law_X}. In the last line of definition, the integration is with respect to the environment distribution $\mathcal{P}_{\gamma}$, and each possible environment $\gamma$ is weighted with the likelihood $\mathbb{P}(H|\gamma)$.

We refer to $\texttt{Alg}^*$ as the Bayes-optimal because its definition follows Bayes' law (as seen in the last line of \eqref{eqn: alg_star}). And the following proposition, which relates $\texttt{Alg}^*$ with the pre-training objective, further justifies why it is called the optimal decision function.  Recall that the decision function $\tilde{f}$ is used to generate the actions in $H_t$.

\begin{proposition}
The following holds for any decision function $\tilde{f}$:
$$\texttt{Alg}^*(\cdot) \in \argmin_{f\in\mathcal{F}} \mathbb{E}_{\gamma\sim\mathcal{P}_\gamma}\left[\sum_{t=1}^T \mathbb{E}_{(H_t,a_t^{*})\sim\mathcal{P}_{\gamma,\tilde{f}}} \left[ l\left(f \left(H_t\right),a_t^*\right)\right]\right]$$
where $\mathcal{F}$ is the family of all measurable functions (on a properly defined space that handles variable-length inputs). 
\label{prop:BO}
\end{proposition}

\begin{figure}[ht!]
\centering
  \begin{subfigure}[b]{0.47\textwidth}
    \centering
    \includegraphics[width=\textwidth]{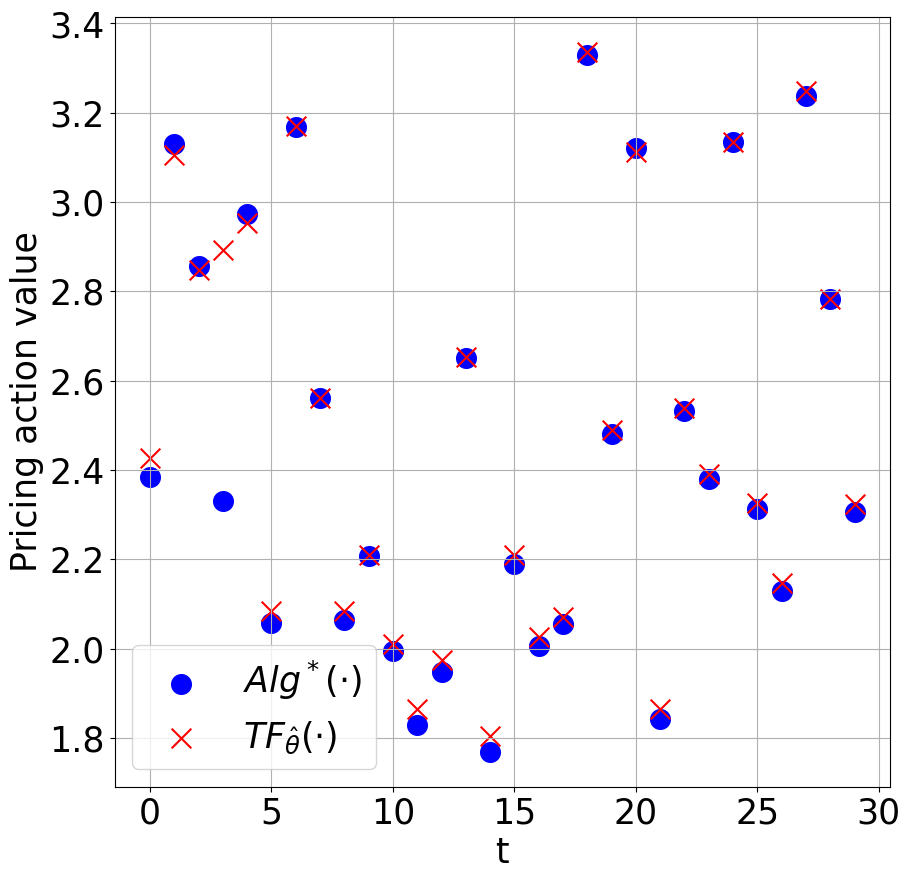}
    \caption{Sample Path}
    \label{fig:Bayes_DP}
  \end{subfigure} \hfill
  \begin{subfigure}[b]{0.47\textwidth}
    \centering
    \includegraphics[width=\textwidth]{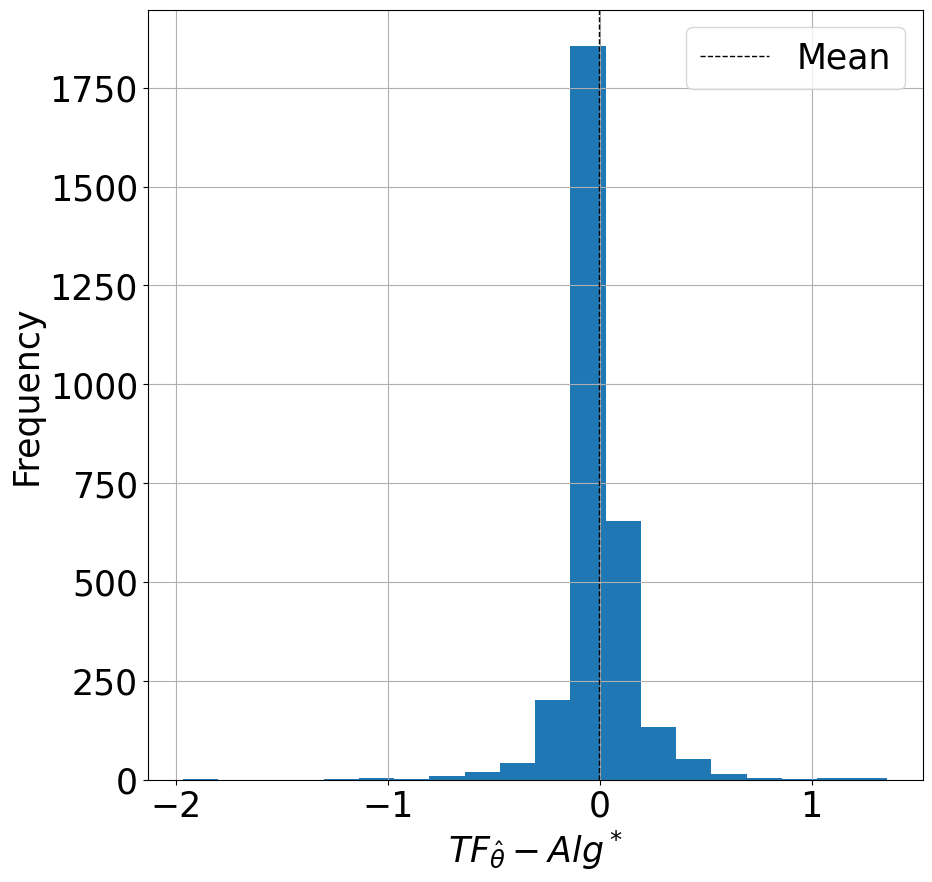}
    \caption{Histogram}
    \label{fig:Bayes_DP_diff}
  \end{subfigure}
  \caption{\small  $\texttt{TF}_{\hat{\theta}}$ nearly matches the optimal decision function $\texttt{Alg}^*$ in dynamic pricing tasks with concentrated deviations. Figure (a) shows one decision trial for $\texttt{Alg}^*$ and $\texttt{TF}_{\hat{\theta}}$ ,where the optimal actions change over time because $X_t$'s are different for different time $t$. Figure (b) shows the histogram of $\texttt{TF}_{\hat{\theta}}-\texttt{Alg}^*$ over different environments. The experiment setup and more results are deferred to Appendix \ref{appx:bayes_match_exp}. }
  \label{fig:Bayes_match}
\end{figure}

Proposition \ref{prop:BO} states that the function $\texttt{Alg}^*$ is one minimizer of the expected loss under any decision function $\tilde{f}$ (given $\mathcal{P}_{\gamma}$). To see this, $\texttt{Alg}^*$ is defined in a pointwise manner for every possible $H.$ The probability $\mathcal{P}_{\gamma,\tilde{f}}$ defines a distribution over the space of $H$. The pointwise optimality of $\texttt{Alg}^*$ naturally leads to its optimality for any decision function $\tilde{f}$. Furthermore, this implies that $\texttt{Alg}^*$ is also the optimal solution to the expected training loss (i.e., the expectation of the objective function of \eqref{eqn:erm_theta}). This means that if the OMGPT class $\{\texttt{TF}_{\theta}\}_{\theta \in \Theta}$ is sufficiently rich to cover $\texttt{Alg}^*$ (for instance, by increasing the number of layers and embedding dimensions), and with an infinite amount of pre-training data, the supervised pre-training will result in $\texttt{Alg}^*$ at best.  Thus, we can interpret the property of the supervised pre-trained $\texttt{TF}_{\hat{\theta}}$ by $\texttt{Alg}^*$ as a proxy. Figure \ref{fig:Bayes_match} shows an example of that $\texttt{Alg}^*$ and $\texttt{TF}_{\hat{\theta}}$ are closely matched.  This notion of a Bayes-optimal function and the Bayesian perspective has been explored in various settings within the in-context learning literature for large language models and GPTs  \citep{xie2021explanation,zhang2023and, jeon2024information}.

In addition, the corollary below characterizes the exact behavior of $\texttt{Alg}^*$ (and thus approximately of  $\texttt{TF}_{\hat{\theta}}$) under different loss functions $l$ in \eqref{eqn: alg_star}. This corollary can serve as a guide when pre-training OMGPT to target one specific algorithm.

\begin{corollary}
    $\texttt{Alg}^*$ behaves as 
    \begin{enumerate}
        \item[(a)] posterior sampling under the cross-entropy loss;
        \item[(b)] posterior averaging under the squared loss; 
        \item[(c)] posterior median under the absolute loss;
    \end{enumerate}
where the posterior distribution is with respect to $\gamma$ and it is defined by $\mathcal{P}(\gamma|H)\propto \mathbb{P}(H|\gamma)\cdot\mathcal{P}_{\gamma}(\gamma)$. Detailed definitions of posterior sampling, averaging, and median are provided in Appendix \ref{appx:posterior_oracle}.
\label{cor:post}
\end{corollary}

\subsection{Regret Analysis}

The previous section analyzes the behavior of $\texttt{TF}_{\hat{\theta}}$. In this subsection, we study its performance in terms of regret, as defined in \eqref{eqn:regret_def}. 

\subsubsection{Surrogate Property of the Prediction Error}  We begin with the following proposition, which establishes a relationship between the prediction error and the regret of any decision function $f$ (where $f=\texttt{TF}_{\hat{\theta}}$ is not necessarily required). Recall we define 
$$L\left(f;\gamma\right)=\sum_{t=1}^T \mathbb{E}_{(H_t^{(\gamma)},a^{(\gamma)*}_t)} \left[l(f(H^{(\gamma)}_t),a^{(\gamma)*}_t)\right]$$
as the testing prediction error of $f$  for a specific environment $\gamma$.

\begin{proposition}[Surrogate property] We say the loss function $l(\cdot, \cdot)$ satisfies the \textit{surrogate property} if there exists a constant $C>0$ such that 
$$\mathrm{Regret}(f;\gamma)
\le C \cdot L\left(f;\gamma\right)$$
holds for any sequence function $f$ and $\gamma$. Then under mild conditions (see Appendix \ref{appx:proofs}), We have
\begin{enumerate}
\item[(a)] The cross-entropy loss satisfies the surrogate property for the multi-armed bandits problem.
\item[(b)] The squared loss satisfies the surrogate property for  the dynamic pricing problem.
\item[(c)] The absolute loss satisfies the surrogate property for  the linear bandits and the newsvendor problem.
\end{enumerate}
\label{prop:surro}
\end{proposition}

We provide the definitions of these problems in Appendix \ref{appx:envs}. Proposition \ref{prop:surro} gives a first validity of the supervised pre-training approach. The surrogate property is an attribute of the loss function and is independent of the distribution $\mathcal{P}_{\gamma}$ or the decision function $f.$ Specifically, it establishes an upper bound on the regret in terms of the action prediction loss. This result is intuitive, as a more accurate prediction of the optimal action naturally leads to lower regret. One key implication is that this result guides the selection of the appropriate loss function for different underlying problems.

 %Part (a) has been directly employed as Assumption 1 in \cite{lee2024supervised} for their theoretical developments. Indeed, part (a) corresponds to a discrete action space such as multi-armed bandits while part (b) and part (c) correspond to the continuous action space such as linear bandits, pricing, and newsvendor problems. 

\subsubsection{Linear Regret of $\texttt{Alg}^*$} 

Although Proposition \ref{prop:surro} provides an upper bound on the regret using the action prediction error for each individual environment $\gamma$, it does not imply that the minimizer $\texttt{Alg}^*$ of the expected prediction error with respect to some distribution $\mathcal{P}_{\gamma}$ will have sub-linear regret in the horizon $T$. This is illustrated by the following proposition:

\begin{proposition}
There exists a linear bandits problem instance and a dynamic pricing problem instance, i.e., an environment distribution $\mathcal{P}_{\gamma}$, such that the optimal decision function learned under the squared loss incurs a linear regret for every $\gamma$, i.e., $\mathrm{Regret}(\texttt{Alg}^*;\gamma)=\Omega(T)$ for every $\gamma$.
\label{prop:lin_reg}
\end{proposition}

Proposition \ref{prop:lin_reg} states a negative result that the optimal decision function $\texttt{Alg}^*$, albeit being optimal in a prediction sense (for predicting the optimal action), does not serve as a good algorithm for sequential decision making problems.  To see the intuition, the decision function is used recursively in the testing phase (see \eqref{eqn:test_dynamics}). While it gives the best possible prediction, it does not conduct \textit{exploration} which is critical for the concentration of the posterior measure.  This result does not contradict Proposition \ref{prop:surro}. While Proposition \ref{prop:surro} provides the action prediction error as an upper bound on the regret, in the instances described in Proposition \ref{prop:lin_reg}, both the prediction error and the upper bound are linear in $T$.

This highlights a key difference between sequential decision making and single decision making problems (e.g., \cite{chen2022using,han2023deep,ban2019big,feng2023framework}) when applying machine learning or deep learning tools. In single decision making problems, where only a single decision needs to be inferred for a single time (rather than an ``entire'' algorithm for a horizon), the collected training samples (analogous to the context-action-observation tuples in $H_t$) are i.i.d. generated, and this results in a convergence/concentration of the posterior to the true model/environment $\gamma$. However, in sequential decision making problems, the dynamics of \eqref{eqn:test_dynamics} induce dependencies across the history $H_t$, and it may result in a non-concentration behavior for the posterior. This is the key insight behind constructing the linear-regret example in Proposition \ref{prop:lin_reg} and the examples are also inspired from and share the same spirit as the discussions in \cite{harrison2012bayesian}. It is also worth noting that such a lack of exploration may not occur in discrete action spaces, as studied in \cite{lee2024supervised}. This is because for discrete action space, the output $\texttt{TF}_{\hat{\theta}}$ (or $\texttt{Alg}^*$) can give the posterior distribution as a distribution over the action space. The distribution output will lead to a randomized action and thus can automatically perform exploration.

\subsubsection{(Sub-linear) Regret Analysis on $\texttt{TF}_{\hat{\theta}}$}

There are simple remedies to address the linear regret behavior of the optimal decision function, such as using an $\epsilon$-greedy version of it and explicitly adding some exploration. However, even without any further modifications to the pre-trained $\texttt{TF}_{\hat{\theta}}$, we do not observe this linear regret behavior in any of our numerical experiments (as will be demonstrated in Section \ref{sec:numerical}). In this section,  we provide a theoretical framework to explain the observed sub-linear regrets of $\texttt{TF}_{\hat{\theta}}$. 

We begin by informally stating the regret upper bound of $\texttt{TF}_{\hat{\theta}}$ to provide intuition on why it performs well without explicit exploration. We assume $\mathcal{P}_{\gamma}$ is a uniform distribution over a bounded support $\Gamma \in \mathbb{R}^d$ and recall $\gamma\in \Gamma$ is the sampled testing environment, which generates $H_t$ through \eqref{eqn:test_dynamics} and decides the reward function $r$. For simplicity, we denote the conditional distribution of the observation $O_t$ given the history $H_t$ and $a_t=\texttt{TF}_{\hat{\theta}}(H_t)$ (or $a_t\sim \texttt{TF}_{\hat{\theta}}(H_t)$)  as $\mathbb{P}_{\gamma'}(\cdot|H_t, \texttt{TF}_{\hat{\theta}})$ for any environment $\gamma'\in \Gamma$ (which is not necessary the true testing environment $\gamma$).  Under conditions on restricting smoothness of $\mathbb{P}_{\gamma'}(\cdot|H_t, \texttt{TF}_{\hat{\theta}})$, we can establish the following result:

\begin{theorem}[Regret Upper Bound (Informal)]
\label{thm:informal}
Suppose there exists constants  $\Delta_{\text{Exploit}}$ and $\Delta_{\text{Explore}}$ such that for any $\gamma\in \Gamma\subset \mathbb{R}^d$ and $H_t$, $\texttt{TF}_{\hat{\theta}}$ and $\texttt{Alg}^*(H_t)$ satisfies 
\begin{itemize}
    \item $\mathbb{E}[r(X_t,\texttt{Alg}^*(H_t))-r(X_t,\texttt{TF}_{\hat{\theta}}(H_t))|H_t]\le \Delta_{\text{Exploit}}$, where the expectation is taken with respect to the possible randomness in $\texttt{Alg}^*$ and $\texttt{TF}_{\hat{\theta}}$. 
    \item The KL divergence
$$\text{KL}\left(\mathbb{P}_{\gamma}(\cdot|X_t, \texttt{TF}_{\hat{\theta}}(H_t))\|\mathbb{P}_{\gamma'}(\cdot|X_t, \texttt{TF}_{\hat{\theta}}(H_t))\right)\ge \Delta_{\text{Explore}}\|\gamma-\gamma'\|_2^2$$
for any $\gamma' \in \Gamma$, and the log-likelihood ratio between $\mathbb{P}_{\gamma}(\cdot|X_t, \texttt{TF}_{\hat{\theta}}(H_t))$ and $\mathbb{P}_{\gamma'}(\cdot|X_t, \texttt{TF}_{\hat{\theta}}(H_t))$ is sub-Gaussian
for any $\gamma' \in \Gamma$.
    %\item $\mathbb{E}[r(X_t,a^*_t)-r(X_t,\texttt{Alg}^*(H_t))|H_t]\le O\left(\mathbb{E}_{\gamma'}[\|\gamma-\gamma'\|_2^2|H_t]\right)$, where the expectation on the left side is taken with respect to the possible randomness in $\texttt{Alg}^*$ and the expectation $\mathbb{E}_{\gamma'}[\cdot]$ on the right side is with respect to the posterior distribution over the possible testing environment $\gamma'$ given $H_t$.
\end{itemize}
Then we have
\begin{align*}
    \mathrm{Regret}(\texttt{TF}_{\hat{\theta}};\gamma)
    = O\left(\Delta_{\text{Exploit}} T+\frac{d}{\Delta_{\text{Explore}}}\right)
\end{align*}
for any $\gamma\in \Gamma$, where the notation $O(\cdot)$ omits logarithm terms in $T$ and $d$ and the parameters in assumptions (e.g., boundedness parameter) .    
\end{theorem}

We should first note that Theorem \ref{thm:informal} provides a worst-case regret bound since it holds for any possible testing environment $\gamma \in \Gamma$. Thus, it also serves as an upper bound for the expected regret $\mathbb{E}_{\gamma\sim \mathcal{P}_{\gamma}} \left[\mathrm{Regret}(\texttt{TF}_{\hat{\theta}};\gamma)\right]$. To better understand Theorem \ref{thm:informal}, we first examine its three key conditions. The first condition is that $\texttt{TF}_{\hat{\theta}}$ behaves closely to $\texttt{Alg}^*$ and the (reward) deviation between these two is bounded by $\Delta_{\text{Exploit}}$. The deviation could arise due to finite pre-training data or the limited function space $\mathcal{F}$ during the pre-training phase. This condition can be justified by Figure \ref{fig:Bayes_match}. The second condition characterizes the exploration intensity of $\texttt{TF}_{\hat{\theta}}$, ensuring it can distinguish between the true environment and any other environment by bounding their KL divergence with  $\Delta_{\text{Explore}}$. The last condition characterizes the performance of $\texttt{Alg}^*$ in terms of the posterior distribution of the possible testing environment given $H_t$. As the posterior distribution concentrates on the true testing environment $\gamma$, the reward of  $\texttt{Alg}^*(H_t)$ converges to that of the optimal action. 

Under these conditions, intuitively,  $\texttt{TF}_{\hat{\theta}}$ can be viewed as the optimal decision function $\texttt{Alg}^*$ plus some deviation. These deviations can be considered as random noises: while they may distort the performance of $\texttt{TF}_{\hat{\theta}}$ when the posterior is concentrated (first and third conditions), they can play the role of exploration and helps the posterior to concentrate (second condition). In particular, $\texttt{TF}_{\hat{\theta}}$ deviates from $\texttt{Alg}^*$ by an amount of $\Delta_{\text{Exploit}}$ in terms of the reward, and this deviation accumulates over time into the first term in the regret bound. On the other hand, this deviation also gives an exploration whose intensity is measured by $\Delta_{\text{Explore}}$, and the second term captures the regret caused during the concentration of the posterior onto the true environment. 

These conditions serve as a stylized model for analyzing the behavior and performance of $\texttt{TF}_{\hat{\theta}}$.  In many online learning algorithms, such as UCB or Thompson sampling, the constants $\Delta_{\text{Exploit}}$ and $\Delta_{\text{Explore}}$ should have a dependency on $t$. Nevertheless, these conditions give us a perspective to understand how the pre-trained $\texttt{TF}_{\hat{\theta}}$ works as a decision algorithm and why it can have sub-linear regret. For example, when $\Delta_{\text{Exploit}}=\Delta_{\text{Explore}}=O(1/\sqrt{T})$, this regret upper bound is in the order of $O(\sqrt{T})$. In the following part, we will elaborate on the assumptions required for these conditions and provide a formal result for the regret upper bound.
 
\paragraph{Assumptions} We first present the assumptions needed for the regret analysis.
\begin{assumption}[Environment Space]
We assume there exists a space $\Gamma\subset \mathbb{R}^{d}$ of environment parameters such that
\begin{itemize}
    \item There exists a constant $\bar{\gamma}$ such that $\sup_{\gamma \in \Gamma} \|\gamma\|_2\leq \bar{\gamma}$.
    \item There exists a constant $\bar{r}$ such that  $\sup_{ x\in\mathcal{X},a\in\mathcal{A},\gamma\in \Gamma}r(x,a;\gamma)\leq \bar{r}$, where we use $r(x,a;\gamma)$ to denote the reward function of environment $\gamma$.
    \item $\mathcal{P}_{\gamma}$ is a uniform distribution over $\Gamma$.
\end{itemize}
\label{assmp:bounded}
\end{assumption}

The above assumptions characterize the space of possible environments. The first two assumptions bound the space $\Gamma$ and the maximum possible reward, which are commonly applied in the literature (e.g., \cite{den2015dynamic, huh2009nonparametric}). The last assumption concerns the decision maker's prior distribution over the testing environment. We remark that this assumption can be relaxed to allow for any general distribution. However, such a relaxation would introduce an additional multiplier in the resulting regret bound. This multiplier would be proportional to the upper bound of the likelihood ratio between two possible environments in $\Gamma$. Intuitively, this represents the (largest) effort needed to correct an incorrect prior. A larger ratio indicates that more exploration is required to concentrate the posterior distribution on the true environment in the worst-case, leading to higher regret.

\begin{assumption}[Observation Distribution]

Denote $\mathbb{S}^d_{+}$ as the space of positive semidefinite $d\times d$ matrices and recall $\gamma$ is the true environment. Then we assume for any $X\in \mathcal{X}$, $a\in \mathcal{A}$ and $\gamma'\in\Gamma$:
\begin{itemize}
    \item  There exists a function $\Delta:(\mathcal{X},\mathcal{A})\rightarrow \mathbb{S}^d_{+}$ such that
    \begin{equation}
\text{KL}\left(\mathbb{P}_{\gamma}(\cdot\big \vert X,a)\| \mathbb{P}_{\gamma'}(\cdot\big \vert X,a) \right)\geq \|\gamma-\gamma'\|^2_{\Delta(X,a)},
    \end{equation}
where $\|z\|_{\Delta(X,a)}=\sqrt{z^T\Delta(X,a) z}$ for $z\in \mathbb{R}^d$. 
\item There exists a constant $C_{\sigma^2}$ such that for any $\gamma'\in\Gamma$, the log-likelihood ratio  $\log  \left(\frac{\mathbb{P}_{\gamma}(O\vert X,a)}{\mathbb{P}_{\gamma'}(O\vert X,a)}\right)$, where $O\sim \mathbb{P}_{\gamma}(\cdot \big \vert X,a)$, is $C_{\sigma^2}\|\gamma-\gamma'\|_{\Delta(X,a)}$-sub-Gaussian.
%\item The sequence $\left\{\sum_{\tau=1}^t \log  \left(\frac{\mathbb{P}_{\gamma}(O_{\tau}\vert X_{\tau},a_{\tau})}{\mathbb{P}_{\gamma'}(O_{\tau}\vert X_{\tau},a_{\tau})}\right)-\mathbb{E}_{O_{\tau}}\left[\log  \left(\frac{\mathbb{P}_{\gamma}(O_{\tau}\vert X_{\tau},a_{\tau})}{\mathbb{P}_{\gamma'}(O_{\tau}\vert X_{\tau},a_{\tau})}\right)\right]\right\}$ is a martingale, where $O_{\tau},X_{\tau},a_{\tau}$ is generated by \eqref{eqn:test_dynamics} under environment $\gamma$ and the expectation $\mathbb{E}_{O_{\tau}}[\cdot]$ is with respect to $O_{\tau}\sim \mathbb{P}_{\gamma}(\cdot \big \vert X_{\tau},a_{\tau})$.
\item  $\log \mathbb{P}_{\gamma'}(O\big \vert X,a)$ is concave in $\gamma'\in \Gamma$. Further, there exists $C_{\Gamma}> 0$ such that for all   $X\in\mathcal{X}$, $a\in \mathcal{A}$ and $O\in \mathcal{O}$, 
    $$\|\nabla^2_{\gamma'} \log \mathbb{P}_{\gamma'}(O\big \vert X,a)\|^2_2\leq C_{\Gamma}.$$
\end{itemize}
\label{assmp:llh}
\end{assumption}
The first two assumptions above characterize the statistical properties required for the log-likelihood ratio, $\log  \left(\frac{\mathbb{P}_{\gamma}(O\vert X,a)}{\mathbb{P}_{\gamma'}(O\vert X,a)}\right)$, and the last one assumes the smoothness of the log-likelihood function $\log \mathbb{P}_{\gamma'}(O\big \vert X,a)$. Specifically, in the first two assumptions, $\sum_{\tau=1}^t\Delta(X_{\tau},a_{\tau})$ can be interpreted as the information matrix from $(X_{\tau},a_{\tau})$'s. For example, in dynamic pricing with Gaussian noise and a linear demand function (i.e., $d(X,a)=\alpha^\top X+\beta^\top X\cdot a$ where $\gamma=(\alpha,\beta)$), we have $\Delta(X_{\tau},a_{\tau})=(X_{\tau},a_{\tau} X_{\tau})(X_{\tau},a_{\tau} X_{\tau})^\top$. The second assumption is necessary to bound the rate at which the posterior distribution concentrates around the true environment. The last assumption requires the log-likelihood to be smooth with respect to the environment parameters.  These assumptions are commonly satisfied in tasks like dynamic pricing and bandits with generalized linear demands or rewards under Gaussian noise.

\begin{assumption}[Reward Gaps]
We assume for any $\gamma \in \Gamma$ with its corresponding reward function $r$:
\begin{itemize}
    \item There exists a constant $\Delta_{\text{Exploit}}$ such that $\texttt{TF}_{\hat{\theta}}$ satisfies $$\mathbb{E}\left[r(X_t,\texttt{Alg}^*(H_t))-r(X_t,\texttt{TF}_{\hat{\theta}}(H_t)|H_t\right]\leq \Delta_{\text{Exploit}} $$
for all possible $H_t$, where the expectation is taken with respect to the possible randomness in $\texttt{Alg}^*$ and $\texttt{TF}_{\hat{\theta}}$.

\item There exists a constant $C_r$ such that $\texttt{Alg}^*$ satisfies
$$\mathbb{E}[r(X_t,a^*_t)-r(X_t,\texttt{Alg}^*(H_t))|H_t]\leq C_r \mathbb{E}_{\gamma'}[\|\gamma-\gamma'\|_2^2|H_t],$$  where the expectation on the left side is taken with respect to the possible randomness in $\texttt{Alg}^*$ and the expectation $\mathbb{E}_{\gamma'}[\cdot|H_t]$ on the right side is with respect to the posterior distribution over the possible testing environment $\gamma'$ given $H_t$.
\end{itemize}
\label{assmp:reward}
\end{assumption}
The first condition bounds the reward deviation between  $\texttt{Alg}^*$ and $\texttt{TF}_{\hat{\theta}}$, which can be potentially caused by the finite pre-training samples and is numerically justified by Figure \ref{fig:Bayes_match}. The second condition indicates the performance of $\texttt{Alg}^*$ in relation to the posterior distribution (which is further decided by the history $H_t$): as the posterior distribution of $\gamma'$ under $H_t$ becomes more concentrated around the true testing environment $\gamma$, the performance of $\texttt{Alg}^*$ improves.

\paragraph{Regret Upper Bound}
\begin{theorem}[Regret Upper Bound]
\label{thm:reg_bound}
    Denote  $\lambda_{t}=\lambda_{\min}\left(\sum_{\tau=1}^t \Delta(X_{\tau},a_{\tau})\right)$ as the minimum eigenvalue of the matrix $\sum_{\tau=1}^t \Delta(X_{\tau},a_{\tau})$ for each $t=1,\ldots,T$, and $t_0=\min\{t=2,\ldots,T|\lambda_{t-1}>0\}$ (here we let $t_0=T+1$ when $\{t=2,\ldots,T|\lambda_{t-1}>0\}=\emptyset$). Under Assumption \ref{assmp:bounded}, \ref{assmp:llh}, \ref{assmp:reward}, we have for any $\gamma \in \Gamma$
    \begin{align*}
        \text{Regret}(\texttt{TF}_{\hat{\theta}};\gamma)&\leq \Delta_{\text{Exploit}}T+\bar{r}\mathbb{E}[t_0-1]+C_r\mathbb{E}\left[\sum_{t=t_0}^T\epsilon_{t}\right]+16C_r(4C_{\sigma_2}+1)\mathbb{E}\left[\sum_{t=t_0}^T\frac{\log\left(2K_tT\right)}{\lambda_{t-1}}\right]+4C_r\bar{\gamma}^2\\
        &=O\left(\Delta_{\text{Exploit}}T+\mathbb{E}\left[t_0+\sum_{t=t_0}^T \frac{d}{\lambda_{t-1}}\right]\right)
    \end{align*}
    where $\epsilon_t=\frac{16d}{\lambda_{t-1}}\log\left(\frac{32\bar{\gamma}\sqrt{d}(C_{\Gamma}(t-1)+\lambda_{t-1})}{\lambda_{t-1}}\right)$,  $K_t=\max\left\{1,\left(\frac{32\bar{\gamma}\sqrt{d}(C_{\Gamma}(t-1)+\lambda_{t-1})}{\lambda_{t-1}\min\{\sqrt{\epsilon_t},1\}}\right)^d\right\}$, and the expectation is with respect to the potential randomness in $\lambda_t$ and $t_0$.
\end{theorem}

 Theorem \ref{thm:reg_bound} follows the same spirit as Theorem \ref{thm:informal} and can also be decomposed into two terms: the first term $O\left(\Delta_{\text{Exploit}}T\right)$ captures the regret caused by the deviation of $\texttt{TF}_{\hat{\theta}}$ from the optimal decision $\texttt{Alg}^*$. The second term $O\left(\mathbb{E}\left[t_0+\sum_{t=t_0}^T \frac{d}{\lambda_{t-1}}\right]\right)$ represents the regret incurred during the concentration of the posterior distribution towards the true environment $\gamma$. Specifically, $\lambda_t$ is the minimum eigenvalue of the empirical ``information matrix'' $\sum_{\tau=1}^t \Delta(X_{\tau},a_{\tau})$, which controls the rate of the posterior distribution convergence to the true testing environment (see Corollary \ref{corr:cts_prob_bound}). The random time $t_0$ is required for the availability of $\lambda_t>0$ and is influenced by both $\texttt{TF}_{\hat{\theta}}$ and the generation of $X_t$.  The proof of Theorem \ref{thm:reg_bound} and further discussions on its technical contributions are deferred to Appendix \ref{appx:proofs}.

 Our analysis offers a complementary understanding of how the pre-trained $\texttt{TF}_{\hat{\theta}}$ works in literature.   While \citet{lin2023transformers} provide a statistical learning framework for analyzing the performance of transformer-style models in decision makings, they show that when (i) $a_t^{(\gamma_i)*}$'s in the training data are no longer the optimal actions $a_t^*$ in our setting but generated from some algorithm $\texttt{Alg}$ and (ii) the decision function $\tilde{f}=\texttt{Alg}$ in the distribution $\mathcal{P}_{\gamma_i,\tilde{f}}$, the pre-trained transformer $\texttt{TF}_{\hat{\theta}}\rightarrow \texttt{Alg}$ as the number of training samples goes to infinity. We note that this only implies that the transformer model can \textit{imitate} an existing algorithm $\texttt{Alg}.$ In other words, such setting brings the theoretical benefits of consistency, but it excludes the possibility of using the optimal $a_t^*$ in the pre-training and thus makes it impossible for the pre-trained  $\texttt{TF}_{\hat{\theta}}$ to perform better than the algorithm $\texttt{Alg}$. To better see it, as shown in Proposition \ref{prop:lin_reg}, $\texttt{Alg}^*$ can incur linear regret (for numerical results, see Appendix \ref{appx:posterior_oracle}), a behavior that is not numerically observed with $\texttt{TF}_{\hat{\theta}}$. This discrepancy cannot be explained by \citet{lin2023transformers} when setting $\texttt{Alg}=$\texttt{Alg}$^*$. Our assumptions and analysis provide a stylized explanation for such a sub-linear regret behavior of $\texttt{TF}_{\hat{\theta}}$, as shown in the following example.  

\paragraph{Example: Dynamic Pricing} Here we adopt the contextual dynamic pricing problem \citep{qiang2016dynamic,ban2021personalized} as an example to show the application of Theorem \ref{thm:reg_bound}.

\begin{example}
\label{example:cts_DP}
Consider a dynamic pricing problem, where the demand is given by
$D_t=\alpha^\top X_t+\beta^\top X_t \cdot a_t+\epsilon_t.$
Here  $\epsilon_t\overset{\mathrm{i.i.d.}}{\sim} \mathcal{N}(0,1)$ and $\gamma=(\alpha,\beta) \in \Gamma \subset \mathbb{R}^{2d}$.  We suppose that  $\texttt{Alg}^*$ performs posterior averaging. In addition, for each time $t$, we assume with some constant $C$,  $\texttt{TF}_{\hat{\theta}}(H_t)=\texttt{Alg}^*(H_t)+\Delta_t$, where
\begin{equation*}
  \Delta_t=
    \begin{cases}
      CT^{-1/4} & \text{w.p.} \ 1/2, \\
      -CT^{-1/4} & \text{w.p.} \ 1/2 
    \end{cases}       
\end{equation*}
 for any possible history $H_t$ and $t=1,\ldots, T$, and $\Delta_t$ is independent of $H_t, \texttt{Alg}^*(H_t),\gamma$, and previous $\Delta_{\tau}$ for $\tau=1,\ldots,t-1$. If we further assume some boundedness
on the parameters, then by Theorem \ref{thm:reg_bound}, we have
$$\text{Regret}( \texttt{TF}_{\hat{\theta}},\gamma)=O(d\sqrt{T})$$
for all $\gamma \in \Gamma$.
\end{example}

We make the following remarks for Example \ref{example:cts_DP}: (i) The derived upper regret upper bound matches the regret lower bound in terms of the order of $T$, as established in \cite{keskin2014dynamic}. This illustrates that, with an appropriate deviation between $\texttt{Alg}^*$ and $\texttt{TF}_{\hat{\theta}}$, $\texttt{TF}_{\hat{\theta}}$ can achieve an asymptotically optimal (worst-case) regret bound.  (ii) The assumption $\texttt{TF}_{\hat{\theta}}(H_t)=\texttt{Alg}^*(H_t)+\Delta_t$  introduces a (random) perturbation term $\Delta_t$ to  $\texttt{Alg}^*(H_t)$.  These perturbations act as explorations to learn $\gamma$, and the assumed rate $\Theta(T^{-1/4})$ effectively balances the exploration and the exploitation to bound the regret.

%While \citet{lin2023transformers} provide a statistical learning framework for analyzing the performance of transformer-style models in decision makings, their result focuses on bounding the performance gap between $\texttt{TF}_{\hat{\theta}}$ and an expert algorithm (i.e., $\texttt{Alg}^*$), with the gap asymptotically converging to zero. However, as shown in Proposition \ref{prop:lin_reg}, $\texttt{Alg}^*$ can incur linear regret, a behavior that is not numerically observed with $\texttt{TF}_{\hat{\theta}}$. This discrepancy cannot be explained by \citet{lin2023transformers}. Our assumptions and analysis provide a stylized explanation for such sub-linear regret behavior of $\texttt{TF}_{\hat{\theta}}$ (which can indeed outperform well-known benchmark online algorithms, as shown in the next section). 
\paragraph{Regret Upper Bound on Finite Environment Space}

In the following, we present the regret upper bound for a special case where $\Gamma$ contains only a finite set of environments. 

\begin{theorem}[Regret Upper Bound on Finite Environment Space]
\label{thm:reg_bound_finite}
    Assume $\Gamma=\{\gamma_1,\ldots,\gamma_n\}$. Under Assumption \ref{assmp:bounded}, \ref{assmp:llh}, \ref{assmp:reward}, we have for any $\gamma \in \Gamma$:
    \begin{itemize}
        \item \textbf{(Problem-dependent bound).} We denote $t_i=\min\{t=2,\ldots,T|\sum_{\tau=1}^{t-1}\|\gamma_i-\gamma\|_{\Delta(X_{\tau},a_{\tau)}}^2\geq 16C_{\sigma^2}\log(nT^2)\}$ (we let $t_i=T+1$ when the set is empty), then
          \begin{equation}
        \label{eqn:reg_dependent}
                \text{Regret}(\texttt{TF}_{\hat{\theta}};\gamma)<4C_r\bar{\gamma}^2+\Delta_{\text{Exploit}}T+\bar{r}C_r\sum_{\gamma_i\neq \gamma}\mathbb{E}[t_i-1]+  C_r\sum_{\gamma_i\neq \gamma}\mathbb{E}\left[\sum_{t=t_i}^T\frac{4\|\gamma_i-\gamma\|_2^2}{\sum_{\tau=1}^{t-1}\|\gamma_i-\gamma\|_{\Delta(X_{\tau},a_{\tau})}^2}\right].
    \end{equation}
    \item \textbf{(Problem-independent bound).} Recall $\lambda_{t}=\lambda_{\min}\left(\sum_{\tau=1}^t \Delta(X_{\tau},a_{\tau})\right)$ as the minimum eigenvalue of the matrix $\sum_{\tau=1}^t \Delta(X_{\tau},a_{\tau})$ for each $t=1,\ldots,T$ and $t_0=\min\{t=2,\ldots,T|\lambda_{t-1}>0\}$, then
    \begin{equation}
        \label{eqn:reg_ind}
        \text{Regret}(\texttt{TF}_{\hat{\theta}};\gamma)<4C_r\bar{\gamma}^2+\Delta_{\text{Exploit}}T+\bar{r}C_r\mathbb{E}[t_0-1]+nC_r\mathbb{E}\left[\sum_{t=t_0}^T\frac{16C_{\sigma^2}\log(nT^2)+4}{\lambda_{t-1}}\right].
    \end{equation}
    \end{itemize}
    Here the expectation is with respect to the potential randomness in $\lambda_t$ and $t_{i}$, $i=0,\ldots,n$.
\end{theorem}

Compared to Theorem \ref{thm:reg_bound}: (i) The regret bounds in Theorem \ref{thm:reg_bound_finite}  depend on the number of environments $n$ rather than the dimension of the environment $d$. This is because, with a finite structure, we only need to identify $\gamma$ from a finite set, rather than from an infinite set of possible environments as in Theorem \ref{thm:reg_bound}, whose complexity is controlled by $d$. Technically, the shift from $n$ to $d$ arises from using a covering number technique in Theorem \ref{thm:reg_bound} to discretize a bounded continuous space into a discrete space, where the covering number depends on $d$. (ii) It is difficult to say whether Theorem \ref{thm:reg_bound_finite} provides sharper bounds than Theorem \ref{thm:reg_bound} without specifying the underlying problem or the structure of $\Gamma$. For instance, when $n \gg d$, the bounds in Theorem \ref{thm:reg_bound_finite} can be worse.

We refer to \eqref{eqn:reg_dependent}
 as the problem-dependent bound and \eqref{eqn:reg_ind} as the problem-independent bound. This is because \eqref{eqn:reg_dependent} accounts for specific candidate environments (e.g., the term $\|\gamma_i-\gamma\|_2^2$), whereas \eqref{eqn:reg_ind} does not depend on such specific problem structure. These two bounds are similar to the problem-dependent and problem-independent bounds in multi-armed bandits \citep{lattimore2020bandit}. However, it is important to note that in the multi-armed bandits problem, there is a finite action space, and the problem-dependent bound relates to the action rewards. In contrast, here we assume a finite number of environments, and the bound relates to these environments.
 
The following example shows that, given a specific problem structure, the problem-dependent bound can sometimes provide a sharper regret upper  bound than the problem-independent bound. 

\begin{example}
\label{example:finite_DP}
Consider a dynamic pricing problem without context, where the demand is given by $D_t=\alpha+\beta\cdot a_t+\epsilon_t$. Here $\epsilon_t\overset{\mathrm{i.i.d.}}{\sim} \mathcal{N}(0,1)$ and $\gamma=(\alpha,\beta) \in \Gamma \subset \mathbb{R}^{2}$. We further assume that  $\texttt{TF}_{\hat{\theta}}=\texttt{Alg}^*$  and $\texttt{Alg}^*$  performs posterior averaging. If we suppose $\Gamma=\{(1,-1),(1,-2)\}$ and assume some boundedness on the parameters, then by applying \eqref{eqn:reg_dependent}, we have
$$\text{Regret}( \texttt{TF}_{\hat{\theta}},\gamma)=O(\log T)$$
for all $\gamma \in \Gamma$.
\end{example}

This example illustrates that a regret bound of $O(\log T)$ can be achieved by applying \eqref{eqn:reg_dependent}, which is smaller than the typical  regret upper bound of $O(\sqrt{T})$ found in dynamic pricing problems (e.g., \cite{broder2012dynamic, keskin2014dynamic, wang2021dynamic}). Indeed, through a similar analysis as in Example \ref{example:cts_DP}, Theorem \ref{thm:reg_bound}  would also result in a regret upper bound of $O(\sqrt{T})$ for Example \ref{example:finite_DP}.

To understand the $O(\log T)$ regret, intuitively, there are no ``uninformative prices'' \citep{broder2012dynamic} for identifying $\gamma$ due to the specific structure of $\Gamma$. Consequently,  the regret term $\frac{4\|\gamma_i-\gamma\|_2^2}{\sum_{\tau=1}^{t-1}\|\gamma_i-\gamma\|_{\Delta(X_{\tau},a_{\tau})}^2}$, which typically arises from uncertainty of $\gamma$, will decrease at a rate of $O(1/t)$ for this specific space $\Gamma$, even when making greedy optimal decisions with $\texttt{Alg}^*(H_t)$ that involve no exploration. This example also illustrates how  $\texttt{TF}_{\hat{\theta}}$ (or $\texttt{Alg}^*$) can outperform benchmark algorithms due to the special structure of $\Gamma$.

\section{Numerical Experiments}
\label{sec:numerical}
In this section, we conduct numerical experiments to evaluate OMGPT's performance from various perspectives and investigate its internal mechanisms when making decisions. All experiment details can be found in Appendix \ref{appx:numerical}. In all figures, where applicable, the shaded area represents the $90\%$ empirical confidence interval, and the reported values are based on 100 runs.

\subsection{Performance Evaluation} 

We first assess OMGPT across several operational decision making tasks and examine its generalization ability in testing environments with longer horizons than those used in the pre-training. Additionally, we evaluate its performance in environments that are generated in a way different from the pre-training ones. We further test its capacity to handle non-stationary environments and environments with multiple demand types, highlighting its potential as a solution to the problem of model misspecification. Lastly, we show that, beyond predicting optimal actions, OMGPT can also serve as a predictor for some side information, such as the demand in dynamic pricing problems, to provide valuable auxiliary insights.

\subsubsection{Basic Performance} 
We evaluate OMGPT across a variety of operational decision making tasks, including dynamic pricing, the newsvendor problem (with censored demand), queuing control, and revenue management. The settings of these problems are deferred to Appendix \ref{appx:envs}.  It is important to note that both the queuing control and revenue management problems feature state-dependent contexts, as defined in Section \ref{sec:disc_state_dependent}. In these tasks, the context or state—such as queue length in queuing control or available resources in revenue management—depends on both the previous context and the actions taken. These tasks are used to validate OMGPT’s ability to effectively handle operational decision making problems with state-dependent contexts.

\begin{figure}[ht!]
  \centering
  \begin{subfigure}[b]{0.23\textwidth}
    \centering
\includegraphics[width=\textwidth]{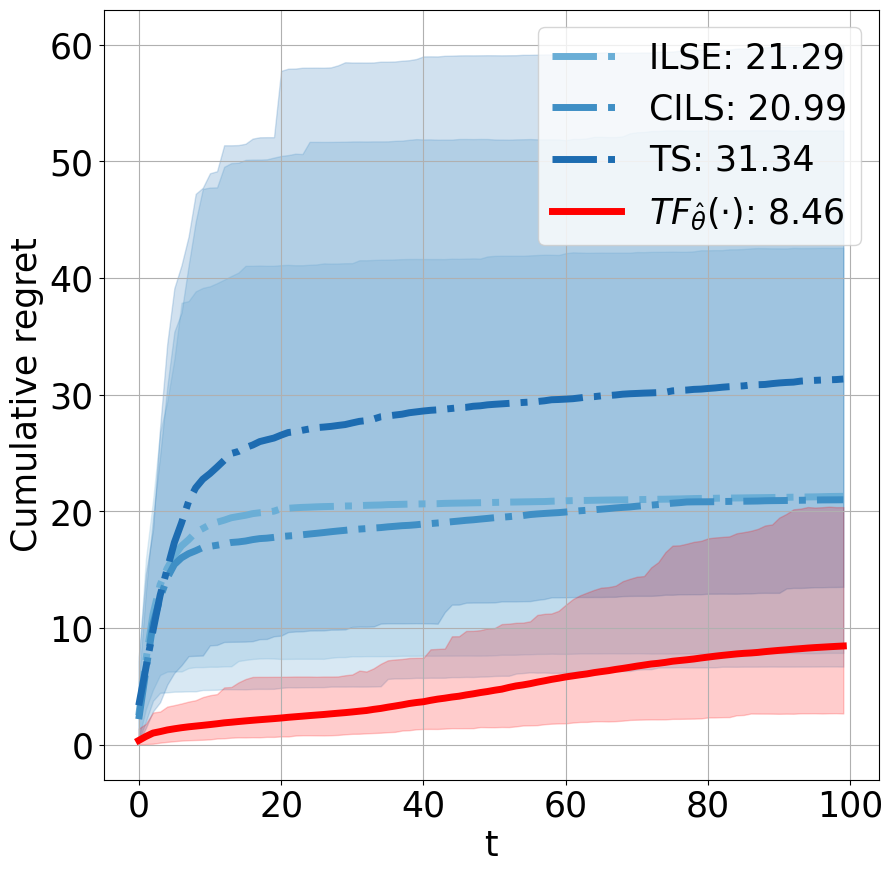} % Replace with your image
    \caption{Dynamic pricing}
  \end{subfigure}
    \hfill
    \begin{subfigure}[b]{0.23\textwidth}
    \centering
\includegraphics[width=\textwidth]{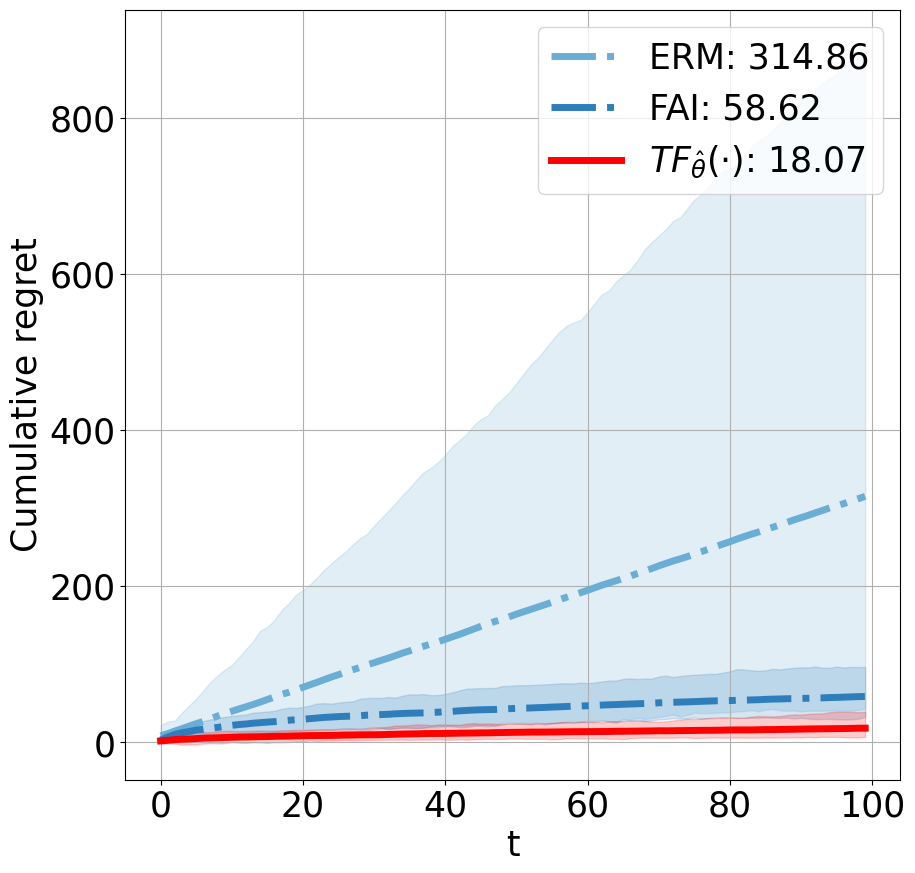}
    \caption{Newsvendor }
  \end{subfigure}
  \hfill  
  \begin{subfigure}[b]{0.23\textwidth}
    \centering
\includegraphics[width=\textwidth]{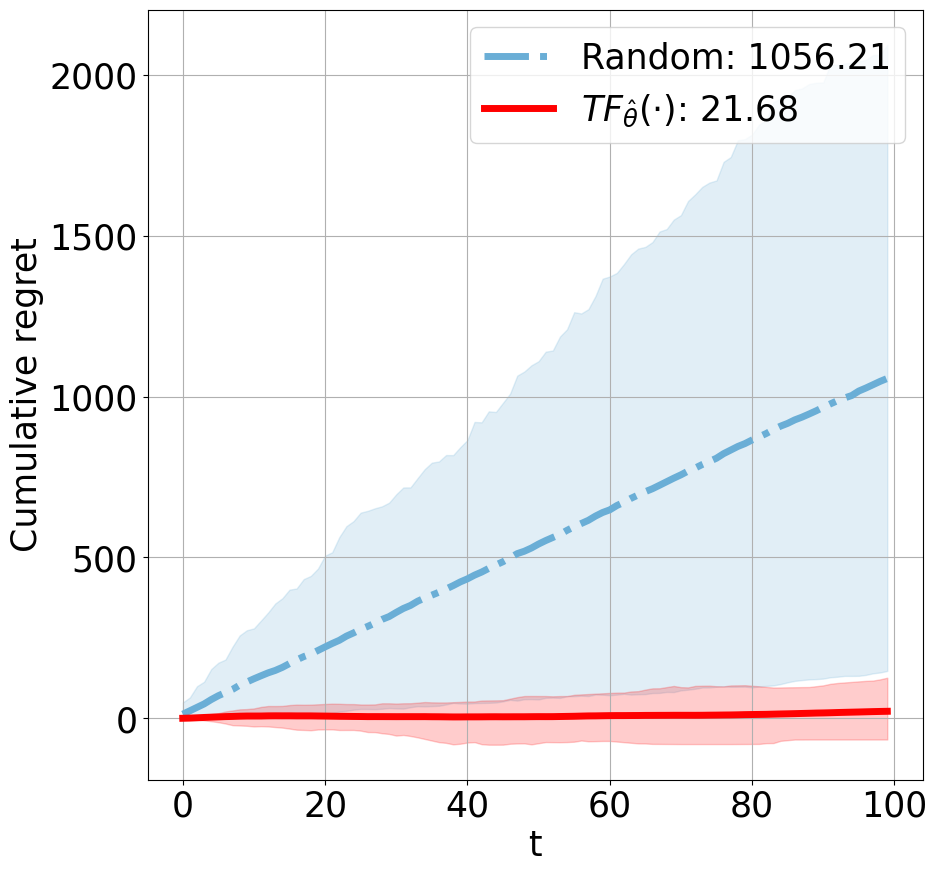}
    \caption{Queuing control}
  \end{subfigure}
  \hfill  
  \begin{subfigure}[b]{0.23\textwidth}
    \centering
\includegraphics[width=\textwidth]{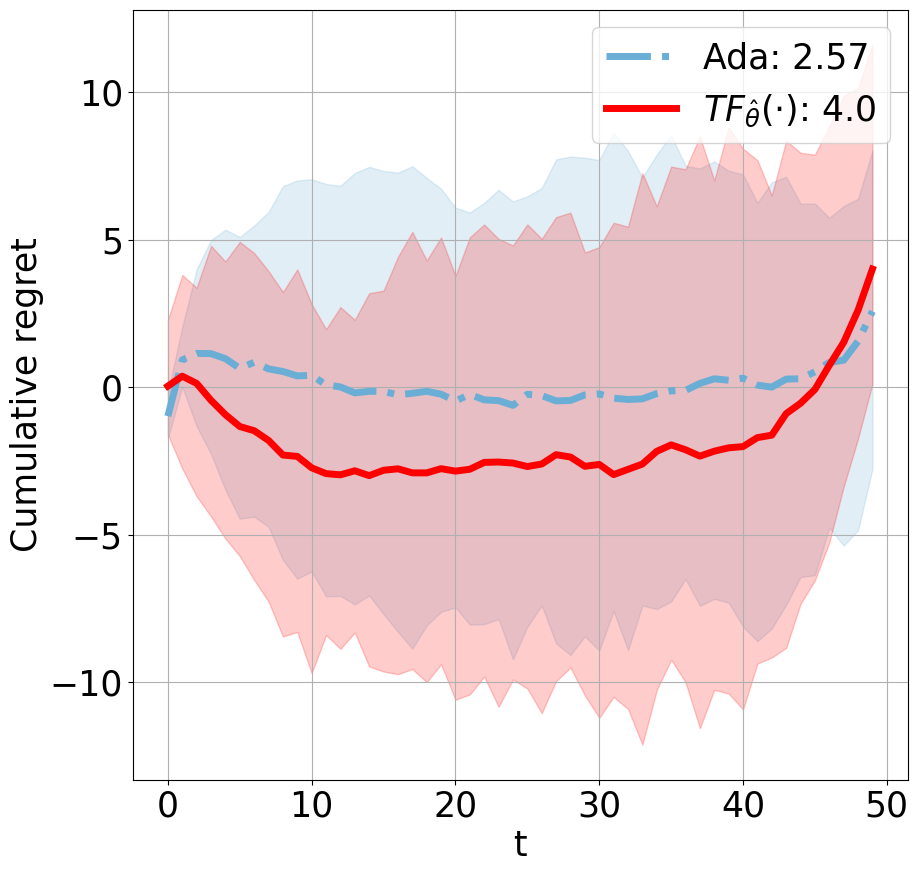}
    \caption{Revenue management}
  \end{subfigure}
  \begin{subfigure}[b]{0.23\textwidth}
    \centering
\includegraphics[width=\textwidth]{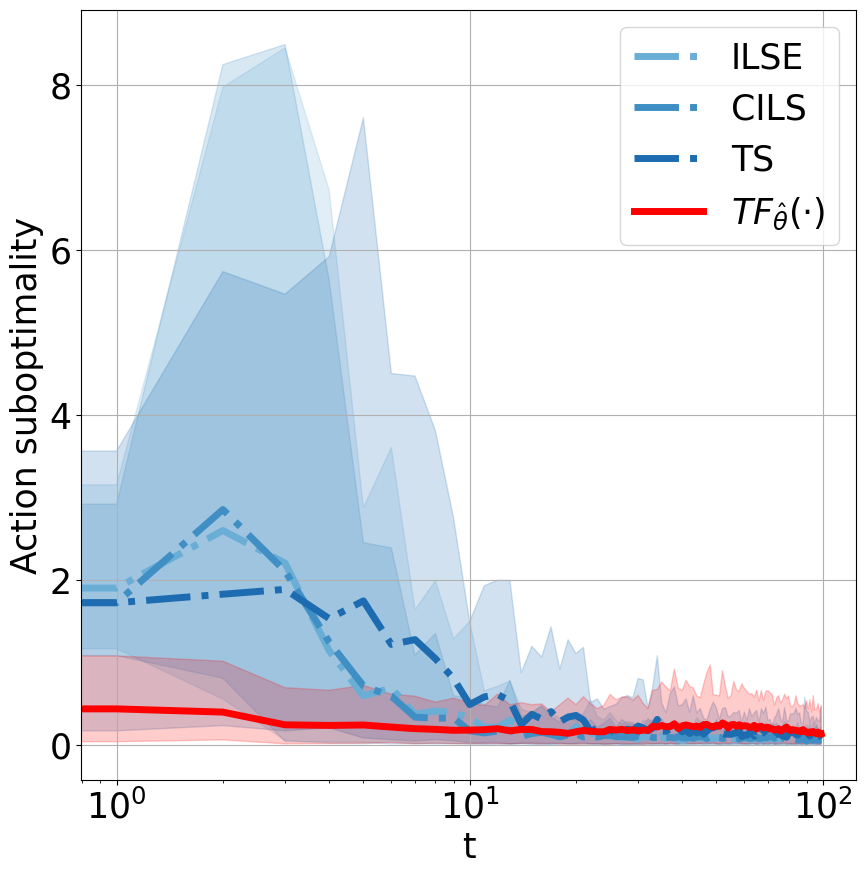} % Replace with your image
    \caption{Dynamic pricing}
  \end{subfigure}
    \hfill
    \begin{subfigure}[b]{0.23\textwidth}
    \centering
\includegraphics[width=\textwidth]{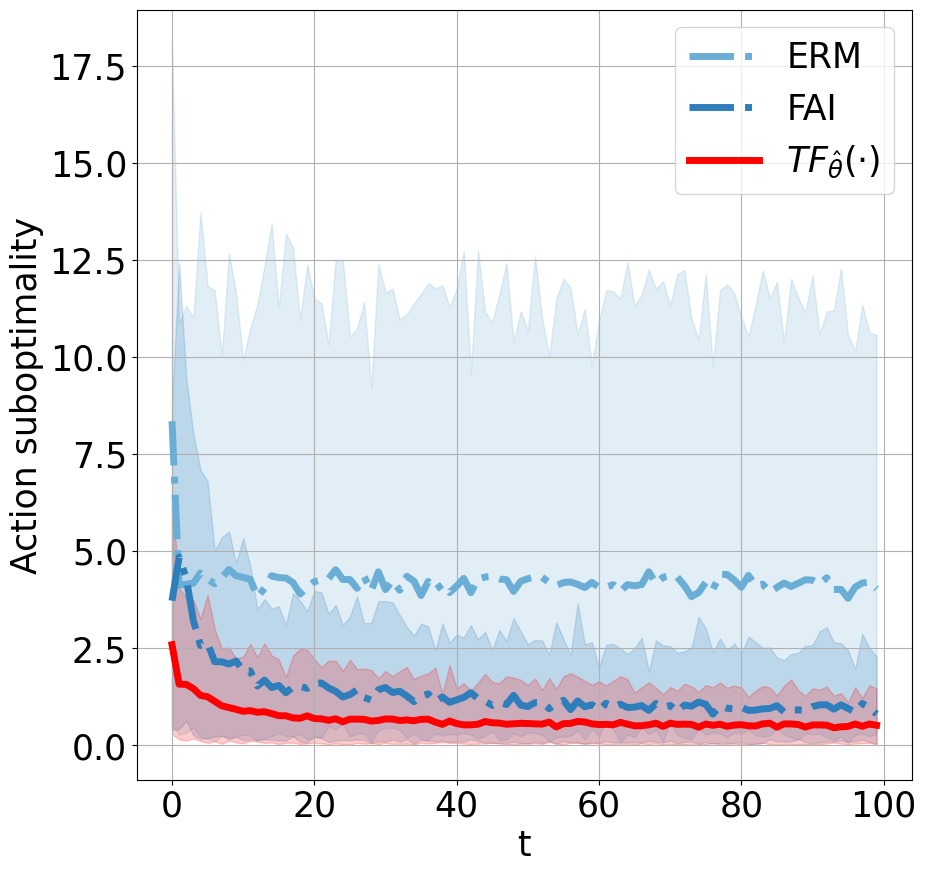}
    \caption{Newsvendor }
  \end{subfigure}
  \hfill  
  \begin{subfigure}[b]{0.23\textwidth}
    \centering
\includegraphics[width=\textwidth]{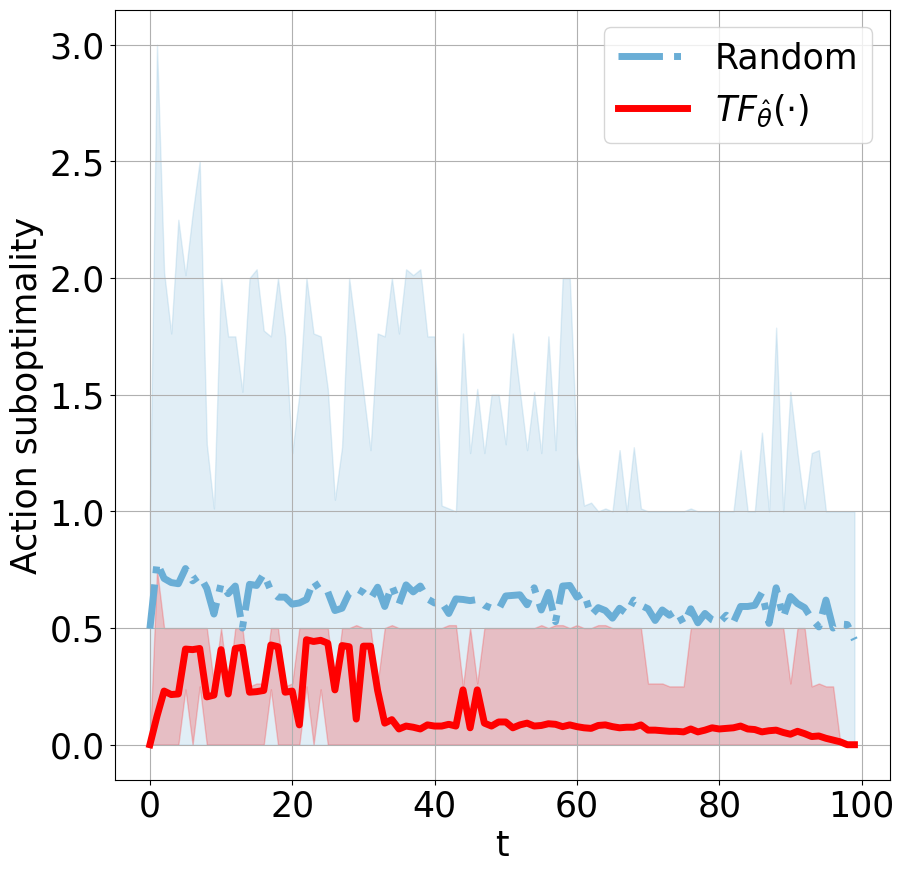}
    \caption{Queuing control}
  \end{subfigure}
  \hfill  
  \begin{subfigure}[b]{0.23\textwidth}
    \centering
\includegraphics[width=\textwidth]{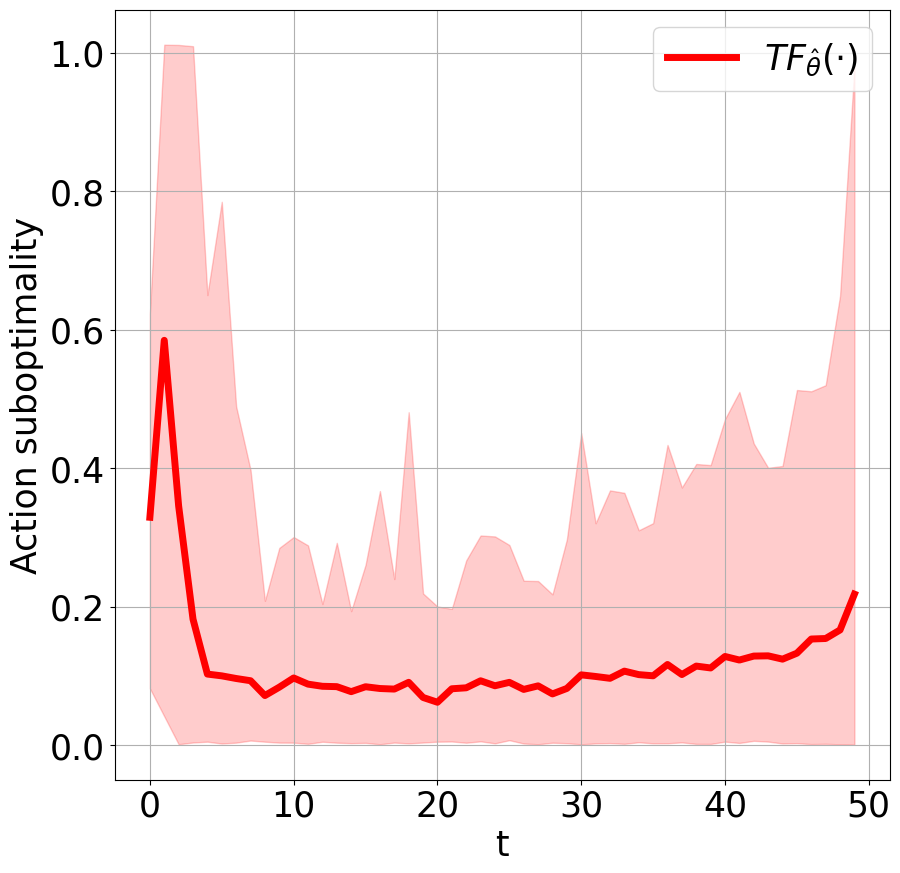}
    \caption{Revenue management}
  \end{subfigure}
\caption{\small The average out-of-sample regret (first row) and action suboptimality, i.e., $|a^*_t - \texttt{Alg}(H_t)|$, (second row) of $\texttt{TF}_{\hat{\theta}}$ against benchmark algorithms (see details in Appendix \ref{appx:benchmark}). The numbers in the legend bar are the final cumulative regret.  The last two tasks can have negative regret during the horizon since they are MDP problems where the optimal actions may achieve a lower single-period reward than the applied algorithms, although they maximize the final cumulative expected reward. For revenue management, since the exact optimal actions are computed at a high computation cost, we use the actions from the Adaptive Allocation Algorithm (Ada) from \cite{chen2024improved}, which can achieve constant regret, to approximate optimal actions in both the pre-training and testing (and thus not show the suboptimality of the Ada algorithm in (h)), and use their upper bound of the optimal cumulative reward to compute (an upper bound) of the regret used in (d).}

  \label{fig:main_reg}
\end{figure}

The pre-trained OMGPT, $\texttt{TF}_{\hat{\theta}}$, is compared against well-established benchmark algorithms for each task.  Figure \ref{fig:main_reg} presents a summary of the results and more details (e.g., experiment setup and benchmark algorithms) are deferred to Appendix \ref{appx:numerical}. It is important to note that while the testing environment is sampled from the decision maker’s prior distribution, $\mathcal{P}_\gamma$, which also generates the pre-training data for $\texttt{TF}_{\hat{\theta}}$, the distribution is assumed to have infinite support. This implies that the testing environment, $\gamma$, differs from the pre-training environments, $\gamma_{i}$'s, with probability 1. This setup ensures a fair comparison and provides a more challenging scenario to assess OMGPT’s robustness and generalization ability.

From Figure \ref{fig:main_reg}, we observe that OMGPT, $\texttt{TF}_{\hat{\theta}}$, consistently outperforms the structured benchmark algorithms across all tasks. What makes this result particularly remarkable is that $\texttt{TF}_{\hat{\theta}}$ does not merely replicate or imitate the performance of the benchmark algorithms; rather, it uncovers a new and more effective approach. This highlights the effectiveness of using the optimal actions, $a_t^*$, as the target during pre-training.

We further attribute the OMGPT's advantage over the benchmark algorithms to two key factors: the incorporation of prior knowledge from pre-training data and its more greedy exploitation strategy. First, the OMGPT effectively leverages the pre-training data, making it well-suited to the decision maker's prior knowledge of the testing environment (see Section \ref{sec:Bayes_decision_maker} and Appendix \ref{appx:prior_benefit}). In contrast, all the benchmark algorithms face a cold-start problem.  Second, while the benchmark algorithms are designed for asymptotic optimality, potentially sacrificing short-term performances, OMGPT takes a greedier approach, similar to $\texttt{Alg}^*$, which does not explicitly encourage exploration. This greedier behavior can lead to better performances when the time horizon $T$ is short, particularly when $T \leq 100$ as shown in Figure \ref{fig:main_reg}.

\subsubsection{Impact of Model Scaling and Task Complexity}

\begin{figure}[ht!]
    \centering
    \includegraphics[width=0.5\linewidth]{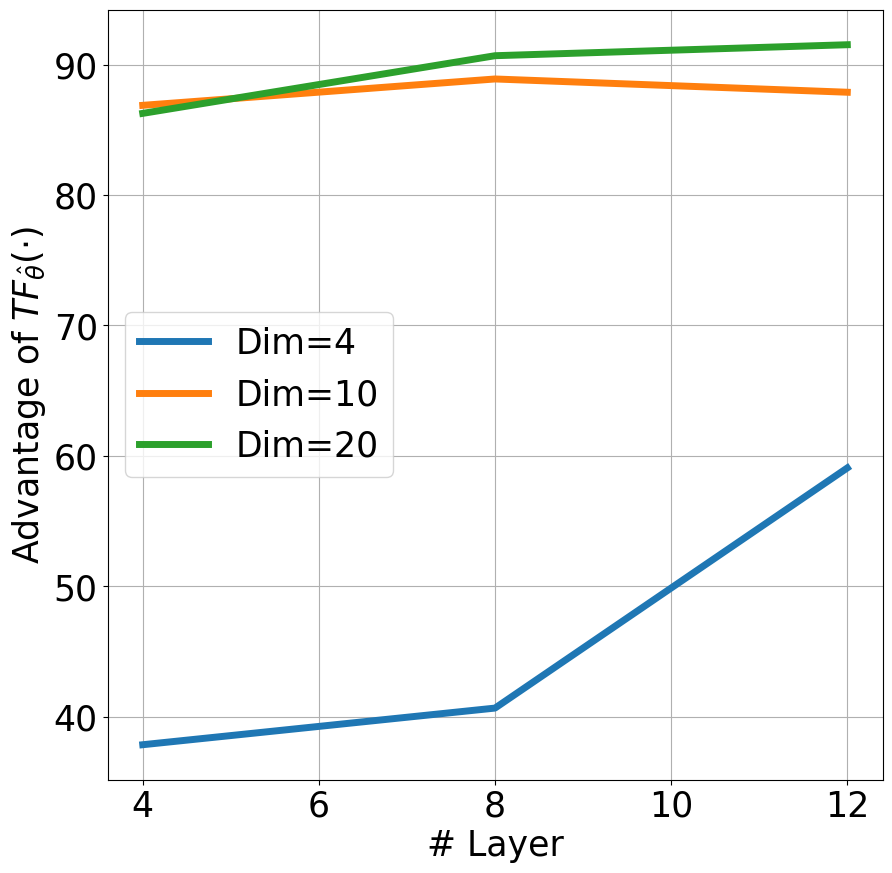}
    \caption{The advantage of OMGPT compared to the best benchmark algorithm in dynamic pricing, across different model size (number of layers) and problem complexity (problem dimensions).}
    \label{fig:aba_size}
\end{figure}

 In this section, we explore the performance changes when tuning both model size and task complexity. Specifically, we adjust the number of layers to control the model size and the problem dimension (the dimension of the context) in a dynamic pricing task to control the task complexity, where the number of unknown parameters is twice the context dimension. Since the optimal reward may scale differently across dimensions, we provide a relatively fair comparison by evaluating the advantage of OMGPT relative to the best-performing benchmark algorithm. This advantage is defined as the reward improvement rate of OMGPT compared to the best benchmark algorithm (the one among ILSE, CILS, and TS that achieves the highest reward). Thus, a positive value indicates the pre-trained OMGPT $\texttt{TF}_{\hat{\theta}}$ outperforms all benchmark algorithms, and a larger value indicates a  better performance of $\texttt{TF}_{\hat{\theta}}$. We test three model sizes (4, 8, and 12 layers) and three problem dimensions (4, 10, and 20), while keeping other hyperparameters the same. The results are presented in Figure \ref{fig:aba_size}.

From Figure \ref{fig:aba_size}, we observe the following: (i) $\texttt{TF}_{\hat{\theta}}$ consistently outperforms the benchmark algorithms, and this advantage grows as the complexity of the problem increases.  (ii) Increasing model size consistently improves the performance of $\texttt{TF}_{\hat{\theta}}$ across all problem dimensions, indicating that larger models are generally preferred.  This highlights $\texttt{TF}_{\hat{\theta}}$'s superior ability to handle more complex tasks and these observations align with the scaling law \citep{kaplan2020scaling} for large language models.

\subsubsection{Performance under Horizon Generalization} 

In practice, the testing environment may have a different time horizon from the pre-training data. In this part, we evaluate OMGPT’s generalization ability when the testing horizon is longer than that seen during pre-training. As for a shorter testing horizon, OMGPT can simply run until the end of the horizon, and Figure \ref{fig:main_reg} demonstrates strong performances in such cases, particularly when the horizon is less than 100.

\begin{figure}[ht!]
  \centering
  \begin{subfigure}[b]{0.45\textwidth}
    \centering
\includegraphics[width=\textwidth]{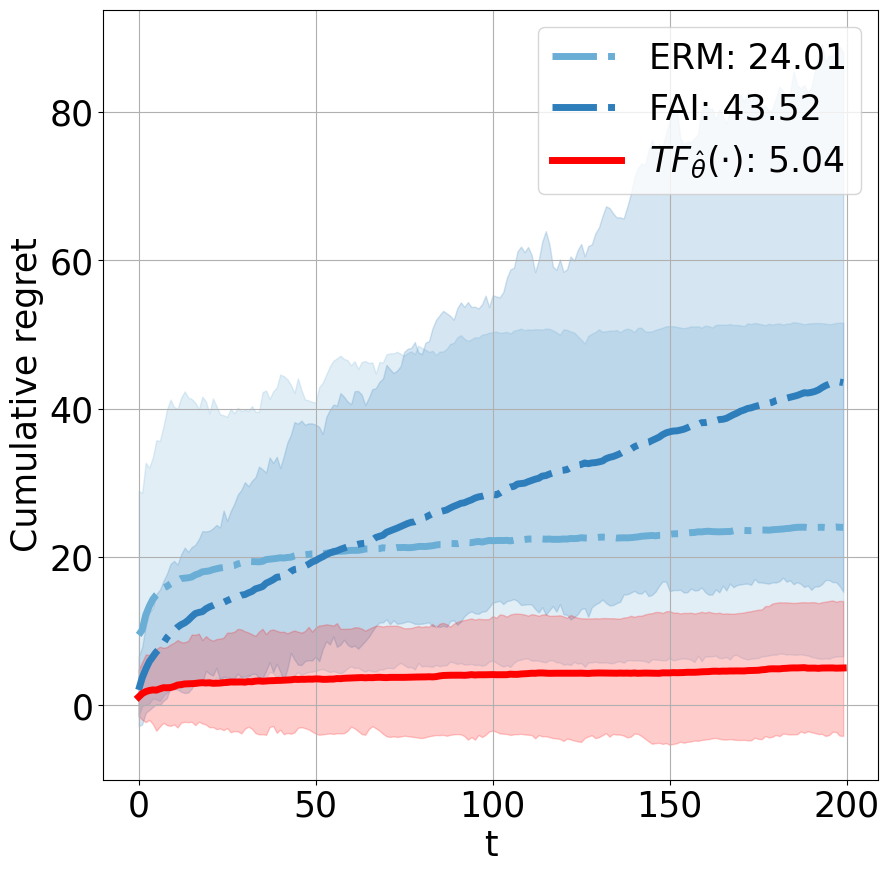}
    \caption{Uncensored demand}
  \end{subfigure}
    \hfill
    \begin{subfigure}[b]{0.45\textwidth}
    \centering
\includegraphics[width=\textwidth]{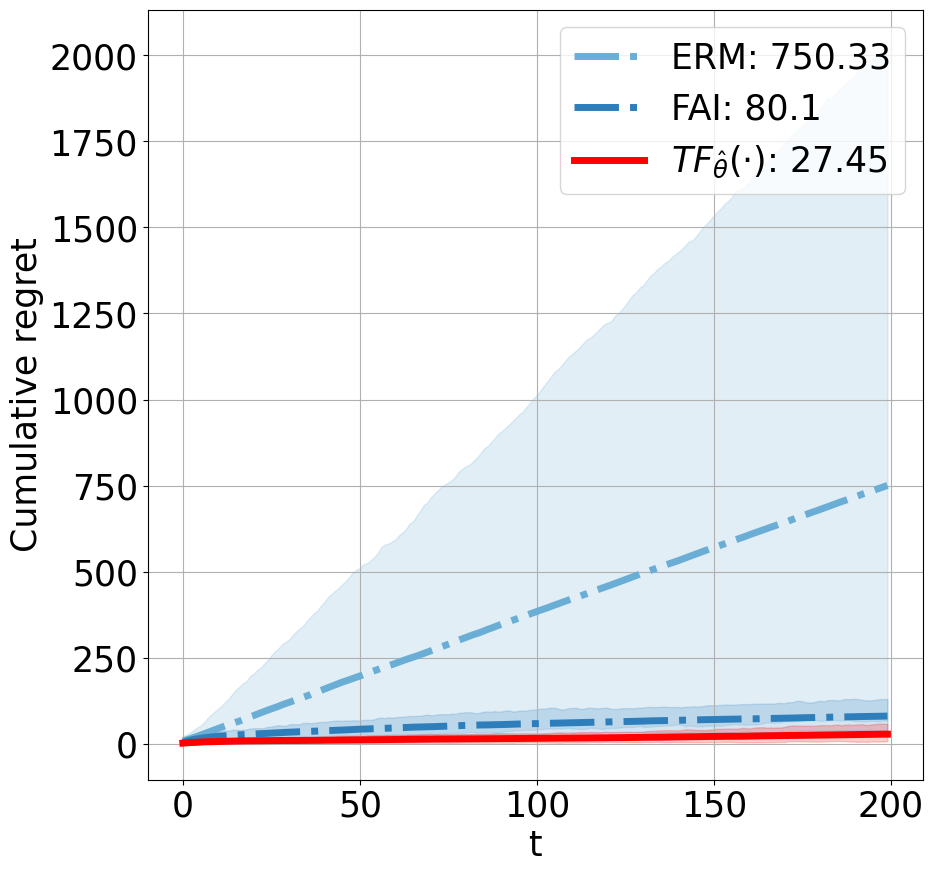}
    \caption{Censored demand}
  \end{subfigure}
  \centering
  \begin{subfigure}[b]{0.45\textwidth}
    \centering
\includegraphics[width=\textwidth]{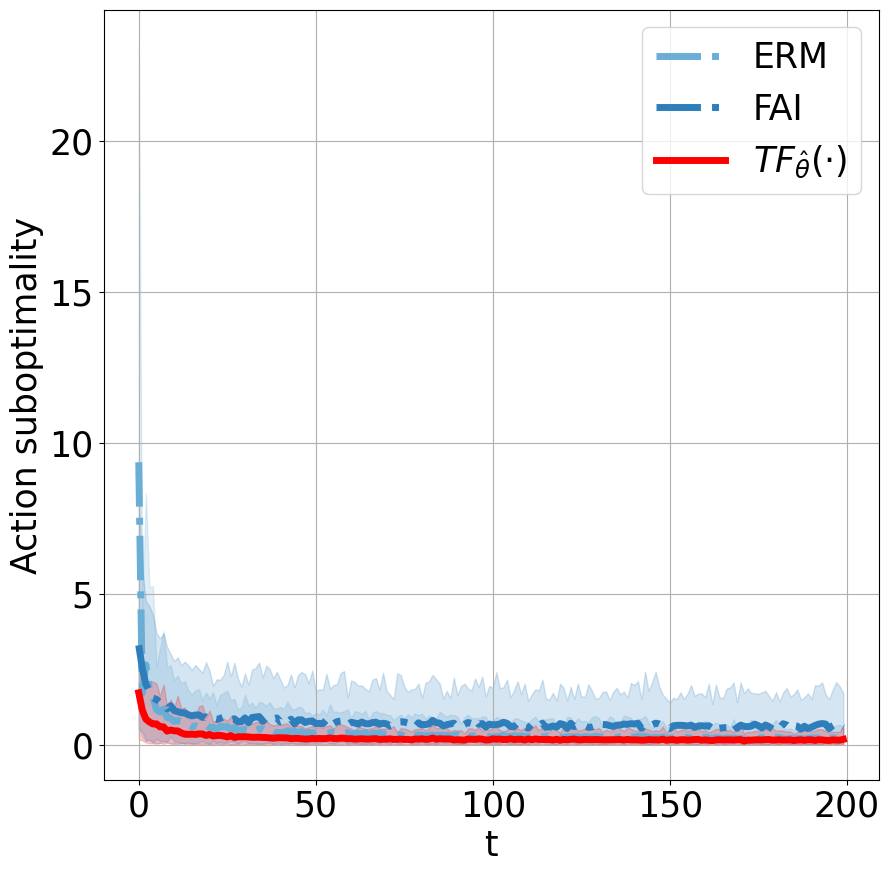}
    \caption{Uncensored demand}
  \end{subfigure}
    \hfill
    \begin{subfigure}[b]{0.45\textwidth}
    \centering
\includegraphics[width=\textwidth]{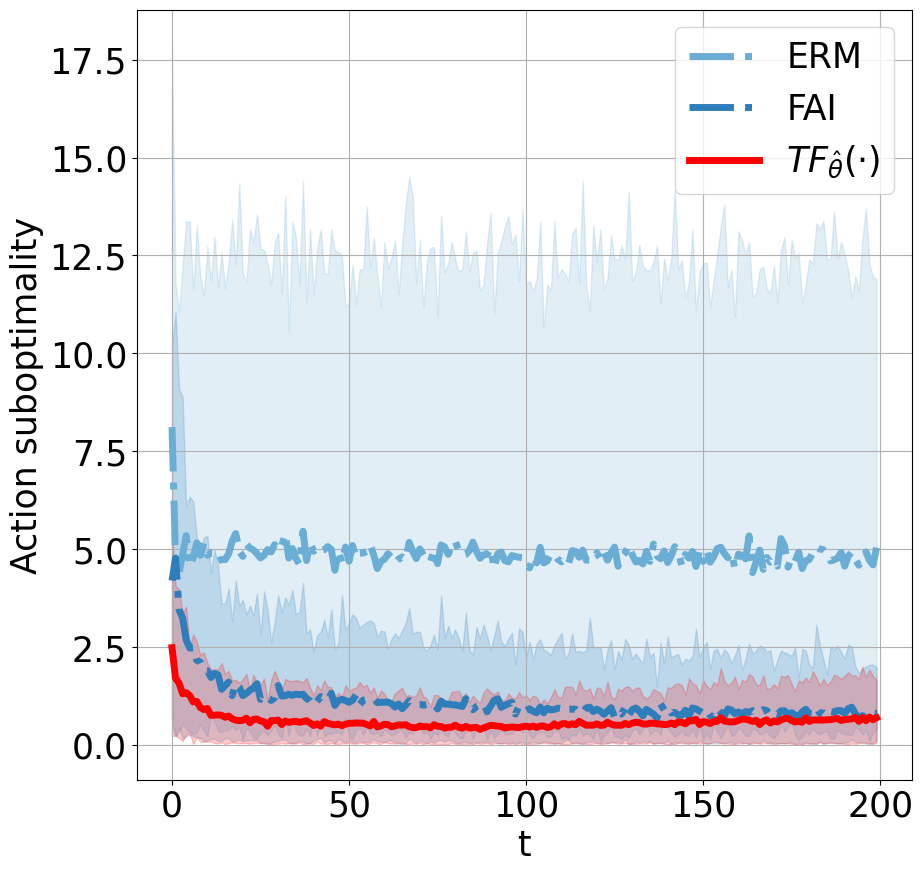}
    \caption{Censored demand}
  \end{subfigure}
  \caption{Performance under horizon generalization in newsvendor problems with uncensored and censored demands, where the pre-training samples have a horizon of 100. It shows the average out-of-sample regret (first row) and action suboptimality, i.e., $|a^*_t - \texttt{Alg}(H_t)|$, (second row) of $\texttt{TF}_{\hat{\theta}}$ against benchmark algorithms. The numbers in the legend bar are the final regret at $t=200$. }
  \label{fig:horizon_gen}
\end{figure}

Technically, during testing for $t > 100$, we limit the input sequence $H_t$ to the last 100 timesteps to predict $a^*_t$. As discussed in Section \ref{sec:disc_longer_horizon}, this restriction aligns with the input sequence length used during pre-training, enabling OMGPT to effectively handle longer testing horizons. Figure \ref{fig:horizon_gen} shows the performance when the testing horizon is extended from 100 (pre-training horizon) to 200 in the newsvendor problem with uncensored demands (first column) or censored demands (second column). In the censored setting, which requires exploration of actions as discussed in Section \ref{sec:seq_dec_mak}, we indeed evaluate whether OMGPT can maintain good performance despite the limited input sequence (i.e., the last 100 timesteps) possibly leading to forgotten explorations. The results demonstrate that OMGPT generalizes well in this extended horizon scenario, even in the censored demand setting where exploration is essential. Its actions remain nearly optimal beyond $t = 100$, resulting in lower regret compared to benchmark algorithms.

\subsubsection{Performance under Shifted Environment Distribution} 
In practice, the prior knowledge of the decision maker, $\mathcal{P}_{\gamma}$, may not exactly match the distribution of the testing environment. This is often referred to as the out-of-distribution (OOD) problem. In this section, we evaluate OMGPT’s generalization ability under such conditions, using dynamic pricing problems as the test case. Specifically, we define the demand functions in the pricing problems as $D_t = \alpha^\top X_t - \beta^\top X_t \cdot a_t + \epsilon_t$, and adjust the generation of testing environments by altering the distributions of the noise term $\epsilon_t$ and the (generation of) parameters $(\alpha, \beta)$ to deviate from those in the pre-training environments.

\begin{figure}[ht!]
  \centering
  \begin{subfigure}[b]{0.31\textwidth}
    \centering
\includegraphics[width=\textwidth]{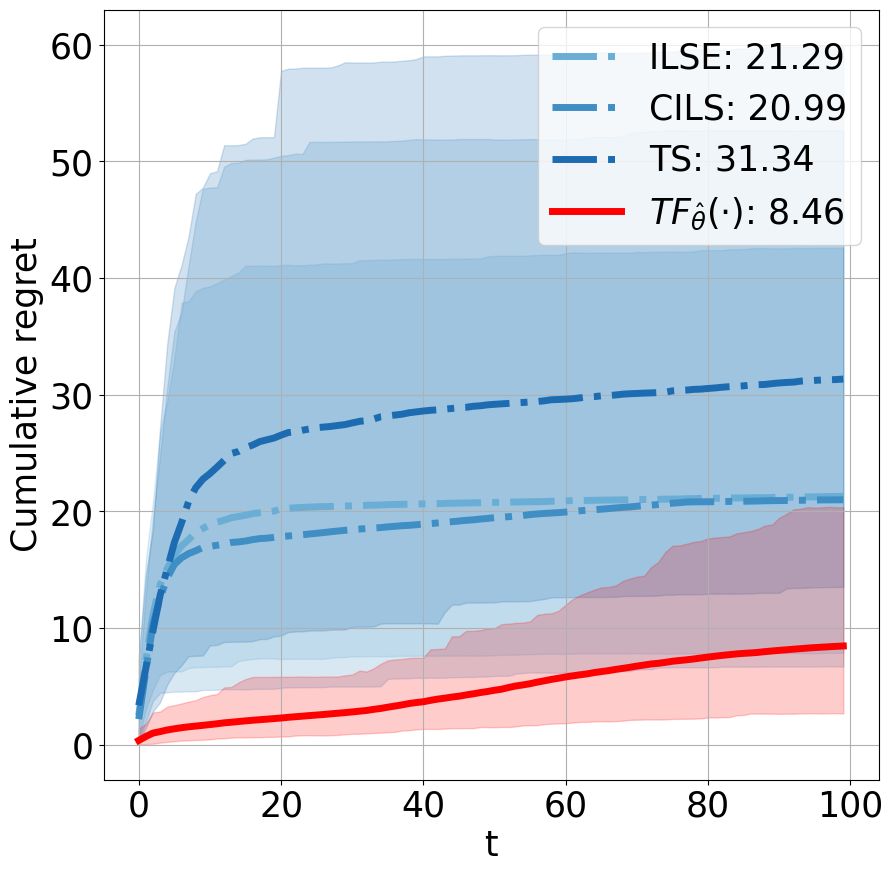}
    \caption{$\sigma^2=0.2$ ($=$ pre-training)}
  \end{subfigure}
    \hfill
    \begin{subfigure}[b]{0.31\textwidth}
    \centering
\includegraphics[width=\textwidth]{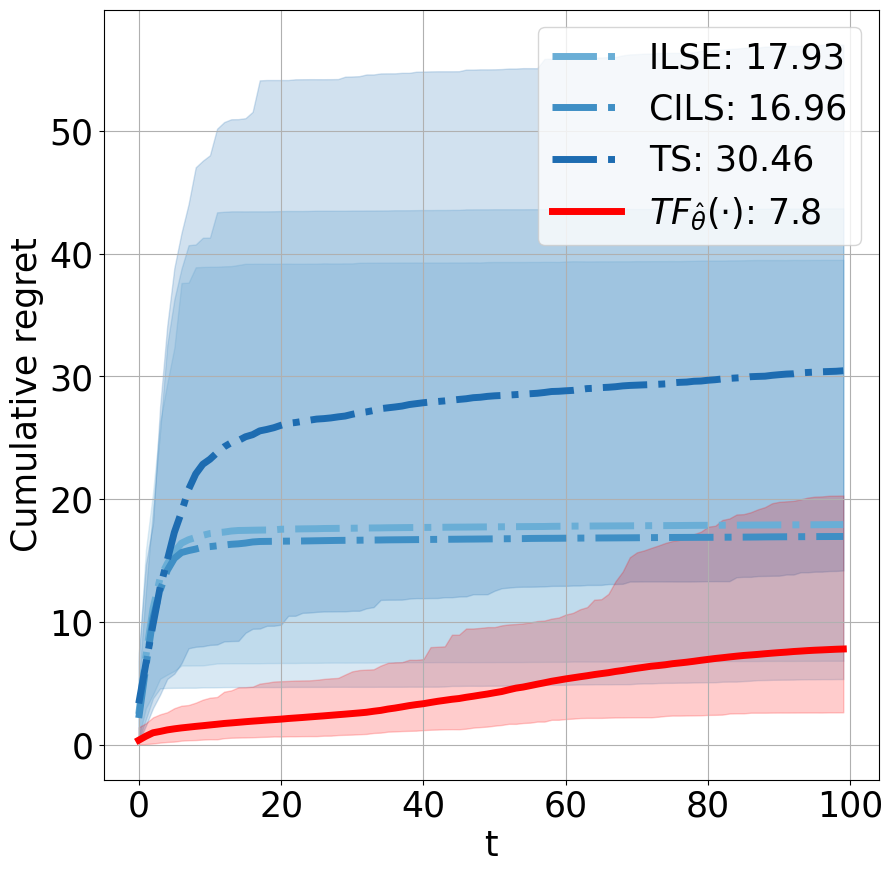}
    \caption{$\sigma^2=0.1$ ($<$ pre-training)}
  \end{subfigure}
    \hfill
    \begin{subfigure}[b]{0.31\textwidth}
    \centering
\includegraphics[width=\textwidth]{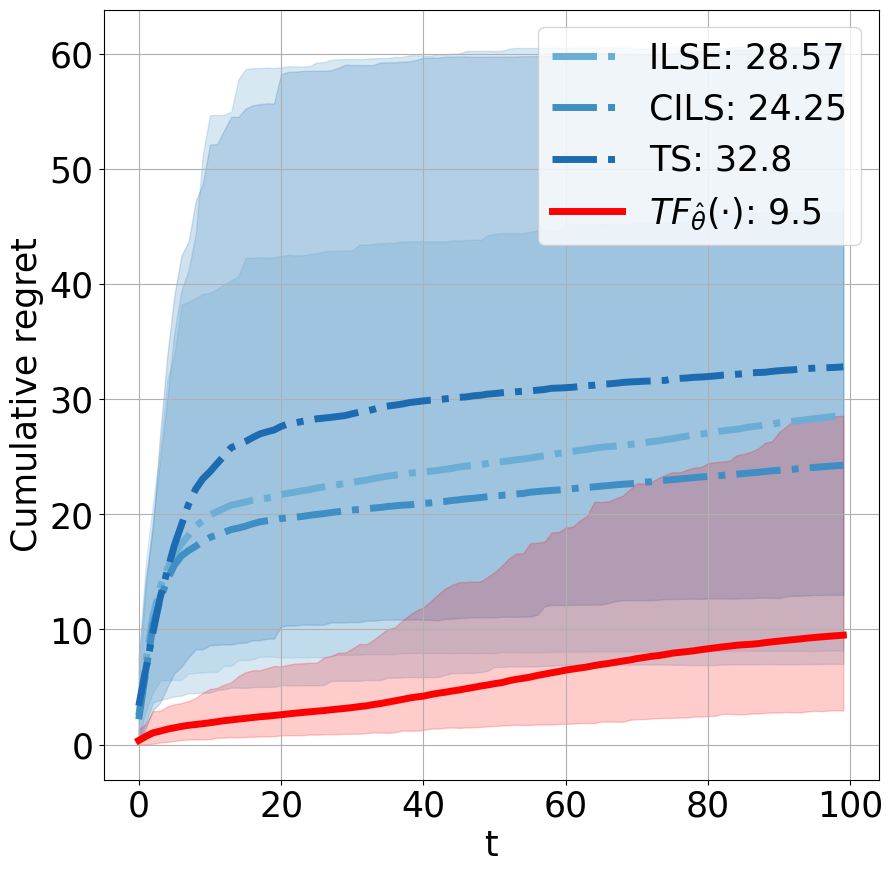}
    \caption{$\sigma^2=0.3$ ($>$ pre-training)}
  \end{subfigure}
  \begin{subfigure}[b]{0.31\textwidth}
    \centering
\includegraphics[width=\textwidth]{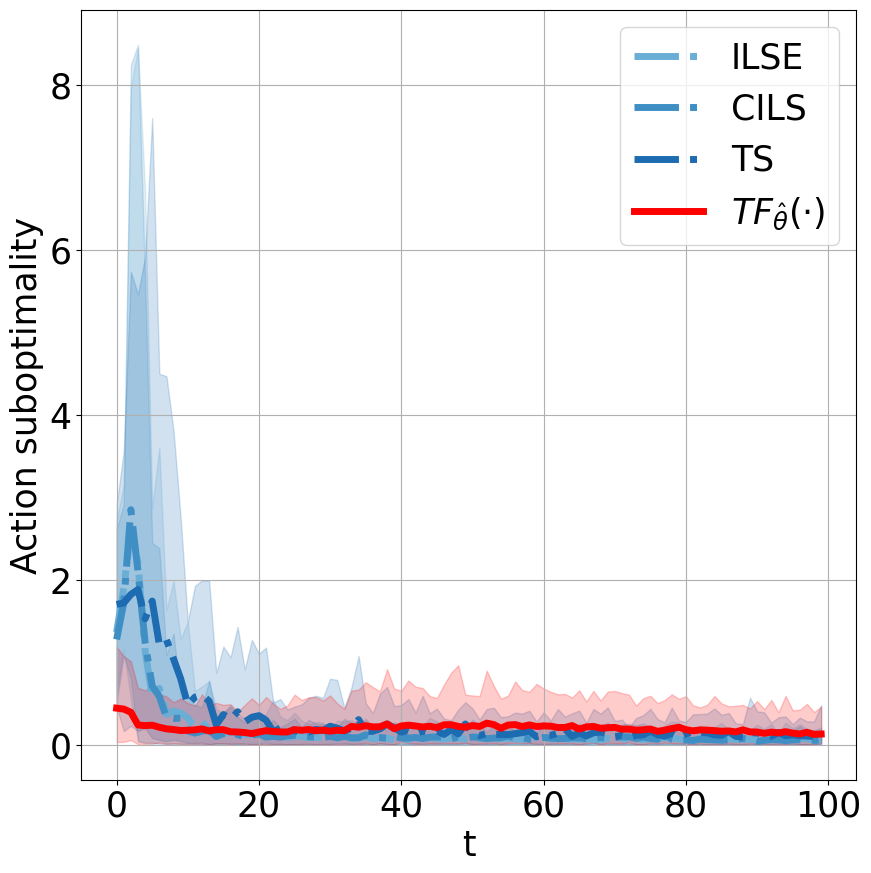}
    \caption{$\sigma^2=0.2$ ($=$ pre-training)}
  \end{subfigure}
    \hfill
    \begin{subfigure}[b]{0.31\textwidth}
    \centering
\includegraphics[width=\textwidth]{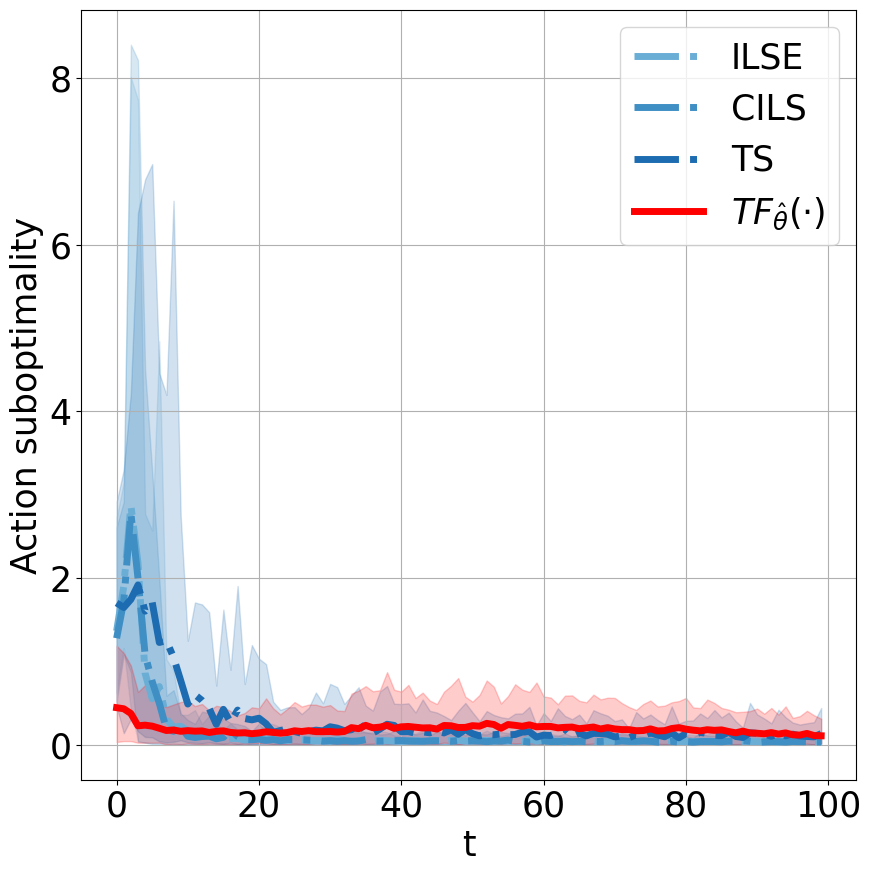}
    \caption{$\sigma^2=0.1$ ($<$ pre-training)}
  \end{subfigure}
    \hfill
    \begin{subfigure}[b]{0.31\textwidth}
    \centering
\includegraphics[width=\textwidth]{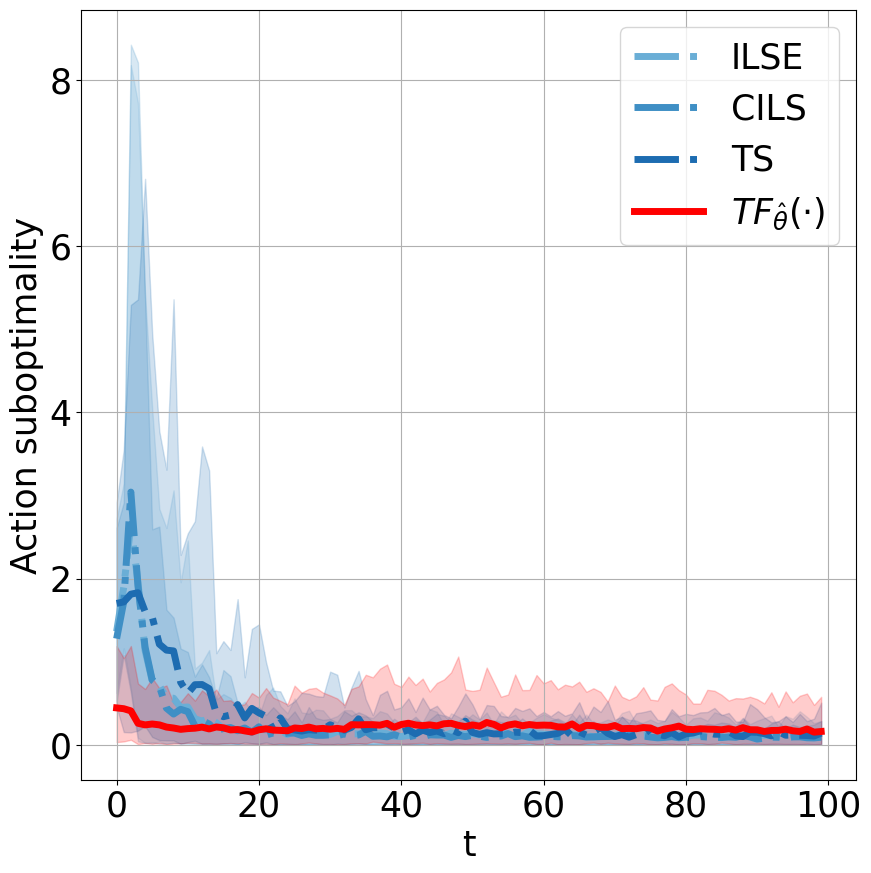}
    \caption{$\sigma^2=0.3$ ($>$ pre-training)}
  \end{subfigure}
  \caption{Performance under different testing noise variances, which may deviate from the pre-training variance of $\sigma^2 = 0.2$.  It shows the average out-of-sample regret (first row) and action suboptimality, i.e., $|a^*_t - \texttt{Alg}(H_t)|$, (second row) of $\texttt{TF}_{\hat{\theta}}$ against benchmark algorithms. The numbers in the legend bar are the final regret at $t=100$. }
  \label{fig:dist_gen}
\end{figure}

We first test OMGPT in three different problem settings with varying testing noise distributions $\epsilon_t\sim\mathcal{N}(0,\sigma^2)$ while keeping the pre-training noise $\epsilon_t\sim\mathcal{N}(0,0.2)$ for all of them: $\sigma^2 = 0.2$, which matches the noise variance used in pre-training; $\sigma^2 = 0.1$, which is smaller than that used in pre-training; and $\sigma^2 = 0.3$, which is larger. Figure \ref{fig:dist_gen} provides the results. We do not observe any significant sign of failure in OMGPT’s OOD performance: across all three settings, the benchmark algorithms consistently incur higher regret than OMGPT. Although OMGPT shows some variation in mean regret across OOD settings (specifically, lower regret when $\sigma^2 = 0.2$ and higher regret when $\sigma^2 = 0.3$), this is partially due to the varying ``difficulty'' of the underlying tasks. As expected, higher noise variances, which need more data for accurately estimating the demand function compared to lower variance cases,  lead to worse performance for all algorithms tested. However, these variations should not be interpreted as evidence of OMGPT failing to handle OOD issues. In fact, all benchmark algorithms show similar performance fluctuations under these conditions, further demonstrating OMGPT's robustness.

\begin{figure}[ht!]
  \centering
  \begin{subfigure}[b]{0.23\textwidth}
    \centering
\includegraphics[width=\textwidth]{new_figs/DP/_inf_4d_123_std_0.2_horizon_100T_False_Reg.png}
    \caption{No shift}
  \end{subfigure}
    \hfill
    \begin{subfigure}[b]{0.23\textwidth}
    \centering
\includegraphics[width=\textwidth]{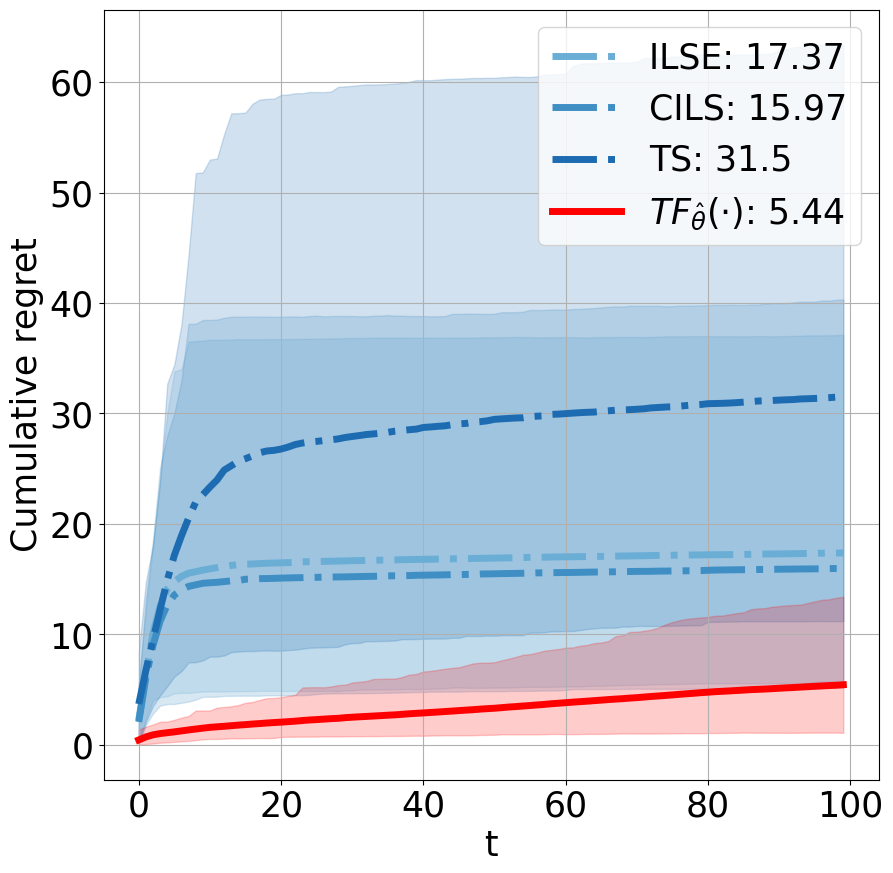}
    \caption{$\mu_{\text{shift}}=0.1$}
  \end{subfigure}
    \hfill
    \begin{subfigure}[b]{0.23\textwidth}
    \centering
\includegraphics[width=\textwidth]{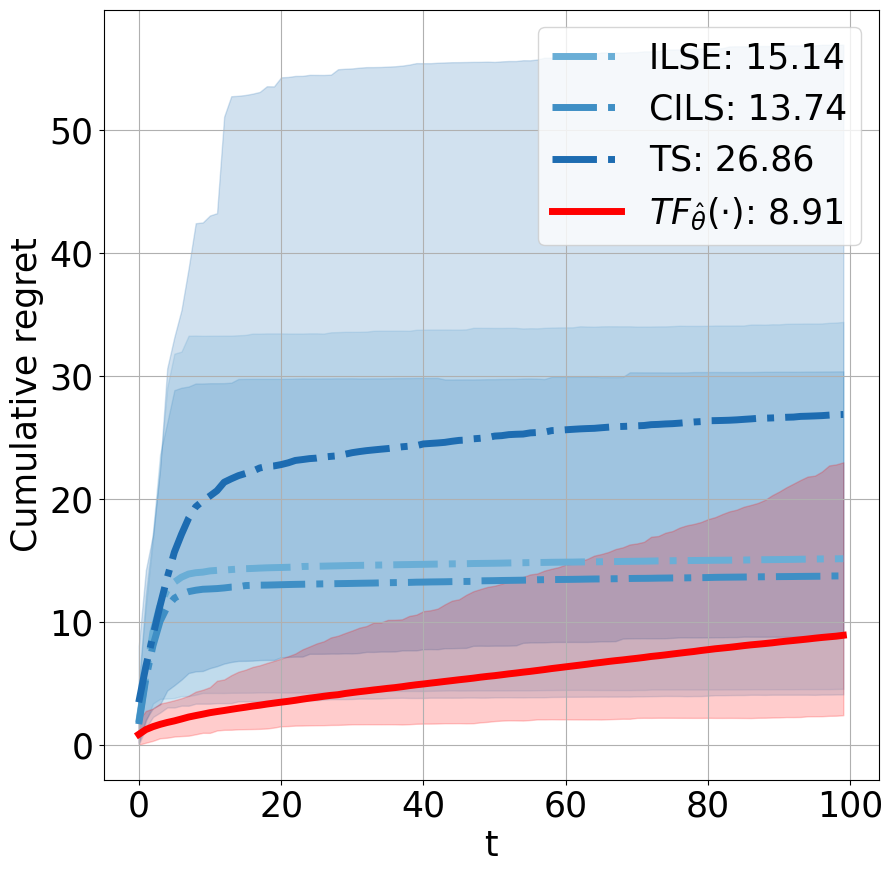}
    \caption{$\mu_{\text{shift}}=0.5$}
  \end{subfigure}
    \hfill
    \begin{subfigure}[b]{0.23\textwidth}
    \centering
\includegraphics[width=\textwidth]{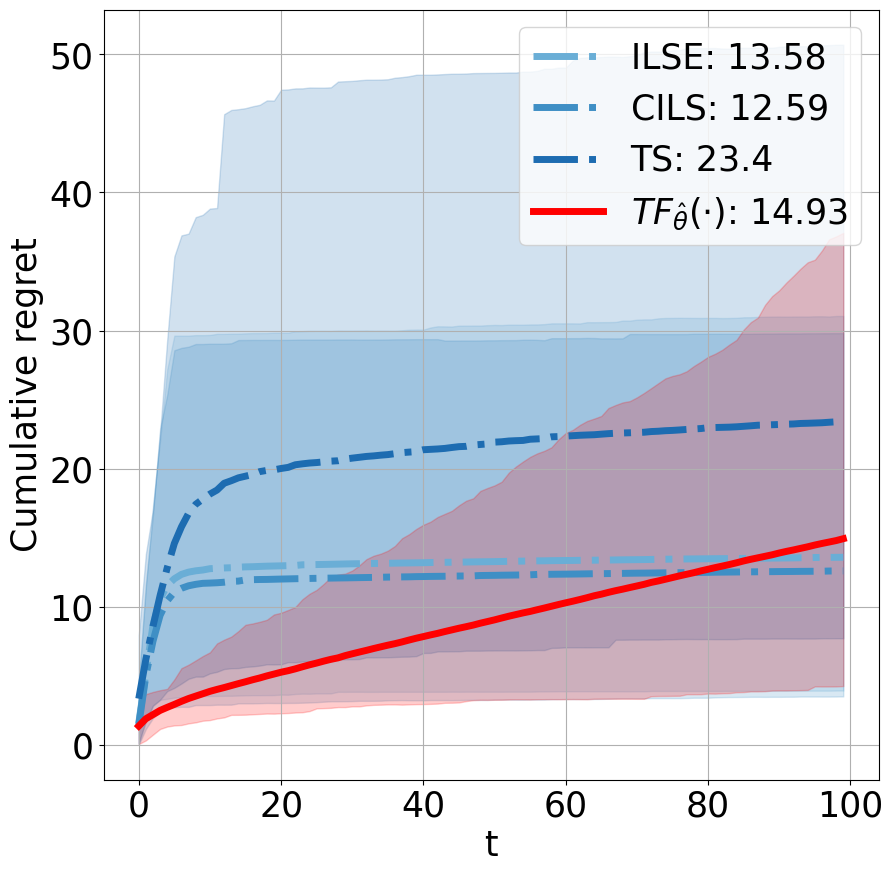}
    \caption{$\mu_{\text{shift}}=1$}
  \end{subfigure}
  \begin{subfigure}[b]{0.23\textwidth}
    \centering
\includegraphics[width=\textwidth]{new_figs/DP/_inf_4d_123_std_0.2_horizon_100T_False_Act.png}
    \caption{No shift }
  \end{subfigure}
    \hfill
    \begin{subfigure}[b]{0.23\textwidth}
    \centering
\includegraphics[width=\textwidth]{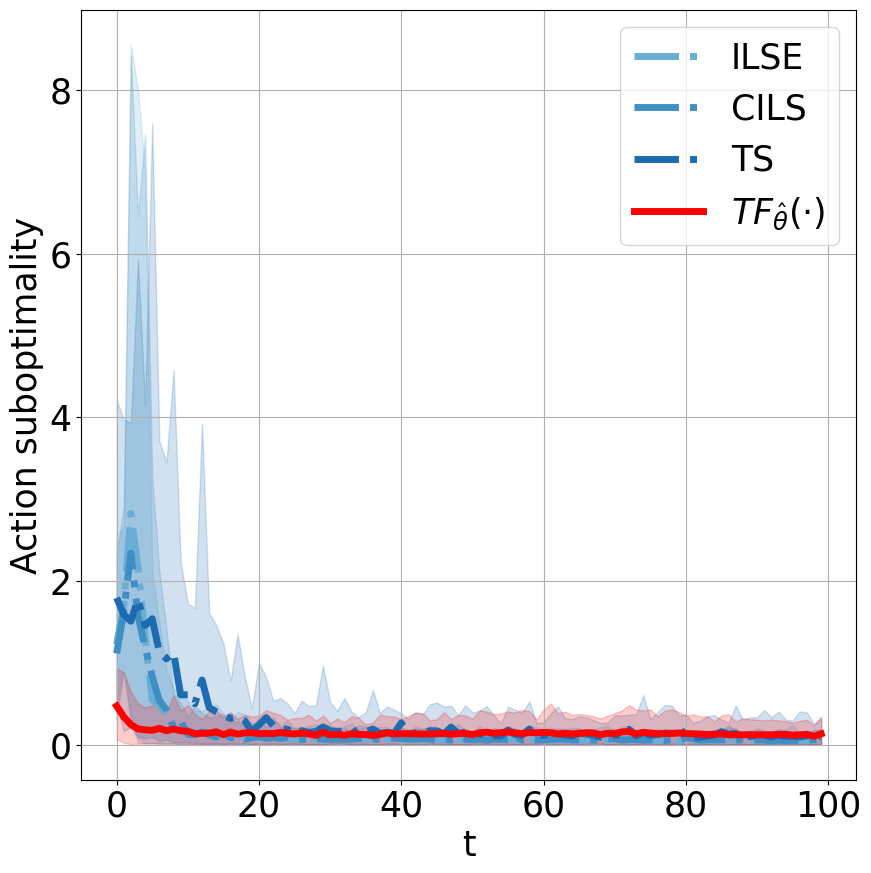}
    \caption{$\mu_{\text{shift}}=0.1$}
  \end{subfigure}
    \hfill
    \begin{subfigure}[b]{0.23\textwidth}
    \centering
\includegraphics[width=\textwidth]{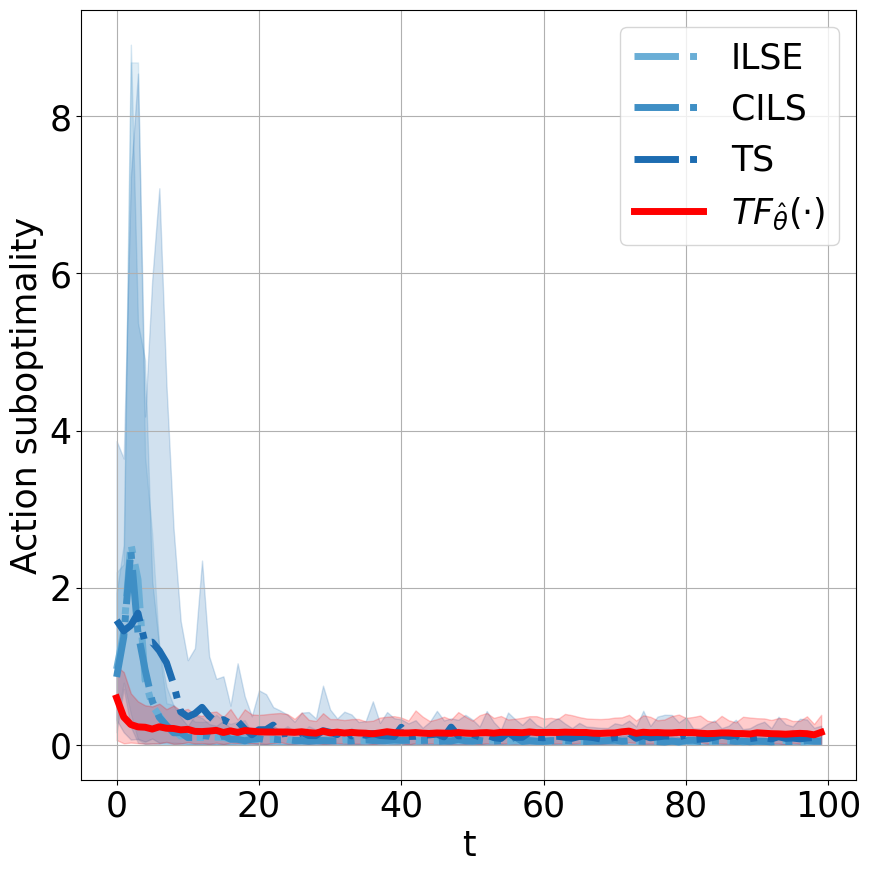}
    \caption{$\mu_{\text{shift}}=0.5$}
  \end{subfigure}
    \hfill
    \begin{subfigure}[b]{0.23\textwidth}
    \centering
\includegraphics[width=\textwidth]{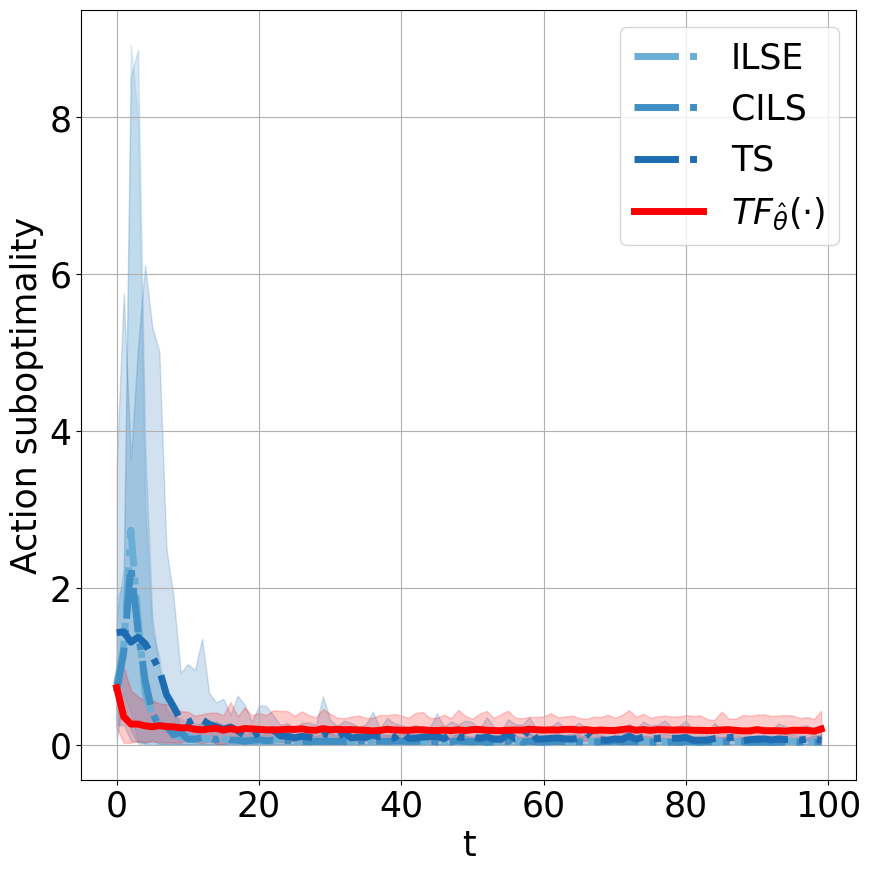}
    \caption{$\mu_{\text{shift}}=1$}
  \end{subfigure}
  \caption{Performance under the different levels of out-of-domain shifts $\mu_{\text{shift}}$ in the testing demand function generation.  It shows the average out-of-sample regret (first row) and action suboptimality, i.e., $|a^*_t - \texttt{Alg}(H_t)|$, (second row) of $\texttt{TF}_{\hat{\theta}}$ against benchmark algorithms. The numbers in the legend bar are the final regret at $t=100$. }
  \label{fig:dist_para_OOD_outdomain}
\end{figure}

\begin{figure}[ht!]
  \centering
  \begin{subfigure}[b]{0.23\textwidth}
    \centering
\includegraphics[width=\textwidth]{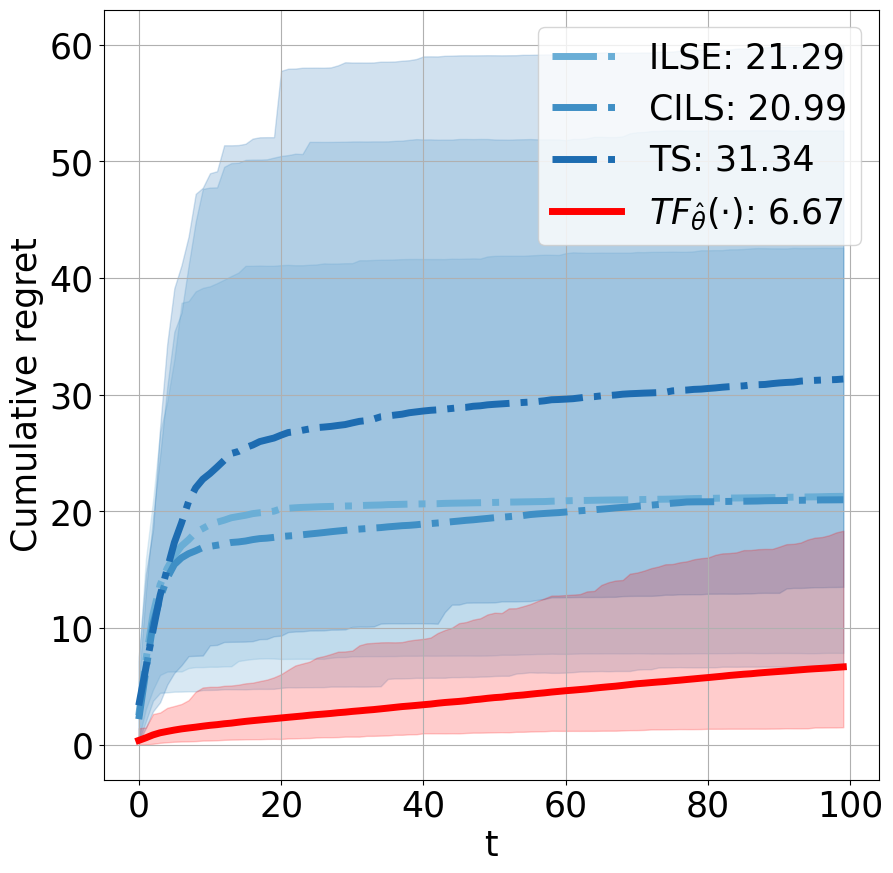}
    \caption{No shift }
  \end{subfigure}
    \hfill
    \begin{subfigure}[b]{0.23\textwidth}
    \centering
\includegraphics[width=\textwidth]{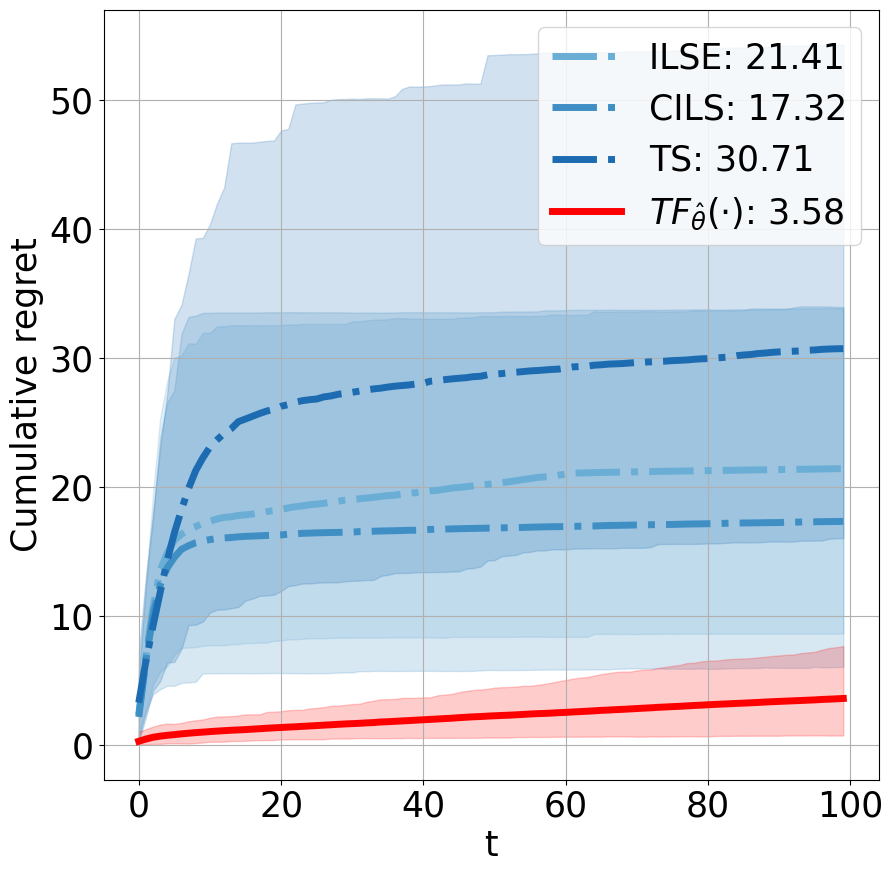}
    \caption{$\mu_{\text{shift}}=0.1$}
  \end{subfigure}
    \hfill
    \begin{subfigure}[b]{0.23\textwidth}
    \centering
\includegraphics[width=\textwidth]{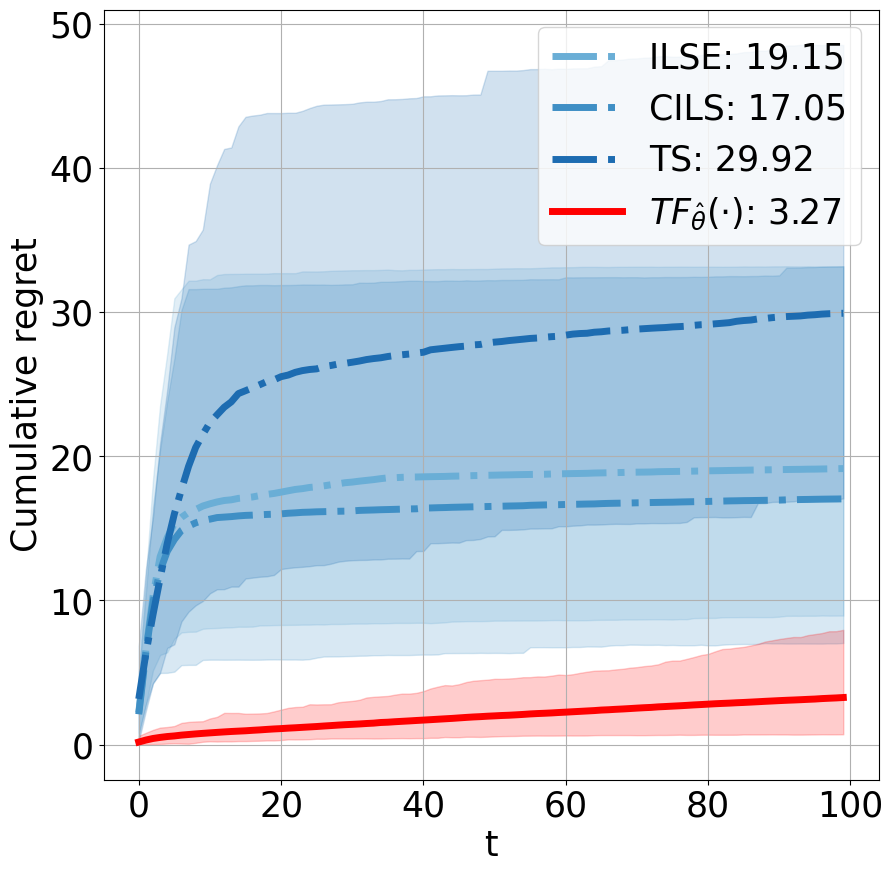}
    \caption{$\mu_{\text{shift}}=0.2$}
  \end{subfigure}
    \hfill
    \begin{subfigure}[b]{0.23\textwidth}
    \centering
\includegraphics[width=\textwidth]{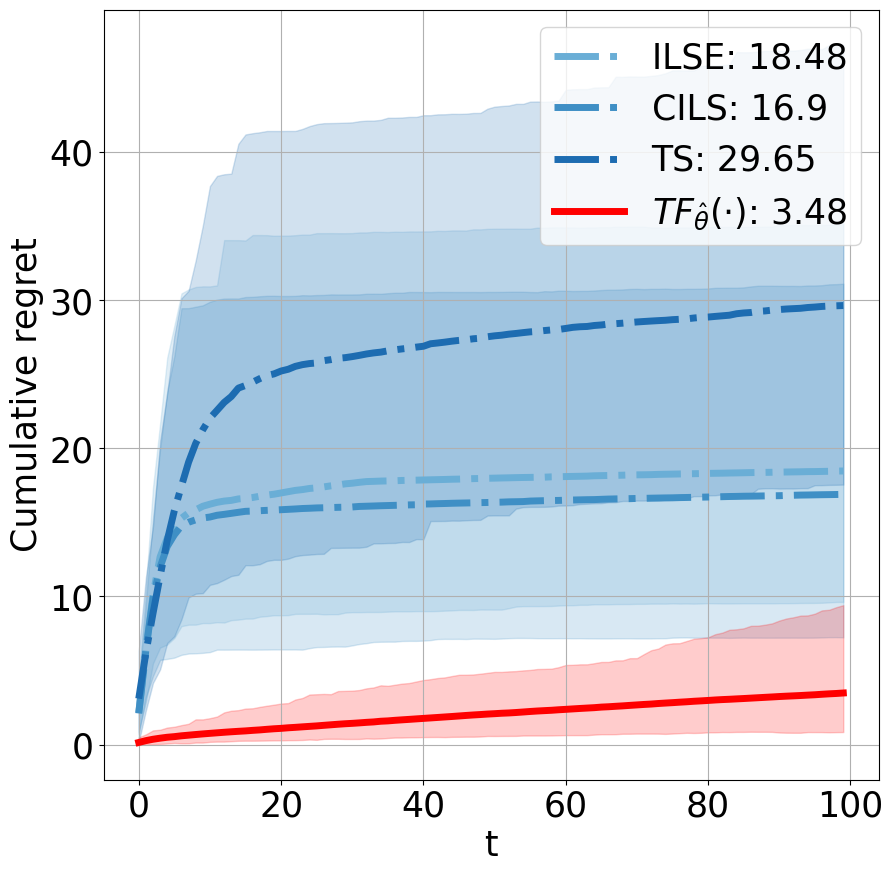}
    \caption{$\mu_{\text{shift}}=0.3$}
  \end{subfigure}
  \begin{subfigure}[b]{0.23\textwidth}
    \centering
\includegraphics[width=\textwidth]{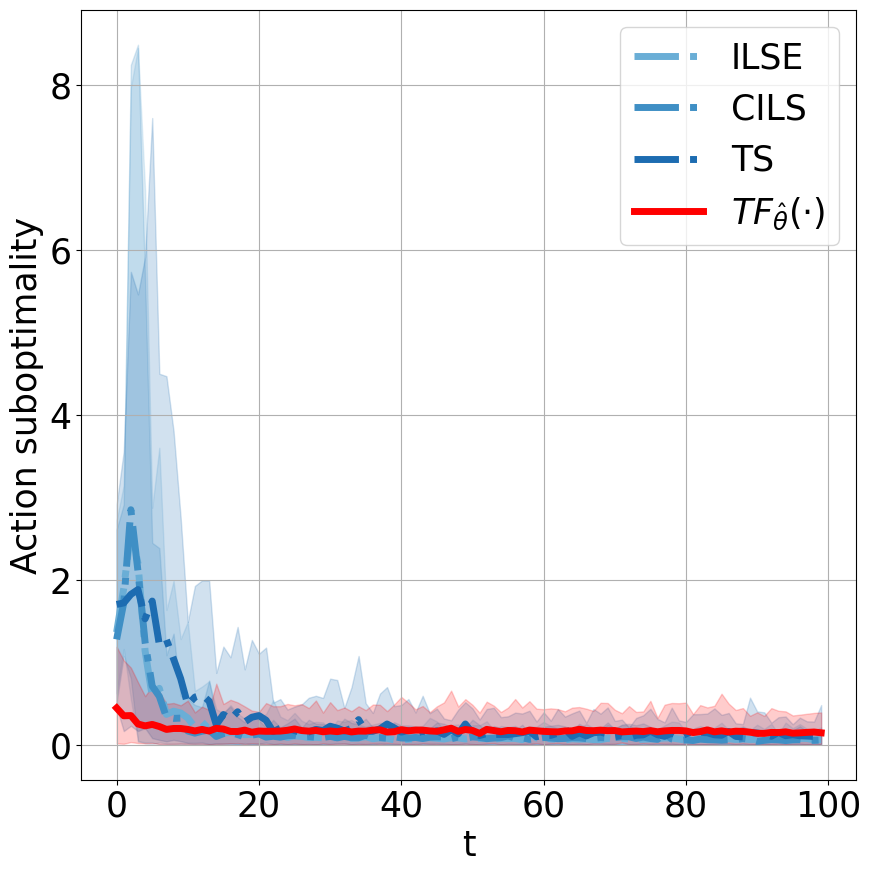}
    \caption{No shift }
  \end{subfigure}
    \hfill
    \begin{subfigure}[b]{0.23\textwidth}
    \centering
\includegraphics[width=\textwidth]{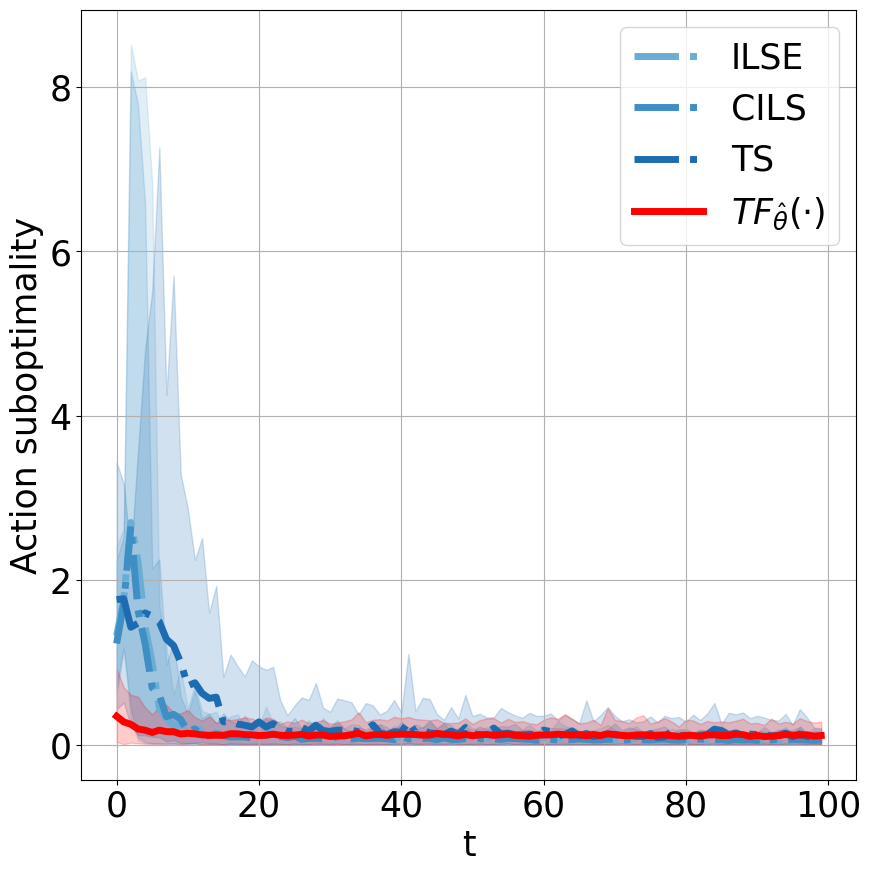}
    \caption{$\mu_{\text{shift}}=0.1$}
  \end{subfigure}
    \hfill
    \begin{subfigure}[b]{0.23\textwidth}
    \centering
\includegraphics[width=\textwidth]{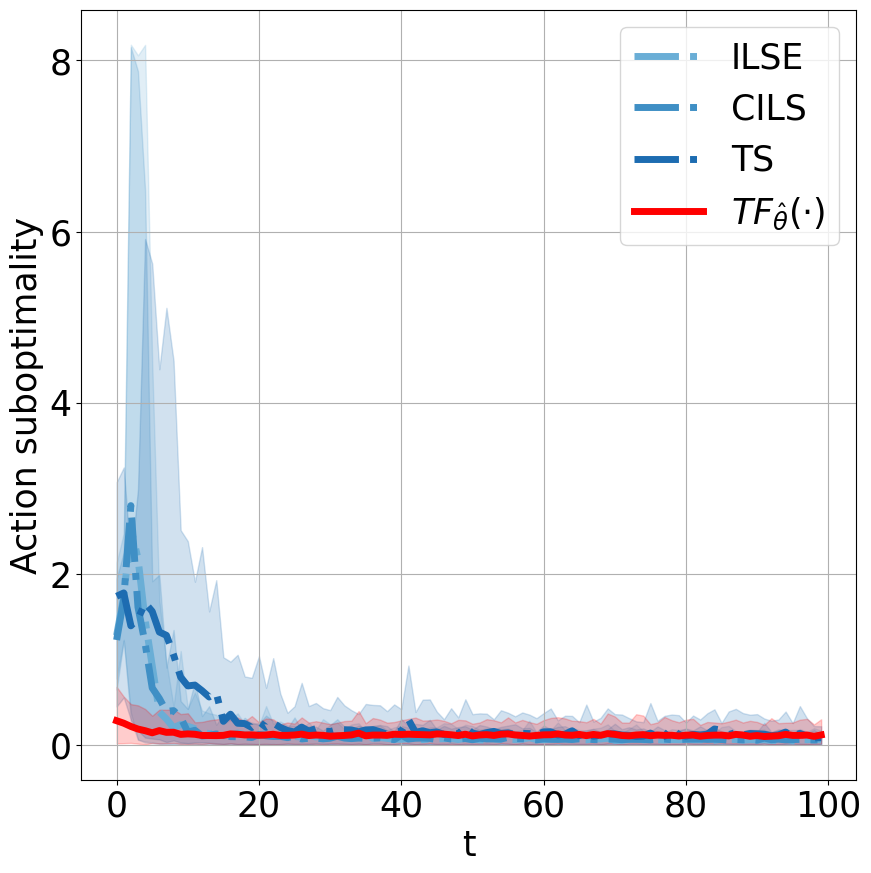}
    \caption{$\mu_{\text{shift}}=0.2$}
  \end{subfigure}
    \hfill
    \begin{subfigure}[b]{0.23\textwidth}
    \centering
\includegraphics[width=\textwidth]{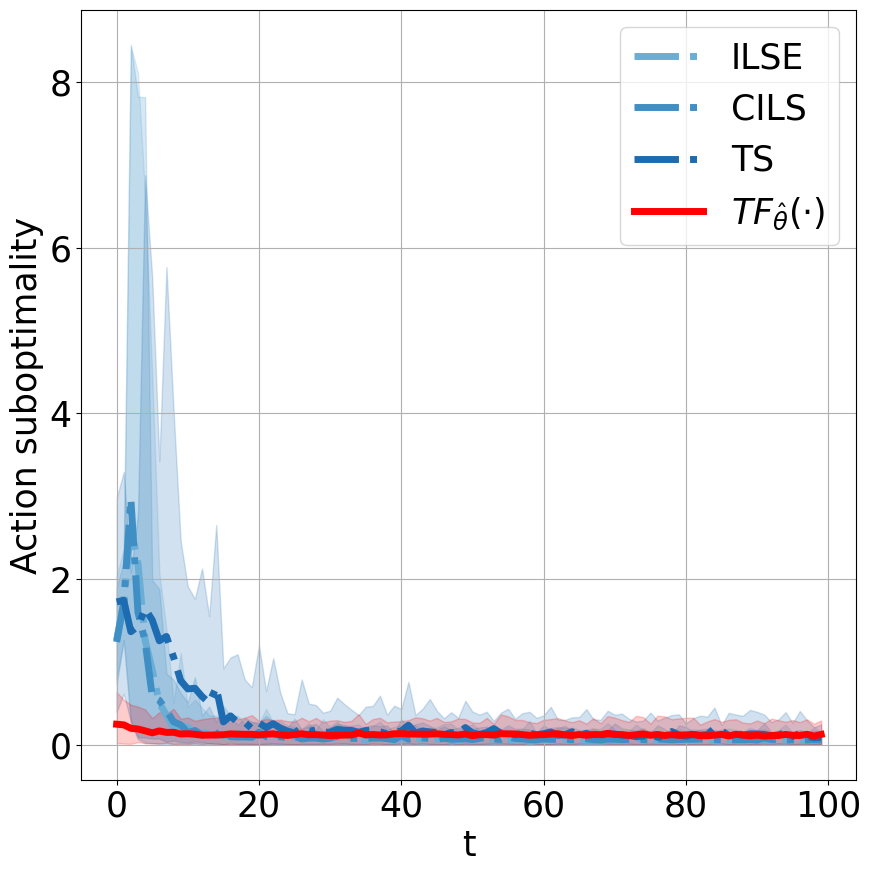}
    \caption{$\mu_{\text{shift}}=0.3$}
  \end{subfigure}
  \caption{Performance under the different levels of in-domain shifts $\mu_{\text{shift}}$ in the testing demand function generation.  It shows the average out-of-sample regret (first row) and action suboptimality, i.e., $|a^*_t - \texttt{Alg}(H_t)|$, (second row) of $\texttt{TF}_{\hat{\theta}}$ against benchmark algorithms. The numbers in the legend bar are the final regret at $t=100$. }
  \label{fig:dist_para_OOD}
\end{figure}

We further evaluate OMGPT under distributional shifts in the generation of parameters $(\alpha, \beta)$, considering both out-of-domain and in-domain shifts. During pre-training, these parameters are sampled per environment as $\alpha \sim \text{Unif}([0.5, 1.5]^6)$ and $\beta \sim \text{Unif}([0.05, 1.05]^6)$. In the testing phase, two types of shifts are introduced, parameterized by the shift level $\mu_{\text{shift}}$:
\begin{itemize}
    \item \textbf{Out-of-domain shifts}: Here, the test parameters $(\alpha, \beta)$ can be sampled beyond the original training ranges to simulate scenarios where prior knowledge fails to encompass the true environment space. Specifically, we generate $\alpha \sim \text{Unif}([0.5+\mu_{\text{shift}}, 1.5+\mu_{\text{shift}}]^6)$ and $\beta \sim \text{Unif}([0.05+\mu_{\text{shift}}, 1.05+\mu_{\text{shift}}]^6)$. Four shift levels are evaluated: $\mu_{\text{shift}}=0$ (matching the pre-training distribution) and $\mu_{\text{shift}}=0.1, 0.5, 1$. 
    \item \textbf{In-domain shifts}: For these shifts,  sub-intervals of length $(1 - \mu_{\text{shift}})$ within the original training ranges are randomly selected, and parameters are sampled uniformly within these sub-intervals. Specifically, for a given $\mu_{\text{shift}}$, we sample $\kappa_1 \sim \text{Unif}([0.5, 1.5 - (1 - \mu_{\text{shift}})])$ and $\kappa_2 \sim \text{Unif}([0.05, 1.05 - (1 - \mu_{\text{shift}})])$. Then, we draw $\alpha \sim \text{Unif}([\kappa_1, \kappa_1 + 1 - \mu_{\text{shift}}]^6)$ and $\beta \sim \text{Unif}([\kappa_2, \kappa_2 + 1 - \mu_{\text{shift}}]^6)$. This approach simulates a scenario in which the prior knowledge is conservative, assuming a broader feasible space than the true space that generates the testing environment. A larger $\mu_{\text{shift}}$ corresponds to more conservative prior knowledge, implying a larger expected feasible parameter space than the actual one. We consider four shift levels: $\mu_{\text{shift}}=0$ (matching the pre-training distribution) and $\mu_{\text{shift}}=0.1, 0.2, 0.3$. 
\end{itemize}

Figure \ref{fig:dist_para_OOD_outdomain} and Figure \ref{fig:dist_para_OOD}
illustrate OMGPT’s performance under out-of-domain and in-domain shifts, respectively. For out-of-domain shifts (Figure \ref{fig:dist_para_OOD_outdomain}), we observe a performance decline at higher shift levels ($\mu_{\text{shift}}=0.5, 1$) in $(\alpha, \beta)$. This effect is reasonable, as $(\alpha, \beta)$ directly determines the optimal price, $\frac{\alpha^\top X_t}{2\beta^\top X_t}$. Therefore, the parameters sampled outside the training distribution's support may lead to optimal decisions that the model has never encountered during training, resulting in performance degradation. For in-domain shifts (Figure \ref{fig:dist_para_OOD}), we observe no significant indications of failure of OMGPT in this case.   Interestingly, the regret performance of OMGPT and benchmark algorithms improves slightly under these shifts. This enhancement may stem from the shifts leading to $(\alpha, \beta)$ values that are more centered within the range, thereby reducing the occurrence of parameters near the original boundaries that would otherwise be too large or too small. As a result, this shift reduces the likelihood of encountering ``corner'' environments where optimal decisions involve extreme values. 
 
In practice, we recommend adopting more conservative prior knowledge to generate a broader range of training environments, as this approach minimizes potential distortions in out-of-domain performance, as observed in Figure \ref{fig:dist_para_OOD_outdomain}, while only potentially incurring little sacrifice as suggested by Figure \ref{fig:dist_para_OOD}.

\subsubsection{Solution to Model Misspecification}

Most sequential decision making algorithms, including all benchmark algorithms used in our experiments, rely on structural or model assumptions about the underlying tasks. For example, in pricing and newsvendor problems, demand functions are typically assumed to be linear with respect to the context. When these algorithms are applied to misspecified environments, where such assumptions do not hold, the performance can degrade significantly.

In contrast, OMGPT provides a potential solution to model misspecification. For example, for newsvendor problems where both linear and non-linear demand functions can appear, by generating pre-training samples from both types of environments, a single pre-trained $\texttt{TF}_{\hat{\theta}}$ can leverage its large capacity to make near-optimal decisions across different types of environments.

In this section, we test OMGPT in two different scenarios: non-stationary environments and environments with multiple demand types. Specifically, we pre-train and test $\texttt{TF}_{\hat{\theta}}$  on environments where (i) demand functions may abruptly change at a random time $\tau \leq T$ (non-stationary setting) or (ii) there are two types of demand functions: linear and square (detailed in Appendix \ref{appx:envs}). We compare $\texttt{TF}_{\hat{\theta}}$ with benchmark algorithms that are well-studied but specifically designed for stationary, linear demand cases.

\begin{figure}[ht!]
  \centering
  \begin{subfigure}[b]{0.45\textwidth}
    \centering
\includegraphics[width=\textwidth]{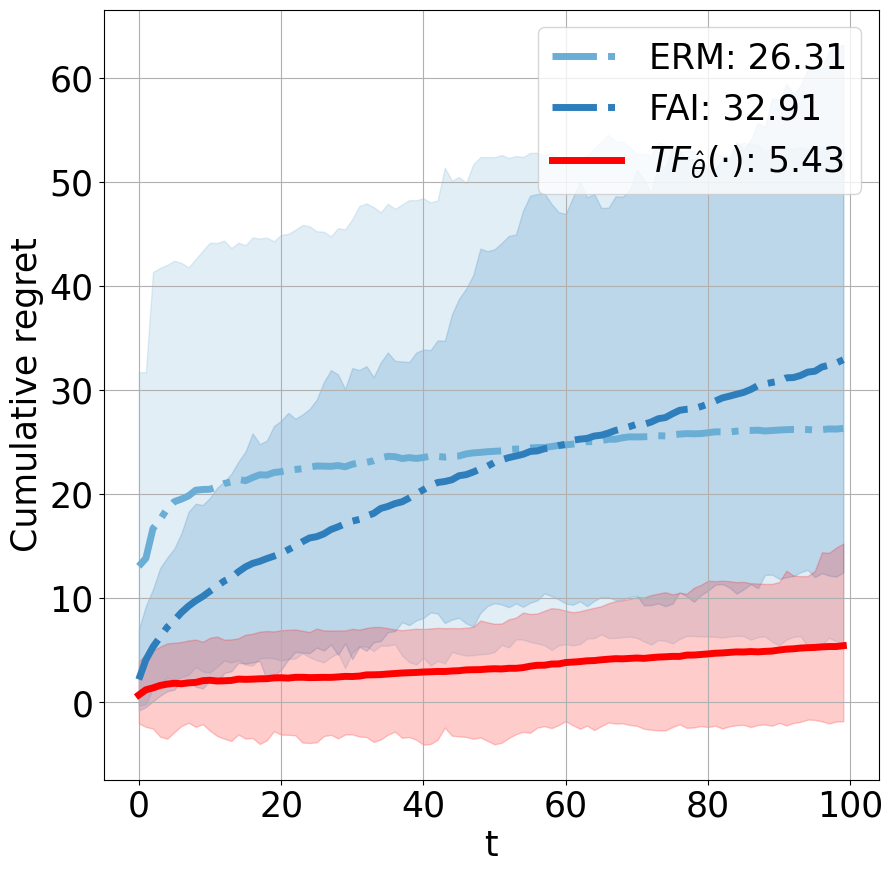}
    \caption{Stationary environment (no misspecification)}
  \end{subfigure}
  \hfill
  \begin{subfigure}[b]{0.45\textwidth}
    \centering
\includegraphics[width=\textwidth]{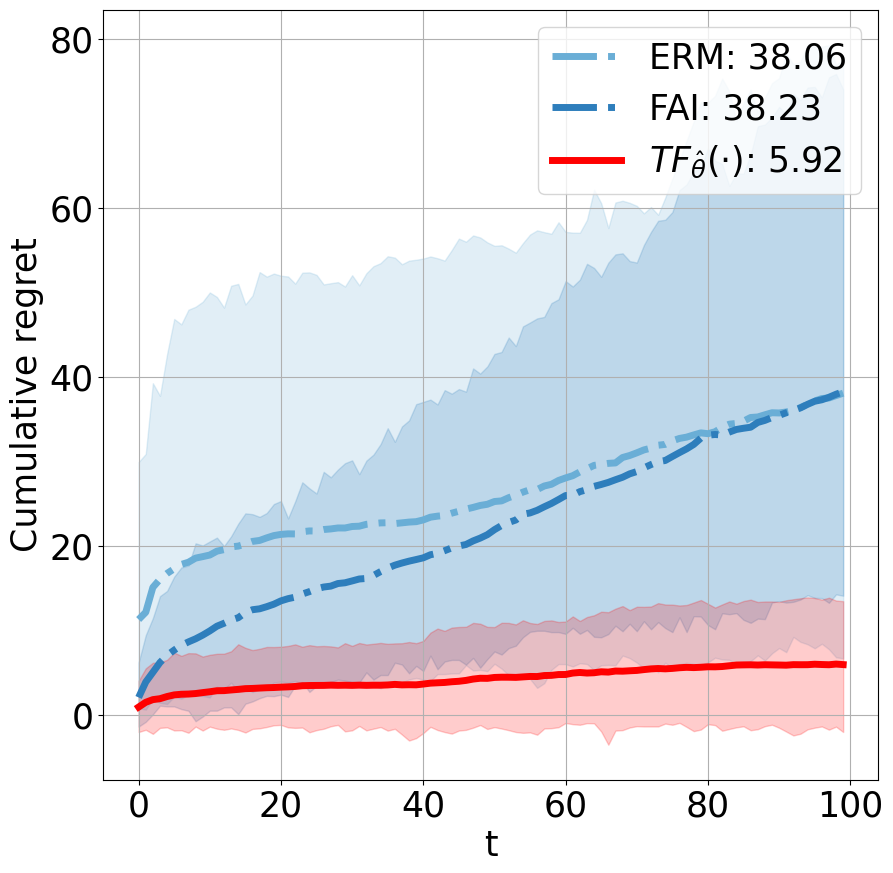}
    \caption{Nonstationary environment (misspecification)}
  \end{subfigure}

  \begin{subfigure}[b]{0.45\textwidth}
    \centering
\includegraphics[width=\textwidth]{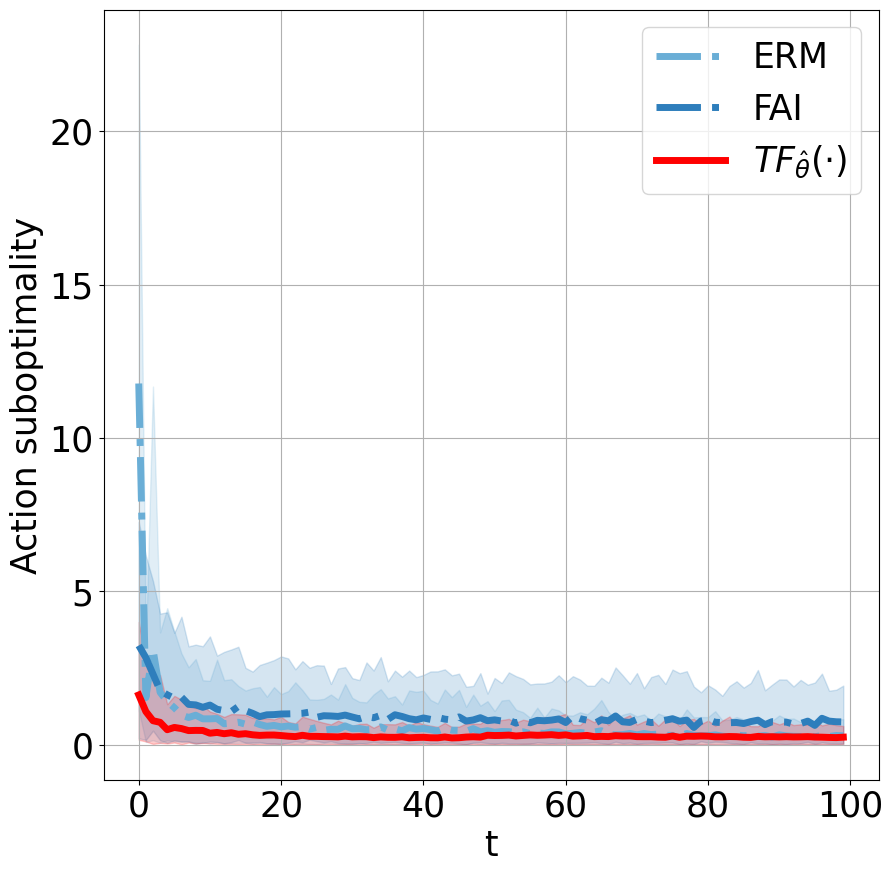}
    \caption{Stationary environment (no misspecification)}
  \end{subfigure}
  \hfill
  \begin{subfigure}[b]{0.45\textwidth}
    \centering
\includegraphics[width=\textwidth]{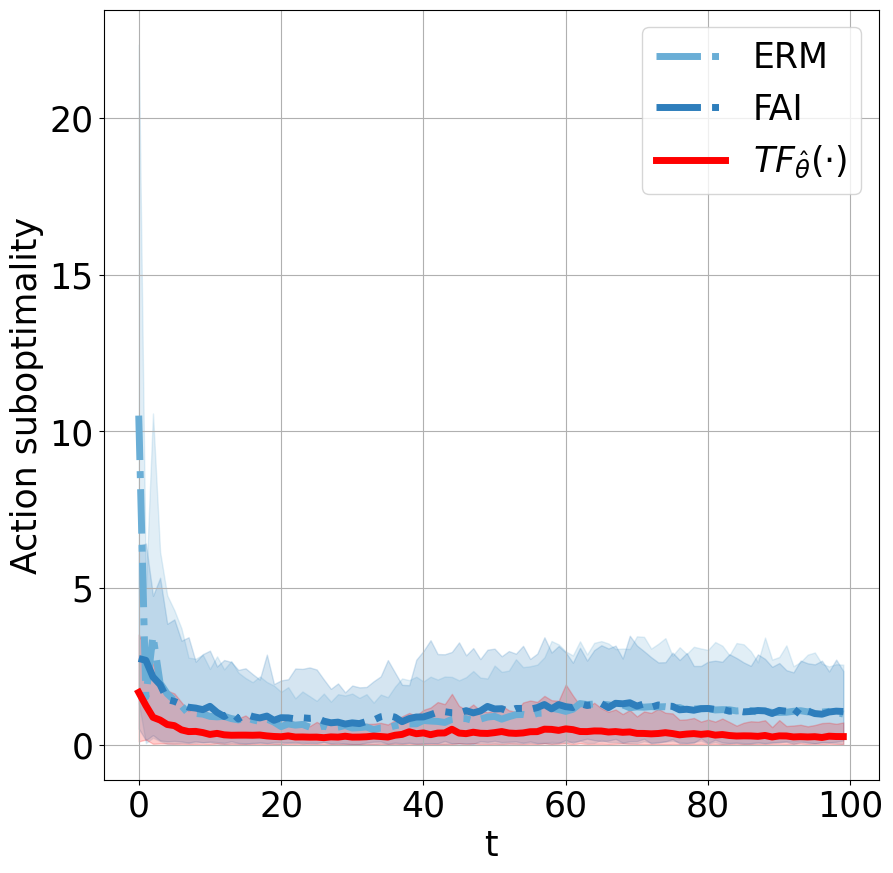}
    \caption{Nonstationary environment (misspecification)}
  \end{subfigure}
  \caption{Performance in stationary environments ((a), (c)) and non-stationary environments ((b), (d)), where benchmark algorithms encounter model misspecification issues. The same $\texttt{TF}_{\hat{\theta}}$ is applied in all figures. It shows the average out-of-sample regret (first row) and action suboptimality, i.e., $|a^*_t - \texttt{Alg}(H_t)|$, (second row) of $\texttt{TF}_{\hat{\theta}}$ against benchmark algorithms. The numbers in the legend bar are the final regret at $t=100$.}
  \label{fig:modelmis_station}
\end{figure}

\begin{figure}[ht!]
  \centering
  \begin{subfigure}[b]{0.45\textwidth}
    \centering
\includegraphics[width=\textwidth]{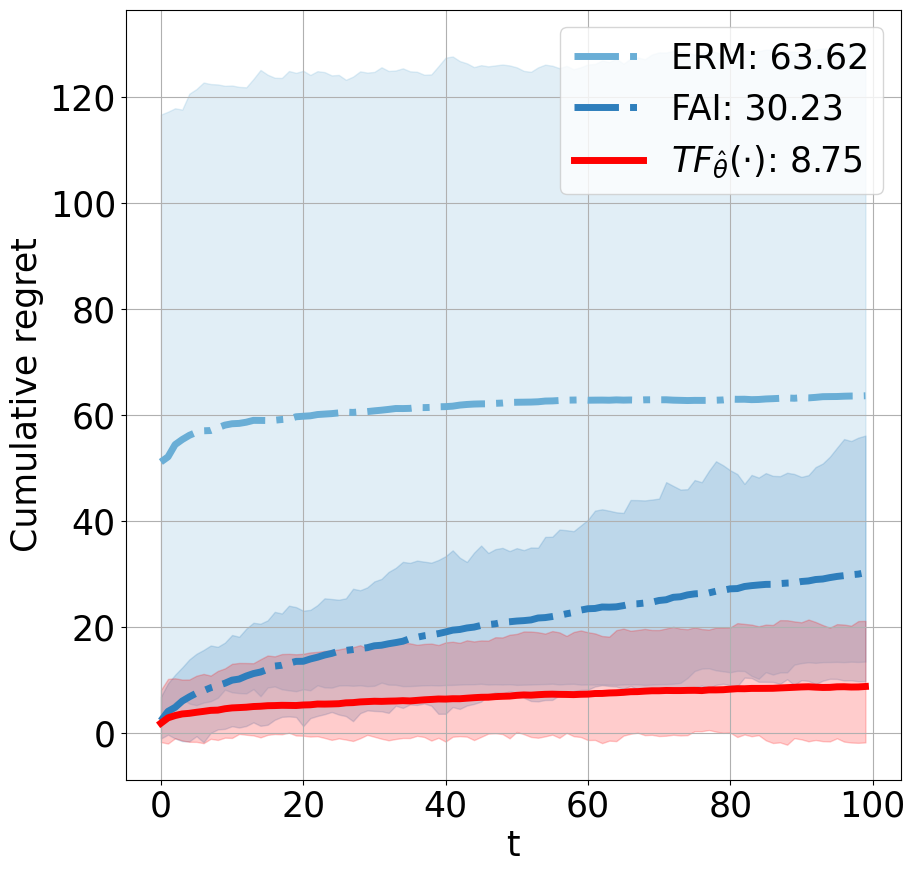}
    \caption{Linear demand (no misspecification)}
  \end{subfigure}
  \hfill
  \begin{subfigure}[b]{0.45\textwidth}
    \centering
\includegraphics[width=\textwidth]{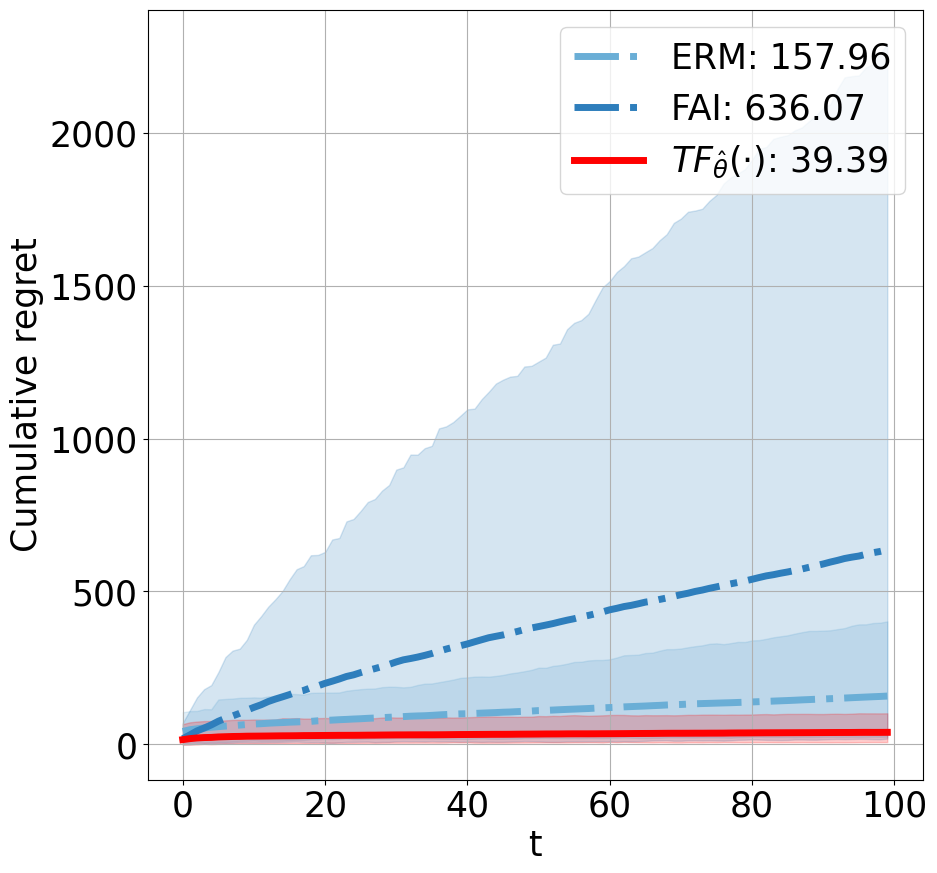}
    \caption{Square demand (misspecification)}
  \end{subfigure}
  \begin{subfigure}[b]{0.45\textwidth}
    \centering
\includegraphics[width=\textwidth]{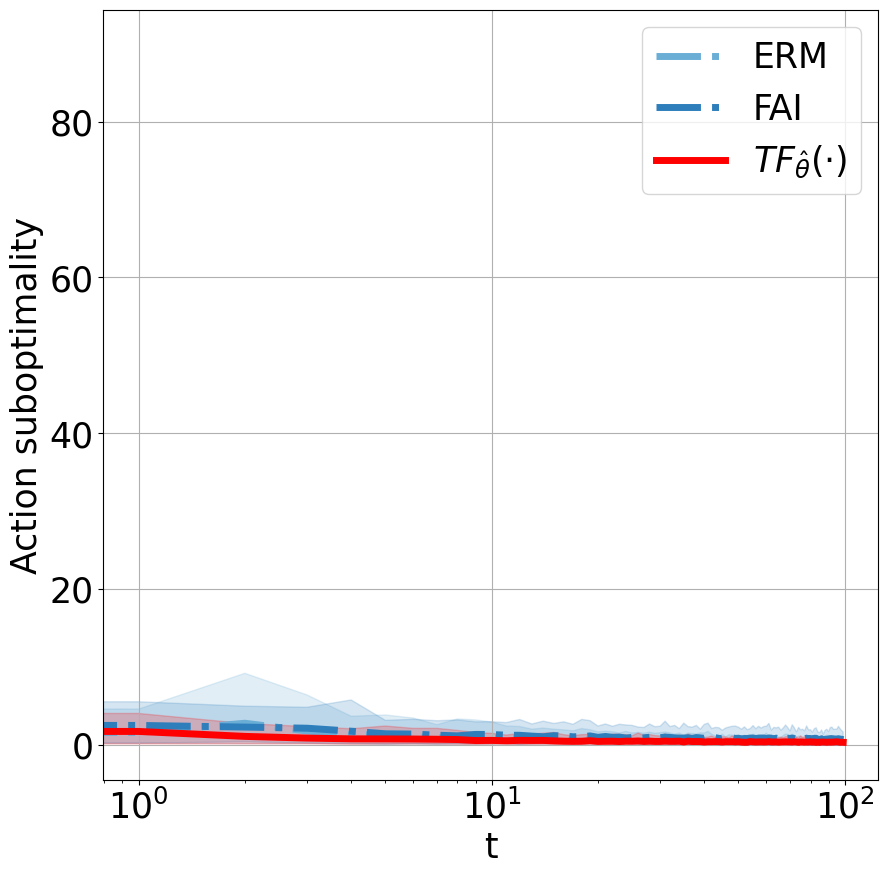}
    \caption{Linear demand (no misspecification)}
  \end{subfigure}
  \hfill
  \begin{subfigure}[b]{0.45\textwidth}
    \centering
\includegraphics[width=\textwidth]{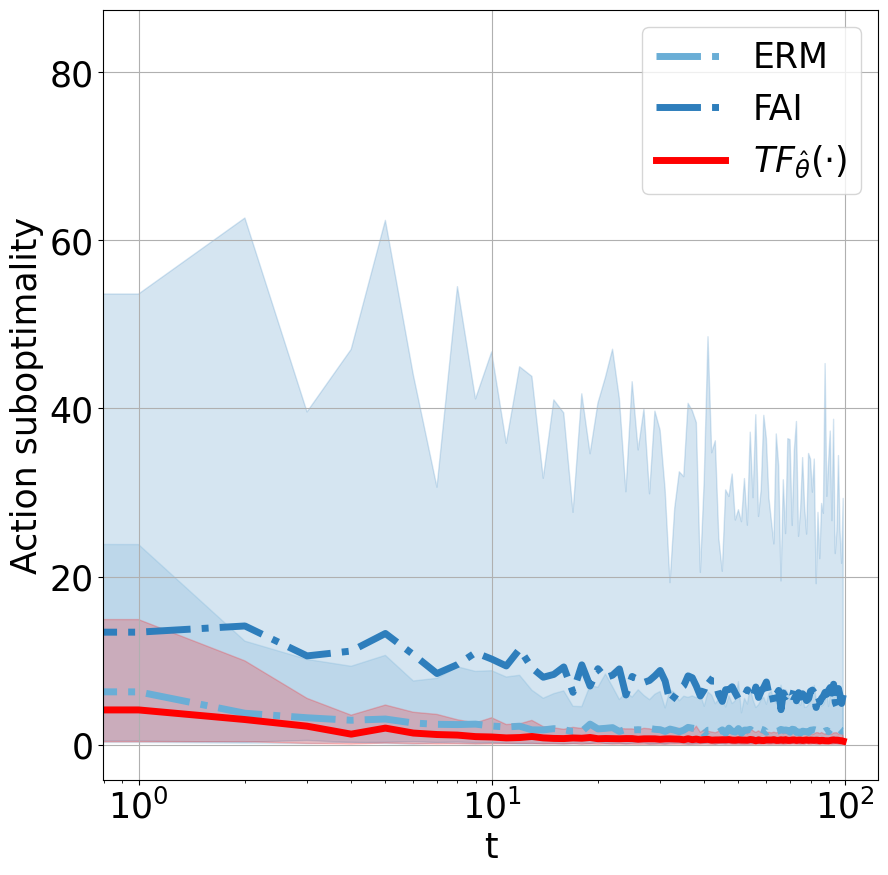}
    \caption{Square demand (misspecification)}
  \end{subfigure}

  \caption{Performance in environments with linear demand ((a), (c)) and environments with square demand ((b), (d)), where benchmark algorithms encounter model misspecification issues. The same $\texttt{TF}_{\hat{\theta}}$ is applied in all figures. It shows the average out-of-sample regret (first row) and action suboptimality, i.e., $|a^*_t - \texttt{Alg}(H_t)|$, (second row) of $\texttt{TF}_{\hat{\theta}}$ against benchmark algorithms. The numbers in the legend bar are the final regret at $t=100$.}
  \label{fig:modelmis_2demands}
\end{figure}

Figure \ref{fig:modelmis_station} illustrates the performance in the newsvendor problem with uncensored demand, tested in environments where (i) the demand functions remain consistent throughout the horizon (first column) and (ii) the demand functions are non-stationary and may change unpredictably during the horizon, causing benchmark algorithms to face model misspecification issues (second column). Figure \ref{fig:modelmis_2demands} shows the performance in newsvendor tasks in environments where demand functions are either (i) linear-type (first column) or (ii) square-type, where benchmark algorithms also encounter model misspecification issues (second column). The results demonstrate that benchmark algorithms, which are designed for stationary, linear demand environments, struggle when faced with model misspecifications. In contrast, OMGPT leverages its large capacity and pre-training to learn and adapt to changes in the environment or different demand types, effectively handling them during testing.  These findings highlight the potential of $\texttt{TF}_{\hat{\theta}}$ to effectively address model misspecifications and perform robustly in diverse environments.

\subsubsection{Demand Prediction as By-product}
In practice, decision makers may require additional side information beyond the optimal actions to better understand the testing environment. For instance, in dynamic pricing, the decision maker might want to learn the demand given a particular price to manage inventory or plan manufacturing. In this section, we evaluate OMGPT’s ability to predict demand in a dynamic pricing scenario (with 8 unknown parameters), showcasing its potential to provide useful side information.

\begin{figure}[ht!]
  \centering
  \begin{subfigure}[b]{0.31\textwidth}
    \centering
\includegraphics[width=\textwidth]{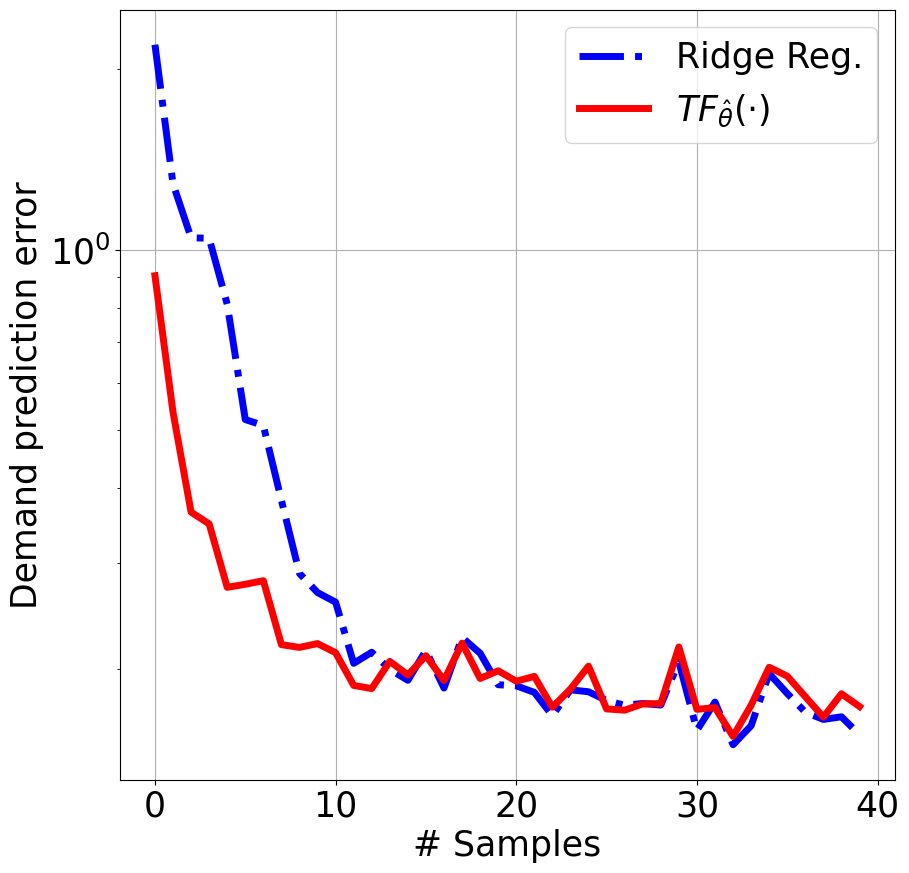}
    \caption{Single demand type,
    linear }
    \label{fig:demand_pred1}
  \end{subfigure}
  \hfill
  \begin{subfigure}[b]{0.31\textwidth}
    \centering
\includegraphics[width=\textwidth]{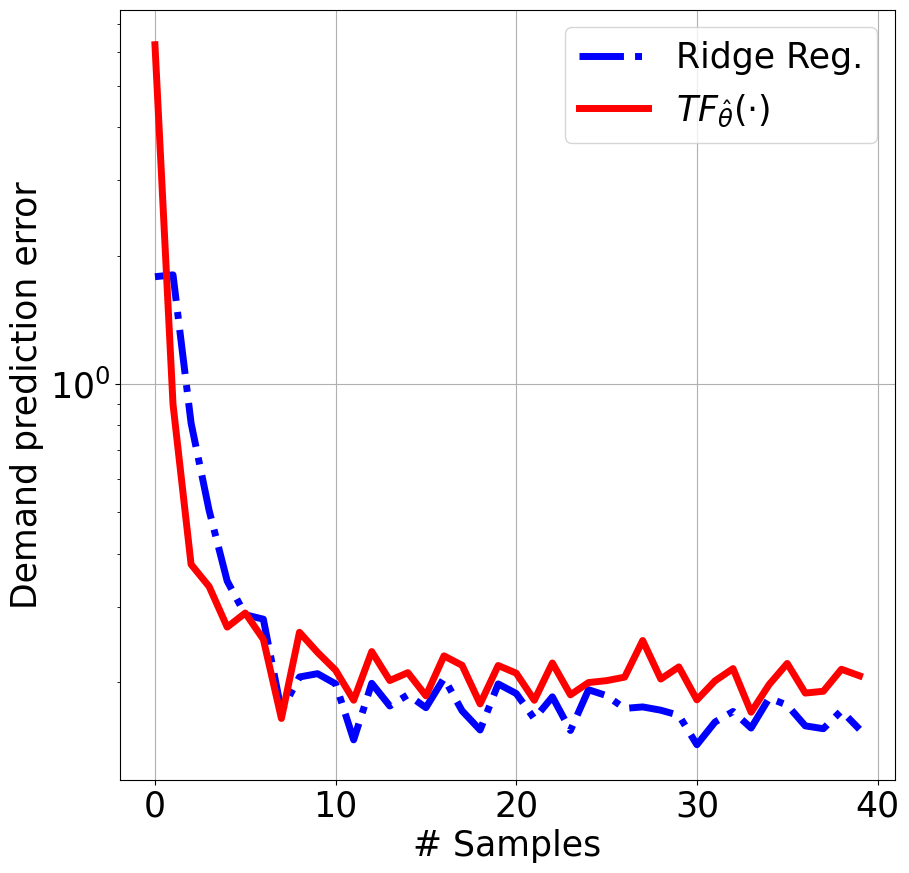}
    \caption{Two demand types, linear }
    \label{fig:demand_pred2}
  \end{subfigure}
  \hfill
  \begin{subfigure}[b]{0.31\textwidth}
    \centering
\includegraphics[width=\textwidth]{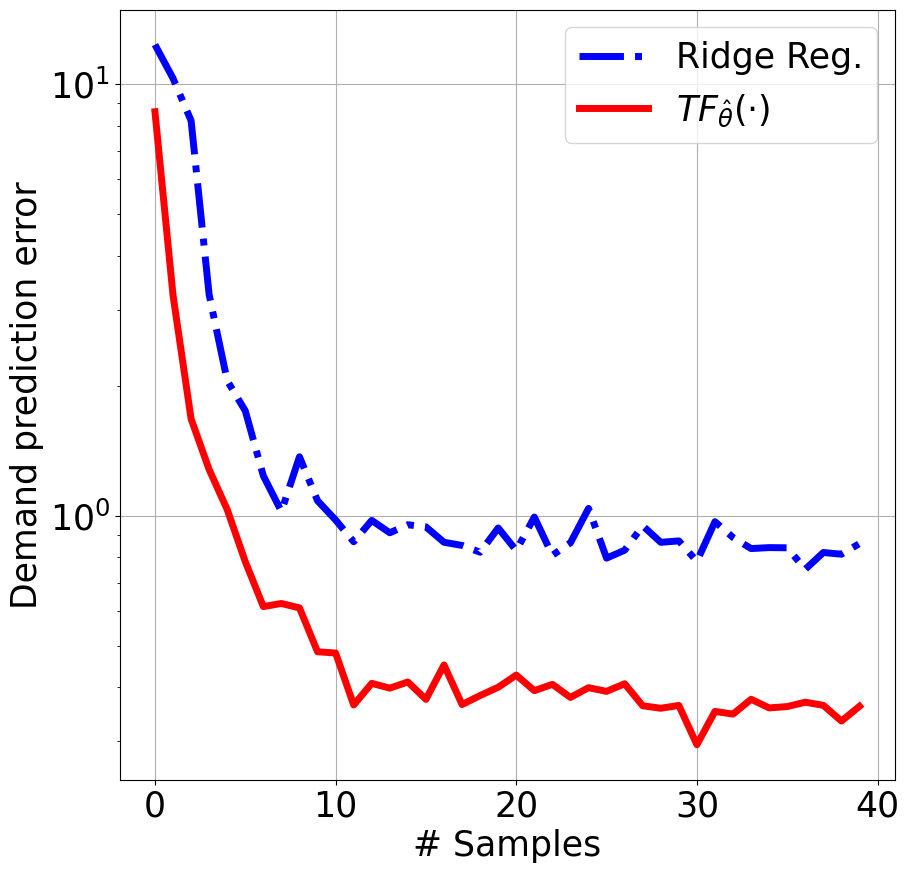}
    \caption{Two demand types, square}
    \label{fig:demand_pred3}
  \end{subfigure}
  \caption{Out-of-sample demand prediction error comparison between OMGPT ($\texttt{TF}_{\hat{\theta}}$) and ridge regression across different scenarios. (a): Single demand type, where $\texttt{TF}_{\hat{\theta}}$ is pre-trained in environments with linear demand, and both methods are tested on linear demand environments. (b) and (c): Two demand types, where $\texttt{TF}_{\hat{\theta}}$ is pre-trained in environments with a mix of linear and square demand ($50\%$ each). Both methods are then tested on either linear demand (b) or square demand (c).}
  \label{fig:demand_pred}
\end{figure}

We modify OMGPT to predict demand as discussed in Section \ref{sec:dics_obs_predict} and compare its performance to ridge regression (with a regularization term of 0.1) as a benchmark. We use ridge regression instead of ordinary least squares since the number of parameters can exceed the number of samples in some cases. We first test the models in environments with linear demands. Figure \ref{fig:demand_pred1} shows the prediction errors with varying input sequence lengths, and consequently, varying numbers of observed $((X_t, a_t), O_t)$ samples in the input (as the training samples in the demand regression task). The results indicate that OMGPT performs similarly to ridge regression when the number of samples exceeds 10. However, with fewer than 10 samples, OMGPT outperforms ridge regression. We attribute this advantage to OMGPT's ability to leverage prior knowledge from its pre-training.

To further evaluate the capability of (a single) pre-trained OMGPT, we pre-train and test it in a more complex scenario where demand functions can be either square-type or linear-type, each with a probability of $1/2$. Figure \ref{fig:demand_pred2} and \ref{fig:demand_pred3} presents the prediction errors with varying numbers of observed $((X_t, a_t), O_t)$ samples in the input sequence for different demand types, but using the same $\texttt{TF}_{\hat{\theta}}$. While ridge regression slightly outperforms OMGPT in the linear-type demand cases (Figure \ref{fig:demand_pred2}), it performs significantly worse in the square-type demand cases (Figure \ref{fig:demand_pred3}), where a linear model fails to capture the square demand function accurately. This highlights one single pre-trained $\texttt{TF}_{\hat{\theta}}$’s superior ability to predict observations across different types of environments.

\subsection{Mechanisms Interpretation}
 In Section \ref{sec:analysis}, we analytically demonstrate that the pre-trained $\texttt{TF}_{\hat{\theta}}$ behaves similarly to the Bayes-optimal decision maker $\texttt{Alg}^*$ and numerically observe such alignment between them (see Figure \ref{fig:Bayes_match} and Appendix \ref{appx:bayes_match_exp}). In this section, we further explore and interpret the internal mechanisms of $\texttt{TF}_{\hat{\theta}}$ when predicting optimal actions.

Specifically, recall that the Bayes-optimal decision maker $\texttt{Alg}^*$ is defined as: 
 
$$ \texttt{Alg}^*(H) = \argmin_{a\in\mathcal{A}} \int_{\gamma} l(a, a^{(\gamma)*}_t) \mathcal{P}(\gamma|H) \mathrm{d} \gamma.$$

It selects the optimal action that minimizes the expected divergence from the true optimal action $a^{(\gamma)*}_t$, with the expectation with respect to the posterior distribution of the environment $\mathcal{P}(\gamma|H)$. This formulation consists of two key components: (i) an environment inference part, which infers the testing environment based on the history $H$ (i.e., $\mathcal{P}(\gamma|H)$), and (ii) an optimal decision inference part, which infers the optimal action $a^{(\gamma)*}_t$ for each environment $\gamma$ based on the current context $X_t$.

In this section, we demonstrate that when the pre-trained $\texttt{TF}_{\hat{\theta}}$ predicts optimal decisions, it implicitly mirrors these two components. This suggests that $\texttt{TF}_{\hat{\theta}}$ not only (finally) produces actions that align with the Bayes-optimal decision maker $\texttt{Alg}^*$, but its internal mechanisms also operate in a similar manner. In essence, these findings provide further numerical support for the analysis presented in Section \ref{sec:analysis}.

\subsubsection{Environment Inference}
\begin{figure}[ht!]
  \centering
  \begin{subfigure}[b]{0.31\textwidth}
    \centering
\includegraphics[width=\textwidth]{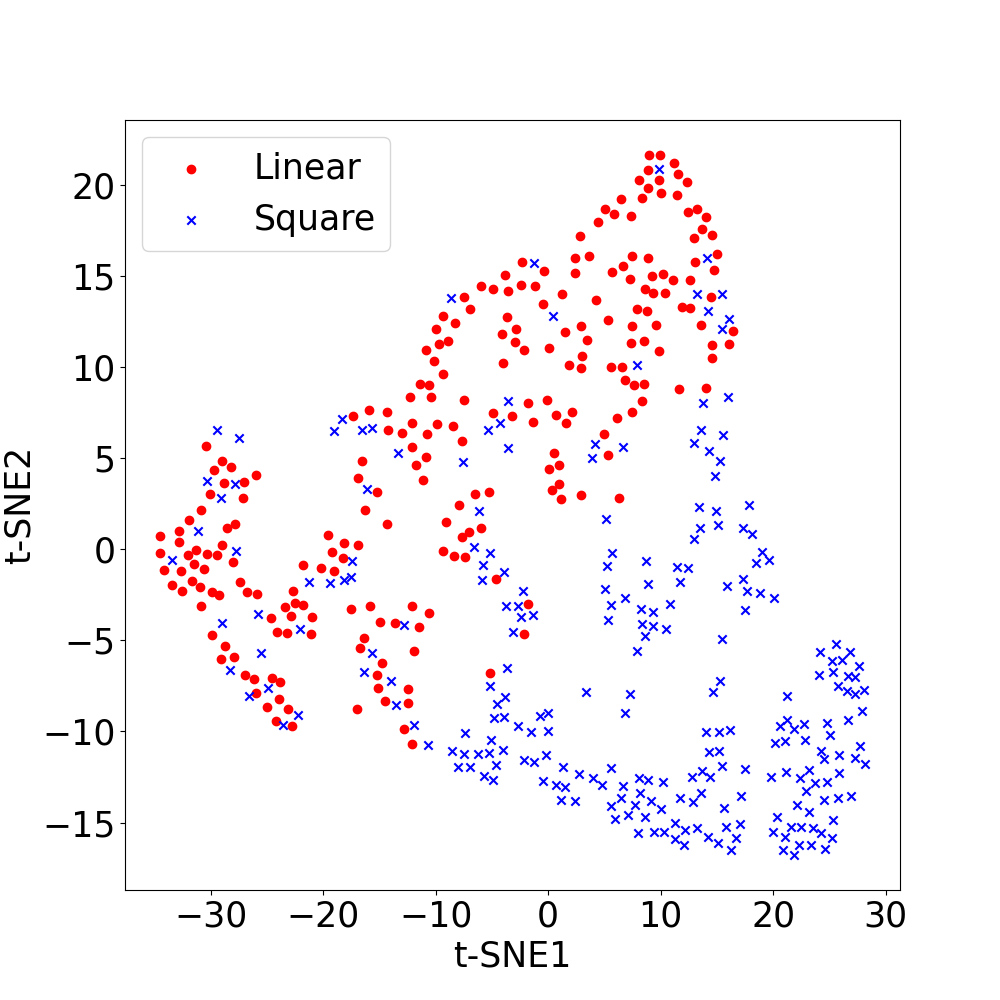}
    \caption{Layer 0 }
  \end{subfigure}
  \hfill
  \begin{subfigure}[b]{0.31\textwidth}
    \centering
\includegraphics[width=\textwidth]{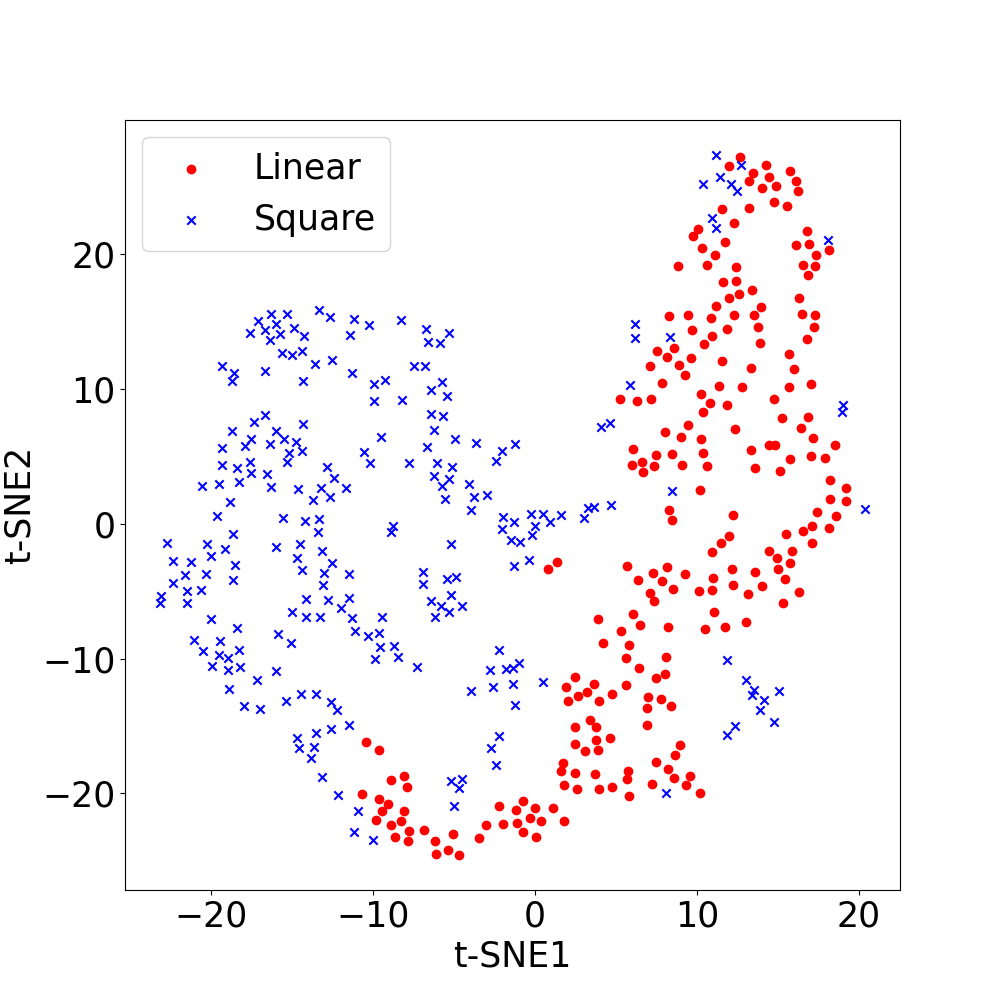}
    \caption{Layer 1 }
  \end{subfigure}
  \hfill
  \begin{subfigure}[b]{0.31\textwidth}
    \centering
\includegraphics[width=\textwidth]{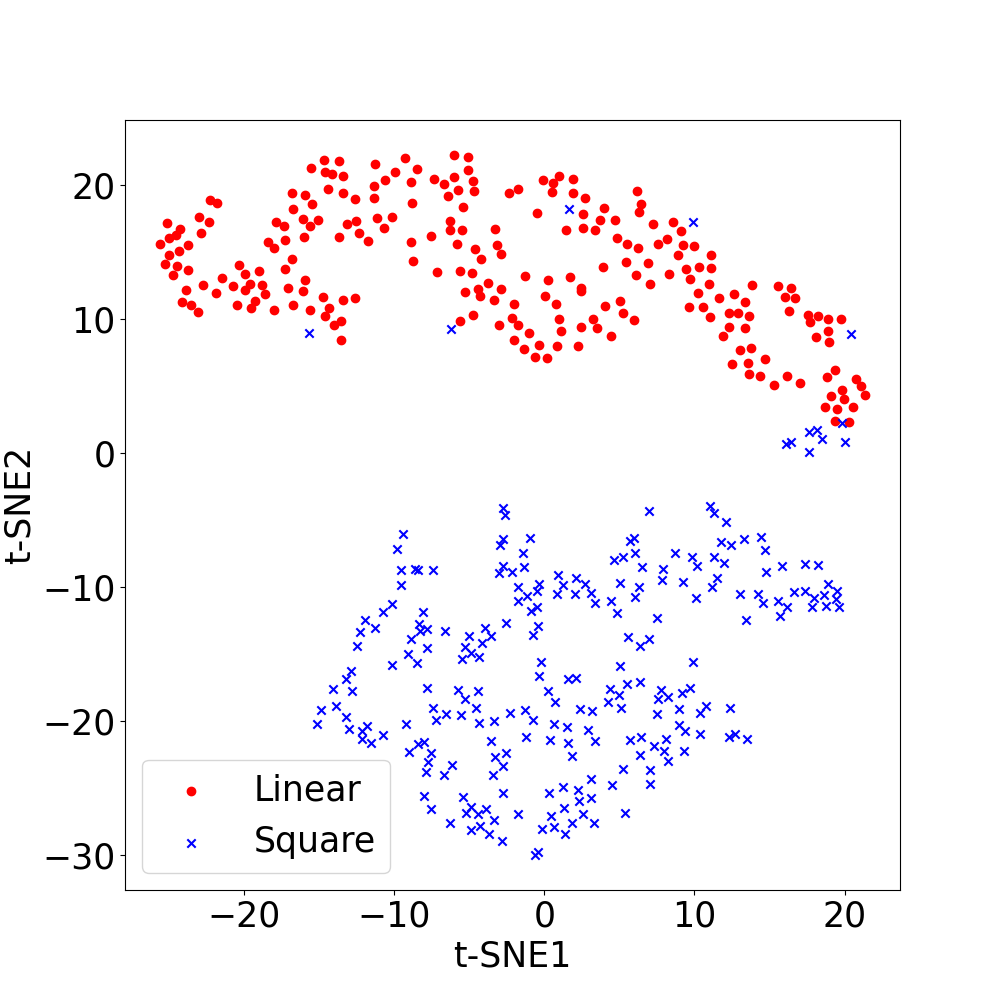}
    \caption{Layer 2}
  \end{subfigure}
  \begin{subfigure}[b]{0.31\textwidth}
    \centering
\includegraphics[width=\textwidth]{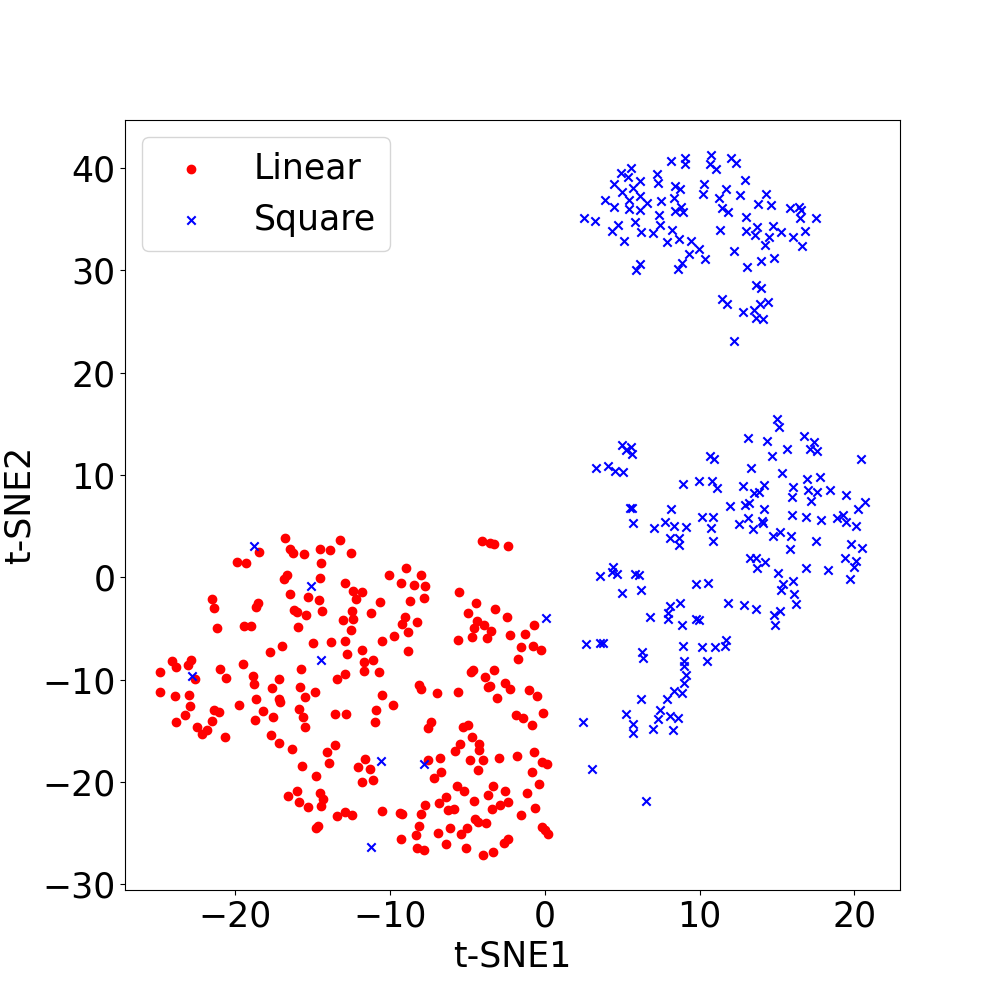}
    \caption{Layer 10 }
  \end{subfigure}
  \hfill
  \begin{subfigure}[b]{0.31\textwidth}
    \centering
\includegraphics[width=\textwidth]{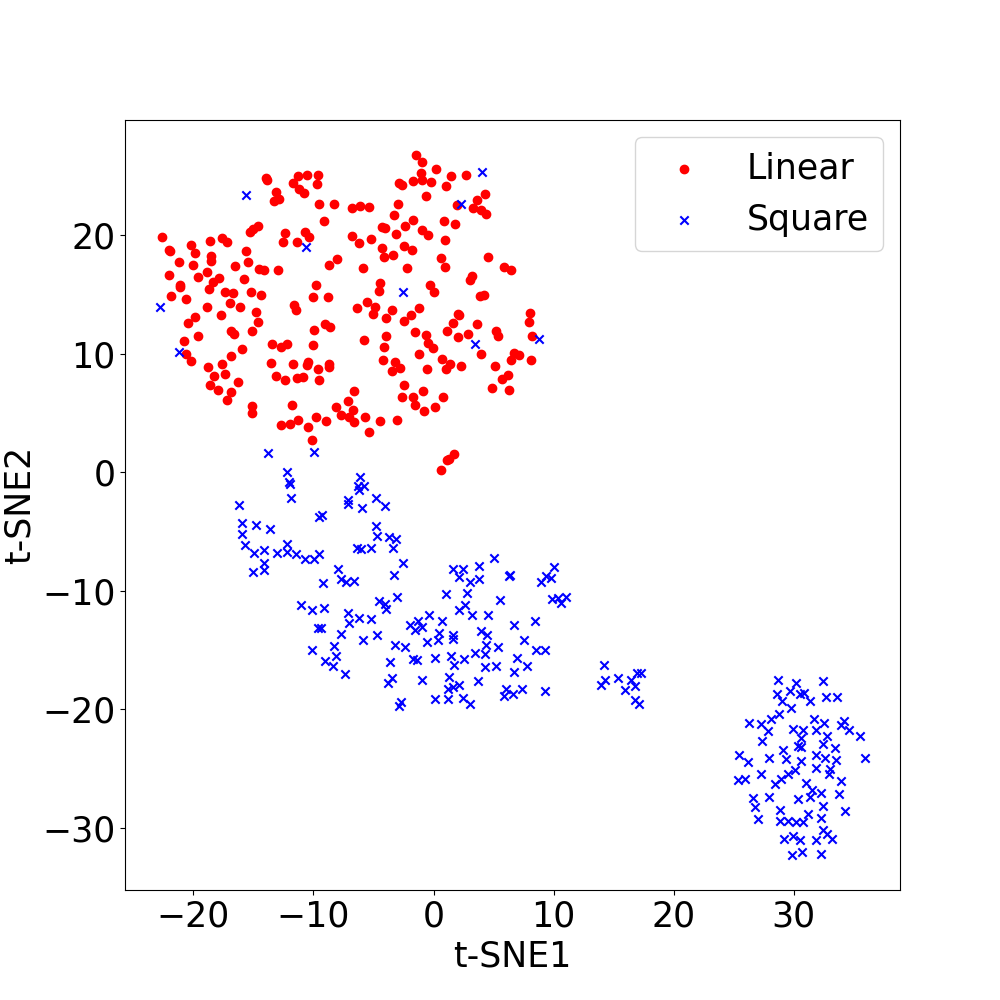}
    \caption{Layer 11 }
  \end{subfigure}
  \hfill
  \begin{subfigure}[b]{0.31\textwidth}
    \centering
\includegraphics[width=\textwidth]{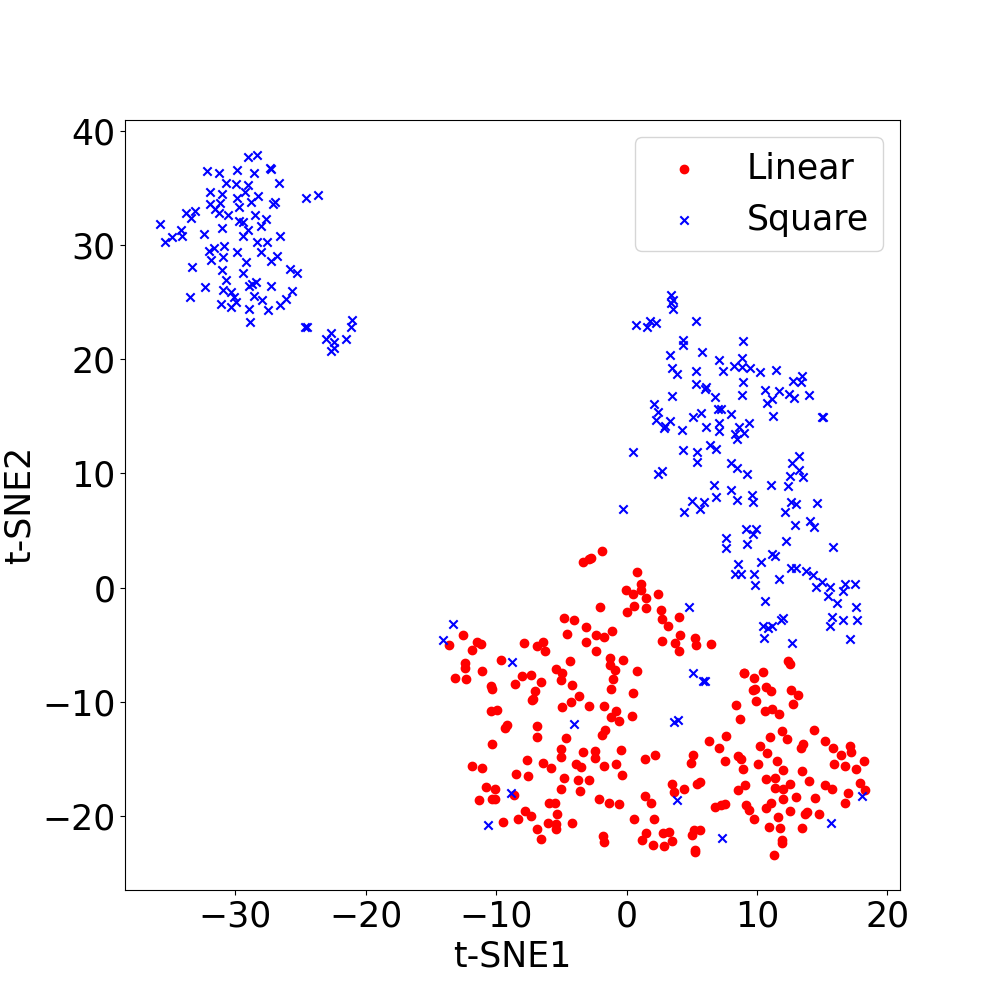}
    \caption{Layer 12}
  \end{subfigure}
  \caption{Visualization of output embedding vectors from different layers. The vectors corresponding to different demand function types become clearly separated starting from layer 2, indicating that environment inference occurs during the processing.   Additional visualizations for other layers are provided in Appendix \ref{appx: tSNE}.}
  \label{fig:probe_tSNE}
\end{figure}

\paragraph{There Exists Environment Inference}
We first investigate whether environment inference is conducted by $\texttt{TF}_{\hat{\theta}}$. To explore this, we pre-train a 12-layer $\texttt{TF}_{\hat{\theta}}$ on dynamic pricing tasks where the demand functions can either be linear or square, each appearing with a $50\%$ probability. After pre-training, we visualize the output embedding vectors from each layer using the t-SNE method (with the 0-layer representing the input vectors to the GPT model, see Appendix \ref{appx:architecture}). The visualization is based on 500 randomly sampled testing histories, each with a length of $T=100$, using the embedding vector corresponding to $X_{100}$. Among the 500 samples, half have linear demand and the other half have square demand.  Figure \ref{fig:probe_tSNE} presents the results. We observe that while the embedding vectors from the input layer and the first layer (Figures (a) and (b)) do not show a clear separation between linear and square demand functions, by layer 2, the vectors are well-separated. This indicates that as the input sequence is processed through the layers, the pre-trained model begins to infer the environment information.

\paragraph{Optimal Decision Prediction Aligns with Environment Inference}
Next, we investigate whether the pre-trained model utilizes the inferred environment when predicting optimal decisions. For this, we use the linear probing technique \citep{alain2016understanding}, a commonly used method to understand the internal workings of deep learning models. Similar to visualizing embedding vectors from intermediate layers, linear probing tests whether intermediate layer outputs contain specific information by linearly regressing or classifying quantities of interest (as the target) based on the intermediate embedding vectors (as the input) and analyzing the prediction error. Intuitively, smaller errors indicate that the embedding vectors from that layer can recover the relevant information through simple linear regression/classification, suggesting that the embedding vectors contain such information. 

\begin{figure}[ht!]
    \centering
    \includegraphics[width=0.5\linewidth]{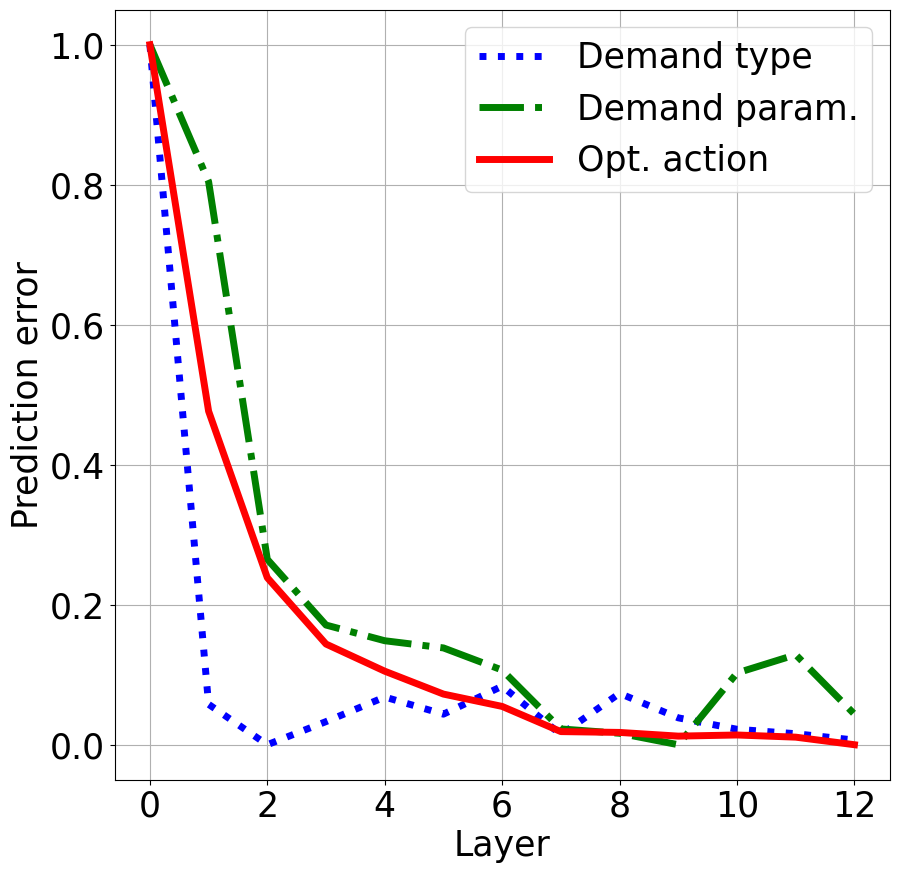}
    \caption{Normalized prediction errors of demand type, demand parameters, and optimal actions across layers, using linear probing. The prediction errors for demand/environment parameters (green, dashed line) and optimal actions (red, solid line) follow similar trends, which suggests a strong correlation between environment inference and optimal action prediction. }
    \label{fig:probe}
\end{figure}

We apply linear probing to the same pre-trained $\texttt{TF}_{\hat{\theta}}$ from Figure \ref{fig:probe_tSNE} and probe three values: demand function type (i.e., linear or square), demand function parameters (i.e., the unknown parameters of the demand function), and optimal actions. The linear probes are trained using 1024 randomly sampled environments and histories. The input embedding vectors to train linear probes correspond to $X_{100}$'s in the histories. Among the training samples, half have linear demand and the other half have demand. The reported values are the averages over 256 randomly sampled (testing) histories, with half having linear demand and the other half having square demand. Figure \ref{fig:probe} shows the normalized prediction errors from linear probing, with values scaled between 0 and 1 for comparison. The results show that:

\begin{itemize}
    \item A minor observation is that the prediction error for the demand function type becomes relatively small after layer 1, aligning with our observations in Figure \ref{fig:probe_tSNE}. This occurs earlier than the layer where the demand function parameter prediction error becomes small (after layer 7), which corresponds to the natural flow of estimation—first identifying the model class (linear or square), then estimating the model parameters.
    \item More importantly, the prediction errors for both the parameters and optimal actions follow a similar trend: both decrease quickly before layer 7 (with similar rates across layers) and then stabilize at a relatively low level. It is important to note that the pre-training only requires the OMGPT to predict the optimal actions, not the exact demand parameters. For example, in a linear demand model, the optimal price can be computed using $\frac{\alpha^\top X}{-2\beta^\top X}$ (where $\alpha,\beta$ are the unknown demand parameters), so exact values for $\alpha$ and $\beta$ are unnecessary. (The divergence in prediction errors between the demand parameters and optimal actions after layer 9 can be attributed to this ``distillation'': model ``distills'' unnecessary information to better predict optimal actions.) This similar trend between the predictions of parameters and optimal actions suggests that optimal decision prediction is highly correlated with parameter/environment inference. While we cannot establish a causal relationship, the result suggests that optimal decision prediction is closely related to environment inference.
\end{itemize}

\subsubsection{Optimal Decision Inference}
\label{sec:decision_inference}
In this section, we investigate whether the pre-trained $\texttt{TF}_{\hat{\theta}}$ performs optimal decision inference. Specifically, we rewrite the input sequence of $\texttt{TF}_{\hat{\theta}}$ as follows:
$$H_t=\{(X_1,a_1,O_1),\ldots,(X_{t-1},a_{t-1},O_{t-1})\}\cup \{X_t\}=\tilde{H}_{t-1}\cup \{X_t\}$$
where $\tilde{H}_{t-1}=\{(X_1,a_1,O_1),\ldots,(X_{t-1},a_{t-1},O_{t-1})\}$ contains all previously collected samples before $t$ and can be used to infer the unknown environment $\gamma$ (which generates $H_t$). Assuming $X_t$ is generated independently of $\gamma$ conditional on $\tilde{H}_{t-1}$, we want to verify whether:
\begin{equation}
    \texttt{TF}_{\hat{\theta}}(H_t)=\hat{a}^*(X_t,\hat{\gamma}(\tilde{H}_{t-1}))
    \label{eqn:act_infer}
\end{equation}
holds for some functions $\hat{\gamma}$ (representing environment inference) and $\hat{a}^*$ (representing optimal action inference). If this holds, it implies that the pre-trained $\texttt{TF}_{\hat{\theta}}$ does not simply memorize pre-training data. Instead, it follows a decision rule $\hat{a}^*(X_t, \hat{\gamma})$ that maps from the current context $X_t$ and the inferred environment $\hat{\gamma}$ (based on $\tilde{H}_{t-1})$ to the predicted action, thereby performing optimal decision inference.

To investigate this, we manipulate the history sequence to create $H'_t = \tilde{H}_{t-1} \cup \{X'_t\}$ by altering the original sampled context $X_t$ to $X'_t$ (while keeping  $\gamma$ and $\tilde{H}_{t-1}$ unchanged), and meanwhile we ensure that no similar pairs $(\tilde{H}_{t-1}, X'_t)$ are seen during pre-training. We then compare the behavior of $\texttt{TF}_{\hat{\theta}}$ in these manipulated histories with that in non-manipulated histories (which are generated in the same way as the pre-training data). If $\texttt{TF}_{\hat{\theta}}$ behaves similarly in both cases, predicting nearly optimal actions, it suggests that $\texttt{TF}_{\hat{\theta}}$ performs optimal decision inference rather than simple memorization, as it has not encountered any similar pairs $(\tilde{H}_{t-1}, X'_t)$ during pre-training.

\begin{figure}[ht!]
  \centering
  \begin{subfigure}[b]{0.45\textwidth}
    \centering
\includegraphics[width=\textwidth]{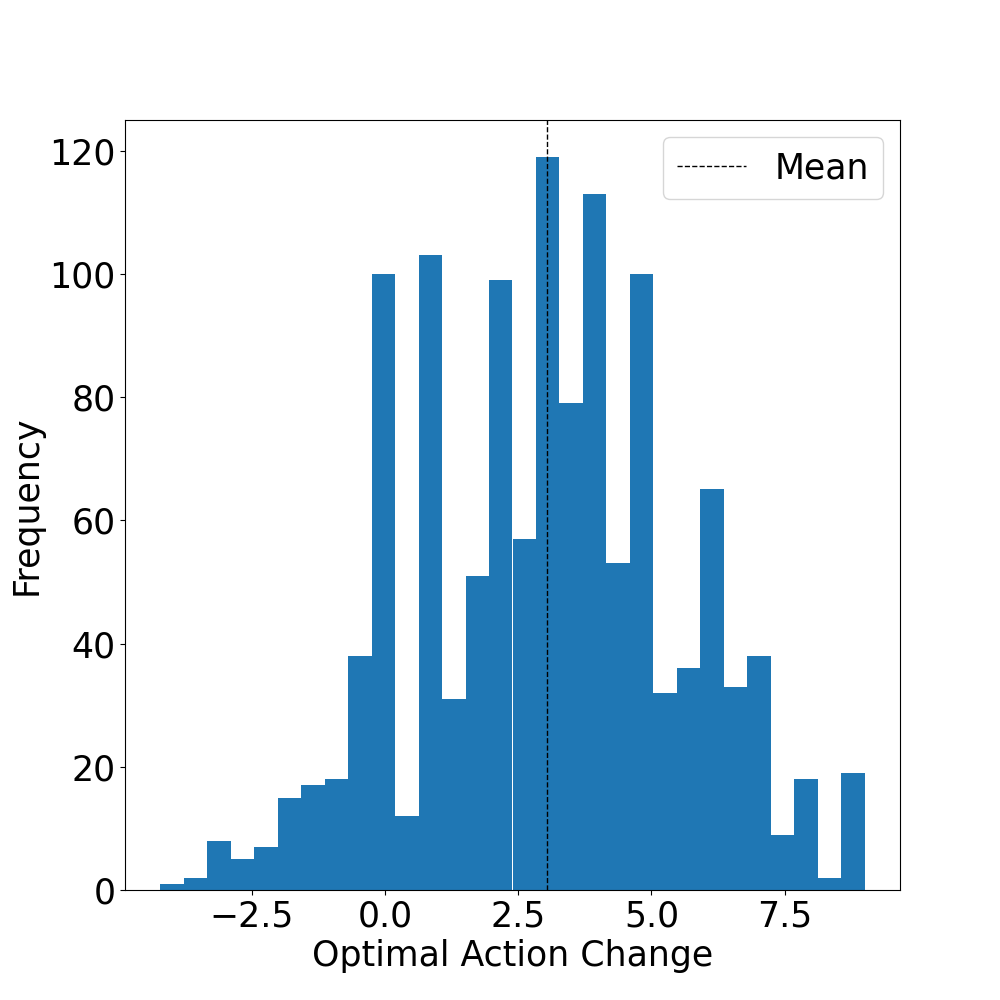}
    \caption{Optimal action change}
    \label{fig:pasting1}
  \end{subfigure}
  \hfill
  \begin{subfigure}[b]{0.45\textwidth}
    \centering
\includegraphics[width=\textwidth]{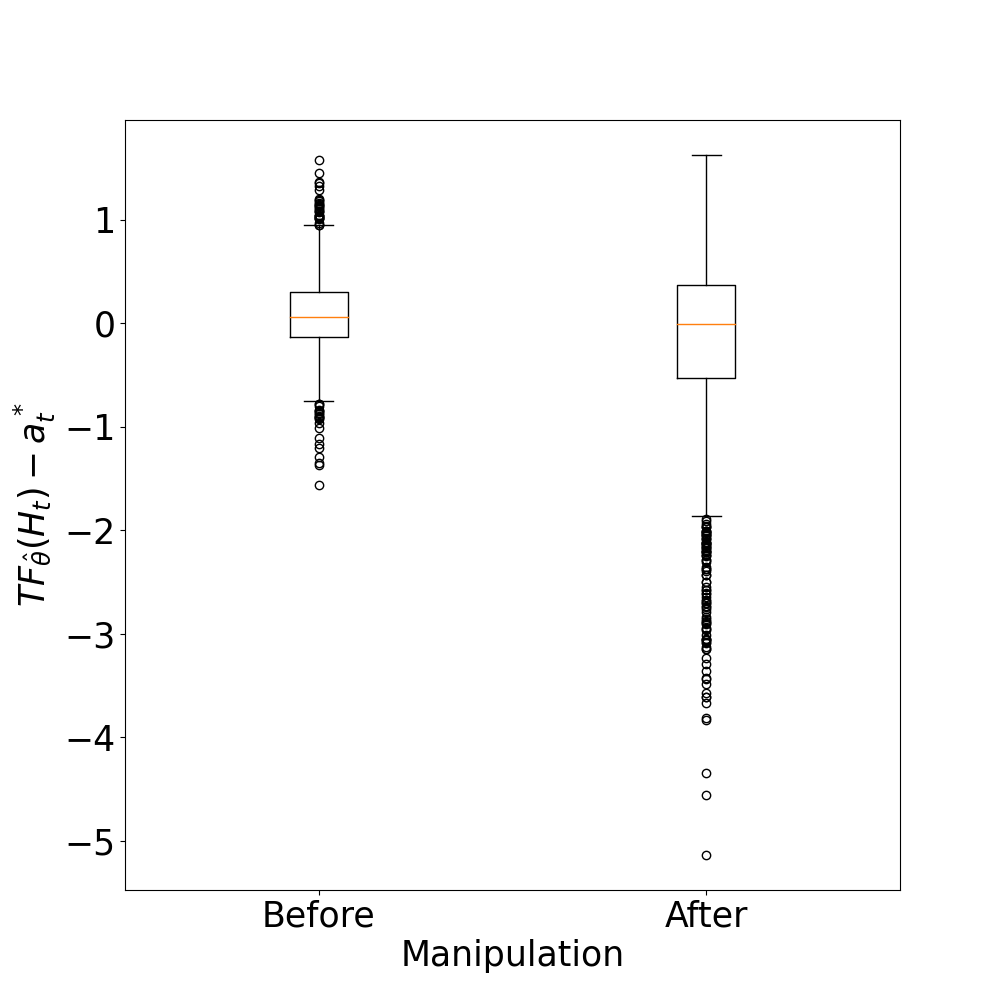}
    \caption{Performance change}
    \label{fig:pasting2}
  \end{subfigure}
  \caption{(a) Histogram of optimal action changes due to the manipulation of the context $X_t$. (b) Boxplots comparing $\texttt{TF}_{\hat{\theta}}$'s behavior ($\texttt{TF}_{\hat{\theta}}(H_t) - a^*_t$) before and after manipulations. }
  \label{fig:pasting}
\end{figure}

In the experiment, we create such manipulated histories in a newsvendor problem with non-perishable products (details provided in Appendix \ref{appx:envs}). The context is generated by $X_t = q(\tilde{H}_{t-1})$, and the optimal action is $a^*_t = a^*(X_t, \gamma)$ with some deterministic functions $q$ and $a^*$. We pre-train OMGPT on un-manipulated data ($X_t = q(\tilde{H}_{t-1})$) and test it on both un-manipulated and manipulated histories. In the manipulated case, we set $X'_t = \epsilon'_t$, where $\epsilon'_t$ is independently sampled from a uniform distribution over the range $[0,10]$ and is independent of $\tilde{H}_{t-1}$. Figure \ref{fig:pasting1} shows the distribution of  $a^*(X'_t,\gamma)-a^*(X_t,\gamma)$ at $t=50$, i.e., the changes of optimal actions after manipulation, over $1000$ randomly sampled $\gamma$'s. It shows most optimal actions are shifted after manipulations (not around 0).

We then compare $\texttt{TF}_{\hat{\theta}}(H_t) - a^*_t$ before $(a^*_t=a^*(X_t,\gamma))$ and after manipulation ($a^*_t=a^*(X'_t,\gamma))$ to check if there exists a significant change in $\texttt{TF}_{\hat{\theta}}$'s behavior after manipulations. Figure \ref{fig:pasting2} presents boxplots of $\texttt{TF}_{\hat{\theta}}(H_t) - a^*_t$.  While the variation increases after manipulation, the predicted actions $\texttt{TF}_{\hat{\theta}}(H_t)$'s are concentrated around the optimal actions in both cases, as opposed to the significant shift observed in the optimal actions in Figure \ref{fig:pasting1}. This indicates that $\texttt{TF}_{\hat{\theta}}$ can still perform nearly optimally, even after the manipulation of contexts. These results demonstrate that $\texttt{TF}_{\hat{\theta}}$ does not rely on memorizing pre-training data but instead (implicitly) performs optimal decision inference.

\section{Conclusion}
\label{sec:conclusion}
In this work, we frame sequential decision making problems as sequence modeling tasks, where we predict optimal actions based on historical data. We propose a supervised learning paradigm that does not assume any analytical model structure and can easily incorporate the decision maker's prior knowledge. Specifically, we (i) design OMGPT, a GPT-based neural network architecture for sequential decision making problems, (ii) introduce a new pre-training scheme tailored to these problems, and (iii) demonstrate through numerical and analytical results that OMGPT exhibits strong performance and capabilities, positioning it as a promising solution for a wide range of operational decision making applications.

\bibliographystyle{plainnat}
\bibliography{main}

\appendix

\section{Model Architecture and $\texttt{Alg}^*$}

\subsection{Model Architecture for OMGPT}
\label{appx:architecture}
\begin{figure}[ht!]
    \centering
    \includegraphics[width=0.7\linewidth]{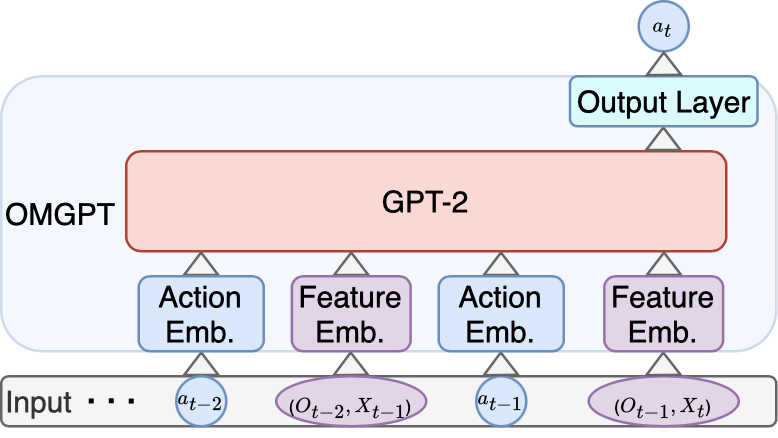}
    \caption{The architecture of OMGPT, where all actions $a_{\tau}$ and the combined ``feature'' elements $(O_{\tau-1}, X_{\tau})$ in $H_t$ pass through the same action embedding layer and feature embedding layer, respectively. The resulting embedding vectors are then processed by a (trainable) GPT-2 model, which summarizes them into a single embedding vector. Finally, this embedding vector passes through the output layer, transforming it into the predicted action $a_t$.}
    \label{fig:architecture}
\end{figure}
We adopt the transformer architecture for regression tasks from \cite{garg2022can} to solve sequential decision making tasks, with modifications tailored to our specific setting. The architecture is summarized in Figure \ref{fig:architecture}.

\textbf{Input Sequence Structure.} Since we have three types of elements (contexts, actions, and observations), and the actions are the ``label type'' we want to predict sequentially, we first combine the contexts and 
 the observations into ``features'', similar to a supervised learning task. Specifically, instead of directly using $H_t$ as the input sequence at time $t$, we construct two types of elements derived from $H_t$:
\begin{itemize}
    \item \textbf{``Feature'' elements} combine the context $X_{\tau} \in \mathbb{R}^{d_X}$ at each time $\tau \leq t$ with the observation $O_{\tau-1} \in \mathbb{R}^{d_O}$ from $\tau-1$ (with $O_0$ set as a zero vector in our experiments). Formally, the feature elements are $\{(O_{\tau-1}, X_{\tau})\}_{\tau=1}^t \subset \mathbb{R}^{d_O + d_X}$, and they serve as the input ``features'' for prediction.
    \item \textbf{``Label''/action elements} are the actions $\{a_{\tau}\}_{\tau=1}^{t-1}$ from $H_t$ that were predicted in the testing phase (or generated in the pre-training phase).
\end{itemize}
Thus, the input sequence becomes $\{(O_0, X_1), a_1, \ldots, (O_{t-2}, X_{t-1}), a_{t-1}, (O_{t-1}, X_t)\}$, which contains $2t-1$ elements in total to predict $a^*_t$.

\textbf{Architecture.} The transformer is based on the GPT-2 family \citep{radford2018improving}. It uses two learnable linear transformations to map each ``feature'' and ``label'' element into vectors in the embedding space respectively. These transformations are referred to as the feature embedding layer and the action embedding layer. The output embedding vectors, which share the same dimension, are processed through the GPT-2's attention mechanism, resulting in a single vector that encapsulates relevant contextual information. This vector then undergoes another learnable linear transformation,  referred to as the output layer, moving from the embedding space to the action $\mathcal{A}$, ultimately resulting in the prediction of $a^*_t$. In our experiments, the GPT-2 model has 12 layers, 16 attention heads, and a 256-dimensional embedding space. The experiments are conducted on 2 A100 GPUs with \texttt{DistributedDataParallel} method of Pytorch.

\subsection{Difference from \citet{lee2024supervised}}

While \citet{lee2024supervised} also propose a supervised learning approach using transformers for general reinforcement learning/Markov decision process problems (e.g., game playing), there are four key differences between their work and ours:
    \begin{itemize}
\item \textbf{Pre-training data generation:} In pre-training, we develop a novel data generation method specifically tailored to these sequential decision making processes, as detailed in Appendix \ref{sec:algo}. In addition, \citet{lee2024supervised} propose constructing input sequences for predicting $a^*_t$ using a set of tuples  $(X_{\tau},a_{\tau},O_{\tau},X_{\tau+1})$, which are sampled from a trajectory and are not contiguous or in chronological order. Intuitively, the goal is for the pre-trained transformer to implicitly learn the transitions and environment dynamics represented by the progression from $(X_{\tau}, a_{\tau}, O_{\tau})$ to $X_{\tau+1}$, thereby ``informing'' its decision making process. In contrast, our approach generates contiguous sequences, i.e., $H_t=(X_1,a_1,O_1,\ldots,X_{t-1},a_{t-1},O_{t-1},X_t)$ to predict $a^*_t$. This approach offers two benefits: (i) It aligns with the testing phase, where decisions are made in chronological order. Such alignment is often necessary in the classic machine learning framework  \citep{shalev2014understanding}. (ii) The chronological order itself can contain important information in operational decision making. For example, in a market where the demand model changes abruptly at some point $t'$ (e.g., a lockdown during a pandemic), the decision maker (and OMGPT) should focus more on the histories after $t'$ to learn the new market conditions.

\item \textbf{Underlying problems:} We focus on sequential decision making problems that do not have a transition probability matrix, unlike Markov decision processes. The advantage here is that the optimal action is either available in closed form or can be efficiently solved. This allows us to use the optimal action as the prediction target in the training data, making the supervised learning paradigm more practical for real-world operational applications.  In addition, \citet{lee2024supervised} focuses primarily on problems with discrete action spaces, whereas we conduct experiments and provide analyses for both discrete and continuous action spaces.
\item \textbf{GPT model architecture:} Due to the different data generation methods and the special structure of the underlying problems, we employ a different GPT architecture.
\item  \textbf{Analysis:}  The distinctions outlined above align our proposed paradigm more closely with sequential decision making in operations management and in-context learning literature \citep{dong2022survey}. This alignment grants us deeper analytical insights into the underlying mechanisms and performance of our OMGPT model. More details are elaborated in Section \ref{sec:analysis}. 
\end{itemize}

\subsection{More Details on $\texttt{Alg}^*$}
\label{appx:posterior_oracle}
In this subsection, we study the behavior of $\texttt{TF}_{\hat{\theta}}$  by comparing regret performances on $\texttt{TF}_{\hat{\theta}}$ and the algorithms utilizing the posterior distribution $\mathcal{P}(\gamma | H)$ of environments. Besides Bayes-optimal decision functions as defined in Section \ref{sec:Bayes_decision_maker}, we can also name the later algorithms as the \textit{oracle posterior algorithms} as they are assumed to know the posterior distribution  $\mathcal{P}(\gamma | H)$ (or equally, the prior distribution $\mathcal{P}_{\gamma}$).

We first present a general framework for these oracle posterior algorithms in Algorithm \ref{alg:posterior}: posterior averaging, sampling, and median. For simplicity of notation, we omit the subscript $t$ and only consider a finite space $\Gamma$ of environments, while the algorithms can be easily extended to the infinite case. For the posterior median algorithm, we assume the action space $\mathcal{A}\subseteq \mathbb{R}$. 
\begin{algorithm}[ht!]
\caption {Posterior averaging/sampling/median at time $t$}
\label{alg:posterior}
\begin{algorithmic}[1] 
\Require  Posterior probability $\mathcal{P}(\gamma | H)$ and optimal action $a^*_{\gamma}$ for each environment $\gamma\in \Gamma$, algorithm  $\texttt{Alg} \in \{\text{posterior averaging, posterior sampling, posterior median}\}$
 \Statex \ \ \ \ \textcolor{blue}{\%\% Posterior averaging }
\If{\texttt{Alg}$=$posterior averaging}
\State{$a=\sum_{\gamma \in \Gamma} \mathcal{P}(\gamma | H) \cdot a^*_{\gamma}$ }
\Statex \ \ \ \ \textcolor{blue}{\%\% Posterior sampling }
\ElsIf{\texttt{Alg}$=$posterior sampling}
\State $a=a^*_{\tilde{\gamma}}$, where $\tilde{\gamma}\sim \mathcal{P}(\cdot | H)$
\Statex \ \ \ \ \textcolor{blue}{\%\% Posterior median }
\ElsIf{\texttt{Alg}$=$posterior median}
\State Sort and index the environments as $\{\gamma_i\}_{i=1}^{|\Gamma|}$   such that the corresponding optimal actions are in ascending order: $a^*_{\gamma_1}\leq a^*_{\gamma_2}\leq \ldots \leq a^*_{\gamma_{|\Gamma|}}$. 
\State Choose $a=\min \left\{ a^*_{\gamma_i} \big \vert \sum_{i'=1}^{i} \mathcal{P}(\gamma_{i'}|H)\geq 0.5\right\}$
\EndIf
\State \textbf{Return}: $a$
\end{algorithmic}
\end{algorithm}

\begin{figure}[ht!]
  \centering
  \begin{subfigure}[b]{0.48\textwidth}
    \centering
\includegraphics[width=\textwidth]{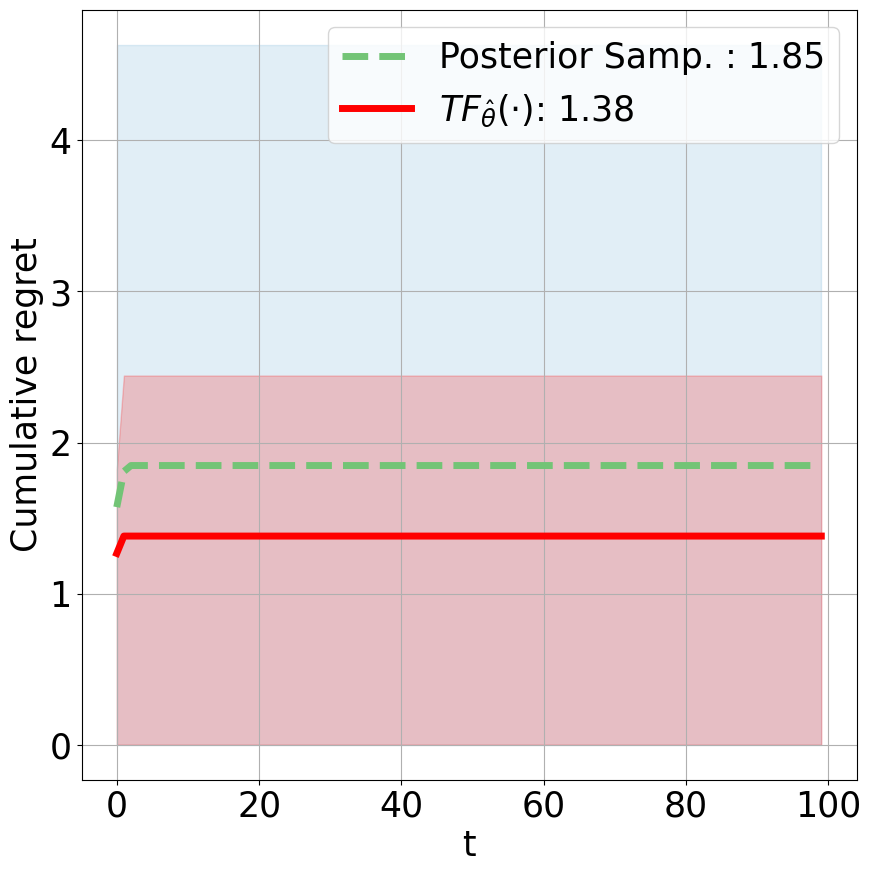}
    \caption{Regret for multi-armed bandits}
  \end{subfigure}
  \hfill
    \begin{subfigure}[b]{0.48\textwidth}
    \centering
\includegraphics[width=\textwidth]{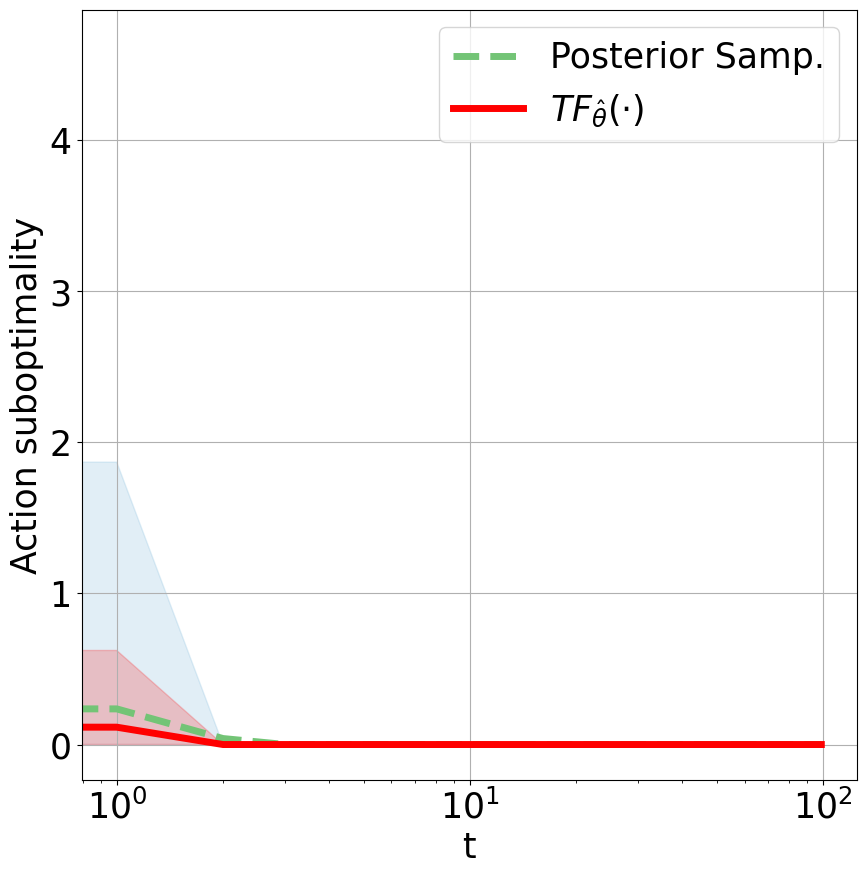}
    \caption{Action suboptimality for multi-armed bandits}
  \end{subfigure}
  
  \begin{subfigure}[b]{0.48\textwidth}
    \centering
\includegraphics[width=\textwidth]{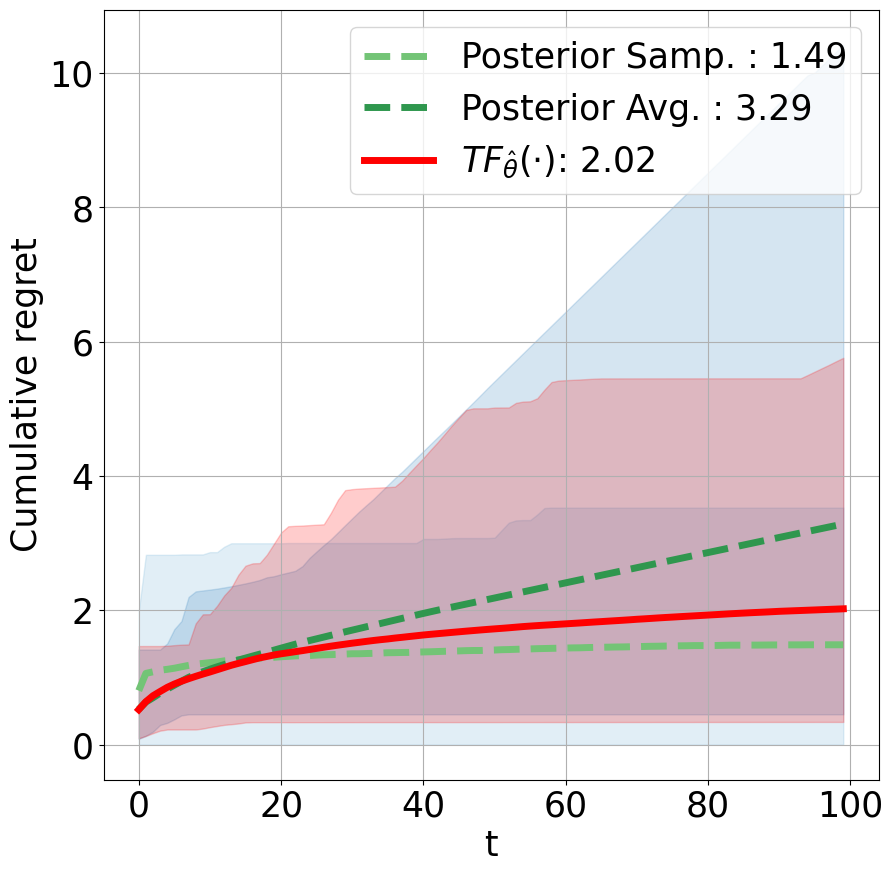}
    \caption{Regret for linear bandits}
  \end{subfigure}
  \hfill
  \begin{subfigure}[b]{0.48\textwidth}
    \centering
\includegraphics[width=\textwidth]{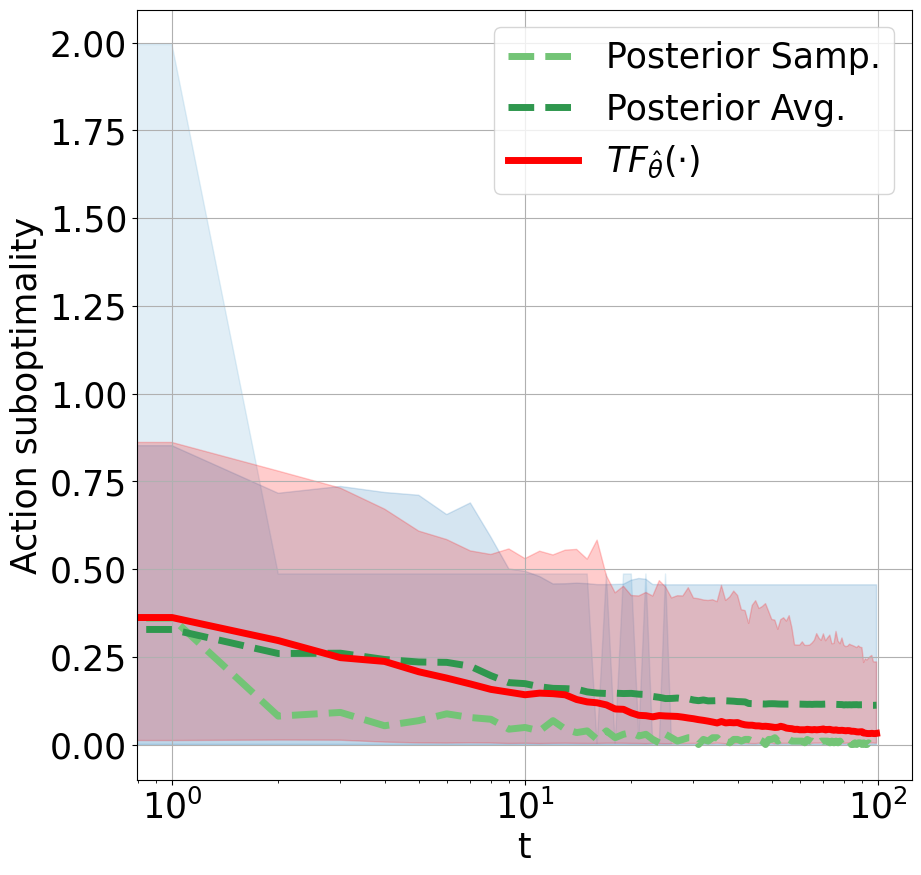}
    \caption{Action suboptimality for linear bandits}
  \end{subfigure}
  \caption{ Performances of $\texttt{TF}_{\hat{\theta}}$ and oracle posterior algorithms. The numbers in the legend bar are the final regret at $t = 100$ and the shaded areas indicate the $90\%$ (empirical) confidence intervals.    }
  \label{fig:appx_orc_post_compare}
\end{figure}

Figure \ref{fig:appx_orc_post_compare} presents the testing regret and action suboptimality for $\texttt{TF}_{\hat{\theta}}$ compared to oracle posterior algorithms. Generally, all algorithms exhibit good regret performance. However, we observe that $\texttt{TF}_{\hat{\theta}}$ is (i) superior to posterior sampling in multi-armed bandits, as shown in (a), and (ii) outperforms posterior averaging in linear bandits, as shown in (c).

Specifically, (b) shows that posterior sampling has higher action suboptimality during the initial time steps compared to $\texttt{TF}_{\hat{\theta}}$. This suggests that the exploration inherent in the sampling step of posterior sampling can introduce additional regret, which might be unnecessary for simpler problems. Conversely, (d) indicates that the decisions from posterior averaging do not converge to the optimal action $a^*_t$ as quickly as those from $\texttt{TF}_{\hat{\theta}}$ and posterior sampling. This suggests that posterior averaging may be too greedy, thus failing to sufficiently explore the environment. This second observation also numerically supports Proposition \ref{prop:lin_reg}.

These observations illustrate that $\texttt{TF}_{\hat{\theta}}$ can outperform oracle posterior algorithms in certain scenarios. This further demonstrates that the advantage of $\texttt{TF}_{\hat{\theta}}$ over benchmark algorithms is not solely due to its use of prior knowledge about the task. Rather, $\texttt{TF}_{\hat{\theta}}$ discovers a new decision rule that achieves better short-term regret than oracle posterior algorithms. It can be more greedy than posterior sampling while exploring more than posterior averaging.

\textbf{Setup.} We consider a multi-armed bandits task (with $20$ arms) and linear bandits task (with $2$-dimensional actions) with $4$ environments. The results are based on $100$ runs. The details of tasks with their specific oracle posterior algorithms are provided in Appendix \ref{appx:numerical}.

\section{More Details on Pre-training}

\label{sec:algo}
\subsection{Generalization Issue in Pre-training}
We first point out a subtle point of the supervised pre-training setting presented in Section \ref{sec:supervised}. Unlike (traditional) supervised learning, the expected loss \eqref{eqn:test_loss} may not be equal to the expectation of the empirical loss \eqref{eqn:erm_theta}. 
As noted earlier, this is because in the sequence $H_t$, the action $a_t=\texttt{TF}_{{\hat{\theta}}}(H_t)$ as in \eqref{eqn:test_dynamics} during the test phase as opposed to that $a_t=\tilde{f}(H_t)$ as in \eqref{eqn:P_ga_g} for the pre-training data $\mathcal{D}_{\text{PT}}$. Taking expectation with respect to $\gamma$ in \eqref{eqn:test_loss}, we can define the expected loss as 
$$L(\texttt{TF}_{\theta}) \coloneqq \mathbb{E}_{\gamma\sim \mathcal{P}_\gamma}\left[L(\texttt{TF}_{\theta};\gamma)\right].$$
However, the expectation of the empirical loss \eqref{eqn:erm_theta} is 
\begin{equation}
\label{eqn:train_expect}
L_{\tilde{f}}(\texttt{TF}_{\theta})  \coloneqq  \mathbb{E}_{\gamma\sim \mathcal{P}_\gamma}\left[L_{\tilde{f}}(\texttt{TF}_{\theta};\gamma)\right]  
\end{equation}
where
$$L_{\tilde{f}}(\texttt{TF}_{\theta};\gamma) \coloneqq \mathbb{E}_{(H_t,a_t^*)\sim\mathcal{P}_{\gamma,\tilde{f}}} \left[\sum_{t=1}^T l\left(\texttt{TF}_{\theta}\left(H_t\right),a_t^*\right)\right].$$
Here $\tilde{f}$ is the pre-specified decision function used in generating the pre-training data $\mathcal{D}_{\text{PT}}$. We note that the discrepancy between $L(\texttt{TF}_{\theta})$ and $L_{\tilde{f}}(\texttt{TF}_{\theta})$ is caused by the difference in the generations of the samples $(H_{t},a_t^*)$, namely, $\mathcal{P}_{\gamma,\texttt{TF}_{\theta}}$ versus $\mathcal{P}_{\gamma,\tilde{f}}$; which essentially reduces to how the actions $a_t$'s are generated in $H_t$. In the training phase, this is generated based on $\tilde{f}$, while measuring the test loss requires that the actions to be generated are based on the transformer $\texttt{TF}_{\theta}.$ This discrepancy is somewhat inevitable because when generating the training data, there is no way we can know the final parameter $\hat{\theta}.$ Consequently, there is no direct guarantee on the loss for $L(\texttt{TF}_{\hat{\theta}};\gamma)$ or $L(\texttt{TF}_{\hat{\theta}})$ with $\hat{\theta}$ being the final learned parameter. This out-of-distribution (OOD) issue motivates the design of our algorithm.

\begin{algorithm}[ht!]
\caption{Supervised pre-training transformers for sequential decision making}\
\label{alg:SupPT}
\begin{algorithmic}[1] 
\Require Iterations $M$, early training phase $M_0\in[1,M]$, number of training sequences per iteration $n$, ratio $\kappa \in[0,1]$, pre-training decision function $\tilde{f}$, prior distribution $\mathcal{P}_\gamma$, initial parameter $\theta_0$
\State Initialize $\theta_1=\theta_0$
\For{$m=1,\ldots,M_0$}
\Statex \ \ \ \ \textcolor{blue}{\%\% Early training phase}
\State Sample $\gamma_1, \gamma_2,\ldots,\gamma_n$ from $\mathcal{P}_\gamma$ and $S_i=\left\{\left(H_1^{(i)},a_1^{(i)*}\right),\ldots,\left(H_T^{(i)},a_T^{(i)*}\right)\right\}$ from $\mathcal{P}_{\gamma_i,\tilde{f}}$
\State Optimize over the generated dataset $\mathcal{D}_m=\{S_i\}_{i=1}^n$ and obtain the updated parameter $\theta_{m+1}$
\EndFor
\For{$m=M_0+1,\ldots,M$}
\Statex \ \ \ \ \textcolor{blue}{\%\% Mixed training phase}
\State Sample $\gamma_1, \gamma_2,\ldots,\gamma_n$ from $\mathcal{P}_\gamma$
\State For $i=1,\ldots,\kappa n$, sample $S_i=\left\{\left(H_1^{(i)},a_1^{(i)*}\right),\ldots,\left(H_T^{(i)},a_T^{(i)*}\right)\right\}$ from $\mathcal{P}_{\gamma_i,\tilde{f}}$
\State For $i=\kappa n,\ldots,n$, sample $S_i=\left\{\left(H_1^{(i)},a_1^{(i)*}\right),\ldots,\left(H_T^{(i)},a_T^{(i)*}\right)\right\}$ from $\mathcal{P}_{\gamma_i,\texttt{TF}_{\theta_m}}$
\State Optimize over the generated dataset $\mathcal{D}_m=\{S_i\}_{i=1}^n$ and obtain the updated parameter $\theta_{m+1}$
\EndFor
\State \textbf{Return}: $\hat{\theta} = \theta_{M+1}$
\end{algorithmic}
\end{algorithm}

Algorithm \ref{alg:SupPT} describes our algorithm to resolve the above-mentioned OOD issue between training and testing. It consists of two phases, an early training phase and a mixed training phase. For the early training phase, the data are generated from the distribution $\mathcal{P}_{\gamma_i,\tilde{f}}$, while for the mixed training phase, the data are generated from both $\mathcal{P}_{\gamma_i,\tilde{f}}$ and $\mathcal{P}_{\gamma_i,\texttt{TF}_{\theta_t}}$ with a ratio controlled by $\kappa$. The decision function $\tilde{f}$  can be chosen as a uniform distribution over the action space or other more complicated options. Intuitively, the benefits of involving the data generated from $\mathcal{P}_{\gamma_i,\texttt{TF}_{\theta_t}}$ is to make the sequences $H_{t}^{(i)}$'s in the training data closer to the ones generated from the transformer. For the mixed training phase, a proportion of the data is still generated from the original $\mathcal{P}_{\gamma_i,\tilde{f}}$ because the generation from $\mathcal{P}_{\gamma_i,\texttt{TF}_{\theta_t}}$ costs relatively more time. For our numerical experiments, we choose $\kappa=1/3.$

\begin{figure}[ht!]
  \centering
  \begin{subfigure}[b]{0.5\textwidth}
    \centering
    \includegraphics[width=\textwidth]{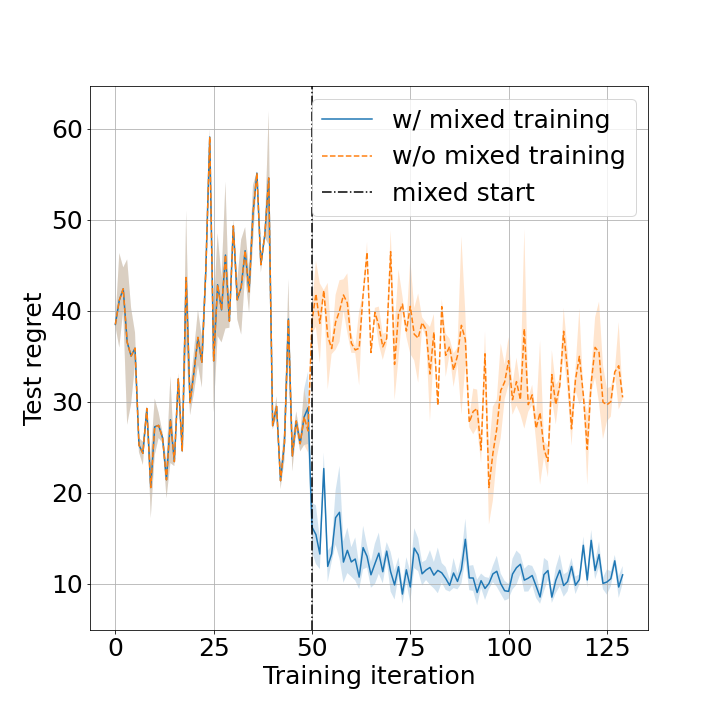}
    \caption{Training dynamics}
    \label{fig:ICL_training_reg}
  \end{subfigure}\hfill
  \begin{subfigure}[b]{0.5\textwidth}
    \centering
    \includegraphics[width=\textwidth]{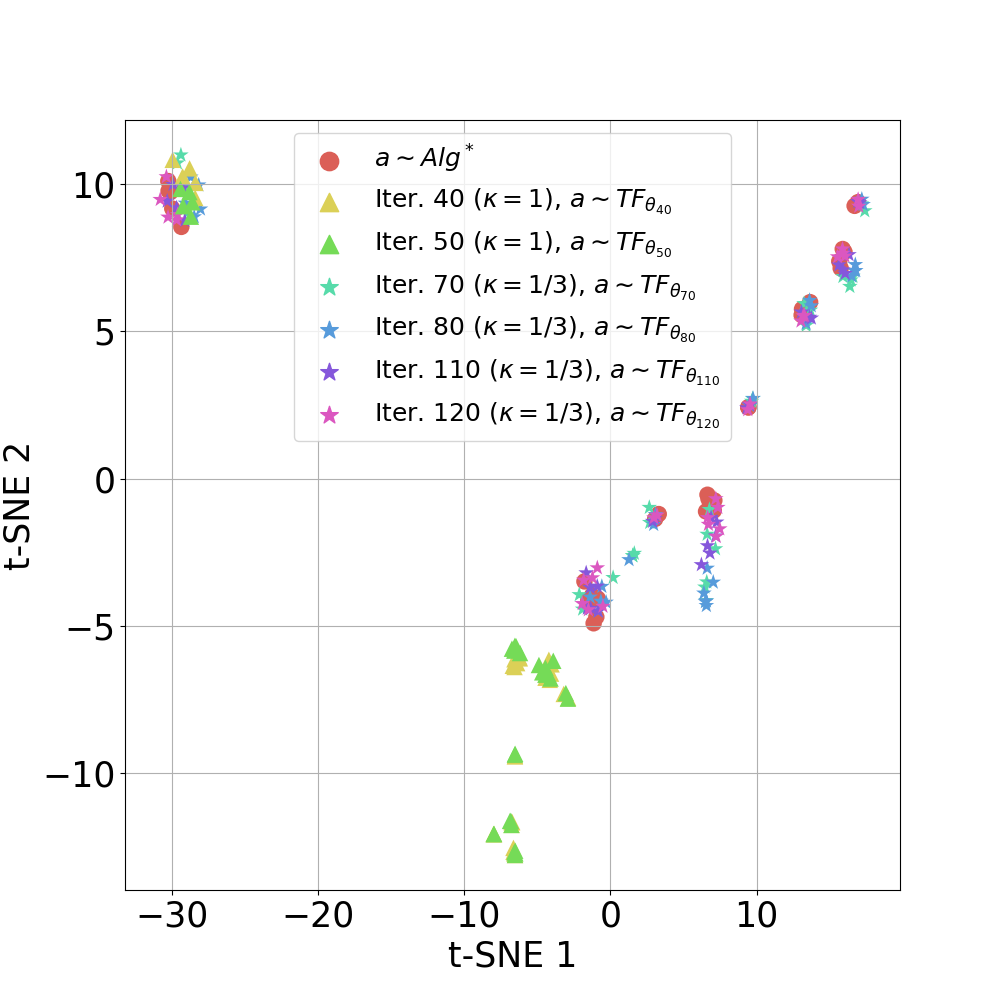}
    \caption{t-SNE of the generated $H_{30}$'s}
    \label{fig:tsne}
  \end{subfigure}
  \caption{(a) Training dynamics. Orange: $M_0=M=130.$ Blue: $M_0=50$ and $M=130$. It shows the effectiveness of injecting/mixing the transformer-generated sequence into the training procedure. (b) A visualization of the $H_t$ with $a_{\tau}$'s in $H_t$ generated from various $\texttt{TF}_{\theta_m}.$ For each $\texttt{TF}_{\theta_m}$, we generate 30 sequences. The decision function $\texttt{Alg}^*$ is defined in the next section. We observe (i) there is a shift over time in terms of the transformer-generated action sequence, and thus the training should adaptively focus more on the recently generated sequence like the design in Algorithm \ref{alg:SupPT}; (ii) the action sequence gradually gets closer to the optimal decision function $\texttt{Alg}^*.$ The experiment details and setups are deferred to the end of this subsection.}
  \label{fig:train_OOD}
\end{figure}

Algorithm \ref{alg:SupPT} from an optimization perspective can be viewed as an iterative procedure to optimize 
$$\mathbb{E}_{\gamma\sim\mathcal{P}_\gamma}\left[\mathbb{E}_{(H_t,a_t^*)\sim\kappa \mathcal{P}_{\gamma,\tilde{f}}+(1-\kappa)\mathcal{P}_{\gamma,\texttt{TF}_{\theta}}} \left[\sum_{t=1}^T l\left(\texttt{TF}_{\theta}\left(H_t\right),a_t^*\right)\right]\right]$$
where both the input data pair $(H_t,a_t^*)$ and the prediction function involve the parameter $\theta$. This falls into the paradigm of performative prediction \citep{perdomo2020performative, izzo2021learn, mendler2020stochastic} where the prediction model may affect the data generated to be predicted. In the literature of performative prediction, a critical matter is the instability issue which may cause the parameter $\theta_{m}$ to oscillate and not converge. \cite{perdomo2020performative} shows that the matter can be solved with strong conditions such as smoothness and strong convexity on the objective function. However, we do not encounter this instability in our numerical experiment, and we make an argument as the following claim. That is, when the underlying function class, such as $\{\texttt{TF}_{\theta}:\theta\in \Theta\}$ is rich enough, one does not need to worry about such instability.

\begin{claim}
\label{claim:pp}
The instability of a performative prediction algorithm and the resultant non-convergence does not happen when the underlying prediction function is rich enough.
\end{claim}

To see this claim, consider the following setup where a feature-target pair $(X,Y)$ is generated from some distribution $\mathcal{P}_{X,Y}(\theta)$. Then we consider the minimization of the following loss
$$\min_{\theta}\mathbb{E}_{\mathcal{P}_{X,Y}(\theta)}[l(f_{\theta}(X),Y)] $$
for some loss function $l:\mathbb{R}\times \mathbb{R}\rightarrow \mathbb{R}.$

Under this formulation, both the data generation distribution $\mathcal{P}_{X,Y}(\theta)$ and the prediction function $f_{\theta}$ are parameterized by $\theta$. This describes the \textit{performative prediction} problem \citep{perdomo2020performative}. 

In addition, we note that in our pre-trained transformer setting, the joint distribution can be factorized in
$$\mathcal{P}_{X,Y}(\theta) = \mathcal{P}_{X}(\theta)\cdot  \mathcal{P}_{Y|X}.$$
That is, the parameter $\theta$ only induces a covariate shift by affecting the marginal distribution of $X$, but the conditional distribution $\mathcal{P}_{Y|X}$ remains the same for all the $\theta$. This exactly matches our setting of pretrained transformer; to see this, different decision functions, $f$ or $\texttt{TF}_{\theta_m}$, only differ in terms of the generation of the actions $a_{\tau}$'s in $H_t$ (for $\tau=1,...,t-1$) which is the features $X$ in this formulation but the optimal action $a_t^*$ will only be affected by $X_t$ in $H_t$.

Also, under this factorization, we define the Bayes-optimal estimator as 
$$f^*(X) = \min_{y} \mathbb{E}_{\mathcal{P}_{Y|X}}[l(y,Y)|X].$$
Then it is easy to note that when $f^*=f_{\theta^*}$ for some $\theta^*$, and have 
$$\theta^* \in \argmin_{\theta}\mathbb{E}_{\mathcal{P}_{X,Y}(\theta')}[l(f_{\theta}(X),Y)] $$
for any $\theta'.$ 

In this light, the oscillating behavior of the optimization algorithms in \cite{perdomo2020performative} will not happen. Because for all the data generation distribution $\mathcal{P}_{X,Y}(\theta')$, they all point to one optimal $\theta^*$. 

Back to the context of the pre-trained transformer, such a nice property is contingent on two factors: (i) the transformer function class is rich enough to cover $f^*$; (ii) there are infinitely many samples, or we can use the expected loss. Also, the above argument is connected to the argument of Proposition \ref{prop:BO}. 

\subsubsection{Details for Figure \ref{fig:train_OOD}}

\paragraph{Figure \ref{fig:train_OOD} (a).} In Figure \ref{fig:train_OOD} (a), we independently implement Algorithm \ref{alg:SupPT} to train two OMGPTs on a dynamic pricing task, which has  $6$-dimensional contexts (more details can be found in Appendix \ref{appx:envs}). For both OMGPTs, the training parameters are identical except for the $M_0$ in Algorithm \ref{alg:SupPT}: the first curve (blue, thick line) uses $M_0=50$, while the second curve (orange, dashed line) uses $M_0=130$, meaning no transformer-generated data is utilized during training. All other parameters follow the configuration detailed in Appendix \ref{appx:alg1_imple}. The figure shows the mean testing regret at each training iteration across 128 randomly sampled environments, with the shaded area representing the standard deviation.

\paragraph{Figure \ref{fig:train_OOD} (b).}  In Figure \ref{fig:train_OOD} (b), we consider a dynamic pricing task with a pool of $8$ linear demand functions and has $6$-dimensional contexts (more details can be found in Appendix \ref{appx:envs}). We generate $30$ sequences from $\texttt{TF}_{\theta_m}$ across different training iterations for this task: before using the transformer-generated data in the training ($m=40, 50$) and after using such data ($m=70,80,110,120$).  All generated sequences share the same context sequence $\{X_t\}_{t=1}^T$ to control the effect of contexts on the chosen actions and the resulting observations. We stack the first $20$ actions and observations into a single sample point and use t-SNE method to visualize these high-dimension points as shown in Figure \ref{fig:train_OOD} (b). We further apply the same method to generate points from $\texttt{Alg}^*$.

We can observe that: (i) compared to the points from $\texttt{TF}$ which have not been trained on self-generated data (i.e., $\texttt{TF}_{\theta_{40}}, \texttt{TF}_{\theta_{50}}$), the points from the trained ones are closer to the  $\texttt{Alg}^*$'s, i.e., the  expected decision rule of a well-trained \texttt{TF}; (ii) When being trained with more self-generated data,  the points from $\texttt{TF}_{\theta_m}$ get closer to the $\texttt{Alg}^*$'s. These observations suggest that utilizing the self-generated data (like in Algorithm \ref{alg:SupPT}) can help mitigate the OOD issue as discussed above: the data from $\texttt{TF}_{\theta_m}$ can get closer to the target samples from $\texttt{Alg}^*$ during the training and thus the pre-training data is closer to the testing data.

\subsection{Pre-training Implementation}
\label{appx:alg1_imple}
During pre-training, we use the AdamW optimizer with a learning rate of $10^{-4}$ and a weight decay of $10^{-4}$. The dropout rate is set to 0.05. For implementing Algorithm \ref{alg:SupPT} in our experiments, we set the total iterations $M=130$ with an early training phase of $M_0=50$. The number of training sequences per iteration is $n=1500 \times 64$, which are randomly split into 1500 batches with a batch size of 64. The transformer's parameter $\theta$ is optimized to minimize the averaged loss of each batch. As suggested in Proposition \ref{prop:surro}, we select the cross-entropy loss for the multi-armed bandits, squared loss for the dynamic pricing, and absolute loss for the linear bandits and newsvendor problem. Further, we select the squared loss for the queuing control and the revenue management.

To reduce computation costs from sampling histories during the early training phase, instead of sampling a new dataset $D_m$ in each iteration (as described in Algorithm \ref{alg:SupPT}), we initially sample $10^6$ data samples before the pre-training phase as an approximation of $\mathcal{P}_{\gamma,\tilde{f}}$. Each sample in the batch is uniformly sampled from these $10^6$ data samples.

To reduce transformer inference costs during the mixed training phase, we reset the number of training sequences to $n=15 \times 64$ per iteration and set the ratio $\kappa=1/3$, meaning $10 \times 64$ samples are from $\mathcal{P}_{\gamma,\texttt{TF}_{\theta_m}}$ and $5 \times 64$ samples are from $\mathcal{P}_{\gamma,\tilde{f}}$ (sampled from the pool of pre-generated $10^6$ samples as before). Consequently, we adjust the number of batches from 1500 to 50 to fit the smaller size of training samples, keeping the batch size at 64.

\paragraph{Decision Function $\tilde{f}$ in Pre-training.} To mitigate the OOD issue mentioned in Section \ref{sec:algo}, we aim for the decision function $\tilde{f}$ to approximate $\texttt{Alg}^*$ or $\texttt{TF}_{\hat{\theta}}$. However, due to the high computation cost of $\texttt{Alg}^*$ and the unavailability of $\texttt{TF}_{\hat{\theta}}$, we set 
\begin{equation}
\tilde{f}(H_t)=a^*_t+\epsilon'_t 
\label{eqn:tilde_f}
\end{equation}
 where $\epsilon'_t$ is random noise simulating the suboptimality of $\texttt{Alg}^*$ or $\texttt{TF}_{\hat{\theta}}$ and independent of $H_t$. As we expect such suboptimality to decrease across $t$, we also reduce the influence of $\epsilon'_t$ across $t$.  Thus, in our experiments, $\epsilon'_t$ is defined as:
\begin{equation*}
  \epsilon'_t \sim
    \begin{cases}
      0 & \text{w.p.} \ \max\{0,1-\frac{2}{\sqrt{t}}\},\\
      \text{Unif}[-1,1] & \text{w.p.} \ \min\{1,\frac{2}{\sqrt{t}}\},
    \end{cases}       
\end{equation*}
where for discrete action spaces we replace $\text{Unif}[-1,1]$ by $\text{Unif}\{-2,-1,1,2\}$ and for continuous discrete action spaces with dimension $d>1$ we apply the uniform distribution of the set $[-1,-1]^d$.   We further project $\tilde{f}(H_t)$ into $\mathcal{A}$ when $\tilde{f}(H_t)\notin \mathcal{A}$.

\paragraph{Curriculum.} Inspired by \citet{garg2022can} in the regression task, we apply curriculum training to potentially speed up Algorithm \ref{alg:SupPT}. This technique uses ``simple'' task data initially and gradually increases task complexity during the training. Specifically, we train the transformer on samples with a smaller horizon $\tilde{T}=20$ (generated by truncating samples from $\mathcal{P}_{\gamma,\tilde{f}}$ or $\mathcal{P}_{\gamma,\texttt{TF}_{\theta_m}}$) at the beginning and gradually increase the sample horizon to the target $T=100$. We apply this curriculum in both the early training phase ($m \leq 50$) and the mixed training phase ($m > 50$). The last 30 iterations focus on training with $\tilde{T}=100$ using non-truncated samples from $\mathcal{P}_{\gamma,\texttt{TF}_{\theta}}$ to let the transformer fit more on the non-truncated samples. The exact setting of $\tilde{T}$ is as follows:
\begin{equation*}
  \tilde{T} =
    \begin{cases}
      20\times (m\%10+1) & \text{when} \ m=1,\ldots,50,\\
      20\times (m\%10-4) & \text{when} \ m=51,\ldots,100,\\
      100 & \text{when} \ m=100,\ldots,130.
    \end{cases}       
\end{equation*}

\subsection{More Experiments on Pre-training}
\label{appx:exps_pretrain}
\begin{figure}[ht!]
  \centering
  \begin{subfigure}[b]{0.45\textwidth}
    \centering
    \includegraphics[width=\textwidth]{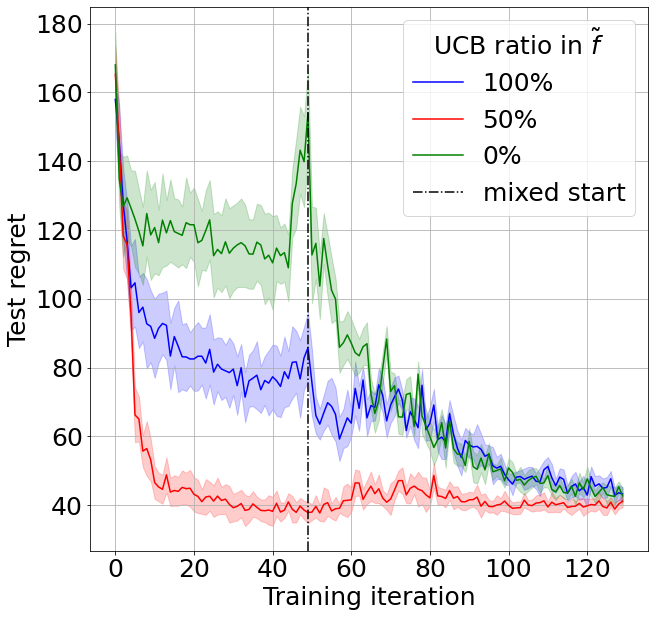}
    \caption{Impact of $\tilde{f}$}
    \label{fig:appx_f}
  \end{subfigure}
  \hfill
  \begin{subfigure}[b]{0.45\textwidth}
    \centering
    \includegraphics[width=\textwidth]{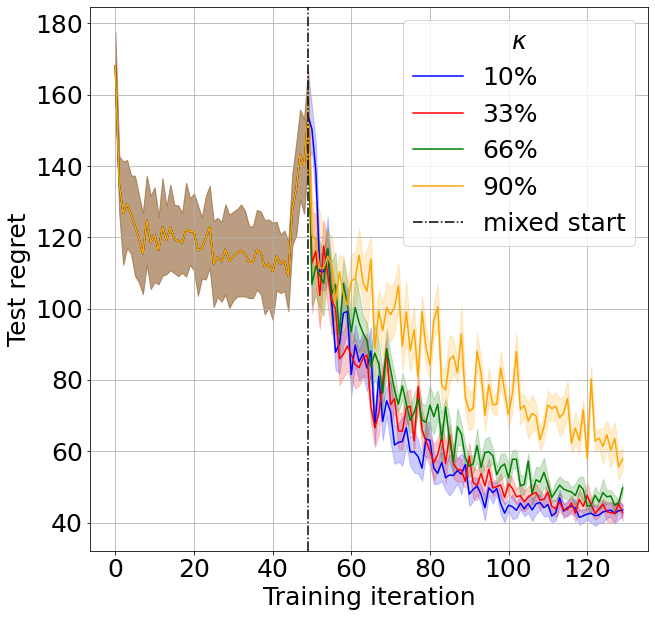}
    \caption{Impact of $\kappa$}
    \label{fig:appx_kappa}
  \end{subfigure}
  \caption{(a) The effect of $\tilde{f}$ (decision function for generating the pre-training data), where we mix the UCB algorithm \citep{lattimore2020bandit} and decision function \eqref{eqn:tilde_f} with different ratios to create the pre-training data.  (b) The effect of $\kappa$ (ratio of samples generated by $\tilde{f}$ in the mixed training phase).}
  \label{fig:appx_train_exps}
\end{figure}

In this section, we present ablation studies to explore the impact of two key factors: the decision function $\tilde{f}$ used for generating pre-training data, and the mix ratio $\kappa$ applied in Algorithm \ref{alg:SupPT}. Figure \ref{fig:appx_train_exps} summarizes the results, where we pre-train and test models in multi-armed bandits problems defined in Appendix \ref{appx:envs}. 

\begin{itemize} \item \textbf{Effect of  $\tilde{f}$}:
We evaluate the influence of different decision functions $\tilde{f}$ on the performance of OMGPT in a multi-armed bandit task. Three types of decision functions are considered: the one we designed, as defined in \eqref{eqn:tilde_f}, the UCB algorithm for multi-armed bandits \citep{lattimore2020bandit}, and a 50/50 mixture of both. The training dynamics for each function are shown in Figure \ref{fig:appx_f}. We observe that all three approaches achieve similar final performance once training ends. While it is possible that some decision functions could lead to faster convergence or better performance after pre-training, we did not encounter any training failures due to the choice of $\tilde{f}$.
\item \textbf{Effect of $\kappa$}:
We also conduct an experiment to numerically investigate the impact of different mix ratios $\kappa$. The results are presented in Figure \ref{fig:appx_kappa}. We test four values of $\kappa$ $(10\%, 33\%, 66\%, 90\%)$, keeping all other hyperparameters constant for a multi-armed bandit task. The findings show that OMGPT’s performance remains robust across different values of $\kappa$, except when $\kappa = 0.9$, where performance declines. This supports our earlier discussion in Algorithm \ref{alg:SupPT}, highlighting the importance of mixing a non-trivial amount of self-generated data (i.e., non-trivial value for $1-\kappa$). 
\end{itemize}

\section{Proofs for Section \ref{sec:analysis}}
\label{appx:proofs}
\subsection{Proofs for Subsection \ref{sec:Bayes_decision_maker}}

\paragraph{Proof of Proposition \ref{prop:BO}}

\begin{proof}
The proof can be done with the definition of $\texttt{Alg}^*.$ Specifically, note $$\mathbb{E}\left[\sum_{t=1}^Tl(f(H_t),a^*_t)\right]=\mathbb{E}\left[\sum_{t=1}^T\mathbb{E}\left[l(f(H_t),a^*_t)|H_t\right]\right]\geq \mathbb{E}\left[\sum_{t=1}^T\mathbb{E}\left[l(\texttt{Alg}^*(H_t),a^*_t)|H_t\right]\right]$$
for any $f\in \mathcal{F}$, we finish the proof.
\end{proof}

\paragraph{Proof of Corollary \ref{cor:post}}
\begin{proof}
Since $\texttt{Alg}^*$ is a function of any possible history $H_t$, we omit the subscrpit $t$ for the optimal actions for notation simplicity, and use $a^{(\gamma)*}$ as the optimal action of $\gamma$.

\begin{itemize}
    \item \textbf{Posterior sampling under the cross-entropy loss.} We assume that $\mathcal{A}=\{a_1,\ldots,a_J\}$ and the output of $\texttt{Alg}^*$ is a probability vector $(p_1,\ldots,p_J)$ such that $\sum_{j=1}^J p_j=1$, meaning the probability of choosing each action. Then  
    \begin{align*}
    \texttt{Alg}^*(H) 
    &= \argmin_{\sum_{j=1}^J p_j=1} \int_{\gamma}-\log\left(p_{a^{(\gamma)*}}\right)\mathbb{P}(H|\gamma)\mathrm{d}\mathcal{P}_{\gamma}=
    \argmin_{\sum_{j=1}^J p_j=1}\ \sum_{j=1}^J\left(-\log\left(p_{j}\right)\right)
    \int_{\left\{\gamma:\ a^{(\gamma)*}=j\right\}}\mathbb{P}(H|\gamma)\mathrm{d}\mathcal{P}_{\gamma}.
\end{align*}

By nothing $ \int_{\left\{\gamma:\ a^{(\gamma)*}=j\right\}}\mathbb{P}(H|\gamma)\mathrm{d}\mathcal{P}_{\gamma}$ is the probability such that action $j$ is the optimal action conditional on $H$ (i.e., the posterior distribution of the optimal action), we are indeed minimizing the cross-entropy of the posterior distribution of the optimal action relative to the decision variables $p_j$. Thus, the optimal solution $p^*_j= \int_{\left\{\gamma:\ a^{(\gamma)*}=j\right\}}\mathbb{P}(H|\gamma)\mathrm{d}\mathcal{P}_{\gamma}$ for each $j$ and  $\texttt{Alg}^*(H_t)$ behaves as the posterior sampling. 
    \item \textbf{Posterior averaging under the squared loss.} By definition,
    \begin{align*}
    \texttt{Alg}^*(H) &= \argmin_{a\in \mathcal{A}} \int_{\gamma} ||a-a^{(\gamma)*}||^2_2 \mathbb{P}(H|\gamma) \mathrm{d}\mathcal{P_{\gamma}}\\
    &=\frac{\int_{\gamma} a^{(\gamma)*}\mathbb{P}(H|\gamma) \mathrm{d}\mathcal{P_{\gamma}}}{\int_{\gamma} \mathbb{P}(H|\gamma) \mathrm{d}\mathcal{P_{\gamma}}}\\
    &=\mathbb{E}_{\gamma}[a^{(\gamma)*}|H]
\end{align*}
where the second line is by the first order condition.
\item \textbf{Posterior median under the absolute loss.} By definition, 
    \begin{align*}
    \texttt{Alg}^*(H) &= \argmin_{a\in \mathcal{A}} \int_{\gamma} |a-a^{(\gamma)*}| \mathbb{P}(H|\gamma) \mathrm{d}\mathcal{P_{\gamma}}.
\end{align*}

Then by zero-subgradient condition,
we can conclude that $\texttt{Alg}^*(H)$ satisfies $$\int_{\left\{\gamma:\  \texttt{Alg}^*(H)\leq a^{(\gamma)*}\right\}} \mathbb{P}(H|\gamma)\mathrm{d}\mathcal{P_{\gamma}}=\int_{\left\{\gamma:\  \texttt{Alg}^*(H)\geq a^{(\gamma)*}\right\}}\mathbb{P}(H|\gamma)\mathrm{d}\mathcal{P_{\gamma}}.$$ Divide both sides of this equation by $\int_{\gamma} \mathbb{P}(H|\gamma)\mathrm{d}\mathcal{P_{\gamma}}$, we can conclude that 
$$\int_{\left\{\gamma:\  \texttt{Alg}^*(H)\leq a^{(\gamma)*}\right\}}\mathrm{d}\mathcal{P}\left({\gamma}| H\right) =\int_{\left\{\gamma:\  \texttt{Alg}^*(H)\geq a^{(\gamma)*}\right\}}\mathrm{d}\mathcal{P}\left({\gamma}| H\right)$$
where $\mathcal{P}\left(\gamma| H_t\right)$ is the posterior distribution of $\gamma$. Hence $\texttt{Alg}^*(H) $ is the posterior median.
\end{itemize}
\end{proof}

\subsection{Proof of Proposition \ref{prop:surro}}

To prove Proposition \ref{prop:surro}, we assume the following mild conditions. First, for all these problems, the noise random variable $\epsilon_t$ has mean 0, is i.i.d. across time, and is independent of the action $a_t$. Second, for the linear bandits problem, the $L_{\infty}$ norm of the unknown parameter $w$ is bounded by $D$. Third, for the dynamic pricing problem, we assume the demand function is differentiable with respect to the action coordinate, and both the absolute value of the first- and second-order derivatives is bounded by $D$. Further details regarding the problem formulations can be found in Appendix \ref{appx:envs}.

\begin{proof}
Since both the regret $\text{Regret}(f; \gamma)$ and the prediction error $L(f; \gamma)$ are the sums of single-step regret and prediction error over the horizon, it is sufficient to prove the single-step surrogate property of the loss function. Therefore, in the following proof, we will focus on proving the single-step surrogate property and omit the subscript $t$ for simplicity.

\begin{itemize}
    \item For the multi-arm bandits problem, without loss of generality, we assume $\mathcal{A}=\{a_1,\ldots,a_J\}$ and the optimal arm is $a_1$. For $1\leq i \leq J$, let $\mu_j$ be the mean of the reward distribution for arm $j$, and define $\Delta_j=\mu_1-\mu_j$ as the reward gap for arm $j$. Define $\Delta_{\max}=\max\left\{\Delta_1,\cdots,\Delta_J\right\}$ as the maximum action gap. Under cross-entropy loss, we denote the output distribution of decision function $f$ is $(p_1,\ldots,p_J)$. Then the single-step regret can be bounded by:
    $$\mathbb{E}\left[r\left(X,a^*\right)-r\left(X,f(H)\right)| H\right]=\mu_1-\sum_{j=1}^{J}p_j\mu_j=\sum_{j=2}^{J}p_j\Delta_j \leq \Delta_{\max}
        (1-p_1)\leq -\Delta_{\max}\cdot\log(p_1),$$
        where the last step is by the inequality $1-x\leq -\log(x)$ for $0<x \leq 1$. Then by the definition of cross-entropy loss, we finish the proof.
    \item For the dynamic pricing problem,  we have 
    $$r(X,a^*)-r(X,f(H))=\nabla_a r(X,a^*)(a^*-f(H))+\nabla_a^2 r(X,\tilde{a}) |f(H)-a^*|^2\leq D|f(H)-a^*|^2$$
    where $\tilde{a}$ is in the line of $f(H)$ and $a^*$. Here, the first step is by the Taylor expansion, and the second step is by the first order condition of $a^*$ and the assumption. Then by noting $|f(H)-a^*|^2$ is the square loss, we finish the proof.
    \item For the linear bandits problem,  we have 
    $$r(X,a^*)-r(X,f(H))=w^\top a^*- w^\top f(H)\leq ||w||_{\infty} ||f(H)-a^*||_1\leq D ||f(H)-a^*||_1,$$
    where the second step is by the Holder's inequality and the last step is by the assumption. Then by noting $||f(H)-a^*||_1$ is the absolute loss, we finish the proof.
    \item For the newsvendor problem, we have 
    \begin{align*}
    r(X,a^*)-r(X,f(H))&\leq \max\left\{h,l\right\}\mathbb{E}_X\left[ |a^*-d(X)-f(H)+d(X)|\right]\\
    &=\max\left\{h,l\right\} |a^*-f(H)|
\end{align*}
where the first line is because the (random) reward function is $\max\left\{h,l\right\}$-Lipschitz in $a-d(X)$. Then by noting $|f(H)-a^*|$ is the absolute loss, we finish the proof. 
\end{itemize}
\end{proof}

\subsection{Proof of Proposition \ref{prop:lin_reg}}

\begin{proof}
\textbf{For the linear bandits problem}, we consider the following example. Suppose we have two environments, ${\gamma_1, \gamma_2}$, each with standard normal distributed noise and unknown parameters $w_{\gamma_1} = (1, 0)$ and $w_{\gamma_2} = (0, 1)$, respectively. Let the action space be $\mathcal{A} = [-1, 1] \times [-1, 1]$. The optimal actions for the two environments are 
$a^{(\gamma_1)*}=(1,0)$, $a^{(\gamma_2)*}=(0,1)$. We assume the prior distribution  $\mathcal{P}_{\gamma}$ is given by $\mathcal{P}_{\gamma}(\gamma_1)=\mathcal{P}_{\gamma}(\gamma_2)=\frac{1}{2}$. 

Then if $\texttt{Alg}^*$ is the posterior averaging, we have 
\begin{align*}
    a_t=\texttt{Alg}^*(H_t) = \mathcal{P}_{\gamma}\left(\gamma_1| H_t\right)\cdot a^{(\gamma_1)*}+\mathcal{P}_{\gamma}\left(\gamma_2| H_t\right)\cdot a^{(\gamma_2)*},
\end{align*}
where the posterior distribution  $\mathcal{P}_{\gamma}\left(\gamma_i| H_t\right)$ is given by
\begin{align*}
    \mathcal{P}_{\gamma}\left(\gamma_i| H_t\right) = \frac{\prod_{\tau=1}^{t-1}\mathbb{P}_{\gamma_i}(O_{\tau}|a_{\tau}) }{\sum_{i'=1}^2\prod_{\tau=1}^{t-1}\mathbb{P}_{\gamma_{i'}}(O_{\tau}|a_{\tau}) },\quad i=1,2.
\end{align*}

We now use induction to show $\mathcal{P}_{\gamma}\left(\gamma_i| H_t\right)=\frac{1}{2}$ for any $t \geq 1$ and $i = 1, 2$, where $H_t$ can be generated by either $\gamma_1$ or $\gamma_2$. This results in $a_t=\frac{1}{2}a^{(\gamma_1)*}+\frac{1}{2}a^{(\gamma_2)*}=\left(\frac{1}{2},\frac{1}{2}\right)$, which does not change with respect to $t$ and will cause regret linear in $T$.

Step 1. Since $\mathcal{P}_{\gamma}\left(\gamma_i| H_0\right)=\mathcal{P}_{\gamma}\left(\gamma_i\right)=\frac{1}{2}$, we have $a_1=\frac{1}{2}a^{(\gamma_1)*}+\frac{1}{2}a^{(\gamma_2)*} = (\frac{1}{2},\frac{1}{2})$, and the conclusion holds for $t=1$.

Step 2. Now assume the conclusion holds for $t$, i.e., $\mathcal{P}_{\gamma}\left(\gamma_i| H_t\right)=\frac{1}{2}$, and $a_t=\left(\frac{1}{2},\frac{1}{2}\right)$. Since $w_{\gamma_1}^Ta_t=w_{\gamma_2}^Ta_t=\frac{1}{2}$, we have $\mathbb{P}_{\gamma_i}(O_t|a_t)=\frac{1}{\sqrt{2\pi}}\exp\left(-\frac{\left(O_t-\frac{1}{2}\right)^2}{2}\right)$ for $i=1,2$. Observe that 

\begin{align*}
    \mathcal{P}_{\gamma}(\gamma_i| H_{t+1})
    =\frac{\mathbb{P}_{\gamma_i}(O_{t+1}|a_{t+1})\mathcal{P}_{\gamma}(\gamma_i| H_{t})}{\sum_{i'=1}^2\mathbb{P}_{\gamma_{i'}}(O_{t+1}|a_{t+1})\mathcal{P}_{\gamma}(\gamma_{i'}| H_{t})}=\frac{1}{2},
\end{align*}
which implies $a_{t+1}=\left(\frac{1}{2},\frac{1}{2}\right)$, and the conclusion holds for $t+1$.

Thus, the conclusion holds for all $t\geq 1$. Then the regret is
\begin{align*}
    \text{Regret}(\texttt{Alg}^*;\gamma_i) =\mathbb{E}\left[\sum_{t=1}^T r(X_{t},a_t^*)-r(X_t,a_t)\right]=\frac{1}{2}T
\end{align*} 
for $i=1,2$.

\textbf{For the dynamic pricing problem}, we can construct a similar example.  Suppose we have two environments without context $X_t$, denoted by $\gamma_1, \gamma_2$. The demands $O_t$ of them are set to be $O_t=2-a_t+\epsilon_t$ and  $O_t=\frac{4}{5}-\frac{1}{5}\cdot a_t+\epsilon_t$ respectively, where $\epsilon_t \overset{i.i.d.}{\sim} \mathcal{N}(0,1)$ and $a_t$ is the price. Accordingly, the optimal actions are then $a^{(\gamma_1)*}=1$ and $a^{(\gamma_2)*}=2$.

Then if $\texttt{Alg}^*$ is the posterior averaging, we have 
\begin{align*}
    a_t=\texttt{Alg}^*(H_t) = \mathcal{P}_{\gamma}\left(\gamma_1| H_t\right)\cdot a^{(\gamma_1)*}+\mathcal{P}_{\gamma}\left(\gamma_2| H_t\right)\cdot a^{(\gamma_2)*},
\end{align*}
where the posterior distribution  $\mathcal{P}_{\gamma}\left(\gamma_i| H_t\right)$ is given by
\begin{align*}
    \mathcal{P}_{\gamma}\left(\gamma_i| H_t\right) = \frac{\prod_{\tau=1}^{t-1}\mathbb{P}_{\gamma_i}(O_{\tau}|a_{\tau}) }{\sum_{i'=1}^2\prod_{\tau=1}^{t-1}\mathbb{P}_{\gamma_{i'}}(O_{\tau}|a_{\tau}) },\quad i=1,2.
\end{align*}
with $\mathbb{P}_{\gamma_{1}}(O_{\tau}|a_{\tau})$ as the normal distribution $\mathcal{N}\left(2-a_t,1\right)$ and $\mathbb{P}_{\gamma_{2}}(O_{\tau}|a_{\tau})$ as $\mathcal{N}\left(\frac{4-a_t}{5},1\right)$.

Observe that $a_1=\frac{1}{2}a^{(\gamma_1)*}+\frac{1}{2}a^{(\gamma_2)*}=\frac{3}{2}$ satisfies $2-a_1=\frac{4-a_1}{5}$. Following a similar analysis as in the linear bandits example, we can conclude that $\mathcal{P}_{\gamma}\left(\gamma_i| H_t\right)=\frac{1}{2}$ for any $t$ and $i=1,2$, where $H_t$ can be generated by either $\gamma_1$ or $\gamma_2$. Therefore, $a_t=\frac{3}{2}$ for all $t\geq 1$ and the regret is
\begin{align*}
    \text{Regret}(\texttt{Alg}^*;\gamma_i) =\mathbb{E}\left[\sum_{t=1}^T r(X_{t},a_t^*)-r(X_t,a_t)\right]=\frac{1}{4}T\cdot\mathbb{I}\left\{\gamma_i=\gamma_1\right\}+\frac{1}{20}T\cdot\mathbb{I}\left\{\gamma_i=\gamma_2\right\},
\end{align*}
for $i=1,2$.

\end{proof}

\subsection{Proof and Discussions of  Theorem \ref{thm:reg_bound}}
We first introduce a few key lemmas for proving Theorem \ref{thm:reg_bound}. We denote $\mathcal{P}_t(\cdot)=\mathcal{P}_{\gamma}(\cdot|H_t)$, i.e., the posterior distribution of the (testing) environment at time $t$, which generates $H_t$ through \eqref{eqn:test_dynamics}, conditional on the history $H_t$. And let $\mathbb{E}_{t}[\cdot]$ denote the corresponding conditional expectation. Then under Assumption \ref{assmp:bounded}, \ref{assmp:llh}, \ref{assmp:reward}, we have the following lemmas. Their proofs can be found in the end of this subsection.

\begin{lemma}[Concentration Inequality]
\label{lemma:concen_inq_cts}
Recall $\lambda_{t}=\lambda_{\min}\left(\sum_{\tau=1}^t \Delta(X_{\tau},a_{\tau})\right)$ is the minimum eigenvalue of the matrix $\sum_{\tau=1}^t \Delta(X_{\tau},a_{\tau})$, and $\gamma\in \Gamma$ is the environment to generate $O_t$ and $X_t$ through \eqref{eqn:law_X}. Then given $t>0$, $\gamma'\in\Gamma$ and $1>\delta>0$,  with probability $1-\delta$,
\begin{align*}
    \sum_{\tau=1}^t \log  \left(\frac{\mathbb{P}_{\gamma}(O_{\tau}\vert X_{\tau},a_{\tau})}{\mathbb{P}_{\gamma'}(O_{\tau}\vert X_{\tau},a_{\tau})}\right)&\geq \frac{1}{2}\sum_{\tau=1}^t\|\gamma-\gamma'\|^2_{\Delta(X_\tau,a_\tau)}-4C_{\sigma^2}\log(1/\delta)\\
    &\geq  \frac{1}{2}\lambda_{t}\|\gamma-\gamma'\|^2_2-4C_{\sigma^2}\log(1/\delta),
\end{align*}
for any possible $a_{\tau}'s\in \mathcal{A}$.
\end{lemma}
This lemma provides a concentration inequality for the log-likelihood ratio between the true environment and any other environment, in terms of the ``exploration intensity'' $\lambda_t$ and the distance between the two environments. The proof is based on the standard martingale concentration inequality, along with some algebraic manipulations.

\begin{lemma}
    For any $\gamma_1,\gamma_2,\gamma'\in\Gamma$ such that $\gamma_1+\gamma', \gamma_2+\gamma'\in\Gamma$, we have
    $$\Bigg \vert \log  \left(\frac{\mathbb{P}_{\gamma_1+\gamma'}(O \big \vert X,a)}{\mathbb{P}_{\gamma_1}(O\big \vert X,a)}\right)-\log  \left(\frac{\mathbb{P}_{\gamma_2+\gamma'}(O \big \vert X,a)}{\mathbb{P}_{\gamma_2}(O\big \vert X,a)}\right)\Bigg \vert \leq C_{\Gamma}\|\gamma'\|_2(\|\gamma'\|_2+\|\gamma_1-\gamma_2\|_2),$$
    for any $X\in\mathcal{X}$, $a\in\mathcal{A}$ and $O\in\mathcal{O}$.
\label{lemma:ratio_diff_bound}
\end{lemma}
This lemma bounds the difference in log-likelihood changes between two environments, $\gamma_1$ and $\gamma_2$, when both are shifted by $\gamma'$. On the one hand, this difference is affected by the shift $\gamma'$, where a smaller shift results in a smaller difference. On the other hand, it is also influenced by the distance between the two environments. The proof is based on the smoothness assumed in Assumption \ref{assmp:llh}.

\subsubsection{Proof of Theorem \ref{thm:reg_bound}}

\begin{proof}
Recall $\mathbb{E}_t[\cdot]=\mathbb{E}_{\gamma'}[\cdot|H_t]$ is the expectation with respect to the posterior distribution over the possible (testing) environment $\gamma'$ conditional on $H_t$. The regret can be decomposed as the following
\begin{align*}
        \text{Regret}(\texttt{TF}_{\hat{\theta}};\gamma)&=\sum_{t=1}^T\mathbb{E}[r(X_t,\texttt{Alg}^*(H_t))-r(X_t,\texttt{TF}_{\hat{\theta}}(H_t))]+\sum_{t=1}^T\mathbb{E}[r(X_t,a^*_t)-r(X_t,\texttt{Alg}^*(H_t))]\\
        &\leq \Delta_{\text{Exploit}}T+\sum_{t=1}^T\mathbb{E}[r(X_t,a^*_t)-r(X_t,\texttt{Alg}^*(H_t))]\\
        &\leq \Delta_{\text{Exploit}}T+\bar{r}\mathbb{E}[t_0-1]+\mathbb{E}\left[\sum_{t=t_0}^TC_r\mathbb{E}_{t}[\|\gamma-\gamma'\|_2^2]\right]\\
        &= \Delta_{\text{Exploit}}T+\bar{r}\mathbb{E}[t_0-1]+\mathbb{E}\left[\sum_{t=t_0}^TC_r\mathbb{E}_{t}\left[\|\gamma-\gamma'\|_2^2\mathbbm{1}_{\{\|\gamma-\gamma'\|_2^2< \epsilon_t\}}\right]+\sum_{t=t_0}^TC_r\mathbb{E}_{t}\left[\|\gamma-\gamma'\|_2^2\mathbbm{1}_{\{\|\gamma-\gamma'\|_2^2\geq  \epsilon_t\}}\right]\right]\\
        &\leq \Delta_{\text{Exploit}}T+\bar{r}\mathbb{E}[t_0-1]+C_r\mathbb{E}\left[\sum_{t=t_0}^T\epsilon_{t}\right]+\mathbb{E}\left[\sum_{t=t_0}^TC_r\mathbb{E}_{t}[\|\gamma-\gamma'\|_2^2\mathbbm{1}_{\{\|\gamma-\gamma'\|_2^2\geq  \epsilon_t\}}]\right]
\end{align*}
where in the last two lines we use a sequence $\{\epsilon_t\}_{t=t_0}^T$ (which is defined in Theorem \ref{thm:reg_bound}) to decompose $\|\gamma-\gamma'\|_2^2$ into $\|\gamma-\gamma'\|_2^2\mathbbm{1}_{\{\|\gamma-\gamma'\|_2^2< \epsilon_t\}}$ and $\|\gamma-\gamma'\|_2^2\mathbbm{1}_{\{\|\gamma-\gamma'\|_2^2\geq \epsilon_t\}}$ at each $t\geq t_0$.  The first term is automatically bounded by $\epsilon_t$. Thanks to the concentration inequality in Lemma \ref{lemma:concen_inq_cts}, for the second term, the tail event $\|\gamma-\gamma'\|_2^2\geq \epsilon$ will have a probability exponentially decreasing in $\epsilon\geq \epsilon_t$, which is shown in Lemma \ref{lemma:cts_prob_bound} below. And the second term, as its expectation, can be accordingly bounded by Corollary \ref{corr:cts_prob_bound}.

\begin{lemma}
\label{lemma:cts_prob_bound}
Conditional on $t\geq t_0$, for $\epsilon\geq \epsilon_t$ and $1>\delta>0$,  with probability $1-\delta$,
$$\mathcal{P}_t(\|\gamma-\gamma'\|_2^2\geq \epsilon)\leq \frac{K_t}{K_t+C_t(\epsilon,\delta)}$$
where
$$C_t(\epsilon,\delta)=\exp\left(-4C_{\sigma^2}\log(K_t/\delta)+\frac{1}{16}\lambda_{t-1}\epsilon \right),$$
and $K_t$ is defined in Theorem \ref{thm:reg_bound} and is an upper bound of the covering number of $\Gamma$ (with a radius as a function of $\epsilon_t$ and $\lambda_{t-1}$ and $L_2$ norm).
\end{lemma}

\begin{corollary}
\label{corr:cts_prob_bound}
Conditional on $t\geq t_0$, for $1>\delta>0$, 
$$\mathbb{E}\left[\mathbb{E}_t\left[\|\gamma-\gamma'\|_2^2\mathbbm{1}_{\{\|\gamma-\gamma'\|_2^2\geq \epsilon_t\}}\right]\right]\leq \mathbb{E}\left[\frac{16}{\lambda_{t-1}}\log\left(2K_t(K_t/\delta)^{4C_{\sigma^2}})\right)\right]+4\bar{\gamma}^2\delta,$$
where the expectation on the right-hand-side is with respect to the potential randomness in $\lambda_{t-1}$ and $K_t$ is defined in Theorem \ref{thm:reg_bound}.
\end{corollary}

By combining all together and choosing $\delta=1/T$, we have 
$$\text{Regret}(\texttt{TF}_{\hat{\theta}};\gamma)\leq \Delta_{\text{Exploit}}T+\bar{r}\mathbb{E}[t_0-1]+C_r\mathbb{E}\left[\sum_{t=t_0}^T\epsilon_{t}\right]+16C_r(4C_{\sigma_2}+1)\mathbb{E}\left[\sum_{t=t_0}^T\frac{\log\left(2K_tT\right)}{\lambda_{t-1}}\right]+4C_r\bar{\gamma}^2.$$
\end{proof}

\paragraph{Discussions on Technique Contributions}

The key challenge in the proof is to uniformly upper bound the likelihood ratio $\frac{p_t(\gamma')}{p_t(\gamma)}$ for all $\gamma' \in \Gamma$. However, the commonly used covering number technique from the machine learning literature, which creates covering balls to discretize and approximate continuous spaces by their centers, encounters several issues when directly applied here. Thus, we developed new techniques for the proof of Theorem \ref{thm:reg_bound}. Specifically:
\begin{itemize}
    \item \textbf{Dependence of the covering set on $\gamma$}: In traditional covering techniques, the covering set (or sets, as in Dudley's theorem, which will be discussed later) is constructed independently of $\gamma$. In this case, $\gamma'$ and $\gamma$ could fall within the same ball, and approximating both with the same center could yield an upper bound for $\frac{p_t(\gamma')}{p_t(\gamma)}$ of 1, which is meaningless. The failure arises because traditional covering techniques focus on a single point (e.g., $\gamma$), while we are interested in two points, $\gamma'$ and $\gamma$. In our proof, we construct covering sets that depend on the underlying environment $\gamma$, allowing us to isolate it from other balls.
    \item \textbf{Larger ball for $\gamma$}: In traditional covering techniques, each ball has the same radius, and this radius is used to uniformly bound the approximation error across all balls when using their centers to approximate any point within them. However, in our case, due to the $\gamma$-dependent covering set, the ball centered at $\gamma$ may overlap with other balls (e.g., $B(\tilde{\gamma}, \epsilon)$), which could lead to large approximation errors. This is because points in $B(\tilde{\gamma}, \epsilon)$ could be arbitrarily close to $\gamma$, resulting in a likelihood ratio of 1, which could be far from the approximation $\frac{p_t(\tilde{\gamma})}{p_t(\gamma)}$. To address this, we adjust the boundaries of $\gamma$'s ball, ensuring that points from other balls are sufficiently distant from $\gamma$.
    \item \textbf{Our ``chaining trick'' differs from Dudley's theorem:} Dudley's theorem (also known as chaining, see Lemma 27.4 of \citep{shalev2014understanding}) is typically used to sharpen bounds, such as the uniform bound of $\frac{p_t(\gamma')}{p_t(\gamma)}$, by constructing a sequence of covering sets with decreasing radii. However, as mentioned earlier, Dudley's covering sets are independent of $\gamma$, whereas our approach requires the dependence. Our strategy is to reformulate the desired result as $\mathbb{E}_t\left[\|\gamma-\gamma'\|_2^2\right]=\int_{\epsilon>0}\mathcal{P}_t(\|\gamma-\gamma'\|_2^2\geq \epsilon)d\epsilon$  and construct a sequence of covering sets related to $\epsilon$. This technique is inspired by \citet{keskin2014dynamic}, which bounds the estimation error of an estimated optimal action rather than the likelihood ratio as we do.
\end{itemize}

\subsubsection{Proof of Example \ref{example:cts_DP}}
\begin{proof}
We first assume there exists constants $\bar{\gamma},\bar{X},\underline{a},\bar{a}>0$ such that $\Gamma=\{\gamma \in \mathbb{R}^{2d}\vert \|\gamma\|_2\leq \bar{\gamma} ] \}$, $\mathcal{X}=\{X \in \mathbb{R}^{d}| \|X\|_2\leq \bar{X} ] \}$, $\mathcal{A}=[\underline{a},\bar{a}]$, and $\mathcal{P}_{\gamma}$ is a uniform distribution in $\Gamma$. We further assume the minimum eigenvalue of the context's covariance matrix is lower bounded by $\underline{\lambda}>0$. Then we verify the assumptions in Theorem \ref{thm:reg_bound} one by one. For notation simplicity, we denote $Z=(X,a\cdot X)^\top$.

\textbf{Assumption \ref{assmp:bounded}.} This is automatically satisfied by the above assumptions.

\textbf{Assumption \ref{assmp:llh}.} We define $\Delta(X,a)=\frac{1}{2}ZZ^\top$. Then we have 
$$\log\left(\frac{\mathbb{P}_{\gamma}(O_t|X,a)}{\mathbb{P}_{\gamma'}(O_t|X,a)}\right)=\frac{-(D_t-\alpha^\top X-\beta^\top X\cdot a)^2+(D_t-\alpha'^\top X-\beta'^\top X\cdot a)^2}{2}=\|\gamma-\gamma'\|^2_{\Delta(X,a)}-(\gamma-\gamma')^\top Z\cdot \epsilon_t,$$
where $\gamma'=(\alpha',\beta')$. Then 
\begin{itemize}
    \item Since $\epsilon_t$ has $0$ mean and is independent from $Z$, we have 
    \begin{align*}
        \text{KL}(\mathbb{P}_{\gamma}(\cdot|X,a)|\mathbb{P}_{\gamma'}(\cdot|X,a))&=\mathbb{E}_{\epsilon_t}\left[\|\gamma-\gamma'\|^2_{\Delta(X,a)}-(\gamma-\gamma')^\top Z\cdot \epsilon_t\right]=\|\gamma-\gamma'\|^2_{\Delta(X,a)}
    \end{align*}
    \item For the term $(\gamma-\gamma')^\top Z\cdot \epsilon_t$, it follows a Normal distribution with variance $\left((\gamma-\gamma')^\top Z\right)^2=\|\gamma-\gamma'\|^2_{ZZ^\top}=2\|\gamma-\gamma'\|^2_{\Delta(X,a)}$. Thus, we have $C_{\sigma^2}=\sqrt{2}$.
    \item $\log\mathbb{P}_{\gamma'}(O|X,a)=\frac{-(D_t-\alpha'^\top X-\beta'^\top X\cdot a)^2}{2}$ which is concave in $\gamma'\in\Gamma$. Further, we have 
     $$\|\nabla^2_{\gamma'} \log \mathbb{P}_{\gamma'}(O\big \vert X,a)\|^2_2= \|ZZ^\top\|_2^2=\|Z\|_2^4\leq (1+\bar{a}^2)^2\bar{X}^4$$
     due to the boundedness assumptions on $X$ and $a$. Thus, we can assign $C_{\Gamma}=(1+\bar{a}^2)^2\bar{X}^4$.
\end{itemize}

\textbf{Assumption \ref{assmp:reward}.} For notation simplicity, we denote $\texttt{Alg}^*(H_t)=\tilde{a}^*_t$
\begin{itemize}
    \item  By assumptions,
    \begin{align*}
        \mathbb{E}\left[r(X_t,\texttt{Alg}^*(H_t))-r(X_t,\texttt{TF}_{\hat{\theta}}(H_t)|H_t\right]&=\mathbb{E}\left[\tilde{a}^*_t(\alpha^\top X_t+\beta^\top X_t\cdot \tilde{a}^*_t)-(\tilde{a}^*_t+\Delta_t)(\alpha^\top X_t+\beta^\top X_t\cdot (\tilde{a}^*_t+\Delta_t)|H_t\right]\\
        &=\mathbb{E}\left[-\Delta_t(\alpha^\top X_t+2\beta^\top X_t\cdot \tilde{a}^*_t)-\beta^\top X_t\Delta_t^2|H_t\right]\\
        &=-\mathbb{E}\left[\beta^\top X_t\Delta_t^2|H_t\right]\\
        &\leq \frac{\bar{X}\bar{\gamma}C^2}{\sqrt{T}}
    \end{align*}
    where the  second last line is because $\Delta_t$ is zero mean and is independent of $H_t,\tilde{a}^*_t$ and $\gamma$, and the last line is because of the boundedness assumptions. Thus, we can assign $\Delta_{\text{Exploit}}=\frac{\bar{X}\bar{\gamma}C^2}{\sqrt{T}}$.
    \item By the proof of Theorem 2 in \cite{ban2021personalized} and Jensen's inequality, there exists a constant $C_0>0$, which depends on the boundedness assumptions, such that
    \begin{align*}
        \mathbb{E}[r(X_t,a^*_t)-r(X_t,\texttt{Alg}^*(H_t))|H_t]
        &=-\beta^\top X_t\mathbb{E}[\left(a^*_t-\tilde{a}^*_t\right)^2|H_t]\\
        &\leq C_0 \bar{\gamma}\bar{X} \mathbb{E}_{\gamma'}[\|\gamma-\gamma'\|^2_2|H_t]
    \end{align*}
 where the last two lines are because of the boundedness assumption. Thus, we can assign $C_r=C_0 \bar{\gamma}\bar{X}.$
\end{itemize}

Therefore, all assumptions in Theorem \ref{thm:reg_bound} are satisfied. We now need to compute  $\mathbb{E}[t_0+\sum_{t=t_0}^T\frac{1}{\lambda_{t-1}}].$ We first bound the value of $\sum_{\tau=2}^{t}(a_{\tau}-\bar{a}_{\tau-1})^2$, where $\bar{a}_t=\frac{\sum_{\tau=1}^t a_{\tau}}{t}$ is the mean of actions/prices. This will later be shown to serve as a lower bound for $\lambda_t$ with high probability.

 Since $a_{\tau}=\texttt{Alg}^*(H_{\tau})+\Delta_{\tau}$, we have 
\begin{align*}
    \sum_{\tau=2}^{t}(a_{\tau}-\bar{a}_{\tau-1})^2&=\sum_{\tau=2}^{t}(\texttt{Alg}^*(H_{\tau})+\Delta_{\tau}-\bar{a}_{\tau-1})^2\\
&=\sum_{\tau=2}^t\Delta_{\tau}^2+2\Delta_{\tau}(\texttt{Alg}^*(H_{\tau})-\bar{a}_{\tau-1})+(\texttt{Alg}^*(H_{\tau})-\bar{a}_{\tau-1})^2
\end{align*}

By assumptions, the sequence $\{\Delta_{\tau}(\texttt{Alg}^*(H_{\tau})-\bar{a}_{\tau-1})\}_{\tau=2}^T$ is a martingale difference sequence with respect to $\mathcal{F}_{\tau-1}=(\texttt{Alg}^*(H_{1}),\Delta_1,\ldots,\texttt{Alg}^*(H_{\tau-1}),\Delta_{\tau-1},\texttt{Alg}^*(H_{\tau}))$. Thus, by Theorem 2.19 in \cite{wainwright2019high} and the union bound, we have that, with probability $1-1/T$, for all $t=2,\ldots,T$,
$$\sum_{\tau=2}^t \Delta_{\tau}(\texttt{Alg}^*(H_{\tau})-\bar{a}_{\tau-1})\geq -\frac{4(\bar{a}+C)\log T\sqrt{t}}{T^{1/4}}.$$
Thus, with probability $1-1/T$, we have for all $t=2,\ldots,T$, 
\begin{align*}
    \sum_{\tau=2}^{t}(a_{\tau}-\bar{a}_{\tau-1})^2&\geq \sum_{\tau=2}^t\Delta_{\tau}^2+2\Delta_{\tau}(\texttt{Alg}^*(H_{\tau})-\bar{a}_{\tau-1})\\
    &\geq \frac{t}{2C^2 \sqrt{T}}-\frac{8(\bar{a}+C)\log T\sqrt{t}}{T^{1/4}}.
\end{align*}
And when $t\geq 512(\bar{a}+C)^2C^4\log^2 T \sqrt{T}$, we have  $\sum_{\tau=2}^{t}(a_{\tau}-\bar{a}_{\tau-1})^2\geq \frac{t}{2C^2 \sqrt{T}}-\frac{8(\bar{a}+C)\log T\sqrt{t}}{T^{1/4}}\geq \frac{t}{4C^2 \sqrt{T}}$. Thus, with probability $1-1/T$, we have $t_0\leq 512(\bar{a}+C)^2C^4\log^2 T \sqrt{T}$.

Further, by Lemma 1 of \cite{ban2021personalized} and the analysis in Example 1 of \cite{keskin2014dynamic}, for each $t$, there exist constants $C_{1}, C_{2}>0$, related to the boundedness of the parameters, such that with probability at least $1-d\exp{\left(-C_{1} \sum_{\tau=2}^{t}(a_{\tau}-\bar{a}_{\tau-1})^2\right)}$,

$$\lambda_{t} \geq C_{2} \sum_{\tau=2}^{t}(a_{\tau}-\bar{a}_{\tau-1})^2.$$

Thus, by the union bound, for all $t\geq \max\{512(\bar{a}+C)^2C^4\log^2 T \sqrt{T},\frac{4C^2\log(dT^2)\sqrt{T}}{C_1}\},$ we have $\lambda_t=\Omega(t/\sqrt{T})$ for all $t=1,\ldots,T$ with probability at least $1-\frac{2}{T}$. With $\Delta_{\text{Exploit}}=O(1/\sqrt{T})$, we can complete the result by Theorem \ref{thm:reg_bound}.
\end{proof}

\subsubsection{Proof of Theorem \ref{thm:reg_bound_finite}}
\begin{proof}
    For both of the two regret bounds, we first decompose the regret:
\begin{align*}
        \text{Regret}(\texttt{TF}_{\hat{\theta}};\gamma)&=\sum_{t=1}^T\mathbb{E}[r(X_t,\texttt{Alg}^*(H_t))-r(X_t,\texttt{TF}_{\hat{\theta}}(H_t))]+\sum_{t=1}^T\mathbb{E}[r(X_t,a^*_t)-r(X_t,\texttt{Alg}^*(H_t))]\\
        &\leq \Delta_{\text{Exploit}}T+C_r\mathbb{E}\left[\sum_{t=1}^T\sum_{\gamma_i\neq \gamma}\mathcal{P}(\gamma_i|H_t)\|\gamma_i-\gamma\|_2^2\right],
\end{align*}
where the last line is by Assumption \ref{assmp:reward}.

Now we apply the union bound and Lemma \ref{lemma:concen_inq_cts} to  $\gamma_i$ for $i=1,\ldots,n$ and for $t=1,\ldots,T$: with probability $1-1/T$, for all $i=1,\ldots,n$, and  $t=1,\ldots,T$
\begin{align*}
    \mathcal{P}(\gamma_i|H_t)&\leq \mathcal{P}(\gamma|H_t) \cdot \exp\left(-\frac{1}{2}\sum_{\tau=1}^{t-1}\|\gamma_i-\gamma\|_{\Delta(X_{\tau},a_{\tau)}}^2+4C_{\sigma^2}\log(nT^2)\right).
\end{align*}

Since  $\mathcal{P}(\gamma_i|H_t)+\mathcal{P}(\gamma|H_t)\leq 1$, we have with probability $1-1/T$, for all $i=1,\ldots,n$, and  $t=1,\ldots,T$
$$ \mathcal{P}(\gamma_i|H_t)\leq \frac{1}{1+\exp\left(\frac{1}{2}\sum_{\tau=1}^{t-1}\|\gamma_i-\gamma\|_{\Delta(X_{\tau},a_{\tau)}}^2-4C_{\sigma^2}\log(nT^2)\right)}.$$

For the first regret bound, recall $t_i=\min\{t=2,\ldots,T|\sum_{\tau=1}^{t-1}\|\gamma_i-\gamma\|_{\Delta(X_{\tau},a_{\tau)}}^2\geq 16C_{\sigma^2}\log(nT^2)\}$, and thus for $i=1,\ldots,n$,
\begin{align*}
\mathbb{E}\left[\sum_{t=1}^T\mathcal{P}(\gamma_i|H_t)\|\gamma_i-\gamma\|_2^2\right]&\leq \bar{r}\sum_{\gamma_i\neq \gamma}\mathbb{E}[t_i-1]+\sum_{\gamma_i\neq \gamma}\mathbb{E}\left[\sum_{t=t_i}^T\mathcal{P}(\gamma_i|H_t)\|\gamma_i-\gamma\|_2^2\right]\\
&\leq 4\bar{\gamma}^2+\bar{r}\sum_{\gamma_i\neq \gamma}\mathbb{E}[t_i-1]+  \sum_{\gamma_i\neq \gamma}\mathbb{E}\left[\sum_{t=t_i}^T\frac{\|\gamma_i-\gamma\|_2^2}{1+\exp\left(\frac{1}{4}\sum_{\tau=1}^{t-1}\|\gamma_i-\gamma\|_{\Delta(X_{\tau},a_{\tau)}}^2\right)}\right]\\
&<4\bar{\gamma}^2+\bar{r}\sum_{\gamma_i\neq \gamma}\mathbb{E}[t_i-1]+  \sum_{\gamma_i\neq \gamma}\mathbb{E}\left[\sum_{t=t_i}^T\frac{4\|\gamma_i-\gamma\|_2^2}{\sum_{\tau=1}^{t-1}\|\gamma_i-\gamma\|_{\Delta(X_{\tau},a_{\tau})}^2}\right]
\end{align*}
where the second inequality is because  $\|\gamma-\gamma_i\|_2^2$ is bounded by $4\bar{\gamma}^2$, and the last inequality is because $x< \exp(x)+1$ for $x>0$. Thus, we can conclude the result.

Now for the second regret bound, we think about two situations for each $i=1,\ldots,n$ and $t=t_0,\ldots,T$:
\begin{itemize}
    \item If $\lambda_{t-1}\|\gamma_i-\gamma\|_2^2< 16C_{\sigma^2}\log(nT^2)$, then since $\mathcal{P}(\gamma_i|H_t)< 1$, we have
    $$\mathcal{P}(\gamma_i|H_t)\|\gamma_i-\gamma\|_2^2< \frac{ 16C_{\sigma^2}\log(nT^2)}{\lambda_{t-1}}.$$
    \item  If $\lambda_{t-1}\|\gamma_i-\gamma\|_2^2\geq 16C_{\sigma^2}\log(nT^2)$. Since $\exp(x)+1> x$ for $x>0$, we have 
     $$\mathcal{P}(\gamma_i|H_t)\|\gamma_i-\gamma\|_2^2< \frac{1}{1+\exp\left(\frac{1}{4}\lambda_{t-1}\|\gamma_i-\gamma\|_2^2\right)} \|\gamma_i-\gamma\|_2^2<\frac{4}{\lambda_{t-1}}.$$
\end{itemize}

Thus, we can conclude that
\begin{align*}
\mathbb{E}\left[\sum_{t=1}^T\sum_{\gamma_i\neq \gamma}\mathcal{P}(\gamma_i|H_t)\|\gamma_i-\gamma\|_2^2\right]&< 4\bar{\gamma}^2+\bar{r}\mathbb{E}[t_0-1]+n\mathbb{E}\left[\sum_{t=t_0}^T\frac{16C_{\sigma^2}\log(nT^2)+4}{\lambda_{t-1}}\right]
\end{align*}
and finish the second regret bound.

\end{proof}

\subsubsection{Proof of Example \ref{example:finite_DP}}
\begin{proof}
We first assume there exists constants $\underline{a},\bar{a}>0$ such that  $\mathcal{A}=[\underline{a},\bar{a}]$ and the potential optimal actions $\frac{1}{2},\frac{1}{4}\in \mathcal{A}$, and $\mathcal{P}_{\gamma}$ is a uniform distribution in $\Gamma$. Then by a similar analysis as in the proof of Example \ref{example:cts_DP},  all the assumptions in Theorem \ref{thm:reg_bound_finite} can be verified, where $\Delta_{\text{Exploit}}=0$, $C_{\sigma^2}=\sqrt{2}$, $\bar{\gamma}=\sqrt{5}$, $\bar{r}=\frac{1}{4}$, and $\Delta(X,a)=\begin{bmatrix}
1 & a\\
a & a^2 
\end{bmatrix}$.

Without loss of generality, we assume $\gamma=\gamma_1=(1,-1)$. Then, we have 
$$\|\gamma_2-\gamma\|_{\Delta(X,a)}^2= a^2\geq \underline{a}^2 $$
for all $a$. Thus, we have $t_2\leq \frac{16\sqrt{2}\log(2T^2)}{\underline{a}^2}$, and for all $t=2,\ldots, T$,
$$\frac{4\|\gamma_2-\gamma\|_2^2}{\sum_{\tau=1}^{t-1}\|\gamma_2-\gamma\|_{\Delta(X_{\tau},a_{\tau})}^2}\leq \frac{4}{(t-1) \underline{a}^2}.$$
Therefore, by \eqref{eqn:reg_dependent}, we have 
\begin{align*}
\text{Regret}(\texttt{TF}_{\hat{\theta}};\gamma)&<4C_r\bar{\gamma}^2+\Delta_{\text{Exploit}}T+\bar{r}C_r\sum_{\gamma_i\neq \gamma}\mathbb{E}[t_i-1]+  C_r\sum_{\gamma_i\neq \gamma}\mathbb{E}\left[\sum_{t=t_i}^T\frac{4\|\gamma_i-\gamma\|_2^2}{\sum_{\tau=1}^{t-1}\|\gamma_i-\gamma\|_{\Delta(X_{\tau},a_{\tau})}^2}\right]\\
&<4\bar{\gamma}^2+\frac{16\sqrt{2}C_r\bar{r}\log(2T^2)}{\underline{a}^2}+  \sum_{t=2}^T\frac{4C_r}{(t-1) \underline{a}^2}\\
&=O(\log T)
\end{align*}
which completes the proof.
\end{proof}

\subsubsection{Proofs of Lemmas}
\paragraph{Proof for Lemma \ref{lemma:concen_inq_cts}}
\begin{proof}
 By the Bernstein-type concentration bound for a martingale difference sequence (Theorem 2.19 in \citep{wainwright2019high}), under Assumption \ref{assmp:llh}, we have for any $\gamma'\in\Gamma$, $t>0$ and $1>\delta>0$ with probability $1-\delta$,
    $$\sum_{\tau=1}^t \log  \left(\frac{\mathbb{P}_{\gamma}(O_{\tau}\vert X_{\tau},a_{\tau})}{\mathbb{P}_{\gamma'}(O_{\tau}\vert X_{\tau},a_{\tau})}\right)-\mathbb{E}\left[\log  \left(\frac{\mathbb{P}_{\gamma}(O_{\tau}\vert X_{\tau},a_{\tau})}{\mathbb{P}_{\gamma'}(O_{\tau}\vert X_{\tau},a_{\tau})}\right)\right]
\geq -\sqrt{2C_{\sigma^2}\sum_{\tau=1}^t\|\gamma-\gamma'\|^2_{\Delta(X_{\tau},a_{\tau})}\log (1/\delta)},$$
where 
$$\sum_{\tau=1}^t\mathbb{E}\left[\log  \left(\frac{\mathbb{P}_{\gamma}(O_{\tau}\vert X_{\tau},a_{\tau})}{\mathbb{P}_{\gamma'}(O_{\tau}\vert X_{\tau},a_{\tau})}\right)\right]=\sum_{\tau=1}^t\text{KL}\left(\mathbb{P}_{\gamma}(\cdot\big \vert X_{\tau},a_{\tau})\| \mathbb{P}_{\gamma'}(\cdot\big \vert X_{\tau},a_{\tau}) \right)\geq \sum_{\tau=1}^t\|\gamma-\gamma'\|^2_{\Delta(X_{\tau},a_{\tau})}.$$
We think about two situations:
\begin{itemize}
    \item If $\sqrt{\sum_{\tau=1}^t\|\gamma-\gamma'\|^2_{\Delta(X_{\tau},a_{\tau})}}\geq 2\sqrt{2C_{\sigma^2}\log(1/\delta)}$, then 
  \begin{align*}
  \sum_{\tau=1}^t \log  \left(\frac{\mathbb{P}_{\gamma}(O_{\tau}\vert X_{\tau},a_{\tau})}{\mathbb{P}_{\gamma'}(O_{\tau}\vert X_{\tau},a_{\tau})}\right)&\geq \sum_{\tau=1}^t\|\gamma-\gamma'\|^2_{\Delta(X_{\tau},a_{\tau})}-\frac{1}{2} \sum_{\tau=1}^t\|\gamma-\gamma'\|^2_{\Delta(X_{\tau},a_{\tau})}\\
  &=\frac{1}{2}\sum_{\tau=1}^t\|\gamma-\gamma'\|^2_{\Delta(X_{\tau},a_{\tau})}\\
  &>\frac{1}{2}\sum_{\tau=1}^t\|\gamma-\gamma'\|^2_{\Delta(X_{\tau},a_{\tau})}-4C_{\sigma^2}\log(1/\delta).
  \end{align*}
  \item If $\sqrt{\sum_{\tau=1}^t\|\gamma-\gamma'\|^2_{\Delta(X_{\tau},a_{\tau})}}< 2\sqrt{2C_{\sigma^2}\log(1/\delta)}$, then
  \begin{align*}
 \sum_{\tau=1}^t \log  \left(\frac{\mathbb{P}_{\gamma}(O_{\tau}\vert X_{\tau},a_{\tau})}{\mathbb{P}_{\gamma'}(O_{\tau}\vert X_{\tau},a_{\tau})}\right)&>
 \sum_{\tau=1}^t\|\gamma-\gamma'\|^2_{\Delta(X_{\tau},a_{\tau})}-4C_{\sigma^2}\log(1/\delta)\\
  &\geq \frac{1}{2}\sum_{\tau=1}^t\|\gamma-\gamma'\|^2_{\Delta(X_{\tau},a_{\tau})}-4C_{\sigma^2}\log(1/\delta).
  \end{align*}
\end{itemize}
Thus we get the first inequality. Since $\sum_{\tau=1}^t\|\gamma-\gamma'\|^2_{\Delta(X_{\tau},a_{\tau})}=(\gamma-\gamma')^\top\left(\sum_{\tau=1}^t\Delta(X_{\tau},a_{\tau})\right)(\gamma-\gamma')\geq \lambda_{t}\|\gamma-\gamma'\|_2^2$, we can get the second inequality.
\end{proof}

\paragraph{Proof for Lemma \ref{lemma:ratio_diff_bound}}
\begin{proof}
    \begin{align*}
        \Bigg \vert \log  \left(\frac{\mathbb{P}_{\gamma_1+\gamma'}(O \big \vert X,a)}{\mathbb{P}_{\gamma_1}(O\big \vert X,a)}\right)-\log  \left(\frac{\mathbb{P}_{\gamma_2+\gamma'}(O \big \vert X,a)}{\mathbb{P}_{\gamma_2}(O\big \vert X,a)}\right)\Bigg \vert &\leq \Bigg \vert\nabla_{\gamma}\log(\mathbb{P}_{\gamma_1}(O\big \vert X,a))^T\gamma'-\nabla_{\gamma}\log(\mathbb{P}_{\gamma_2+\gamma'}(O\big \vert X,a))^T\gamma'\Bigg \vert\\
&=\Bigg \vert\gamma'^{\top}\nabla^2_{\gamma}\log(\mathbb{P}_{\tilde{\gamma}}(O\big \vert X,a))(\gamma_1-\gamma_2-\gamma')\Bigg \vert\\
&\leq C_{\Gamma}\|\gamma'\|_2(\|\gamma'\|_2+\|\gamma_1-\gamma_2\|_2)
    \end{align*}
where the first line is by the concavity under Assumption \ref{assmp:llh}, the second line is by the mean value theorem for vector value functions (see Section 12.11 of \citep{apostol1974mathematical}) with $\tilde{\gamma}$ in the line between $\gamma_1$ and $\gamma_2+\gamma'$, and the last line is by the definition of $C_{\Gamma}$ in Assumption \ref{assmp:llh}.
\end{proof}

\paragraph{Proof of Lemma \ref{lemma:cts_prob_bound}}

\begin{proof}
    With little abusing of notations, we use $p_t(\gamma')$ as the density function of the posterior distribution of possible (testing) environment $\gamma'$ given $H_t$. Recall $H_t$ is generated through \eqref{eqn:test_dynamics} with environment $\gamma$.
    $$\mathcal{P}_t(\|\gamma-\gamma'\|_2^2\geq \epsilon)=\int_{\|\gamma-\gamma'\|_2^2\geq \epsilon}p_t(\gamma')d\gamma'.$$
Let $\tilde{\Gamma}=\{\gamma_1,\ldots,\gamma_k\}$ as the largest packing set of $\Gamma$ including $\gamma$  with $\|\cdot\|_2$ norm and a radius $\epsilon'$ (i.e., all other packing sets including $\gamma$ have a size no larger than $k$). The choice of $\epsilon'$ will be specified later.  Without loss of generality, we suppose $\gamma_1=\gamma$.

Then $\tilde{\Gamma}$ is also a covering set of $\Gamma$ (with the same norm and radius). The proof follows a standard analysis based on definitions: if $\exists \gamma'\in \Gamma$ such that $\|\gamma'-\gamma_i\|_2> \epsilon'$ for all $i=1,\ldots,k$, then we can add $\gamma'$ into $\tilde{\Gamma}$ to construct a new packing set containing $\gamma$, which contradicts $\tilde{\Gamma}$ is the largest.

Thus by the definition of the covering set, for all $\gamma'\in \Gamma$, there exists $\gamma_i\in\tilde{\Gamma}$ such that $\gamma' \in B(\gamma_i,\epsilon')$, where $B(\gamma_i,\epsilon')=\{\gamma'\in\Gamma:\|\gamma'-\gamma_i\|_2\leq \epsilon'\}$, i.e., the ball with center $\gamma_i$ and radius $\epsilon'$.

Since the interested set $\{\gamma'\in\Gamma: \|\gamma-\gamma'\|_2^2\geq \epsilon\}\subset \Gamma$, we can find $\Tilde{\Gamma}'=\{\gamma_1',\ldots,\gamma_{k'}'\}\subset\tilde{\Gamma}$ which is the smallest subset of $\tilde{\Gamma}$  such that $\{\gamma'\in\Gamma: \|\gamma-\gamma'\|_2^2\geq \epsilon\}\subseteq \bigcup_{i=1}^{k'}B(\gamma_i',\epsilon')$. 

Now we apply the union bound and Lemma \ref{lemma:concen_inq_cts} to the density ratios of $\gamma'_{i}, i=1,\ldots,k'$. Recall under Assumption \ref{assmp:bounded}, the prior is a  uniform distribution. Then with probability $1-\delta$,
$$\frac{p_{t}(\gamma'_i)}{p_t(\gamma)}=\prod_{\tau=1}^{t-1} \left(\frac{\mathbb{P}_{\gamma'_i}(O_{\tau}\vert X_{\tau},a_{\tau})}{\mathbb{P}_{\gamma}(O_{\tau}\vert X_{\tau},a_{\tau})}\right)\leq \exp\left(-\frac{1}{2}\lambda_{t-1}\|\gamma'_i-\gamma\|_2^2+4C_{\sigma^2}\log(k'/\delta)\right)\quad \forall i=1,\ldots,k'.$$

Then we can bound the interested probability by applying the above result: with probability $1-\delta$,
\begin{align*}
 \int_{\|\gamma-\gamma'\|_2^2\geq \epsilon}p_t(\gamma')d\gamma'&\leq \sum_{i=1}^{k'}\int_{\|\gamma'-\gamma'_i\|_2\leq \epsilon'} p_t(\gamma') d\gamma' \\
&=\sum_{i=1}^{k'}\int_{\|\gamma'\|_2\leq \epsilon'} p_t(\gamma'_i+\gamma') d\gamma'\\    
&=\sum_{i=1}^{k'}\frac{p_t(\gamma_i')}{p_t(\gamma)}\cdot\int_{\|\gamma'\|_2\leq \epsilon'} p_t(\gamma+\gamma')\cdot \frac{p_t(\gamma_i'+\gamma)}{p_t(\gamma_i')} \cdot \frac{p_t(\gamma)}{p_t(\gamma+\gamma')}d\gamma' \\
    &\leq \sum_{i=1}^{k'}\exp(C_{\Gamma}\epsilon'(t-1)(\|\gamma-\gamma_i'\|_2+\epsilon'))\frac{p_t(\gamma_i')}{p_t(\gamma)}\int_{\|\gamma'\|_2\leq \epsilon' } p_t(\gamma+\gamma')d\gamma' \\
    &\leq \sum_{i=1}^{k'}\exp\left(4C_{\sigma^2}\log(k'/\delta)+C_{\Gamma}\epsilon'(t-1)(\|\gamma-\gamma_i'\|_2+\epsilon')-\frac{1}{2}\lambda_{t-1}\|\gamma-\gamma_i'\|_2^2\right)\int_{\|\gamma-\gamma'\|_2\leq \epsilon'} p_t(\gamma')d\gamma'
\end{align*}
where the second last step is by  Lemma \ref{lemma:ratio_diff_bound} with $\|\gamma'\|_2\leq \epsilon'$.

Now by choosing $\epsilon'=\frac{\lambda_{t-1}\min\{\sqrt{\epsilon},1\}}{8C_{\Gamma}(t-1)+2\lambda_{t-1}}$, we have 
\begin{itemize}
    \item $\epsilon'< \frac{\lambda_{t-1}\min\{\sqrt{\epsilon},1\}}{2\lambda_{t-1}}=\frac{\min\{\sqrt{\epsilon},1\}}{2}$
    \item $2C_{\Gamma}\epsilon'(t-1)< \frac{\lambda_{t-1}}{4}\|\gamma-\gamma'_i\|_2$ for all $i=1,\ldots,k'$. To prove this, we first note that for all $ i=1,\ldots,k'$
    $$\|\gamma-\gamma'_i\|_2\geq \sqrt{\epsilon}-\epsilon'\geq \epsilon'.$$
    Otherwise, if there exists $\gamma'_i$ such that $\|\gamma-\gamma'_i\|_2< \sqrt{\epsilon}-\epsilon'$, then $B(\gamma'_i,\epsilon')\bigcap \{\gamma'\in\Gamma: \|\gamma-\gamma'\|_2^2\geq \epsilon\}=\emptyset$ and we can drop $\gamma'_i$ in $\tilde{\Gamma}'$ but still keep the correspondingly constructed union of balls covering $\{\gamma'\in\Gamma: \|\gamma-\gamma'\|_2^2\geq \epsilon\}$, which contradicts the fact that $\tilde{\Gamma}'$ is the smallest subset. Thus, we have
    \begin{align*}
    \frac{\lambda_{t-1}}{4}\|\gamma-\gamma'_i\|_2-
        2C_{\Gamma}\epsilon'(t-1)&\geq  \frac{\lambda_{t-1}}{4}(\sqrt{\epsilon}-\epsilon')- 2C_{\Gamma}\epsilon'(t-1)\\
        &> \frac{1}{4}\left(\lambda_{t-1}\sqrt{\epsilon}-(8C_{\Gamma}(t-1)+2\lambda_{t-1})\epsilon'\right)\\
        &=\frac{1}{4}\left(\lambda_{t-1}\sqrt{\epsilon}-\lambda_{t-1}\min\{\sqrt{\epsilon},1\}\right)\\
        &\geq 0.
    \end{align*}
\end{itemize}

Thus, with probability $1-\delta$
\begin{align*}
 \int_{\|\gamma-\gamma'\|_2^2\geq \epsilon}p_t(\gamma')d\gamma'
    &\leq \sum_{i=1}^{k'}\exp\left(4C_{\sigma^2}\log(k'/\delta)+2C_{\Gamma}\epsilon'(t-1)\|\gamma-\gamma_i'\|_2-\frac{1}{2}\lambda_{t-1}\|\gamma-\gamma_i'\|_2^2\right)\int_{\|\gamma-\gamma'\|_2\leq \epsilon'} p_t(\gamma')d\gamma'\\
    &\leq \sum_{i=1}^{k'}\exp\left(4C_{\sigma^2}\log(k'/\delta)-\frac{1}{4}\lambda_{t-1}\|\gamma-\gamma_i'\|_2^2\right)\int_{\|\gamma-\gamma'\|_2\leq \epsilon'} p_t(\gamma')d\gamma'\\
    &< k\exp\left(4C_{\sigma^2}\log(k/\delta)-\frac{1}{16}\lambda_{t-1}\epsilon \right)\int_{\|\gamma-\gamma'\|^2_2< \epsilon} p_t(\gamma')d\gamma'
\end{align*}
where the last inequality is because $k'\leq k$ and $\|\gamma-\gamma'_i\|_2\geq \sqrt{\epsilon}-\epsilon'$ with $\epsilon'\leq \frac{\sqrt{\epsilon}}{2}$.

In the following, we bound $k$ which is the size of the constructed packing set $\Tilde{\Gamma}$. We use $N_P(\tilde{\epsilon})$ and $N_C(\tilde{\epsilon})$ to denote the packing number and covering number of $\Gamma$ with $\|\cdot\|_2$ norm and radius $\tilde{\epsilon}$ respectively. 
\begin{align*}
    k&\leq N_P(\epsilon')\\
    &\leq N_C(\frac{\epsilon'}{2})\\
    &= N_C\left(\frac{\lambda_{t-1}\min\{\sqrt{\epsilon},1\}}{16C_{\Gamma}(t-1)+4\lambda_{t-1}}
    \right)\\
    &\leq N_C\left(\frac{\lambda_{t-1}\min\{\sqrt{\epsilon_t},1\}}{16C_{\Gamma}(t-1)+4\lambda_{t-1}}
    \right)\\
    &\leq \max\left\{1,\left(\frac{32\bar{\gamma}\sqrt{d}(C_{\Gamma}(t-1)+\lambda_{t-1})}{\lambda_{t-1}\min\{\sqrt{\epsilon_t},1\}}\right)^d\right\}
\end{align*}
where the first line is from the definition of the packing number, the second line comes from the relationship between covering and packing numbers (Lemma 4.2.8 in \citep{vershynin2018high}), the third line is from the definition of $\epsilon'$, the fourth line is by $\epsilon_t\leq \epsilon$, and the last line is from Example 27.1 in \citet{shalev2014understanding} for upper bounds of covering numbers in Euclidean (sub)space.

Recall the notations 
$$K_t=\max\left\{1,\left(\frac{32\bar{\gamma}\sqrt{d}(C_{\Gamma}(t-1)+\lambda_{t-1})}{\lambda_{t-1}\min\{\sqrt{\epsilon_t},1\}}\right)^d\right\},$$
$$C_t(\epsilon,\delta)=\exp\left(-4C_{\sigma^2}\log(K_t/\delta)+\frac{1}{16}\lambda_{t-1}\epsilon \right).$$
Then we have with probability $1-\delta$,
$$\int_{\|\gamma-\gamma'\|^2_2\geq \epsilon} p_t(\gamma')d\gamma'\leq \frac{K_t}{C_t(\epsilon,\delta)} \int_{\|\gamma-\gamma'\|^2_2< \epsilon} p_t(\gamma')d\gamma'.$$
Since 
$$\int_{\|\gamma-\gamma'\|^2_2\geq \epsilon} p_t(\gamma')d\gamma'+\int_{\|\gamma-\gamma'\|^2_2< \epsilon} p_t(\gamma')d\gamma'=1,$$
by rearranging the term we can conclude the result. 
\end{proof}
\paragraph{Proof of Corollary \ref{corr:cts_prob_bound}}

\begin{proof}
For any given $\epsilon\geq \epsilon_t$ and $1>\delta>0$, we denote the event 
$\mathcal{E}=\left\{\mathcal{P}_t(\|\gamma-\gamma'\|_2^2\geq \epsilon)\leq \frac{K_t}{K_t+C_t(\epsilon,\delta)}\right\}$, which holds with probability $1-\delta$ by Lemma \ref{lemma:cts_prob_bound}, and its complementary event as $\bar{\mathcal{E}}$. Then 
\begin{align*}
    \mathbb{E}\left[\mathbb{E}_t\left[\|\gamma-\gamma'\|_2^2\mathbbm{1}_{\{\|\gamma-\gamma'\|_2^2\geq \epsilon_t\}}\right]\right]&=\mathbb{E}\left[\int_{\epsilon\geq \epsilon_t}\mathcal{P}_t(\|\gamma-\gamma'\|_2^2\geq\epsilon)d\epsilon\right]\\
&=\mathbb{E}\left[\int_{\epsilon\geq \epsilon_t}\mathcal{P}_t(\|\gamma-\gamma'\|_2^2\geq\epsilon)(\mathbbm{1}_{\mathcal{E}}+\mathbbm{1}_{\bar{\mathcal{E}}})]d\epsilon\right]\\
    &\leq \mathbb{E}\left[\int_{\epsilon\geq \epsilon_t}\frac{K_t}{K_t+C_t(\epsilon,\delta)}d\epsilon\right]+\int_{0\leq \epsilon\leq 4\bar{\gamma}^2}\delta d\epsilon\\
    &=  \mathbb{E}\left[\int_{\epsilon\geq \epsilon_t}\frac{K_t}{K_t+C_t(\epsilon,\delta)}d\epsilon\right]+4\bar{\gamma}^2\delta,
\end{align*}
where the third line is because  $\mathcal{P}_t(\|\gamma-\gamma'\|_2^2\geq\epsilon)\leq 1$ and $\mathcal{P}_t(\|\gamma-\gamma'\|_2^2\geq\epsilon)=0$ for $\epsilon>4\bar{\gamma}^2$, with the probability of $\bar{\mathcal{E}}$ being $\delta$. The expectation in the last line is with respect to the potential randomness in $\lambda_{t-1}$.

We now bound the first term. Note that for any $C_1,C_2>0$, we have 
$$\int_{\epsilon\geq \epsilon_t}\frac{C_1}{C_1+\exp(C_2\epsilon)}d\epsilon=\frac{\log(C_1+\exp(C_2\epsilon_t))}{C_2}-\epsilon_t< \frac{\log(C_1+\exp(C_2\epsilon_t))}{C_2}.$$
By plugging $C_t(\epsilon,\delta)$, we have 
$\mathbb{E}\left[\int_{\epsilon\geq \epsilon_t}\frac{K_t}{K_t+C_t(\epsilon,\delta)}d\epsilon\right]\leq \mathbb{E}\left[\frac{16}{\lambda_{t-1}}\log\left(K_t(\frac{K_t}{\delta})^{4C_{\sigma^2}}+\exp(\frac{\lambda_{t-1}\epsilon_t}{16})\right)\right].$

For $\epsilon_t=\frac{16d}{\lambda_{t-1}}\log\left(\frac{32\bar{\gamma}\sqrt{d}(C_{\Gamma}(t-1)+\lambda_{t-1})}{\lambda_{t-1}}\right)$, we have for all $t\geq t_0$
\begin{align*}
    \frac{\lambda_{t-1}\epsilon_t}{16}&= d\log\left(\frac{32\bar{\gamma}\sqrt{d}(C_{\Gamma}(t-1)+\lambda_{t-1})}{\lambda_{t-1}}\right)\\
    &\leq d\log\left(\frac{32\bar{\gamma}\sqrt{d}(C_{\Gamma}(t-1)+\lambda_{t-1})}{\lambda_{t-1}\min\{\sqrt{\epsilon_t},1\}}\right)\\
    &=\log K_t\\
&<\log\left(K_t(K_t/\delta)^{4C_{\sigma}^2}\right)
\end{align*}
where the last inequality is because $K_t\geq 1$ and $\delta<1$. Then we have 
$$\mathbb{E}\left[\int_{\epsilon\geq \epsilon_t}\frac{K_t}{K_t+C_t(\epsilon,\delta)d\epsilon}\right]\leq \mathbb{E}\left[\frac{16}{\lambda_{t-1}}\log\left(2K_t(K_t/\delta)^{4C_{\sigma^2}})\right)\right].$$

\end{proof}

\section{Appendix for Numerical Experiments}
\label{appx:numerical}

\subsection{More Numerical Experiments}
\subsubsection{Matchings of $\texttt{TF}_{\hat{\theta}}$ and  $\texttt{Alg}^*$}
\label{appx:bayes_match_exp}
\begin{figure}[ht!]
  \centering
    \begin{subfigure}[b]{0.31\textwidth}
    \centering
\includegraphics[width=\textwidth]{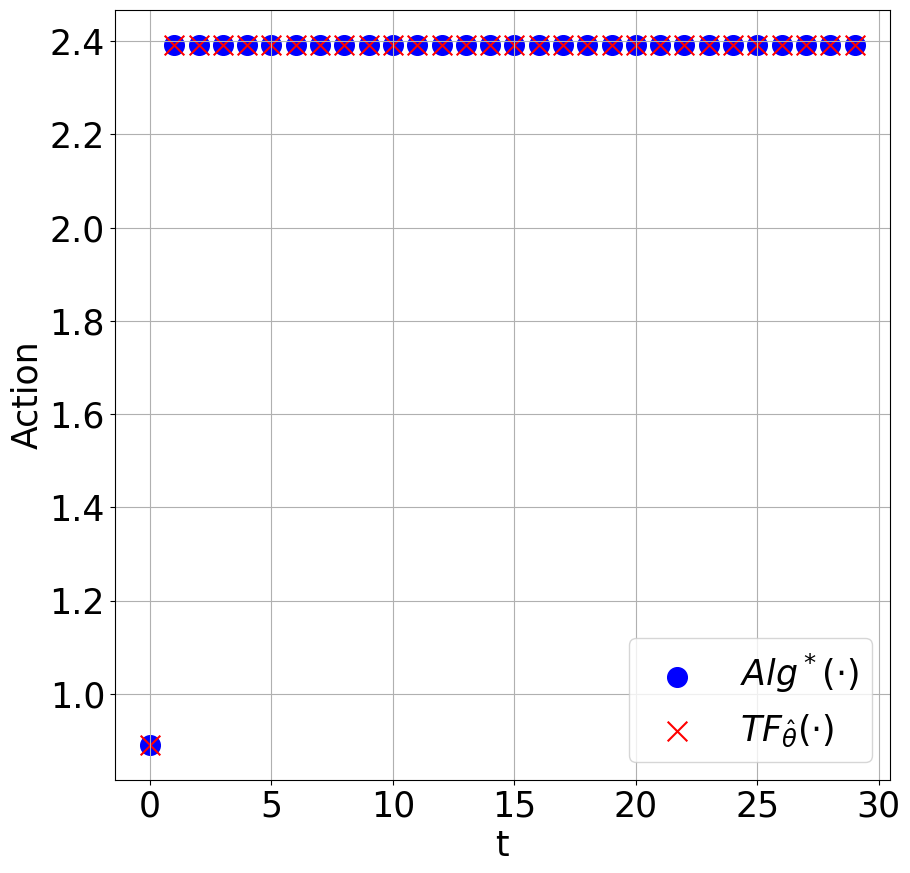}
    \caption{Multi-armed bandits, $4$ environments}
  \end{subfigure}
    \hfill
  \begin{subfigure}[b]{0.31\textwidth}
    \centering
\includegraphics[width=\textwidth]{figs/DP/_4env_4d_132_Act_match_2.png}
    \caption{Dynamic pricing,  $4$ environments}
  \end{subfigure}
    \hfill
  \begin{subfigure}[b]{0.31\textwidth}
    \centering
\includegraphics[width=\textwidth]{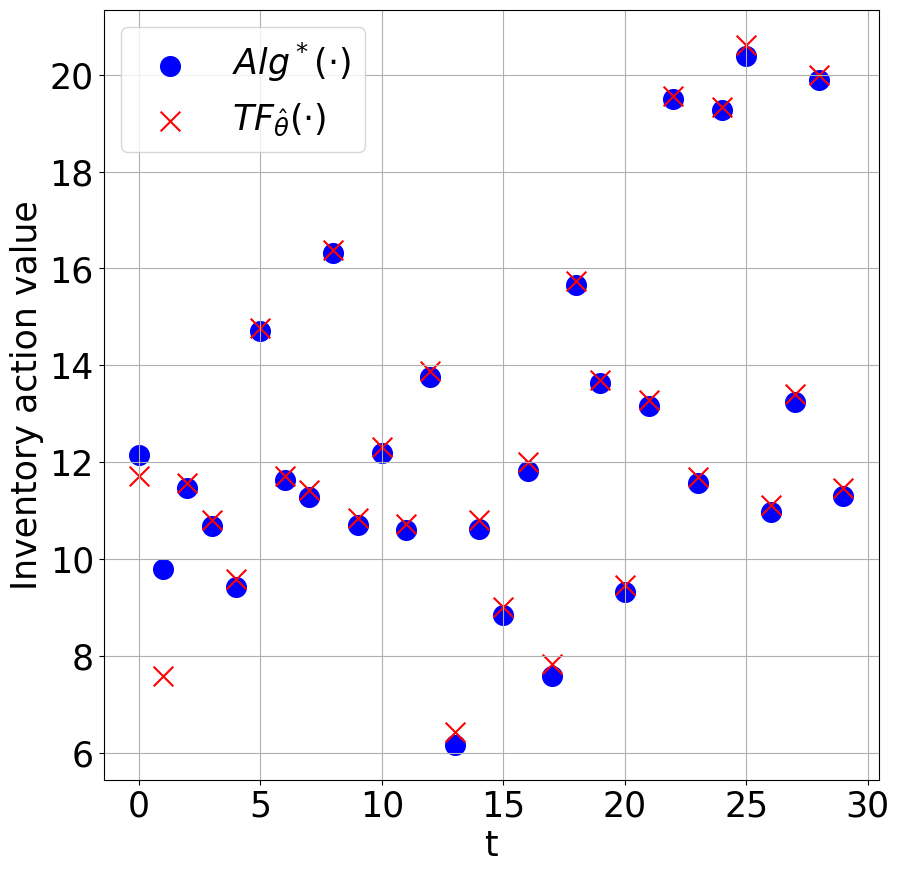}
    \caption{Newsvendor,  $4$ environments}
  \end{subfigure}
    \begin{subfigure}[b]{0.31\textwidth}
    \centering
\includegraphics[width=\textwidth]{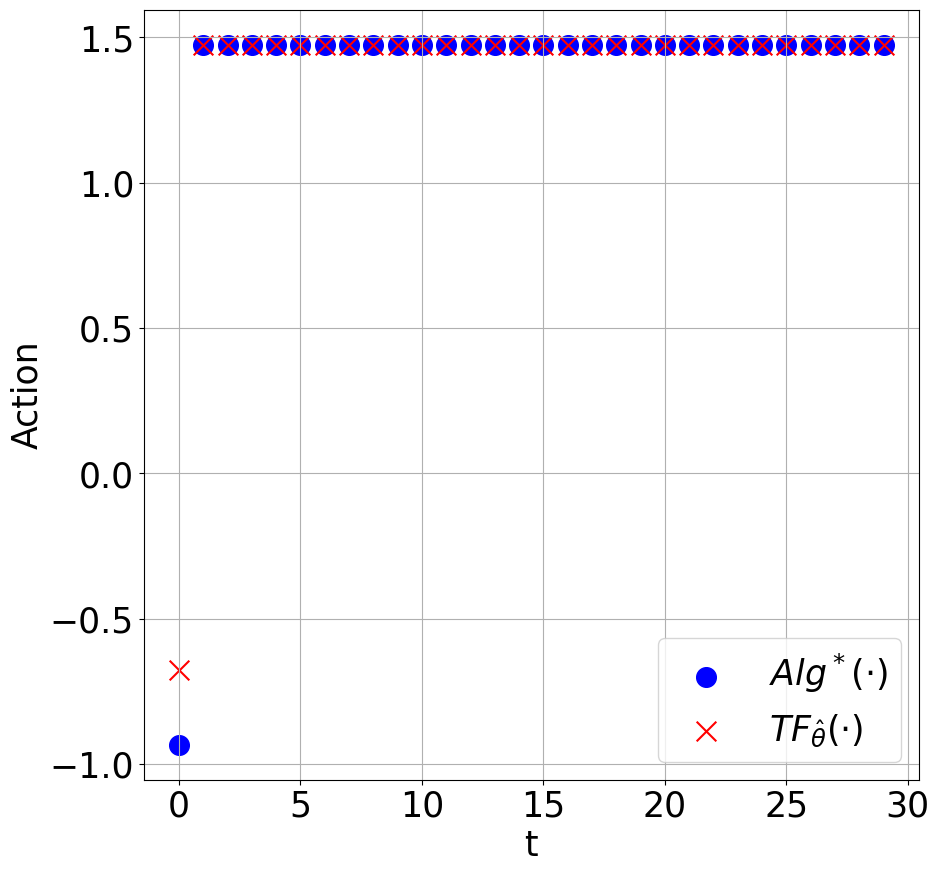}
    \caption{Multi-armed bandits,  $100$ environments}
  \end{subfigure}
    \hfill
  \begin{subfigure}[b]{0.31\textwidth}
    \centering
\includegraphics[width=\textwidth]{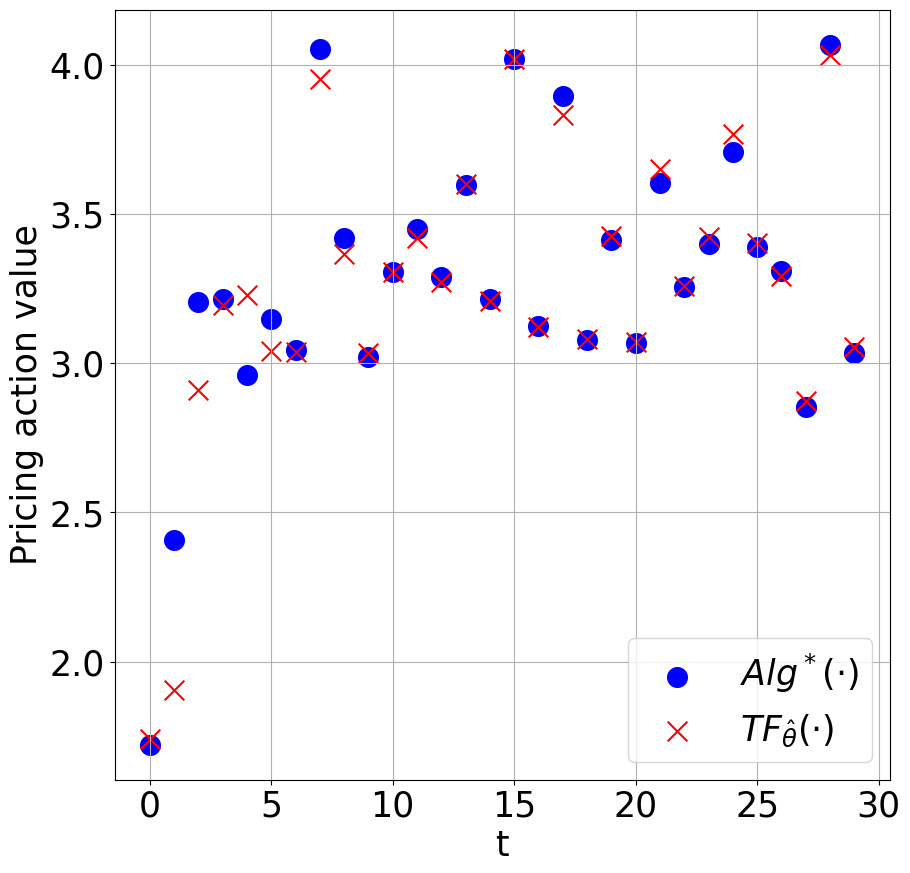}
    \caption{Dynamic pricing,  $100$ environments}
  \end{subfigure}
    \hfill
  \begin{subfigure}[b]{0.31\textwidth}
    \centering
\includegraphics[width=\textwidth]{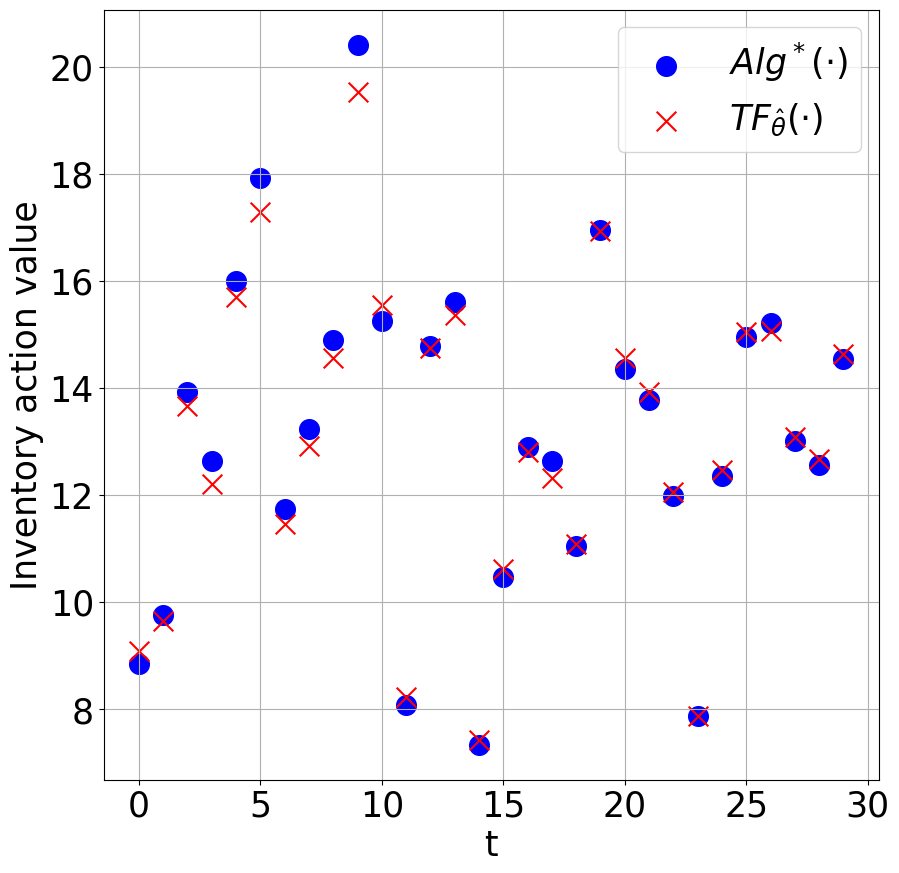}
    \caption{Newsvendor, $100$ environments}
  \end{subfigure}
 
  \caption{Examples to compare the actions from the transformer $\texttt{TF}_{\hat{\theta}}$ and the optimal decision function $\texttt{Alg}^*$.}
  \label{fig:appx_bayesmatch}
\end{figure}

\begin{figure}[ht!]
  \centering
  \begin{subfigure}[b]{0.24\textwidth}
    \centering
\includegraphics[width=\textwidth]{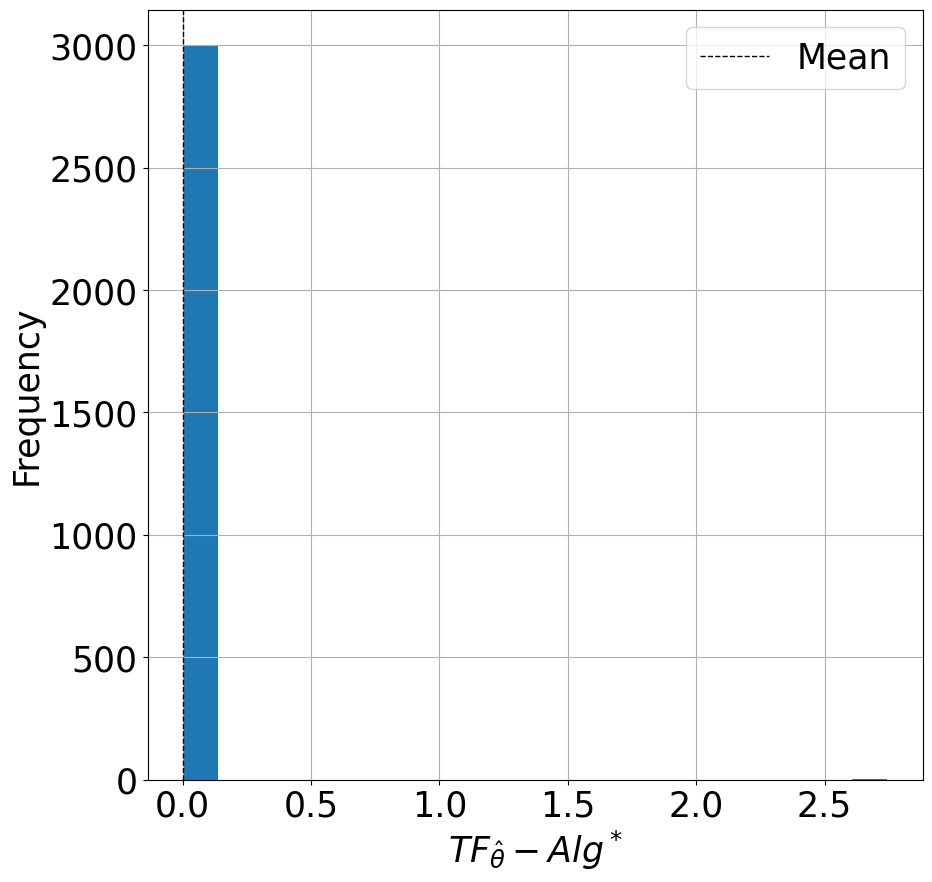}
    \caption{Multi-armed bandits, 4 environments}
  \end{subfigure}
  \hfill
  \begin{subfigure}[b]{0.24\textwidth}
    \centering
\includegraphics[width=\textwidth]{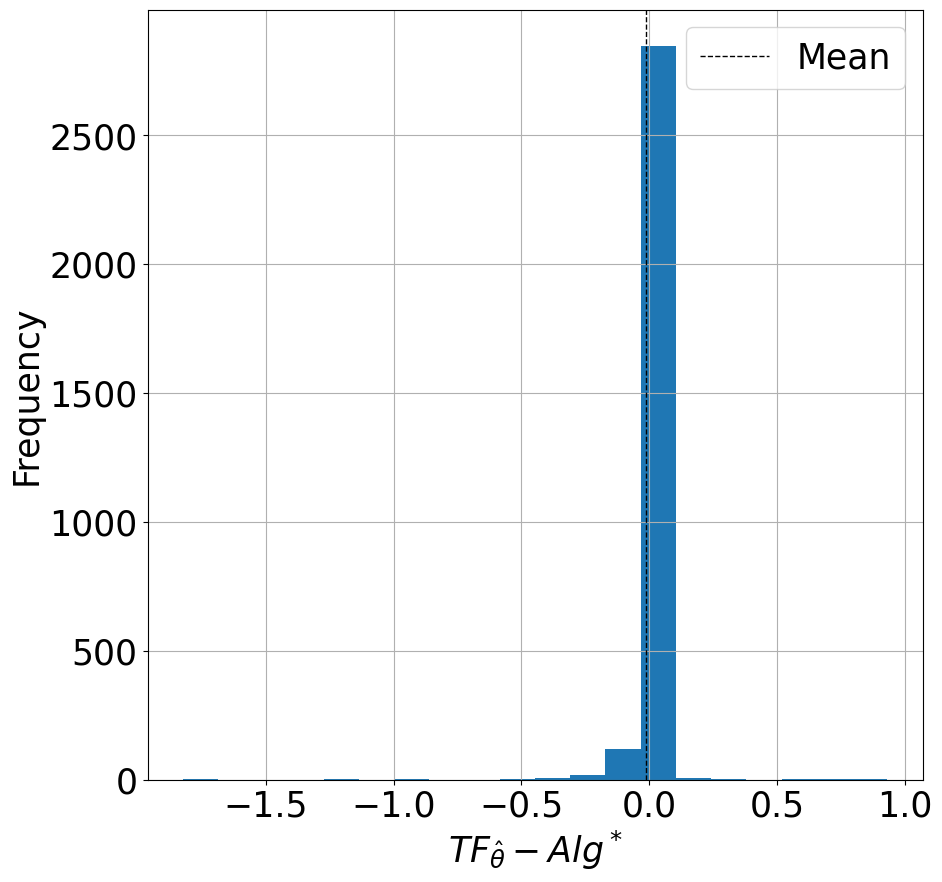}
    \caption{Dynamic pricing, 4 environments }
  \end{subfigure}
    \hfill
  \begin{subfigure}[b]{0.24\textwidth}
    \centering
    \includegraphics[width=\textwidth]{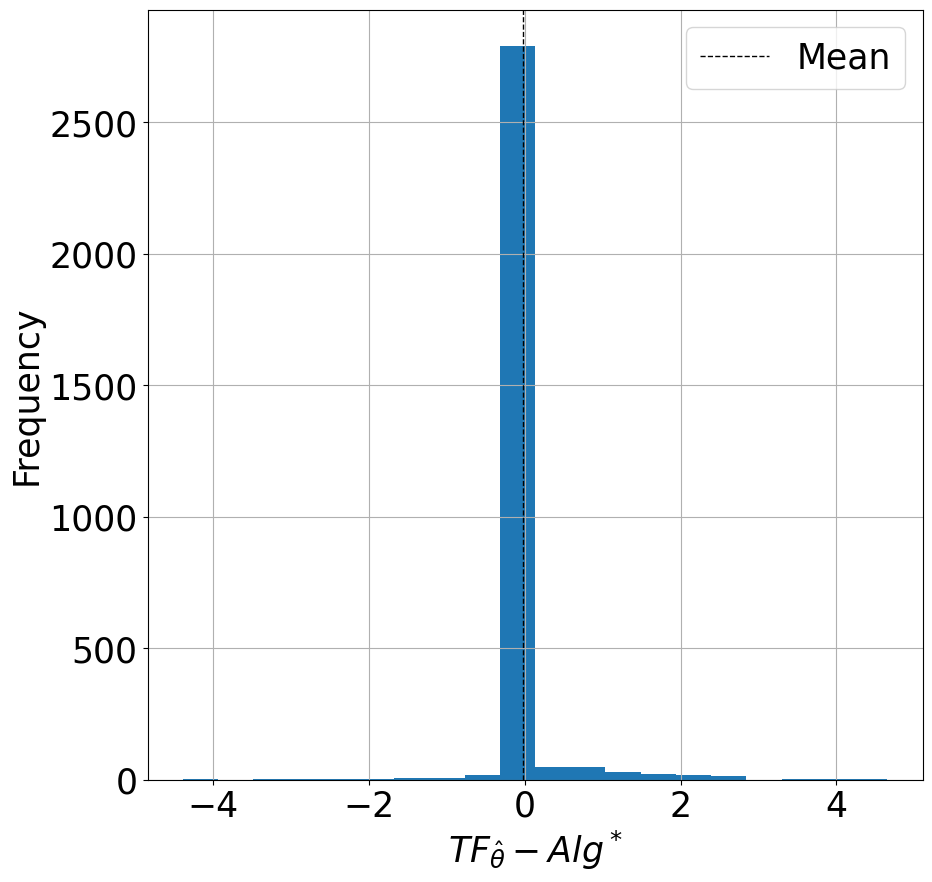}
    \caption{Newsvendor, $4$ environments}
  \end{subfigure}
    \hfill
  \begin{subfigure}[b]{0.24\textwidth}
    \centering
\includegraphics[width=\textwidth]{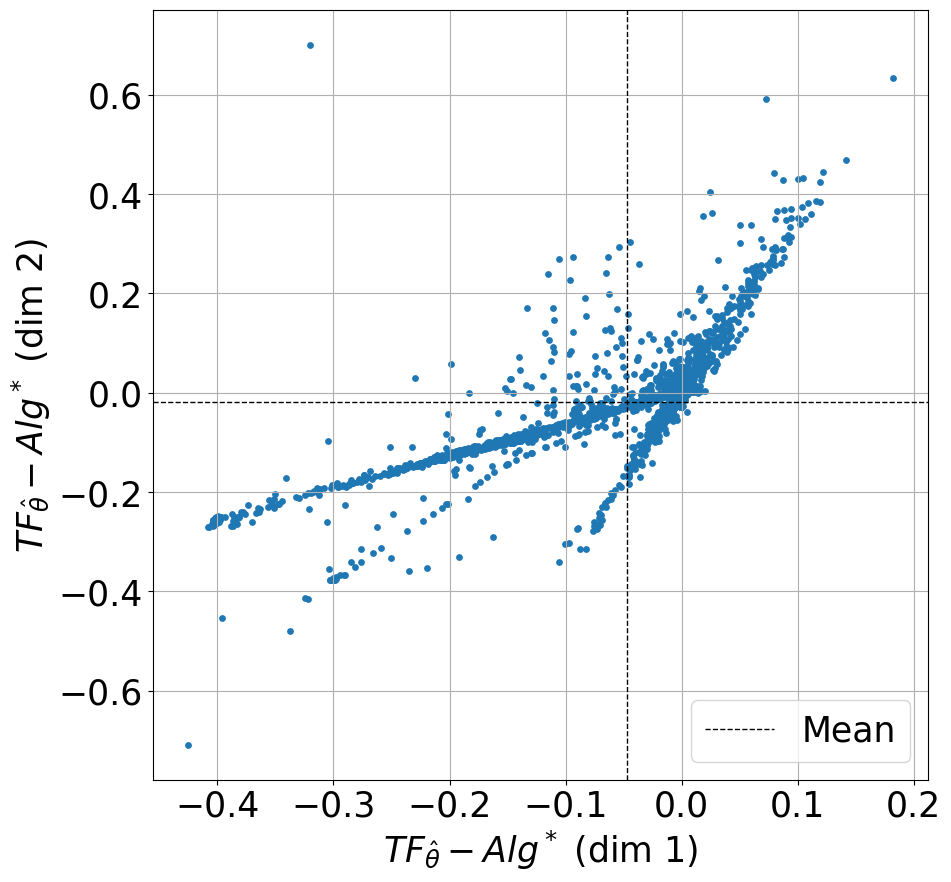}
    \caption{Linear bandits, 4 environments}
  \end{subfigure}

  \begin{subfigure}[b]{0.24\textwidth}
    \centering
\includegraphics[width=\textwidth]{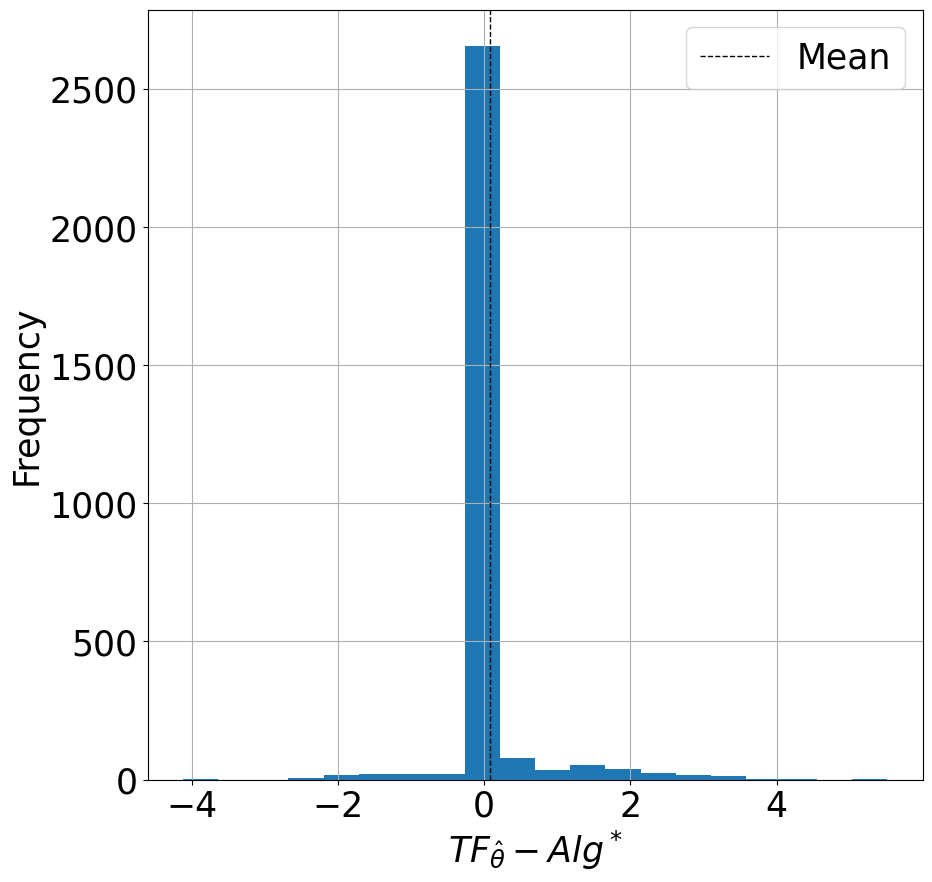}
    \caption{Multi-armed bandits, 100 environments}
  \end{subfigure}
  \hfill
  \begin{subfigure}[b]{0.24\textwidth}
    \centering
\includegraphics[width=\textwidth]{figs/DP/_100env_4d_123_Act_diff.png}
    \caption{Dynamic pricing, 100 environments }
  \end{subfigure}
    \hfill
  \begin{subfigure}[b]{0.24\textwidth}
    \centering
    \includegraphics[width=\textwidth]{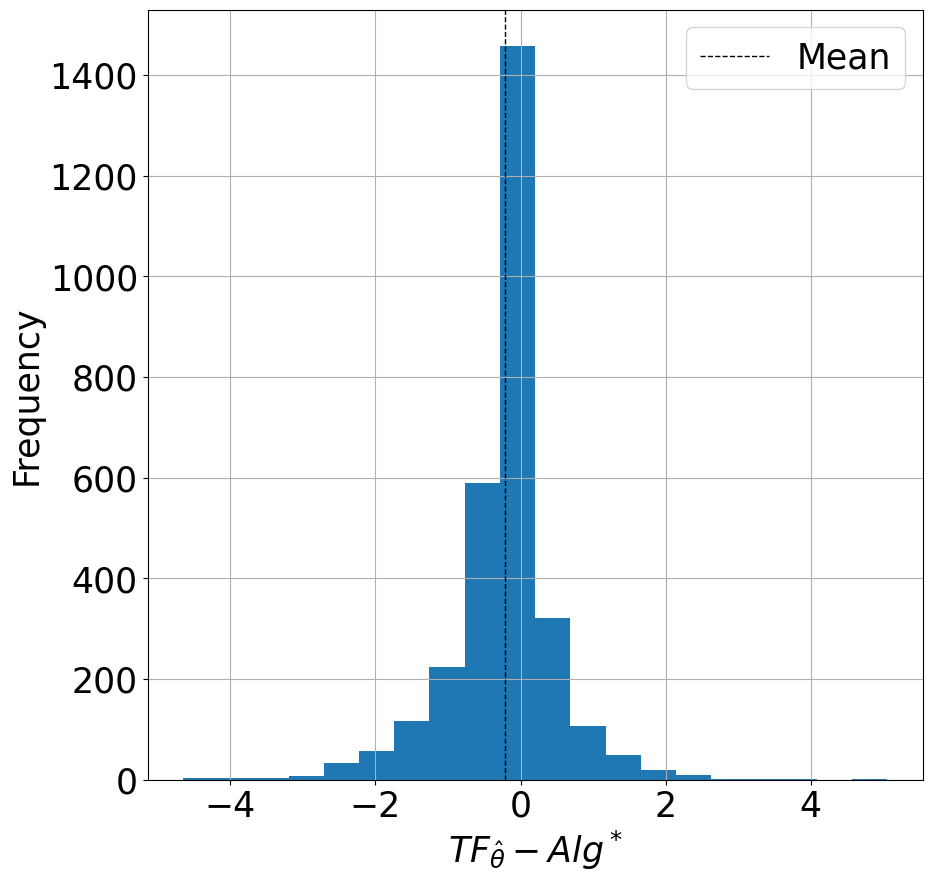}
    \caption{Newsvendor, 100 environments}
  \end{subfigure}
    \hfill
  \begin{subfigure}[b]{0.24\textwidth}
    \centering
\includegraphics[width=\textwidth]{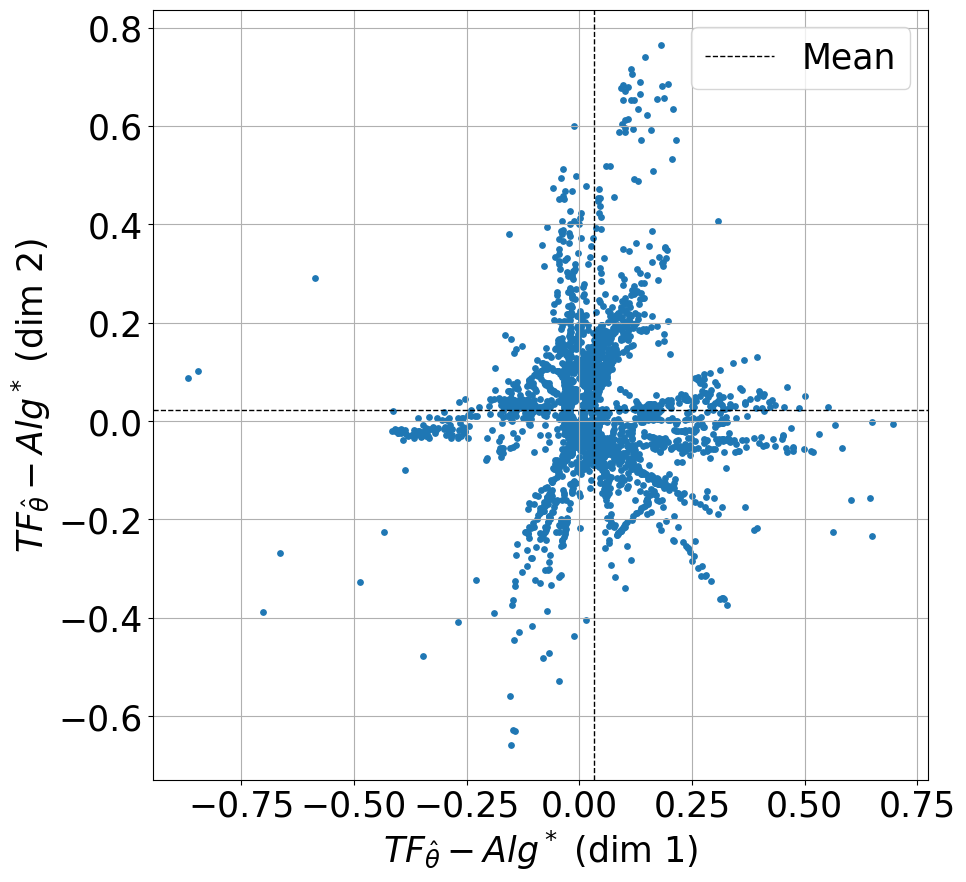}
    \caption{Linear bandits, 100 environments}
  \end{subfigure}
  \caption{$\texttt{TF}_{\hat{\theta}}(H_t)-\texttt{Alg}^*(H_t)$ across different tasks with various numbers of possible environments. More possible environments lead to a harder decision making problem.}
  \label{fig:appx_transformer_noise}
\end{figure}

Here we provide more results regarding the matching of $\texttt{TF}_{\hat{\theta}}$ and $\texttt{Alg}^*$  across different tasks: Figure \ref{fig:appx_bayesmatch} provides more examples comparing decisions made by $\texttt{TF}_{\hat{\theta}}$ and $\texttt{Alg}^*$ at sample path levels, while Figure \ref{fig:appx_transformer_noise} compares them at population levels.

We plot the actions generated from  $\texttt{TF}_{\hat{\theta}}$ and $\texttt{Alg}^*$ on the same environment with the same contexts  in Figure \ref{fig:appx_bayesmatch} across multi-armed bandits, dynamic pricing and newsvendor tasks. Each row has a different number of possible environments with positive probability by $\mathcal{P}_{\gamma}$ (See Appendix \ref{appx:envs} for the definition of such a finite pool of environments). The population-level differences $\{\texttt{TF}_{\hat{\theta}}(H_t)-\texttt{Alg}^*(H_t)\}_{t=1}^T$ are presented in Figure \ref{fig:appx_transformer_noise}.  We observe that across different tasks: (1) $\texttt{TF}_{\hat{\theta}}$ nearly matches $\texttt{Alg}^*$, but (2) the matchings are not perfect. The actions from $\texttt{TF}_{\hat{\theta}}$ and $\texttt{Alg}^*$ are not exactly the same, and their differences increase as the underlying problems become more complex, as seen by comparing the first and second rows in Figure \ref{fig:appx_transformer_noise}.

\textbf{Setup.} 
For Figure \ref{fig:appx_bayesmatch}, each subfigure is based on a sampled environment with a sampled sequence of contexts $\{X_t\}_{t=1}^{30}$ from the corresponding task. The data generation process follows the description detailed in Appendix \ref{appx:envs}.  Both the pre-training and testing samples are drawn from \(\mathcal{P}_{\gamma}\), i.e., from these fixed 16/100 environments. The remaining setups for these tasks follow the methods provided in Appendix \ref{appx:envs}.

For each task shown in Figure \ref{fig:appx_transformer_noise}, we generate $100$ sequences of  $\{\texttt{TF}_{\hat{\theta}}(H_t)-\texttt{Alg}^*(H_t)\}_{t=1}^{100}$, i.e., each subfigure is based on $10000$ samples of $\texttt{TF}_{\hat{\theta}}(H_t)-\texttt{Alg}^*(H_t)$. For dynamic pricing and newsvendor,  actions are scalars and thus $\texttt{TF}_{\hat{\theta}}(H_t)-\texttt{Alg}^*(H_t)$ is also a scalar; For multi-armed bandits, the action space is discrete and we use the  reward difference associated with the chosen actions (by $\texttt{TF}_{\hat{\theta}}(H_t)$ or $\texttt{Alg}^*(H_t)$) as the value of $\texttt{TF}_{\hat{\theta}}(H_t)-\texttt{Alg}^*(H_t)$. We use histograms to summarize the samples from them. For linear bandits, the actions are $2$-dimensional and thus we use the scatters to show  $\texttt{TF}_{\hat{\theta}}(H_t)-\texttt{Alg}^*(H_t)$ in Figure \ref{fig:appx_transformer_noise} (d) or (h).

\subsubsection{Benefit of Utilizing Prior Knowledge}
\label{appx:prior_benefit}
\begin{figure}[ht!]
  \centering
  \begin{subfigure}[b]{0.23\textwidth}
    \centering
\includegraphics[width=\textwidth]{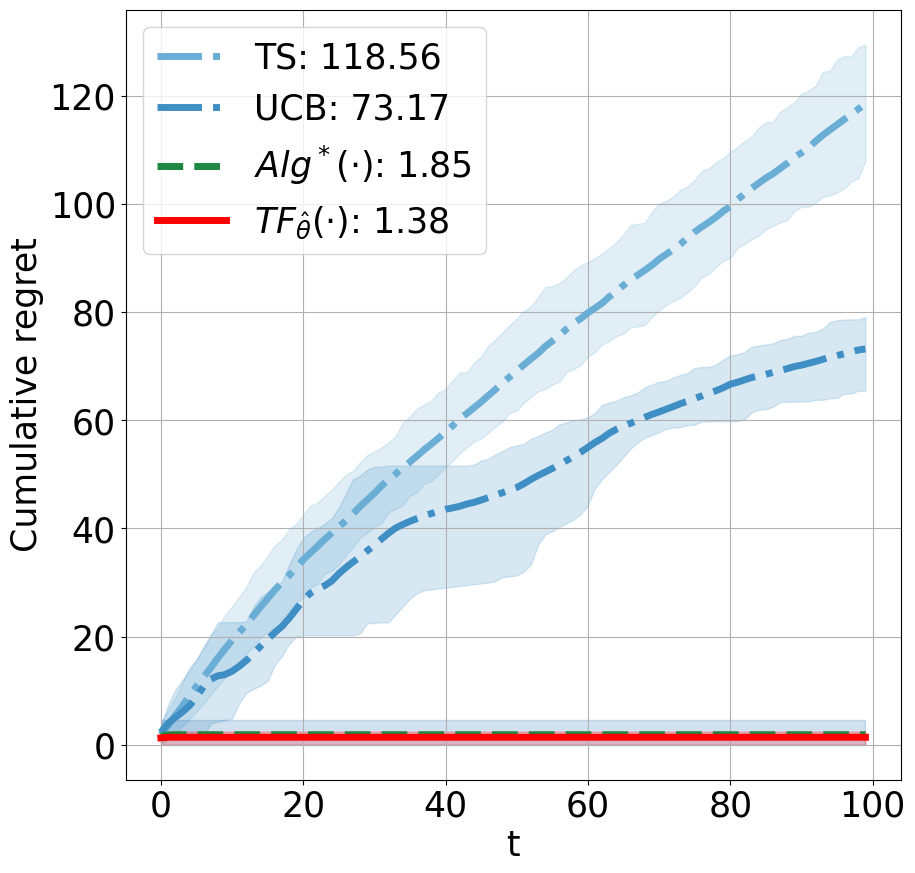}
    \caption{Multi-armed bandits}
  \end{subfigure}
    \hfill
    \begin{subfigure}[b]{0.23\textwidth}
    \centering
\includegraphics[width=\textwidth]{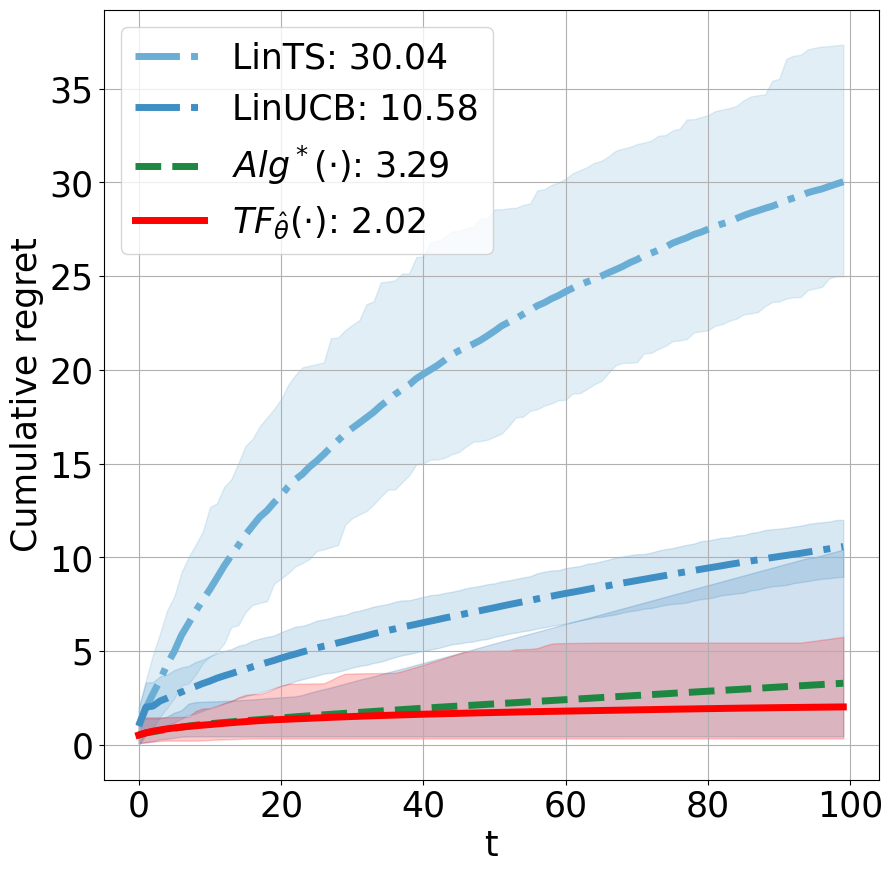}
    \caption{Linear bandits}
  \end{subfigure}
  \hfill  
 \begin{subfigure}[b]{0.23\textwidth}
    \centering
\includegraphics[width=\textwidth]{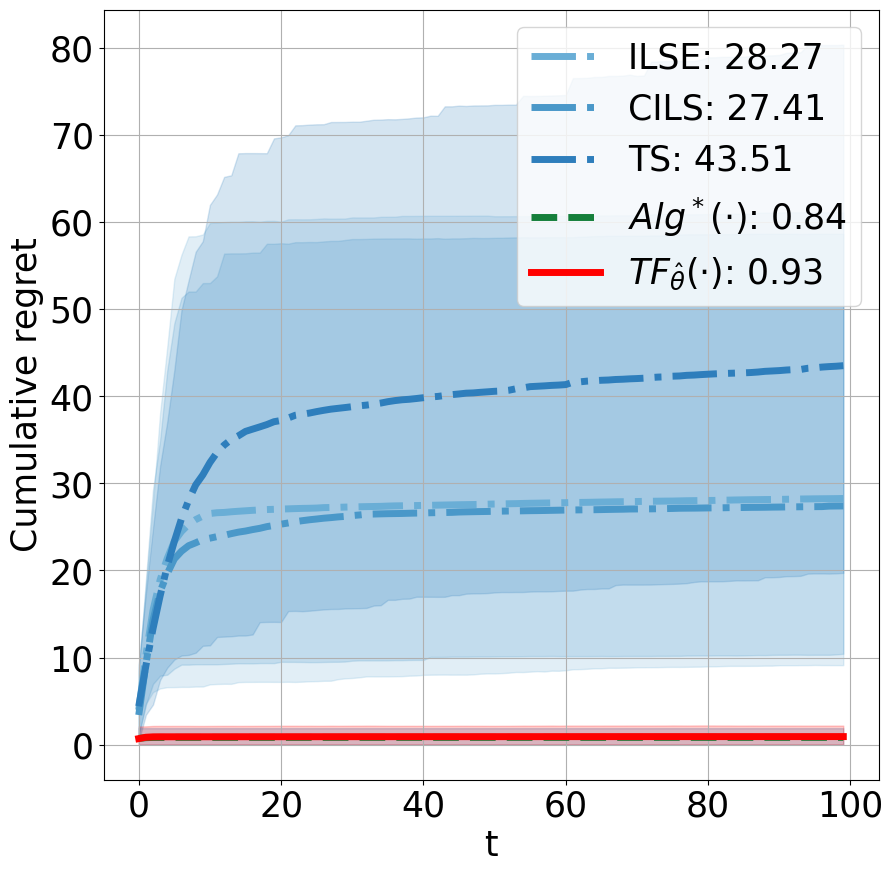}
    \caption{Dynamic pricing}
  \end{subfigure}
    \hfill
  \begin{subfigure}[b]{0.23\textwidth}
    \centering
\includegraphics[width=\textwidth]{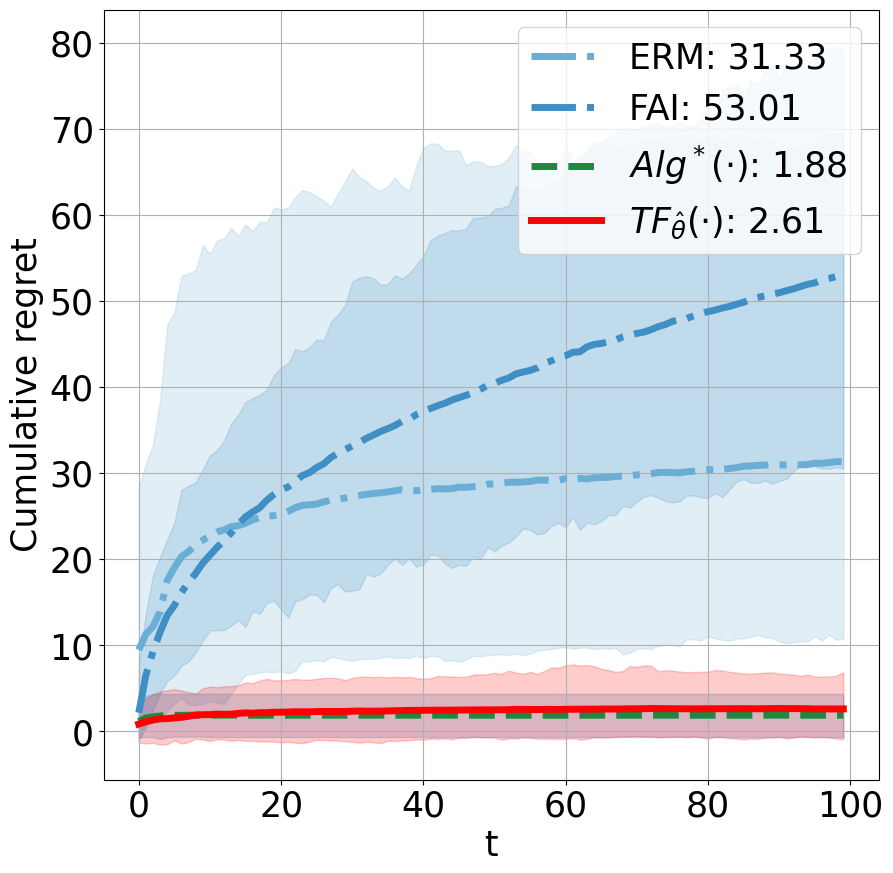}
    \caption{Newsvendor}

  \end{subfigure}
  \caption{ The average out-of-sample regret  on tasks with simpler environments, where each task only has $4$ possible environments.  The numbers in the legend bar are the final regret at $t = 100$ and the shaded areas indicate the $90\%$ (empirical) confidence intervals. The details of benchmarks can be found in Appendix \ref{appx:benchmark}.}
  \label{fig:appx_regret_ez}
\end{figure}

\begin{figure}[ht!]
  \centering
  \begin{subfigure}[b]{0.23\textwidth}
    \centering
\includegraphics[width=\textwidth]{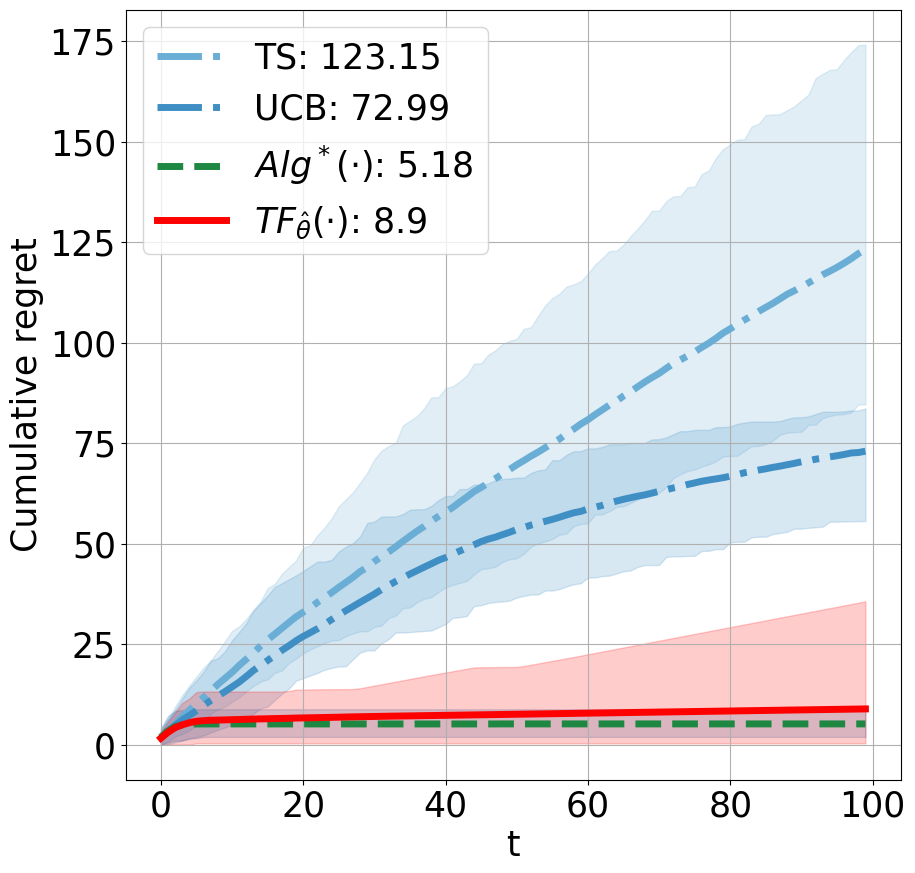}
    \caption{Multi-armed bandits, 100 environments, linear demand}
  \end{subfigure}
    \hfill
    \begin{subfigure}[b]{0.23\textwidth}
    \centering
\includegraphics[width=\textwidth]{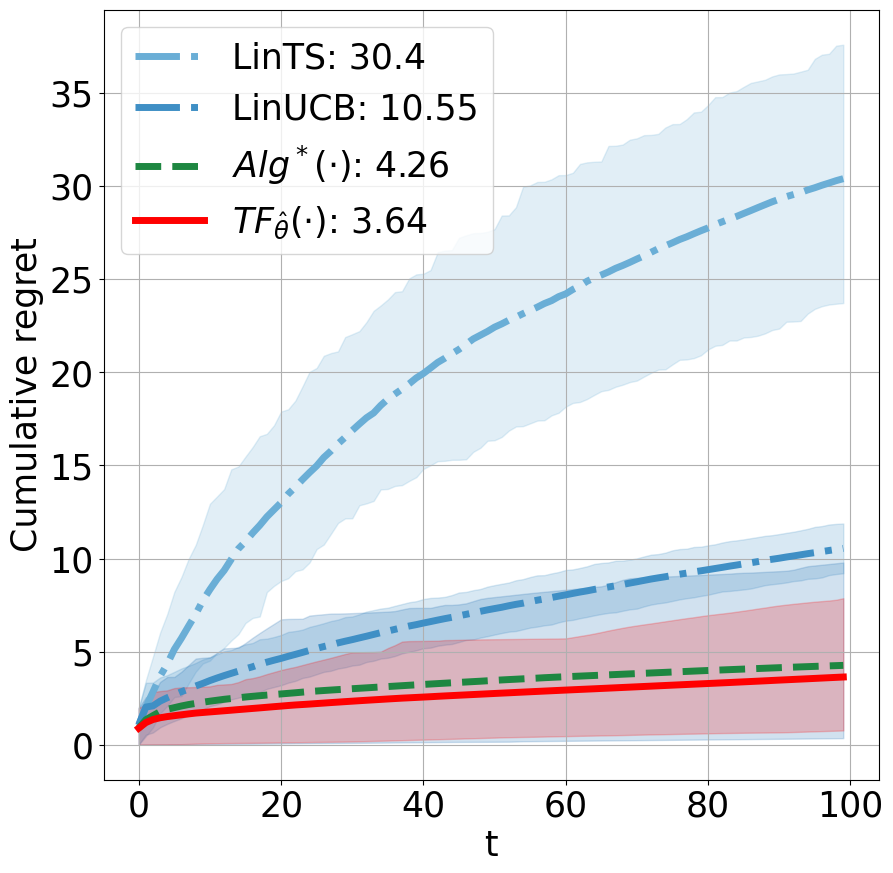}
    \caption{Linear bandits, 100 environments, linear demand}
  \end{subfigure}
  \hfill  
 \begin{subfigure}[b]{0.23\textwidth}
    \centering
\includegraphics[width=\textwidth]{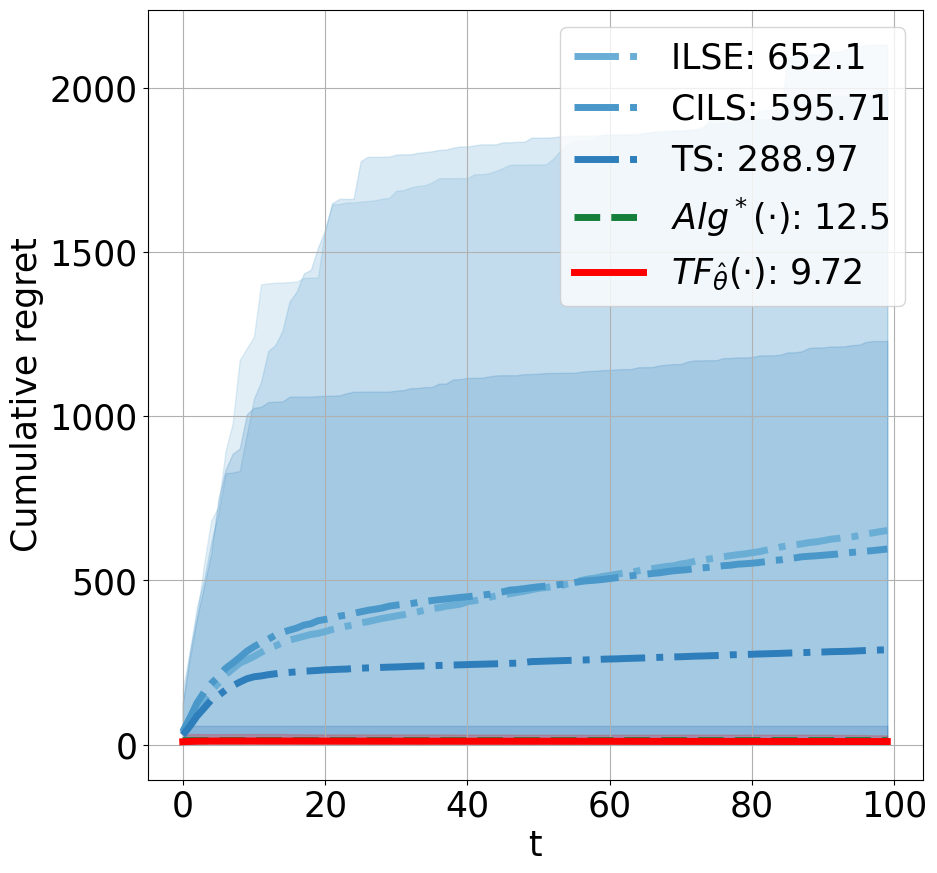}
    \caption{Dynamic pricing, 16 environments, 2 demand types}
  \end{subfigure}
    \hfill
  \begin{subfigure}[b]{0.23\textwidth}
    \centering
\includegraphics[width=\textwidth]{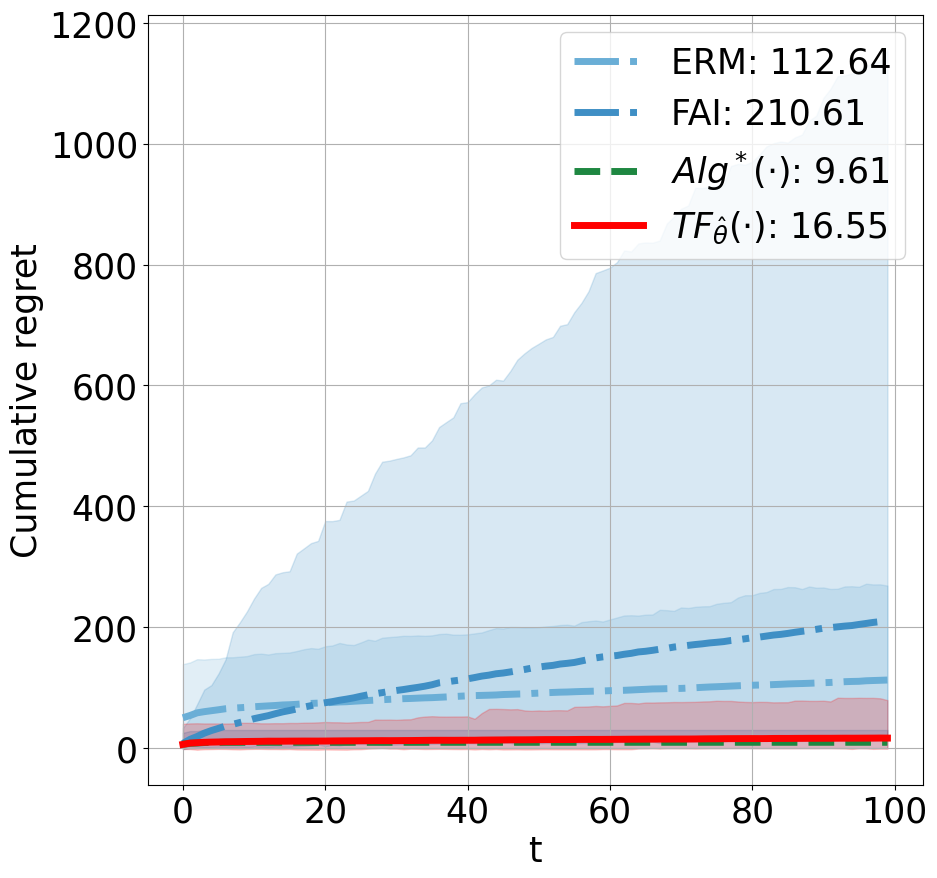}
    \caption{Newsvendor, 16 environments, 2 demand types}

  \end{subfigure}
  \caption{The average out-of-sample regret  on harder tasks: with 100 environments in (a) multi-armed bandits and (b) linear bandits; or with 16 environments, 2 demand types in (c) dynamic pricing and (d) newsvendor.  The numbers in the legend bar are the final regret at $t = 100$ and the shaded areas indicate the $90\%$ (empirical) confidence intervals. The details of benchmarks can be found in Appendix \ref{appx:benchmark}}
  \label{fig:appx_regret_hard}
\end{figure}

Here we present further results on the performance of $\texttt{TF}_{\hat{\theta}}$ to demonstrate the benefit of utilizing prior knowledge $\mathcal{P}_{\gamma}$. Specifically, we compare $\texttt{TF}_{\hat{\theta}}$ with benchmark algorithms on (i) simple tasks, which include only 4 possible environments in both pretraining and testing samples (Figure \ref{fig:appx_regret_ez}), and (ii) more complex tasks, which involve either 100 possible environments (Figure \ref{fig:appx_regret_hard} (a), (b)) or 16 environments with two possible types of demand functions (Figure \ref{fig:appx_regret_hard} (c), (d)). See Appendix \ref{appx:envs} for the definition of finite environments used here.

Figures \ref{fig:appx_regret_ez} and \ref{fig:appx_regret_hard} illustrate the consistently and significantly superior performance of $\texttt{TF}_{\hat{\theta}}$ and $\texttt{Alg}^*$ across all tasks compared to the benchmark algorithms. These results underscore the advantage of leveraging prior knowledge about the tested environments.

\textbf{Setup.} All figures are based on 100 runs. We consider four tasks with 4 environments (i.e., the support of $\mathcal{P}_{\gamma}$ contains only 4 environments) in Figure \ref{fig:appx_regret_ez} and two tasks with 100 environments in Figure \ref{fig:appx_regret_hard} (a), (b). The setup for these tasks follows the generation methods provided in Appendix \ref{appx:envs}.

For the tasks in Figure \ref{fig:appx_regret_hard} (c), (d), there are 16 environments included in the support of $\mathcal{P}_{\gamma}$, where 8 environments have demand functions of the linear type and the other 8 have demand functions of the square type (see Appendix \ref{appx:envs} for the definitions of these two types). Both the pretraining and testing samples are drawn from \(\mathcal{P}_{\gamma}\), i.e., from these 16 environments. The remaining setups for these tasks follow the methods provided in Appendix \ref{appx:envs}.

\subsubsection{Impact of Network Architecture}
\label{appx:LSTM}
Here we investigate the effect of different network architectures on performance. Specifically, we replace the transformer/GPT-2 architecture used in OMGPT with Long Short-Term Memory (LSTM) \citep{hochreiter1997long}, a popular neural network architecture for sequence modeling before the rise of transformers \citep{van2020review}.

We maintain the overall architecture of OMGPT, as described in Appendix \ref{appx:architecture} except for replacing the transformer/GPT-2 module with LSTM. We evaluate two LSTM variants: a 5-layer LSTM and a 12-layer LSTM (for comparison, the tested transformer also has 12 layers). Typically, LSTM architectures should be shallower than transformers in practice, which is why we also include the 5-layer version. Other hyperparameters like embedding space dimension are kept the same across all models. We pre-train and test these models on dynamic pricing problems (see Appendix \ref{appx:envs} for the experimental setup), and the pre-training procedure is identical to the one outlined in Appendix \ref{sec:algo}.

Figure \ref{fig:LSTM_reg} presents the testing regret for the three models. The results indicate no significant difference between the 5-layer and 12-layer LSTM models. However, the transformer consistently performs better than both LSTM variants. This outcome aligns with the transformer’s superior performance in other domains, such as natural language processing, and suggests that transformers are more effective for sequential decision making tasks as well.

\begin{figure}[ht!]
    \centering
    \includegraphics[width=0.7\linewidth]{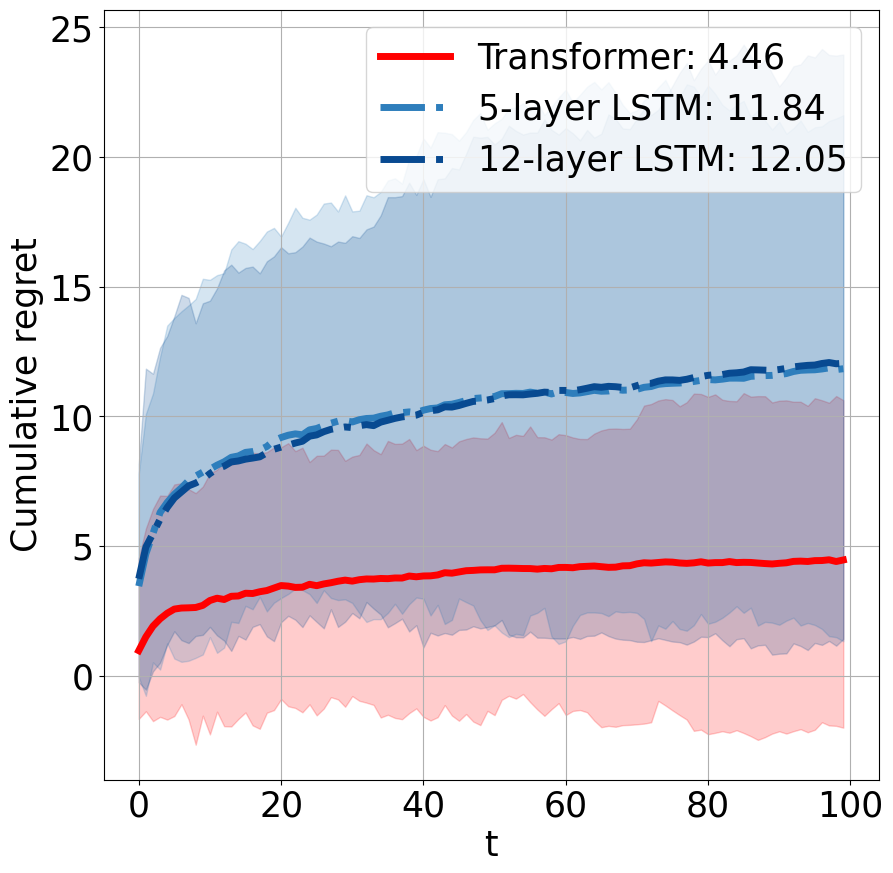}
    \caption{Comparison of the average out-of-sample regret between using the Transformer and LSTM architectures in OMGPT.}
    \label{fig:LSTM_reg}
\end{figure}

\subsubsection{More Results for Figure \ref{fig:probe_tSNE}}
\label{appx: tSNE}
\begin{figure}[ht!]
  \centering
  \begin{subfigure}[b]{0.31\textwidth}
    \centering
\includegraphics[width=\textwidth]{new_figs/probe/hidden_emb_layer_0.png}
    \caption{Layer 0 }
  \end{subfigure}
  \hfill
  \begin{subfigure}[b]{0.31\textwidth}
    \centering
\includegraphics[width=\textwidth]{new_figs/probe/hidden_emb_layer_1.png}
    \caption{Layer 1 }
  \end{subfigure}
  \hfill
  \begin{subfigure}[b]{0.31\textwidth}
    \centering
\includegraphics[width=\textwidth]{new_figs/probe/hidden_emb_layer_2.png}
    \caption{Layer 2}
  \end{subfigure}
    \begin{subfigure}[b]{0.31\textwidth}
    \centering
\includegraphics[width=\textwidth]{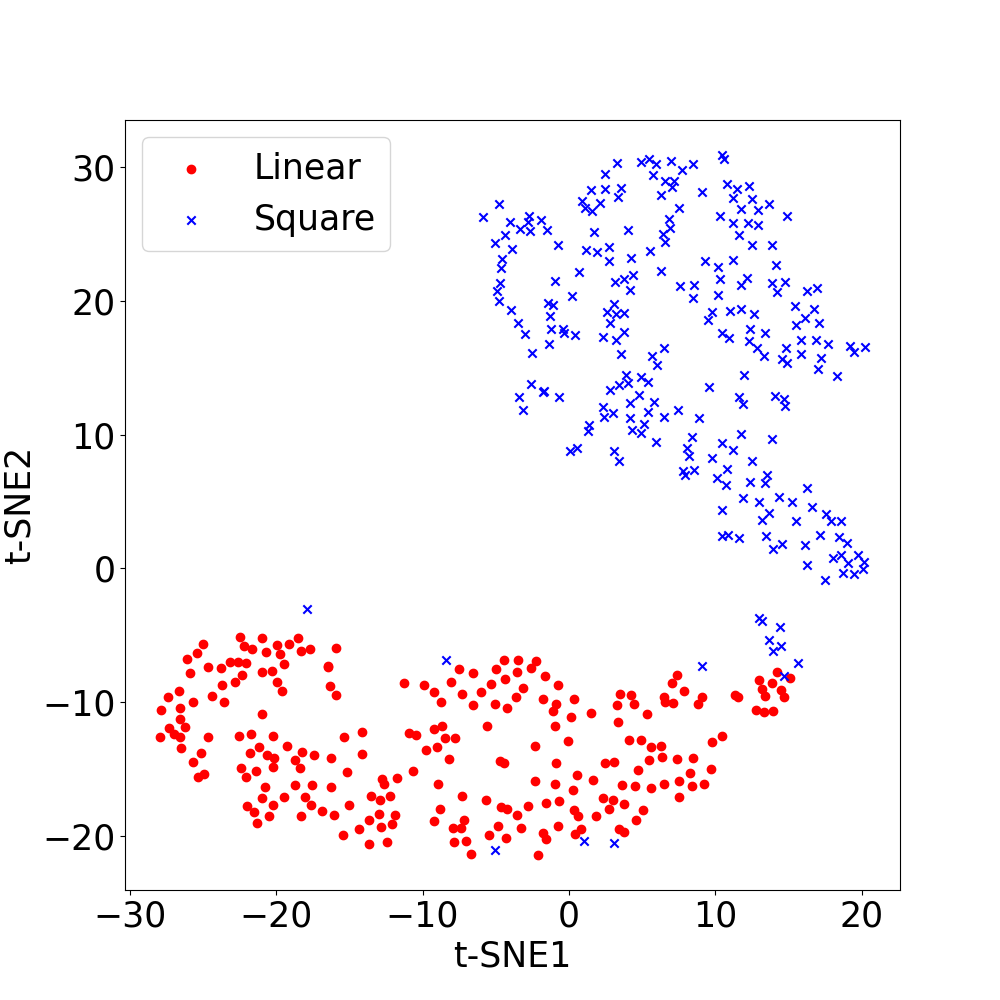}
    \caption{Layer 3 }
  \end{subfigure}
  \hfill
  \begin{subfigure}[b]{0.31\textwidth}
    \centering
\includegraphics[width=\textwidth]{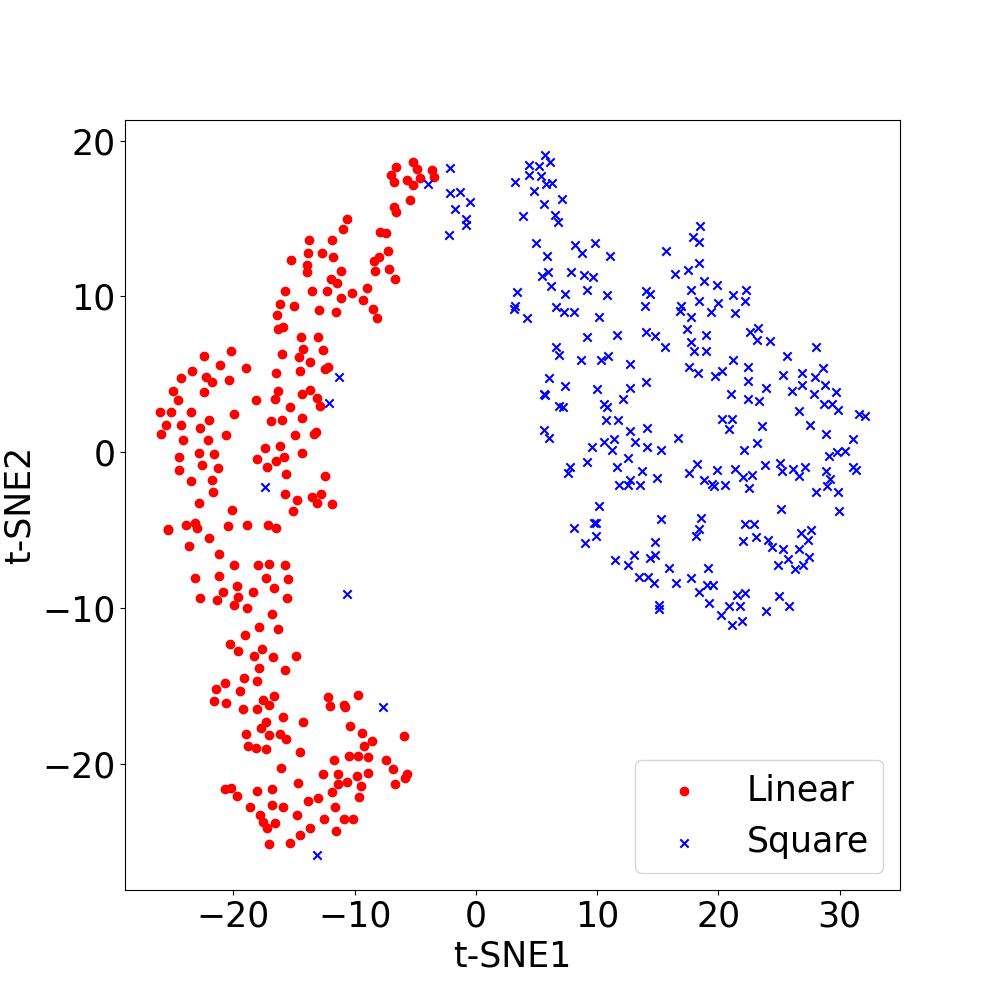}
    \caption{Layer 4 }
  \end{subfigure}
  \hfill
  \begin{subfigure}[b]{0.31\textwidth}
    \centering
\includegraphics[width=\textwidth]{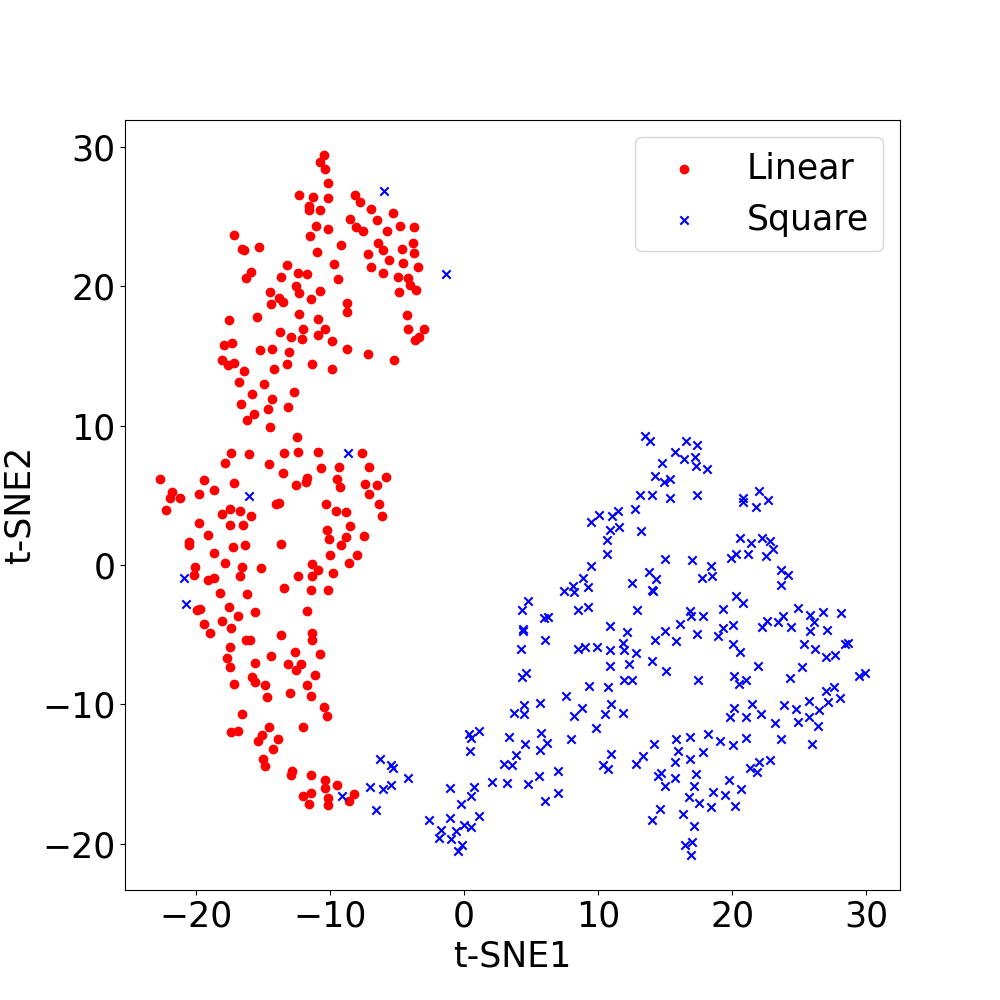}
    \caption{Layer 5}
  \end{subfigure}
    \begin{subfigure}[b]{0.31\textwidth}
    \centering
\includegraphics[width=\textwidth]{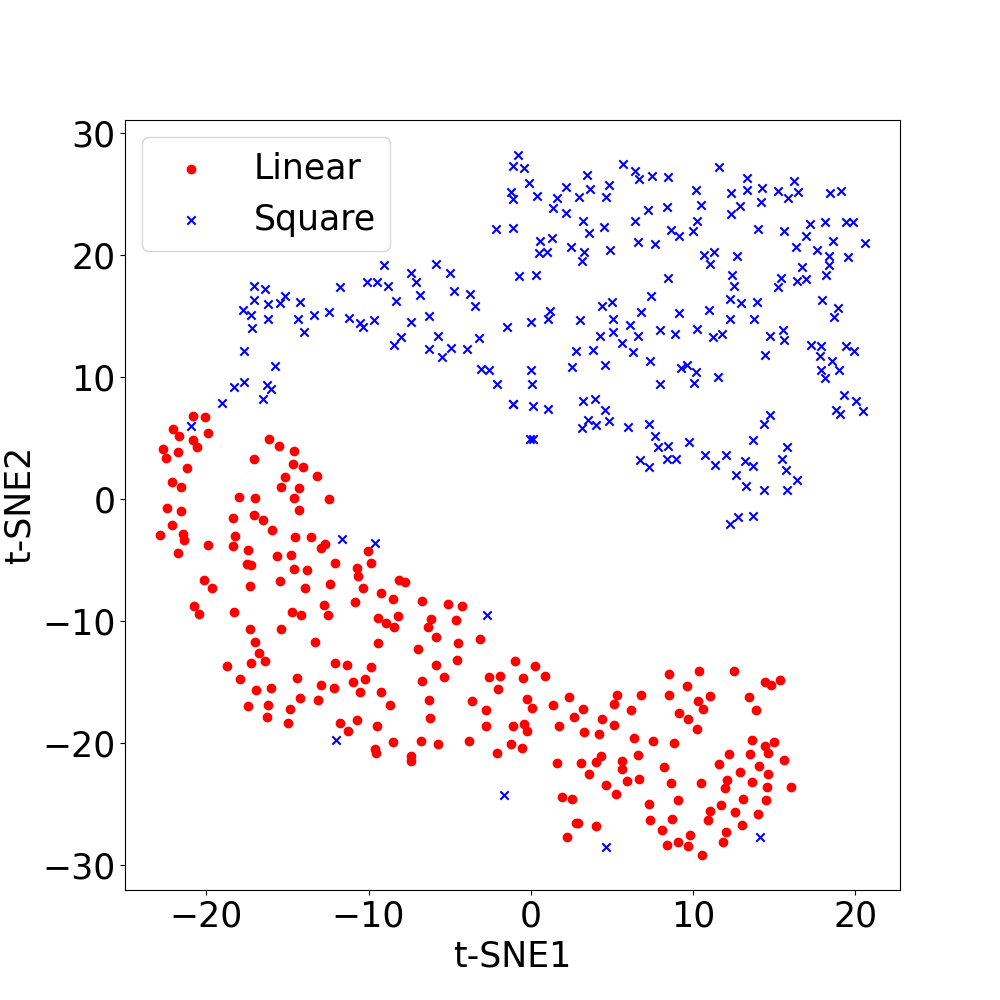}
    \caption{Layer 6 }
  \end{subfigure}
  \hfill
  \begin{subfigure}[b]{0.31\textwidth}
    \centering
\includegraphics[width=\textwidth]{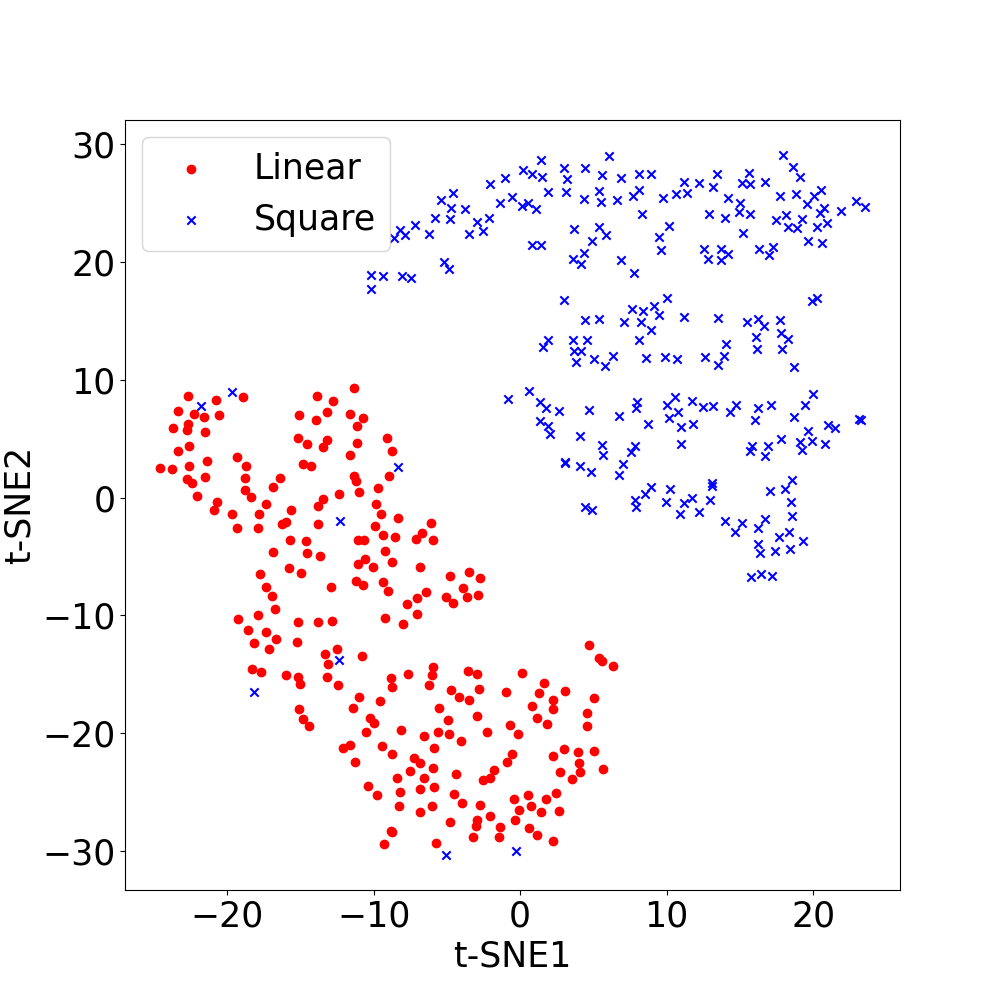}
    \caption{Layer 7 }
  \end{subfigure}
  \hfill
  \begin{subfigure}[b]{0.31\textwidth}
    \centering
\includegraphics[width=\textwidth]{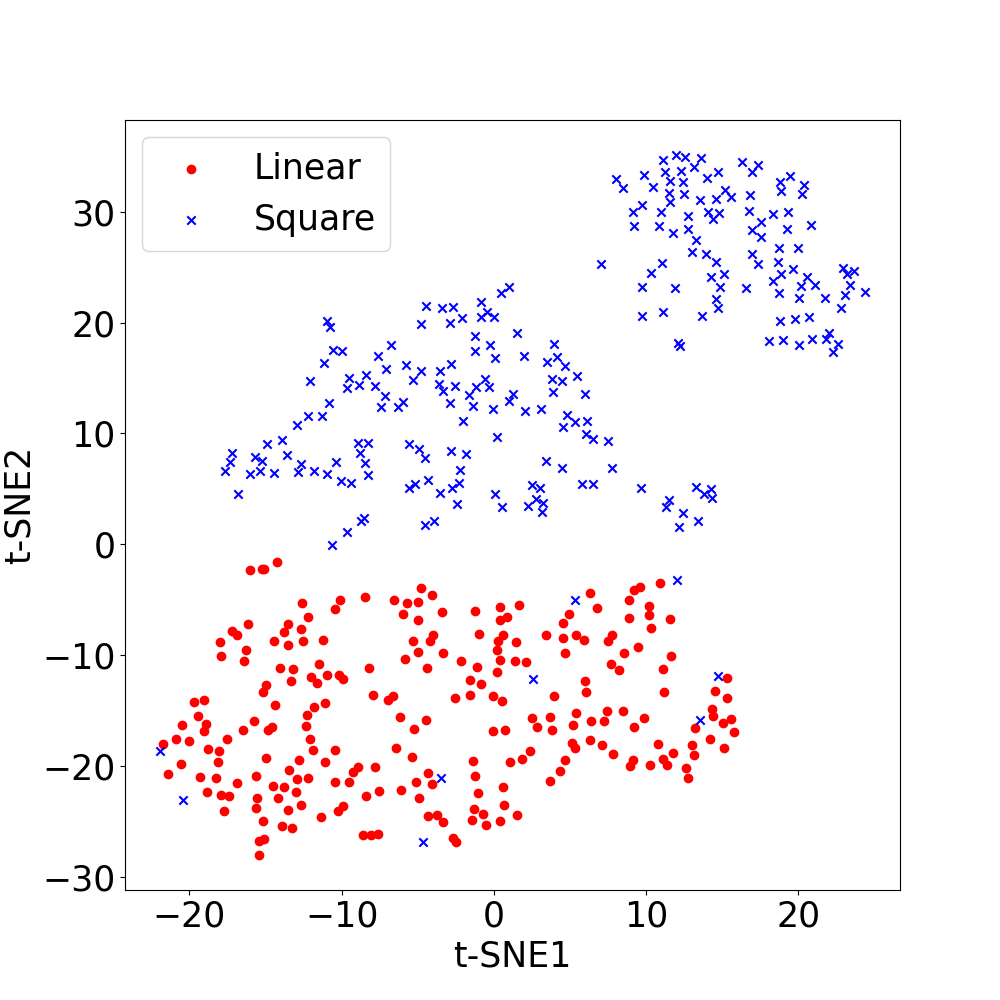}
    \caption{Layer 8}
  \end{subfigure}
  \begin{subfigure}[b]{0.31\textwidth}
    \centering
\includegraphics[width=\textwidth]{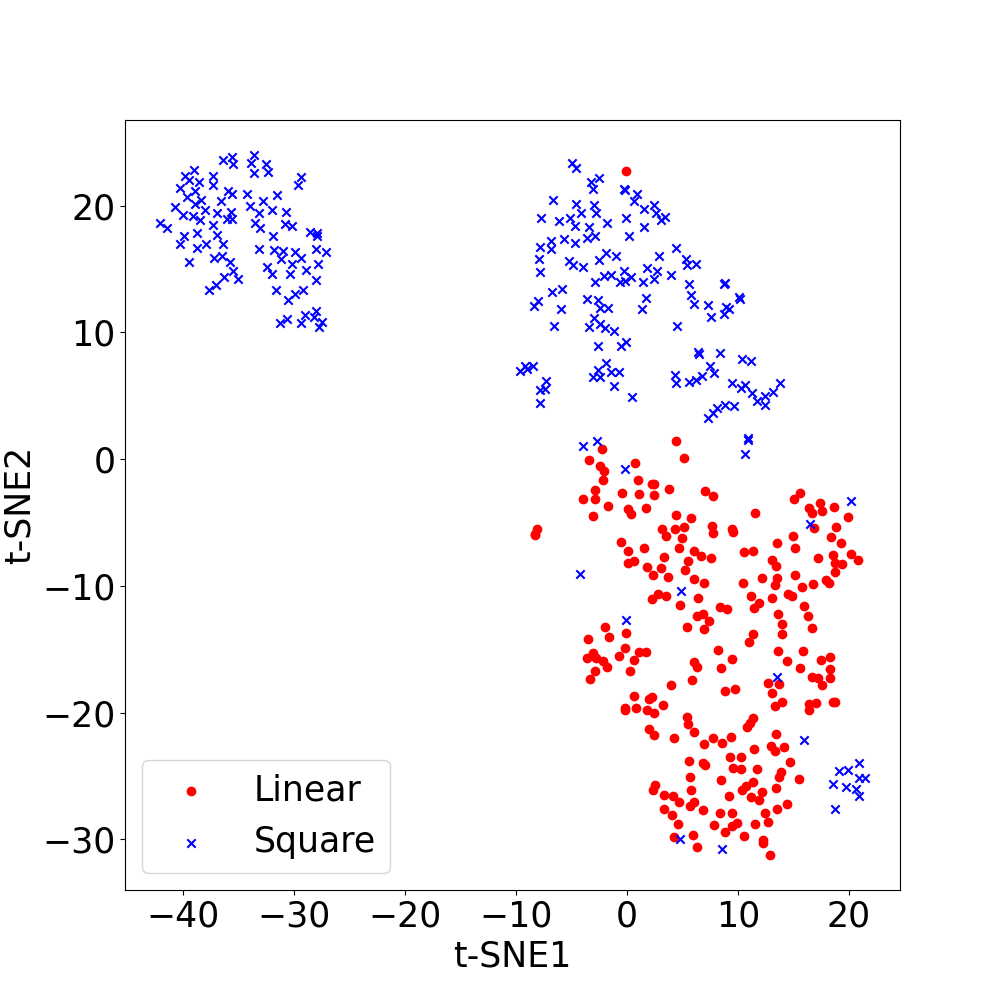}
    \caption{Layer 9 }
  \end{subfigure}
  \hfill
  \begin{subfigure}[b]{0.31\textwidth}
    \centering
\includegraphics[width=\textwidth]{new_figs/probe/hidden_emb_layer_10.png}
    \caption{Layer 10 }
  \end{subfigure}
  \hfill
  \begin{subfigure}[b]{0.31\textwidth}
    \centering
\includegraphics[width=\textwidth]{new_figs/probe/hidden_emb_layer_11.png}
    \caption{Layer 11}
  \end{subfigure}
  \caption{Complete results corresponding to Figure \ref{fig:probe_tSNE}. Visualization of output embedding vectors from different layers. The vectors corresponding to different demand function types become clearly separated starting from layer 2, indicating that environment inference occurs during the processing. We omit the layer 12 here.}
  \label{fig:probe_tSNE_full}
\end{figure}

Figure \ref{fig:probe_tSNE_full} presents the complete results corresponding to Figure \ref{fig:probe_tSNE}.

\subsection{Sequential Decision Making Problems and Environment Generation}
\label{appx:envs}
We provide the formulations of the sequential decision making problems studied in this paper, along with their generation methods in experiments. We set the time horizon $T=100$ for all tasks except Figure \ref{fig:horizon_gen} and the revenue management task, where we set $T=50$ to reduce the computational cost of running the benchmark algorithm and generating the pre-training data.
\paragraph{Stochastic multi-armed bandit} There is no context, i.e., $X_t=\text{null}$ for all $t.$ The action $a_t\in \mathcal{A}=\left\{1,2,\ldots,k\right\}$ denotes the index of the arm played at time $t$.  In experiments, We consider the number of actions/arms $|\mathcal{A}|=20$. The environment parameter $\gamma$ encapsulates the expected reward $r_a$ of each arm $a=1,\ldots,20$, where the reward of arm $a$ is sampled from $\mathcal{N}(r_a,0.2)$. Thus, the environment distribution $\mathcal{P}_{\gamma}$ is defined by the (joint) distribution of $(r_1,\ldots,r_{20})$. We set $r_a \overset{\text{i.i.d.}}{\sim} \mathcal{N}(0,1)$ for each action $a$. The optimal arm is $a^* = \argmax_{a} r_a$.
    \paragraph{Linear bandits} There is no context, i.e., $X_t=\text{null}$ for all $t.$ The action $a_t\in \mathcal{A} \subset \mathbb{R}^{d}$ is selected from some pre-specified domain $\mathcal{A}$. The random reward $R(X_t, a_t)=w^\top a_t+\epsilon_t$ where $\epsilon_t$ is some zero-mean noise random variable and $w\in \mathbb{R}^{d}$ is a vector unknown to the decision maker. The observation $O_t=R(X_t, a_t)$ and the expected reward $r(X_t, a_t)= \mathbb{E}\left[R(X_t, a_t)|X_t,a_t\right]=w^\top a_t$.   In experiments, we consider the dimension of actions/arms $d=2$ and $\mathcal{A}=\mathcal{B}(0,1)$, i.e., a unit ball centered at the origin (with respect to the Euclidean norm). The environment parameter $\gamma$ encapsulates the (linear) reward function's parameter $w \in \{\tilde{w} \in \mathbb{R}^2: \|\tilde{w}\|_2=1\}$, where the reward of arm $a$ is sampled from $\mathcal{N}(w^\top a,0.2)$. Thus, the environment distribution $\mathcal{P}_{\gamma}$ is defined by the distribution of $w$. We set $w$ to be uniformly sampled from the unit sphere, and the corresponding optimal arm being $a^* = w$.

\paragraph{Dynamic pricing} In experiments, we set the noise $\epsilon_t \overset{\text{i.i.d.}}{\sim} \mathcal{N}(0,0.2)$ and the context $X_t$ to be sampled i.i.d. and uniformly from $[0,5/2]^d$, with $d=6$ as the dimension of contexts (except Figure \ref{fig:aba_size}, where we set $d=4,10,20$ and accordingly change the dimension of the following introduced parameters), and $\mathcal{A}=[0,30]$. We consider the linear demand function family $D_t = \alpha^\top X_t - \beta^\top X_t \cdot a_t + \epsilon_t$, where $\alpha,\beta \in \mathbb{R}^d$ are the demand parameters. Thus, the environment distribution $\mathcal{P}_{\gamma}$ is defined by the (joint) distribution of $(\alpha,\beta)$. We set $\alpha$ to be uniformly sampled from $[1/2, 3/2]^6$ and $\beta$ to be uniformly sampled from $[1/20, 21/20]^6$, independent of $\alpha$. The optimal action $a^*_t = \frac{\alpha^\top X_t}{2 \cdot \beta^\top X_t}$. We also consider a square demand function family, which will be specified later, to test the OMGPT's performance on a mixture of different demand-type tasks.

\paragraph{Newsvendor problem} In experiments, we set the context $\tilde{X}_t$ to be sampled i.i.d. and uniformly from $[0,3]^d$ with $d=4$, and $\mathcal{A}=[0,30]$. We consider the linear demand function family $D_t = w^\top \tilde{X}_t + \epsilon_t$, where $w \in \mathbb{R}^4$. Under uncensored demand setting, the observation $O_t=D_t$, while under the censored demand setting, the observation $O_t=\min\{D_t,a_t\}$. The environment parameter $\gamma$ encapsulates (i) the upper bound $\bar{\epsilon}$ of the noise $\epsilon_t$, where $\epsilon_t \overset{\text{i.i.d.}}{\sim} \text{Unif}(0, \bar{\epsilon})$; (ii) the left-over cost $h$ (with the lost-sale cost $l \text{ being } 1$); and (iii) the demand parameter $w$. Accordingly, we set (i) $\bar{\epsilon} \sim \text{Unif}[1,10]$; (ii) $h \sim \text{Unif}[1/2,2]$; and (iii) $w$ uniformly sampled from $[0,3]^4$. The optimal action can be computed by $a^*_t = w^\top \tilde{X}_t + \frac{\bar{\epsilon}}{1+h}$ \citep{ding2024feature}, which is indeed the $\frac{1}{1+h}$ quantile of the the random variable $w^\top \tilde{X}_t+\epsilon_t$. When implementing OMGPTs, we add the information of $h$ into the context (since it is known), i.e., $X_t=(h,\tilde{X}_t)$. We also consider a square demand function family, which will be specified later, to test the OMGPT's performance on a mixture of different demand-type tasks.

In the non-perishable setting applied in Section \ref{sec:decision_inference}, any unsold products can be carried over to the next period, becoming part of the inventory. Specifically, let $I_t$ denote the inventory at the beginning of each period $t$, with $I_1=0$. This leads to several adjustments: (i) the action $a_t$ must satisfy $a_t \geq I_t$ for each $t$; (ii) the context now includes $I_t$, which should be considered when making decisions, i.e., $X_t = (I_t, h, \tilde{X}_t)$; (iii) the optimal decision is now $a^*_t = \max\{0, w^\top \tilde{X}_t + \frac{\bar{\epsilon}}{1+h} - I_t\}$. (iv) $I_{t+1}=\max\{0,a_t-D_t\}$ (under uncensored cases). In the experiment, we follow the same generation method but allow the action space $\mathcal{A}$ to vary over time as $[I_t, 30 + I_t]$.

\paragraph{Queuing control} We consider a service rate controlling task for a single queue, which is adopted by an example of \citet{de2003linear} (Section 6.1). We assume the queue begins with the empty state at $t=1$. At the beginning of each time period $t$, the decision maker can first observe the length of queue, and then decide the service rate $a_t\in \{0,0.2,\ldots, 1\}$ to process the customers in the queue. Then right before the end of the time period, one customer will arrive with an unknown arrival rate $\lambda$. We assume the queue has a maximum length $4$ and set the observation $O_t$ by the length of the queue at the end of the time period $t$, then we have $\mathcal{O}=\{0,1,2,3,4\}$. Given the service rate $a_t$, the transition matrix of the length of the queue is  
\[\bordermatrix{ & 0 & 1 & 2 &3 & 4 \cr
      0 & 1-\lambda & \lambda & 0 &0 &0\cr
      1 & (1-\lambda)a_t & (1-\lambda)(1-a_t)+\lambda a_t & \lambda(1-a_t) &0 &0 \cr
            2 & 0& (1-\lambda)a_t & (1-\lambda)(1-a_t)+\lambda a_t & \lambda(1-a_t) &0 \cr
                  3 & 0& 0& (1-\lambda)a_t & (1-\lambda)(1-a_t)+\lambda a_t & \lambda(1-a_t) \cr
    4 & 0&0& 0 & (1-\lambda)a_t & 1-a_t+\lambda a_t }.
      \]
The reward function is a minus cost function with a known cost parameter $c$:
$$r(X_t,a_t)=-(O_{t-1}+c \cdot a_t^2),$$
where the context $X_t=(O_{t-1},c)$ feeding the information of the cost parameter to the decision marker.
In our experiments, the parameters are uniformly sampled from $\lambda\sim \text{Unif}\left(\{0.1,0.26,0.42,0.58,0.74,0.9\}\right)$, $c \sim \text{Unif}\left(\{5,16, 27,38,49, 60\}\right)$. Thus there are in total $36$ possible environments.

\paragraph{Revenue management} We consider a revenue management problem \citep{talluri2006theory,jasin2015performance} with horizon $T$ and budget constraints on $J$ types of resources. In each time period $t$, a customer arrives, with their type i.i.d. sampled from a pool of $K$ possible types according to an unknown distribution, and requests a certain amount of each resource with a reward $\tilde{r}_t$.  The decision is whether to accept or reject the customer's order. Specifically, the problem is modeled using the following linear program:
\begin{align}
   \max \ \ & \sum_{t=1}^T \tilde{r}_t y_t \label{eqn:OLP}  \\
    \text{s.t. }\ & \sum_{t=1}^T \bm{\tilde{A}}_t y_t \le T \cdot \bm{1} \nonumber  \\ 
    & y_t\in\{0,1\},  \ \ t=1,...,T \nonumber
\end{align}
where $\bm{A}_t = (A_{1,t},...,A_{J,t})^\top \in \mathbb{R}^{J}$ represents the resource consumption and the decision variables are $\bm{y}=\left({y}_1,...,{y}_T\right)^\top\in \{0,1\}^{T}.$ There are $J$ constraints and $T$ decision variables. 

In this decision problem, the action $a_t \in [0,1]$ represents the probability of accepting an order, and $X_t$ represents the revealed customer type, i.e., $X_t = (\tilde{r}_t, \bm{\tilde{A}}_t) \in \mathcal{X}$, where $\mathcal{X}$ contains $K$ distinct elements. The reward is defined as $r(X_t, a_t) = \tilde{r}_t \cdot a_t$, and the observation at time $t$ is $O_t = X_t$.

In our experiments, we set $J = 3$ (the number of resource types) and $K = 3$ (the number of potential customer types). Each vector element in the space $\mathcal{X}$ is independently and identically sampled from a uniform distribution in the range $[1, 2]^4$. Additionally, the unknown distribution for the arrival probability of each customer type is generated as $(\frac{\mu_1}{\mu_1+\mu_2+\mu3},\frac{\mu_2}{\mu_1+\mu_2+\mu3},\frac{\mu_3}{\mu_1+\mu_2+\mu3})$, where $\mu_1, \mu_2, \mu_3$ are independently sampled from a uniform distribution over the range $[0, 1]$.

\paragraph{Other settings} 
\begin{itemize}
    \item \textbf{Finite Pool of Environments.} To study the behavior and test the performance of the OMGPT on finite possible environments (e.g., to see if and how $\texttt{TF}_{\hat{\theta}}$ converges to $\texttt{Alg}^*$), we also consider a finite pool of environments as the candidates of $\gamma$. Specifically, for some tasks we first sample finite environments (e.g., $\gamma_1, \ldots, \gamma_4$) i.i.d. following the sampling rules previously described. We then set the environment distribution $\mathcal{P}_{\gamma}$ as a uniform distribution over the pool of sampled environments (e.g., $\{\gamma_1, \ldots, \gamma_4\}$). We should notice this pool of environments does not restrict the context generation (if any): in both the pre-training and testing, contexts are generated following the rules previously described and are independent of the given finite pool. 
    \item 
\textbf{Tasks with Two Demand Types.} To test the OMGPT's performance on tasks with possibly different demand types, we also consider the square demand function family for the newsvendor tasks. Specifically, besides the linear demand function family, we also consider $d^{\text{sq}}(X_t, a_t) = (w^\top X_t)^2$ in the newsvendor problem. Thus, with a slight abuse of notation, we can augment $\gamma$ such that it also parameterizes the type of demand (linear or square) and the type is unknown to the decision maker. In experiments related to multi-type demands, the probability of each type being sampled is $1/2$, while the distributions over other parameters of the environment remain the same as in the linear demand case.
\item \textbf{Non-stationary environment.} To evaluate OMGPT's performance in non-stationary environments, we modify the demand function in the newsvendor problem, where  $d^{t}(X_t, a_t) = w_t^\top X_t$ varies over time due to changes in the unknown parameter $w_t$. Specifically, we assume there exists a random time $\tau\leq T$, uniformly sampled from $[1,T]$, such that $w_1=w_2=\ldots=w_{\tau}$ and $w_{\tau+1}=w_{\tau+2}=\ldots=w_{T}$. The parameters  $w_1$ and $w_{\tau+1}$ are independently sampled using the generation method described earlier for the newsvendor problem.
\end{itemize}

\subsection{Benchmark Algorithms}
\label{appx:benchmark}

\subsubsection{Multi-armed bandits}
\begin{itemize}
    \item Upper Confidence Bound (UCB) \citep{lattimore2020bandit}: Given $H_t$, the action $a_t$ is chosen by $a_t =\argmax_{a\in \mathcal{A}} \hat{r}_a+\frac{\sqrt{2\log T}}{\min\{1,n_a\}}$, where $n_a$ is the number of pulling times of arm $a$ before time $t$ and $\hat{r}_a$ is the empirical mean reward of $a$ based on $H_t$. 
    \item Thomspon sampling (TS) \citep{russo2018tutorial}: Given $H_t$, the action $a_t$ is chosen by $a_t =\argmax_{a\in \mathcal{A}} \Tilde{r}_a$, where each $\tilde{r}_a\sim \mathcal{N}\left(\hat{r}_a,\frac{\sqrt{2\log T}}{\min\{1,n_a\}}\right)$ is the sampled reward of arm $a$ and $\hat{r}_a, n_a$ follow the same definitions as in UCB.
    \item $\texttt{Alg}^*$: Given $H_t$ from a task with a pool of $|\Gamma|$ environments $\{\gamma_1,\ldots,\gamma_{|\Gamma|}\}$, the action $a_t$ is chosen by the posterior sampling defined in Algorithm \ref{alg:posterior}. The posterior distribution can be computed by 
    $$\mathcal{P}(\gamma_i| H_t)=\frac{\exp(-\frac{1}{\sigma^2}\sum_{\tau=1}^{t-1}(O_{\tau}-r^{i}_{a_{\tau}})^2)}{\sum_{i'=1}^{|\Gamma|}\exp(-\frac{1}{\sigma^2}\sum_{\tau=1}^{t-1}(O_{\tau}-r^{i'}_{a_{\tau}})^2)}, $$
    where $r^i_{a}$ is the expected reward of $a$ in environment $\gamma_i$ and $\sigma^2$ is the variance of the noise (which equals to $0.2$ in our experiments).
\end{itemize}
\subsubsection{Linear bandits}
\begin{itemize}
    \item LinUCB \citep{chu2011contextual}: Given $H_t$, we define $\Sigma_t=\sum_{\tau=1}^{t-1}a_{\tau}a_{\tau}^\top+\sigma^2 I_d$, where $\sigma^2$ is the variance of the reward noise. The action $a_t$ is chosen by $a_t \in \argmax_{a\in \mathcal{A}} \hat{w}_t^\top a+\sqrt{2\log T}\|a\|_{\Sigma_{t}^{-1}}$, where $\hat{w}_t=\Sigma_t^{-1}(\sum_{\tau=1}^{t-1} O_{\tau}\cdot a_{\tau})$ is the current estimation of $w$ from $H_t$.
    \item LinTS \citep{agrawal2013thompson}: Given $H_t$, the action $a_t$ is chosen by $a_t=\argmax_{a\in \mathcal{A}} \tilde{w}_t^\top a$, where $\Tilde{w}_t\sim \mathcal{N}\left(\hat{w}_t, \sqrt{2\log T}\Sigma_t^{-1}\right)$ as the sampled version of $w$ and  $\hat{w}_t, \Sigma_t$ follow the same definitions in LinUCB. 
    \item $\texttt{Alg}^*$: Given $H_t$ from a task with a pool of $|\Gamma|$ environments $\{\gamma_1,\ldots,\gamma_{|\Gamma|}\}$, the action $a_t$ is chosen by the posterior median defined in Algorithm \ref{alg:posterior}. The posterior distribution can be computed by 
    $$\mathcal{P}(\gamma_i| H_t)=\frac{\exp(-\frac{1}{\sigma^2}\sum_{\tau=1}^{t-1}(O_{\tau}-w_i^\top a_{\tau})^2)}{\sum_{i'=1}^{|\Gamma|}\exp(-\frac{1}{\sigma^2}\sum_{\tau=1}^{t-1}(O_{\tau}-w^\top_{i'} a_{\tau})^2)}, $$
    where $w_i$ is the reward function parameter in environment $\gamma_i$ and $\sigma^2$ is the variance of the noise (which equals to $0.2$ in our experiments).
\end{itemize}
\subsubsection{Dynamic pricing}
All the benchmark algorithms presented below assume the demand model belongs to the linear demand function family as defined in Appendix \ref{appx:envs} (which can be mis-specified when we deal with dynamic pricing problems with two demand types).
\begin{itemize}
    \item Iterative least square estimation (ILSE) \citep{qiang2016dynamic}: At each $t$, it  first estimates the unknown demand parameters $(\alpha,\beta)$ by applying a ridge regression based on $H_t$, and chooses $a_t$ as the optimal action according to the estimated parameters and the context $X_t$. Specifically, we denote $(\hat{\alpha}_{t},-\hat{\beta}_{t})=\Sigma_t^{-1}(\sum_{\tau=1}^{t-1}O_{\tau}\cdot z_{\tau})$ as the estimation of $(\alpha,-\beta)$ through a ridge regression, where $z_{\tau}=(X_{\tau},a_{\tau}\cdot X_{\tau})$ serves as the ``feature vector'' and $\Sigma_t=\sum_{\tau=1}^{t-1}z_{\tau}z_{\tau}^\top+\sigma^2 I_d$ ($\sigma^2$ is the variance of the demand noise). Then the action is chosen as the optimal one by treating $(\hat{\alpha}_{t},\hat{\beta}_{t})$ as the true parameter: $a_t=\frac{\hat{\alpha}_{t}^\top X_t}{2\hat{\beta}_{t}^\top X_t}$.
    \item Constrained iterated least squares (CILS) \citep{keskin2014dynamic}: It is similar to the ILSE algorithm except for the potential explorations when pricing. Specifically, we follow the notations in ILSE and denote $\hat{a}_t=\frac{\hat{\alpha}_{t}^\top X_t}{2\hat{\beta}_{t}^\top X_t}$ as the optimal action by treating $(\hat{\alpha}_{t},\hat{\beta}_{t})$ as the true parameter (the chosen action of ILSE), and denote $\Bar{a}_{t-1}$ as the empirical average of the chosen actions so far. Then the CILS chooses 
    \begin{equation*}
a_t=\begin{cases}
 \bar{a}_{t-1}+\text{sgn}(\delta_t)\frac{ t^{-\frac{1}{4}}}{10}, &\text{if } |\delta_t|<\frac{1}{10} t^{-\frac{1}{4}} ,  \\
\hat{a}_t, &\text{otherwise },
\end{cases}
\end{equation*}
where $\delta_t=\hat{a}_t-\bar{a}_{t-1}$. The intuition is that if the tentative price $\hat{a}_t$ stays too close to the history average, it will introduce a small perturbation around the average as price experimentation to encourage parameter learning.
    \item Thomspon sampling for pricing (TS) \citep{wang2021dynamic}: Like the ILSE algorithm, it also runs a regression to estimate $(\alpha,\beta)$ based on the history data, while the chosen action is the optimal action of a sampled version of parameters. Specifically, we follow the notations in ILSE and then TS chooses
    $$a_t=\frac{\tilde{\alpha}}{2\cdot \Tilde{\beta}},$$
    where $(\Tilde{\alpha},\Tilde{\beta})\sim\mathcal{N}((\hat{\alpha}_{t}^\top X_t,\hat{\beta}_{t}\top X_t),\tilde{\Sigma}_t^{-1})\in\mathbb{R}^2$ are the sampled parameters of the ``intercept'' and ``price coefficient'' in the linear demand function and 
    \begin{equation*}
    \Tilde{\Sigma}_{t}=\left( \begin{pmatrix}
X_t&\bm{0}\\
\bm{0}&X_t
\end{pmatrix}^\top \Sigma_{t}^{-1} \begin{pmatrix}
X_t&\bm{0}\\
\bm{0}&X_t
\end{pmatrix} \right)^{-1}
\end{equation*}
is the empirical covariance matrix given $X_t$.
    \item  $\texttt{Alg}^*$: Given $H_t$ from a task with a pool of $|\Gamma|$ environments $\{\gamma_1,\ldots,\gamma_{|\Gamma|}\}$,  the action $a_t$ is chosen by the posterior averaging defined in Algorithm \ref{alg:posterior}. To compute the posterior distribution, we follow the notations in ILSE and denote $w=(\alpha,\beta)$ as the stacked vector of parameters, then   the posterior distribution is 
    $$\mathcal{P}(\gamma_i| H_t)=\frac{\exp(-\frac{1}{\sigma^2}\sum_{\tau=1}^{t-1}(O_{\tau}-w_i^\top z_{\tau})^2)}{\sum_{i'=1}^{|\Gamma|}\exp(-\frac{1}{\sigma^2}\sum_{\tau=1}^{t-1}(O_{\tau}-w^\top_{i'} z_{\tau})^2)}, $$
    where $w_i$ is the demand function parameter in environment $\gamma_i$ and $\sigma^2$ is the variance of the noise (which equals to $0.2$ in our experiments). 
\end{itemize}
\subsubsection{Newsvendor}
All the benchmark algorithms presented below assume the demand model belongs to the linear demand function family as defined in Appendix \ref{appx:envs} (which can be mis-specified when we deal with newsvendor problems with two demand types).
\begin{itemize}
    \item Empirical risk minimization (ERM) \citep{ban2019big}: Since the optimal action $a^*_t$ is the $\frac{1}{1+h}$ quantile  of the random variable $w^\top \tilde{X}_t+\epsilon_t$ \citep{ding2024feature}, ERM conducts a linear quantile regression based on the observed contexts and demands $\{(\tilde{X}_{\tau},O_{\tau})\}_{\tau=1}^{t-1}$ to predict the  $\frac{1}{1+h}$ quantile on $\tilde{X}_t$.
    \item Feature-based adaptive inventory algorithm (FAI) \citep{ding2024feature}: FAI is an online gradient descent style algorithm aiming to minimize the cost $\sum_{t=1}^T h\cdot (a_t-O_t)^+ + l\cdot (O_t-a_t)^+$ (we set $l=1$ in our experiments). Specifically,  it chooses 
    $a_t=\Tilde{w}_t^\top \tilde{X}_{t}$, where      \begin{equation*}
\tilde{w}_t=\begin{cases}
\tilde{w}_{t-1}-\frac{h}{\sqrt{t}} \cdot \tilde{X}_{t-1}, &\text{if } O_{t-1}<a_{t-1} ,  \\
\tilde{w}_{t-1}+ \frac{l}{\sqrt{t}} \cdot \tilde{X}_{t-1}, &\text{otherwise},
\end{cases}
\end{equation*}
is the online gradient descent step and $\Tilde{w}_0$ can be randomly sampled in $[0,1]^d$.
\item  $\texttt{Alg}^*$: Given $H_t$ from a task with a pool of $|\Gamma|$ environments $\{\gamma_1,\ldots,\gamma_{|\Gamma|}\}$, the action $a_t$ is chosen by the posterior median as shown in Algorithm \ref{alg:posterior}. To compute the posterior distribution, we denote $\Bar{\epsilon}_{\gamma}$ and $\beta_{\gamma}$ as the noise upper bound and demand function parameter of $\gamma$ at $\tau\leq t-1$, and define the event $\mathcal{E}_{\gamma,\tau}=\left\{ 0 \leq O_{\tau}-\beta_{\gamma}^\top \tilde{X}_{\tau}\leq \Bar{\epsilon} \right \}$ to indicate the feasibility of environment $\gamma$ from $(\tilde{X}_{\tau},O_{\tau})$, and denote $\bar{\mathcal{E}}_{\gamma,t}=\bigcap_{\tau=1}^{\tau-1} \mathcal{E}_{\gamma,\tau}$ to indicate the feasibility at $t$. Then the posterior distribution of the underlying environment is 
    $$\mathcal{P}(\gamma_i| H_t)=\frac{\mathbbm{1}_{\bar{\mathcal{E}}_{\gamma_i,t}}\cdot \Bar{\epsilon}_{\gamma_i}^{1-t}}{\sum_{i'=1}^{|\Gamma|} \mathbbm{1}_{\bar{\mathcal{E}}_{\gamma_{i'},t}}\cdot \Bar{\epsilon}_{\gamma_{i'}}^{1-t}}. $$

\end{itemize}
\subsubsection{Queuing Control}
\begin{itemize}
    \item Random: The algorithm samples an action $a_t$ uniformly from the action space $\mathcal{A}$ at each time step $t$.
\end{itemize}
\subsubsection{Revenue Management}
\begin{itemize}
    \item Adaptive Allocation Algorithm (Ada) \citep{chen2024improved}: This is a re-solving-based algorithm that recalculates a linear optimization problem at each time step, using the sample counts as estimates for the unknown arrival probabilities, as outlined in Algorithm \ref{alg:DAA}. 
    \begin{algorithm}[ht!]
\caption{Adaptive Allocation Algorithm\citep{chen2024improved}}
\label{alg:DAA}
\begin{algorithmic}[1]
\State Input: $T, \mathcal{X}= \{(\tilde{r}^k,\bm{\tilde{A}^k})\}_{k=1}^K$
\State Initialize $\bm{B}_1 = T \cdot \bm{1}$, $\bm{b}_1=\bm{B}_1/T$
\State Set $y_1=1$
\For {$t=2,..., T$}
\State Compute $\bm{B}_t=\bm{B}_{t-1}-\bm{\tilde{A}}_{t-1}y_{t-1}$
\State Compute $\bm{b}_t=\bm{B}_{t}/(T-t+1)$
\State Solve the following linear program where the decision variables are $(y'_1,...,y'_K)$:
\begin{align}
   \max \ \ & \sum_{k=1}^K \frac{n_{t-1}(k)}{t-1} \tilde{r}^k y'_k \label{adpt_lp}  \\
    \text{s.t. }\ & \sum_{k=1}^K  \frac{n_{t-1}(k)}{t-1}\bm{\tilde{A}^k} \cdot y'_k \le \bm{b}_t \nonumber  \\
    & 0 \le y'_k \le 1,  \ \ k=1,...,K \nonumber
\end{align}
\State Denote the optimal solution as $\bm{y}^{'*}_t=(y_{1,t}^{'*},...,y_{K,t}^{'*})$
\State Observe $(\tilde{r}_t,\bm{\tilde{A}_t})$ and identify $(\tilde{r}_t,\bm{\tilde{A}_t}) = (\tilde{r}^k,\bm{\tilde{A}^k})$ for some $k$

\State Set 
\begin{align*}
    y_t = \begin{cases}
    1, & \text{ with probability } y_{k,t}^{'*}\\
    0, & \text{ with probability } 1-y_{k,t}^{'*}
    \end{cases}
\end{align*}
when the constraint permits; otherwise set $y_t=0.$ 
\State Update the counts 
\begin{align*}
    n_{t}(k) = \begin{cases}
    n_{t-1}(k)+1, & \text{ if } (\tilde{r}_t,\bm{\tilde{A}_t}) = (\tilde{r}^k,\bm{\tilde{A}^k})\\
    n_{t-1}(k), & \text{ otherwise }
    \end{cases}
\end{align*}
\EndFor
\State Output: $\bm{y} = (y_1,...,y_T)$
\end{algorithmic}
\end{algorithm}
\end{itemize}

%%%%%%%%%%%%%%%%%%%%%%%%%%%%%%%%%%%%%%%%%%%%%%%%%%%%%%%%%%%%

\appendix

\end{document}